%% file: acl_latex.tex
\documentclass[11pt]{article}
\usepackage{amsmath}
\usepackage{amssymb}

\usepackage[final]{acl}

\usepackage{times}
\usepackage{latexsym}

\usepackage{subcaption}

\usepackage[T1]{fontenc}

\usepackage[utf8]{inputenc}

\usepackage{microtype}

\usepackage{inconsolata}

\usepackage{graphicx}
\usepackage{booktabs}  
\usepackage{tabularx}  
\usepackage{array}     
\usepackage{multirow}  
\usepackage{xcolor}    

\title{Towards a Mechanistic Understanding of Propositional \\ Logical Reasoning in Large Language Models}

\author{
  \textbf{Danchun Chen\textsuperscript{1,2}\thanks{Equal Contribution.}} \and
  \textbf{Qiyao Yan\textsuperscript{1,3}\footnotemark[1]} \and
  \textbf{Chenpeng Wang\textsuperscript{5}} \and
  \textbf{Liangming Pan\textsuperscript{1,3,4}\thanks{Corresponding Author.}} \\
  \textsuperscript{1}State Key Laboratory of Multimedia Information Processing, Peking University\\
  \textsuperscript{2}Singapore Management University\\
  \textsuperscript{3}School of Computer Science, Peking University\\
  \textsuperscript{4}Beijing Academy of Artificial Intelligence, Beijing, China\\
  \textsuperscript{5}YiXin-AILab, YIXIN, Beijing, China\\
  \texttt{dcchen.nexus@gmail.com}, \texttt{yqy.1@foxmail.com},
  \texttt{liangmingpan@pku.edu.cn}
}

\usepackage{placeins}
\begin{document}
\maketitle
\begin{abstract}
Understanding how Large Language Models (LLMs) perform logical reasoning internally remains a fundamental challenge. While prior mechanistic studies focus on identifying task-specific circuits, they leave open the question of what computational strategies LLMs employ for propositional reasoning.
We address this gap with a causal mechanistic analysis on \textit{PropLogic-MI}, a controlled benchmark of 11 propositional rules across one- and two-hop tasks, applied to three model families (\textit{Qwen3}, \textit{Llama-3.1}, \textit{Mistral}). Rather than asking which components are necessary, we ask how the reasoning process is organized, and identify four interlocking mechanisms: \textit{Staged Computation}, where early, middle, and late layers take on distinct functional roles; \textit{Information Transmission}, where semantic content aggregates at boundary tokens;
\textit{Fact Retrospection}, where fact tokens stay causally active as binding lookup in middle layers and as sustained access in late layers under hard reasoning loads; and \textit{Specialized Attention Heads}
that structurally implement these patterns. A prompt-order control that places the query before the facts confirms that this organization is model-internal rather than input-layout-induced. These findings show that pretrained LLMs solve propositional reasoning through a structured, layer-organized process that recurs across models, rule categories, and reasoning hops.
\footnote{Code is available at \url{https://github.com/yanqiyao1/Propo-MI}.}
\end{abstract}

\section{Introduction}

The capability of Large Language Models (LLMs) to execute complex logical reasoning has improved significantly \citep{cheng2025empoweringllmslogicalreasoning, openai2024openaio1card, deepseekai2025deepseekr1incentivizingreasoningcapability}. State-of-the-art evaluations demonstrated this trend: \textit{GPT-5.2} achieves 92\% accuracy on \textit{GPQA-Diamond} and \textit{Gemini-3-Pro} reaches 76\% on \textit{SimpleBench}.
However, the underlying computational mechanisms behind such high empirical performance remain poorly understood.
This gap in understanding raises a fundamental question: are these models simply sophisticated pattern matchers relying on surface statistics
\citep{bender2021stochasticparrots, dziri2023faithfatelimitstransformers},
or have they internalized robust algorithms that mirror logical deduction \citep{li2023emergent, nanda2023progress}?
To address this ambiguity, the emerging field of \textit{Mechanistic Interpretability} (MI)
\citep{sharkey2025openproblemsmechanisticinterpretability}
has begun to reverse-engineer transformers to causally uncover the mechanisms underlying model behaviors. Recent efforts have applied MI to reasoning domains including arithmetic \citep{kantamneni2025languagemodelsuse, stolfo2023mechanisticinterpretationarithmeticreasoning, yu2024interpretingarithmeticmechanismcomparative},
syllogistic inference \citep{kim2024reasoningcircuitslanguage}, and multi-hop composition \citep{ye2025transformerslearnimplicit, wang2024languagemodelscan}. Their findings suggest that LLMs may employ structured, interpretable strategies when processing logical reasoning, rather than relying solely on opaque pattern matching.

Propositional logical reasoning, \textit{i.e.}, the capacity to manipulate boolean operators and chain inference rules, represents a cornerstone of formal deduction that underpins more complex reasoning tasks. Despite its foundational role, this domain remains largely unexplored in MI. Behavioral studies reveal that model accuracy degrades significantly with increased proof depth \citep{dziri2023faithfatelimitstransformers}, yet the internal computational mechanisms responsible for these failures remain opaque. A prior mechanistic work, \citet{aimpliesb}, provides circuit-level analysis identifying four attention head families in \textit{Mistral-7B}, \textit{Gemma-2-9B}, and \textit{Gemma-2-27B}. While they demonstrate that similar circuits emerge across these models, their study is constrained to a single operator structure (\textit{i.e.}, implication chains) using simplified one-hop problems. Broader propositional rules, including \textit{De Morgan's Laws}, \textit{Distributivity}, and \textit{Commutativity}, as well as applying multiple different rules in a multi-step reasoning process, remain mechanistically unexplored. Moreover, their circuit-discovery paradigm focuses on identifying \textit{which specific components} are necessary for a given model-task pair, rather than characterizing \textit{how} the reasoning process in computational strategies that may transfer across model architectures and logical structures.

\begin{figure*}[!htbp]
    \centering
    \begin{subfigure}[b]{0.46\textwidth}
        \centering
        \includegraphics[width=\linewidth]{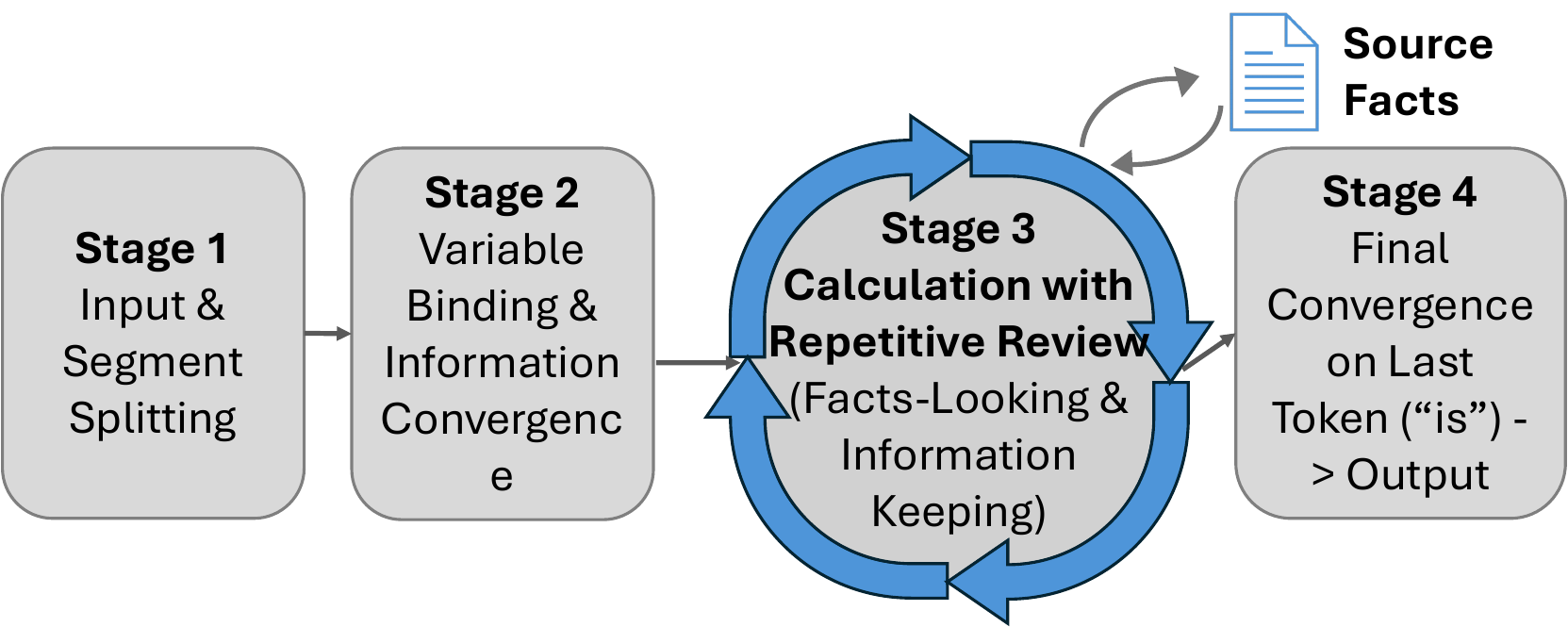}
        \vspace{-0.5em}
        \caption{\textbf{Staged Computation}}
        \label{fig:overview_pipeline}
    \end{subfigure}
    \hspace{0.5em}
    \begin{subfigure}[b]{0.46\textwidth}
        \centering
        \includegraphics[width=\linewidth]{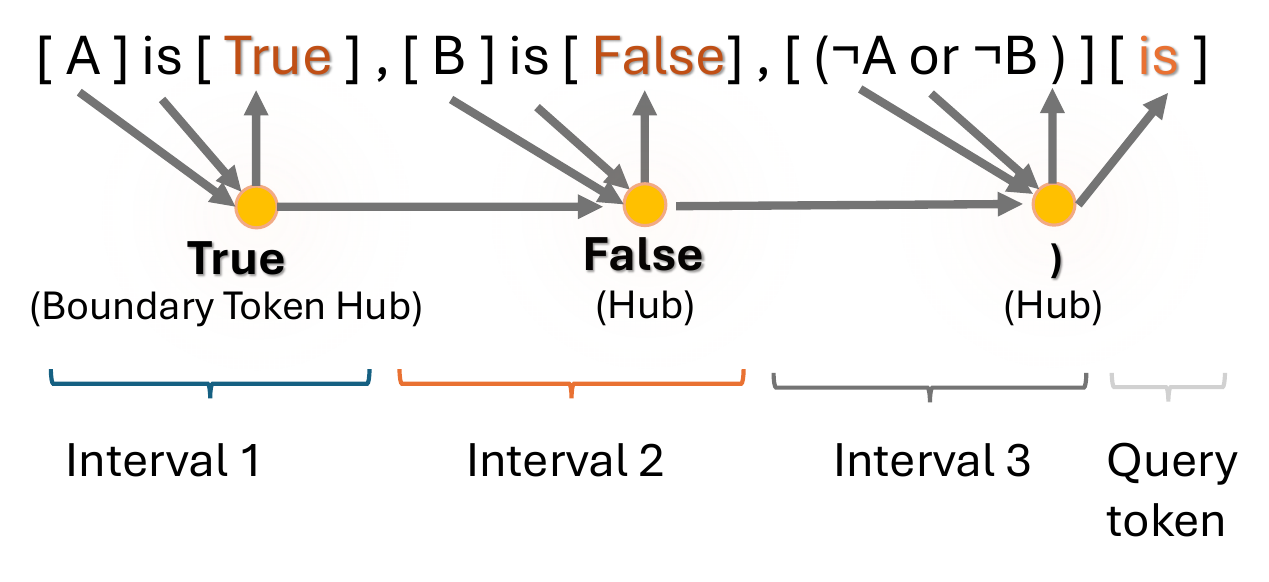}
        \vspace{-1.8em}
        \caption{\textbf{Information Transmission}}
        \label{fig:info_convergence}
    \end{subfigure}

    \vspace{0.9em}

    \begin{subfigure}[b]{0.46\textwidth}
        \centering
        \includegraphics[width=\linewidth]{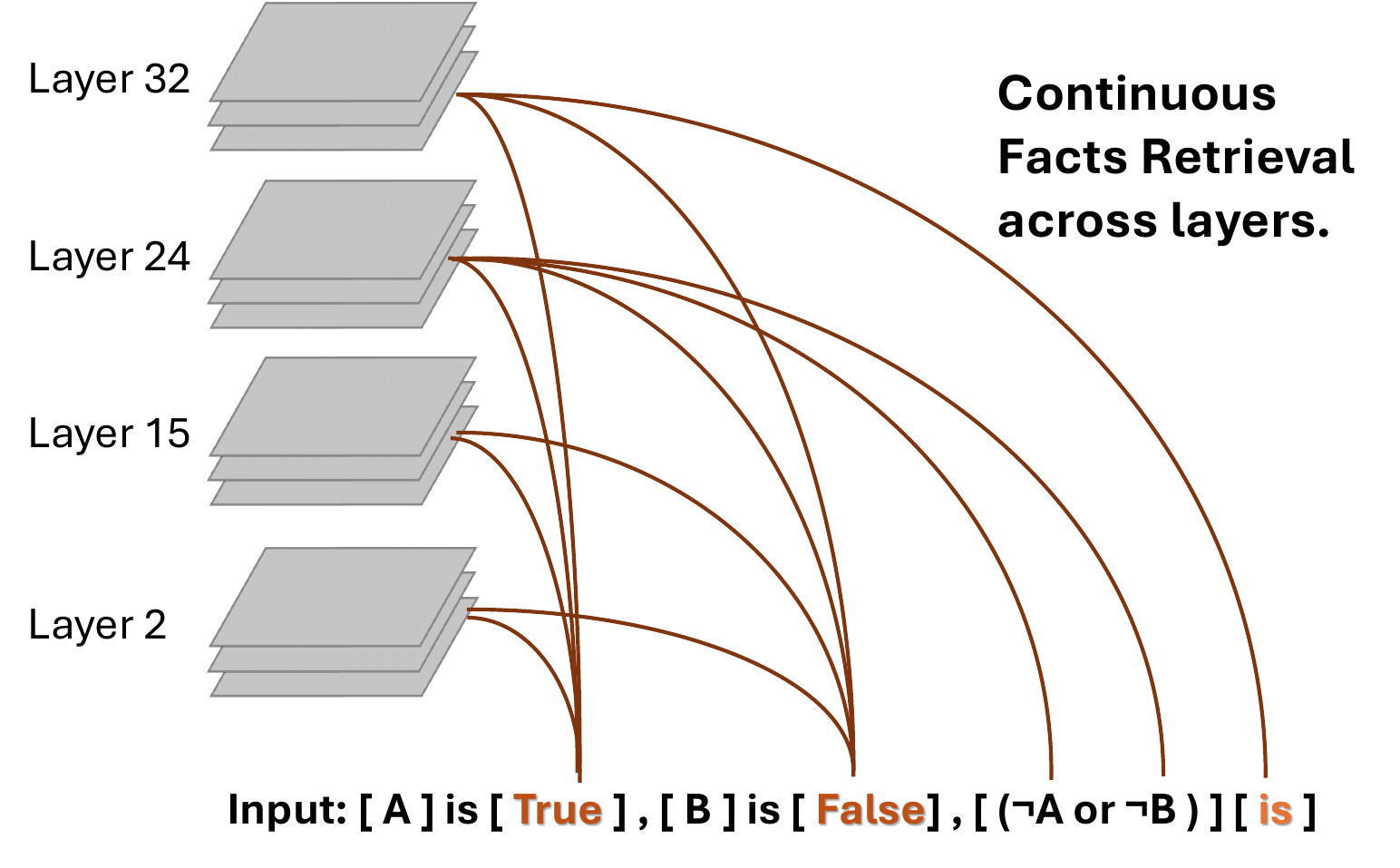}
        \vspace{-0.9em}
        \caption{\textbf{Fact Retrospection}}
        \label{fig:repetitive_looking}
    \end{subfigure}
    \hspace{0.5em}
    \begin{subfigure}[b]{0.46\textwidth}
        \centering
        \includegraphics[width=\linewidth]{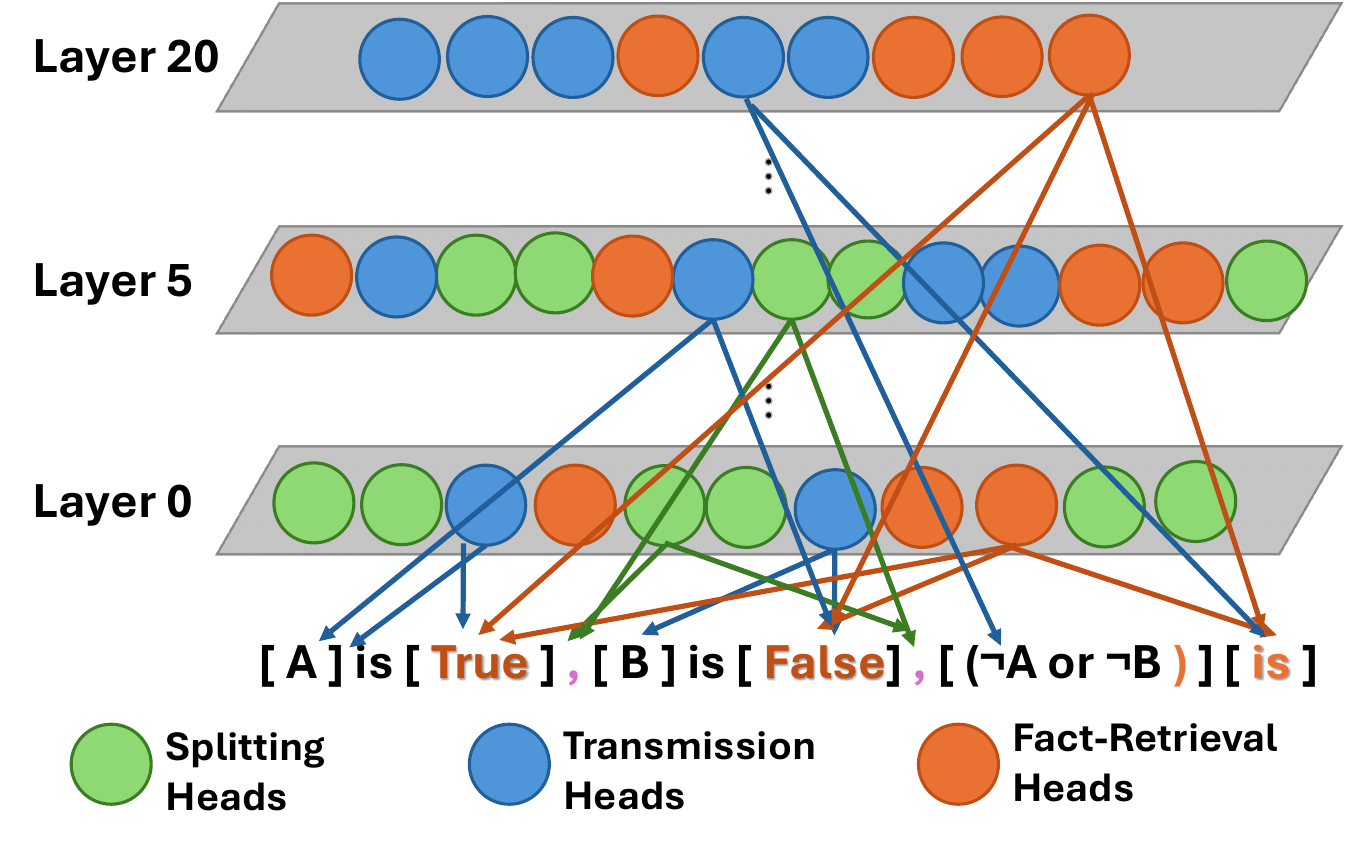}
        \vspace{-0.9em}
        \caption{\textbf{Specialized Attention Heads}}
        \label{fig:attention_heads}
    \end{subfigure}

    \vspace{-0.3em}
    \caption{\textbf{Four interlocking mechanisms of propositional-logic reasoning.} (a) \textbf{Staged Computation}: early, middle, and late layers take on distinct functional roles. (b) \textbf{Information Transmission}: information aggregates at semantic boundary tokens. (c) \textbf{Fact Retrospection}: fact tokens used for middle-layer binding lookup and late-layer access under hard reasoning loads. (d) \textbf{Specialized Attention Heads}: Splitting, Transmission, and Fact-Retrieval heads implement the three patterns above.}
    \vspace{-0.5em}
    \label{fig:overview}
\end{figure*}
In this paper, we address this gap with a more comprehensive analysis.
First, we shift analytical focus from model-specific circuit identification to generalizable mechanistic principles.
As shown in Figure~\ref{fig:overview}, this perspective reveals four interlocking mechanisms: (1) \textit{Staged Computation}, where fact premises, logical expressions and query compute mainly at early, middle, and late layers; (2) \textit{Information Transmission}, where semantic information aggregates at semantic boundary tokens, \textit{i.e.}, positions that close or anchor a meaningful segment such as a factual clause, logical expression, or query suffix; (3) \textit{Fact Retrospection}, in which fact tokens remain causally active beyond early-layer encoding --- as a middle-layer binding lookup and, under hard reasoning loads, as a sustained late-layer signal; and (4) \textit{Specialized Attention Heads}, where heads specialize into Splitting, Transmission, and Fact-Retrieval roles, respectively marking segment boundaries, routing information within a logical block, and looking up bound fact values.
%
Second, we expand the scope of analysis to $11$ propositional logic rule categories across one- and two-hop reasoning on four models (\textit{i.e.}, \textit{Qwen3-8B}, \textit{Qwen3-14B}, \textit{Llama-3.1-8B-Instruct}, and \textit{Mistral-7B-Instruct}), and find the same four mechanisms \textit{generalize} across these axes.
%
Third, we run every analysis under two prompt orderings, one where the facts precede the query and one where they follow it, to test whether the staged organization tracks the surface prompt layout or the underlying logical content. The same staged template emerges in both orderings across all models, indicating that the organization is model-internal rather than input-order-induced.
%
These analyses provide a more complete and generalizable picture of the mechanisms of propositional logical reasoning.

\section{Related Work}
A common three-level abstraction in \textit{Mechanistic Interpretability} (MI) distinguishes features, circuits, and the algorithms they implement \citep{bereska2024mechanisticinterpretabilityaisafety}. Building on this view, we organize prior work on LLM reasoning along two complementary axes.
A \textit{component-level} lens asks \textit{which} heads or layers implement a given sub-computation, while a \textit{strategy-level} lens asks \textit{how} those components are orchestrated into a network-wide reasoning pattern.

\paragraph{Component-Level MI.}
MI has identified specific components responsible for individual computational primitives: induction heads for in-context learning \citep{olsson2022incontextlearninginductionheads}, MLP layers that store factual associations \citep{meng2023locatingeditingfactualassociations, geva2023dissectingrecallfactual, ferrando2025informationflowlarge}, and binding ID vectors through which attention heads link entities to their attributes \citep{feng2024languagemodelsrepresent, wang2023labelwordsanchors}. Causal mediation analysis \citep{vig2020causalinterpretingneuralNLP, hase2024doeslocalizationinform, zhang2024towardsbestpractices, heimersheim2024useinterpretactivation} and circuit-based methods \citep{elhage2021mathematicalframeworktransformer, wang2022interpretabilitywildcircuitindirect} extend this by identifying minimal subgraphs causally necessary for specific tasks.
%
%
For reasoning specifically, \citet{kim2024reasoningcircuitslanguage} decompose syllogistic inference into modular circuits with premise-retrieval and conclusion-assembly components. These studies provide fine-grained causal evidence at the level of individual heads and layers.

However, as \citet{NEURIPS2024_47c7edad} observe, the specific attention heads realizing a functional component may shift across training and scale, yet the overarching algorithm they implement remains stable. Our work contributes primarily on the strategy-level axis while drawing its primitives from the component-level literature.

\paragraph{Strategy-Level MI.}
This line of research focuses on how individual components organize into network-wide reasoning strategies.
\citet{dutta2024thinkstepbystep} trace how \textit{chain-of-thought} (CoT) prompting restructures internal information flow, showing that \textit{explicit} intermediate tokens serve as anchor points for staged computation. Those tokens, however, are generated and re-enter the context as externalized working memory, so the organization built around them may belong to the scaffold instead of LLM itself. Moreover, the verbalized rationale itself need not faithfully reflect the computation that produces the answer \citep{turpin2023language, lanham2023measuring, paul2024making}.
Therefore, we remove the scaffold and ask whether a similar organization appears when LLMs have to do all the reasoning \textit{implicitly}. We find that models still stage the computation temporally across layers and spatially across token positions, anchored by boundary tokens instead of generated ones.

A strategy-level view also recasts primitives identified at the component level. Prior component-level studies show that attention heads link entities to attributes through \textit{binding ID vectors} \citep{feng2024languagemodelsrepresent} and that MLPs localize \textit{single-fact} retrieval to specific layers \citep{meng2023locatingeditingfactualassociations}. In the propositional setting, both observations resurface as part of a single \textit{Fact Retrospection} pattern, in which fact tokens remain causally active beyond their initial encoding: middle layers carry a binding-flow lookup, and under hard reasoning loads (\textit{e.g.}, two-hop reasoning) an additional sustained-access signal appears at late layers.

\paragraph{Mechanistic Studies of Propositional Logic.}

To the best of our knowledge, the only mechanistic study on propositional logic is \citet{aimpliesb}, which decomposes implication reasoning into four functional head types (locating, mover, processing, and decision heads) in \textit{Mistral-7B} and \textit{Gemma-2}, pinpointing the heads responsible for rule retrieval, information movement, and answer selection. Building on this \textit{head-level} decomposition, we characterize \textit{network-wide} computational patterns along three complementary axes: \textit{temporal computation} across depth, \textit{spatial information routing} across token positions, and the underlying \textit{computational operators}. Combined with broader rule coverage and multi-hop scenarios, this strategy-level perspective reveals organizational principles that generalize across diverse logical structures, inference chain lengths, model scales and families.

\section{PropLogic-MI: Task and Dataset}
\label{sec:proplogic_mi}

\subsection{Task Formulation}
\label{sec:task}

We formulate propositional reasoning as \textit{next-token prediction} over a \emph{logical instance} $I=(F,E,Q)$, where $F$ is a set of factual assignments, $E$ is a set of logical expressions,
$Q$ is the queried expression, and the gold label $y \in \{\texttt{True},\texttt{False}\}$ is determined by $I$.
We study two prompt ordering, \textit{i.e.},
$\pi_{\mathrm{FF}}$ listing $F$ before $E$, and $\pi_{\mathrm{EF}}$ placing $E$ first. Both preserve the same logical contents and the same answer, so differences between them reflect input order.
The model is required to emit the truth value immediately after the query for analyzing the hidden computation that precedes the answer.

\subsection{Dataset Construction}
\label{sec:dataset}
\paragraph{PropLogic-MI.}
\textit{PropLogic-MI} is a controlled corpus for mechanistic analysis.
It spans $11$ propositional rule categories and two reasoning depths:
(1) \emph{one-hop} questions can be answered directly from the stated facts, and (2) \emph{two-hop} reasoning, where an intermediate assignment $M=e_{1}$ must first be derived before predicting the final answer.
Each logical sample is instantiated under both prompt orderings (\textit{i.e.}, $\pi_{\mathrm{FF}}$ and $\pi_{\mathrm{EF}}$) with matched clean/corrupted pairs, shared labels, and randomized variable remapping. Full rule definitions, prompt templates, and data generation details can be seen at Appendix~\ref{sec:appendix_dataset}.
\paragraph{Data Statistics.}
Each prompt ordering contains $20{,}000$ clean/corrupted pairs ($9{,}523$ one-hop and $10{,}477$ two-hop), with near-balanced rule coverage;
Table~\ref{tab:dataset_baselines} summarizes clean accuracy and the \emph{dual-correct} rate, namely the fraction of prompts that a model answers correctly on both the clean and corrupted runs. This dual-correct subset defines the sample pool for experiments. \textit{Qwen3} yields larger pools than \textit{Llama} and \textit{Mistral}, and the same four mechanisms recur across families and analysis is restricted to these matched pools. Therefore, we only claim the strategy-level mechanism is shared conditionally on successful reasoning; how a model fails, and whether failure mechanisms coincide across models, is not addressed by this pool.

\begin{table}[!htbp]
\centering
\small
\setlength{\tabcolsep}{3.5pt}
\begin{tabular}{l cc cc c}
\toprule
& \multicolumn{2}{c}{$\pi_{\mathrm{FF}}$} & \multicolumn{2}{c}{$\pi_{\mathrm{EF}}$} & \\
\cmidrule(lr){2-3}\cmidrule(lr){4-5}
\textbf{Model} & Clean & Dual & Clean & Dual & Pool ($\pi_{\mathrm{FF}}$/$\pi_{\mathrm{EF}}$) \\
\midrule
\textit{Qwen3-8B}            & 87.0 & 75.4 & 74.2 & 53.9 & 15{,}087 / 10{,}778 \\
\textit{Qwen3-14B}           & 87.4 & 76.6 & 74.4 & 52.8 & 15{,}310 / 10{,}562 \\
\textit{Llama-3.1-8B}        & 63.9 & 38.4 & 60.3 & 37.0 & \phantom{0}7{,}674 / \phantom{0}7{,}403 \\
\textit{Mistral-7B}          & 65.1 & 35.1 & 66.7 & 40.0 & \phantom{0}7{,}022 / \phantom{0}7{,}995 \\
\bottomrule
\end{tabular}
\caption{\textbf{Data Statistics of \textit{PropLogic-MI}.} Each prompt ordering contains $20{,}000$ prompts. Clean = clean accuracy (\%). Dual = dual-correct rate (\%), \textit{i.e.}, accuracy on both the clean and corrupted runs. The right column gives the pool size used for mechanistic analysis. Hop-wise and model-family breakdowns are deferred to Appendix~\ref{sec:appendix_performance} (including the cross-family summary in Table~\ref{tab:cross_family_summary}).}
\label{tab:dataset_baselines}
\vspace{-0.6em}
\end{table}

\section{Methodology}
\label{sec:methodology}

\paragraph{Causal Interventions.}
\label{sec:primitives}
We use two complementary intervention primitives. The intervention's exact scope (which layers, which token positions, which to average) varies per experiment and is described with experiments in §\ref{sec:results}.
(1) \textbf{Activation patching.}
Following \citet{vig2020causalinterpretingneuralNLP, meng2023locatingeditingfactualassociations}, we form a clean prompt $x_{\text{clean}}$ with gold label $y$ and a minimally corrupted (\textit{e.g.}, flipping the assignment of a fact) prompt $x_{\text{corr}}$ whose prediction flips. We cache an internal activation from the clean run at a chosen layer and token and restore it into the corrupted run, measuring how much the corrupted output moves back toward $y$.
(2) \textbf{Mean ablation.}
We replace a targeted output by a defined mean value computed over a clean reference set. The exact scope of the mean differs across experiments. 

\paragraph{Analytical Metrics.} Let $w_y$ denote the token for the correct label $y \in \{\texttt{True}, \texttt{False}\}$ and $w_{\neg y}$ its opposite. We define \textbf{Probability Difference} as
\begin{equation}
    \mathrm{PD}(x, y) = \mathrm{P}(w_y \mid x) - \mathrm{P}(w_{\neg y} \mid x),
    \label{eq:logit_diff}
\end{equation}
where $\mathrm{P}(w \mid x)$ is the model probability at the answer position. Larger PD means stronger support for the correct answer.
For an intervention $\tilde{x}$, we report \textbf{Probability Difference Shift} as
\begin{equation}
    \mathrm{dPD} = \mathrm{PD}(\tilde{x}, y) - \mathrm{PD}(x_{\text{baseline}}, y),
    \label{eq:dld}
\end{equation}
where $x_{\text{baseline}}$ is the input of the unintervened run (the corrupted prompt for activation patching; the clean prompt for mean ablation).
A positive $\mathrm{dPD}$ indicates that the intervention restores the model toward the correct answer.

\begin{figure}[!htbp]
\centering
\includegraphics[width=\columnwidth]{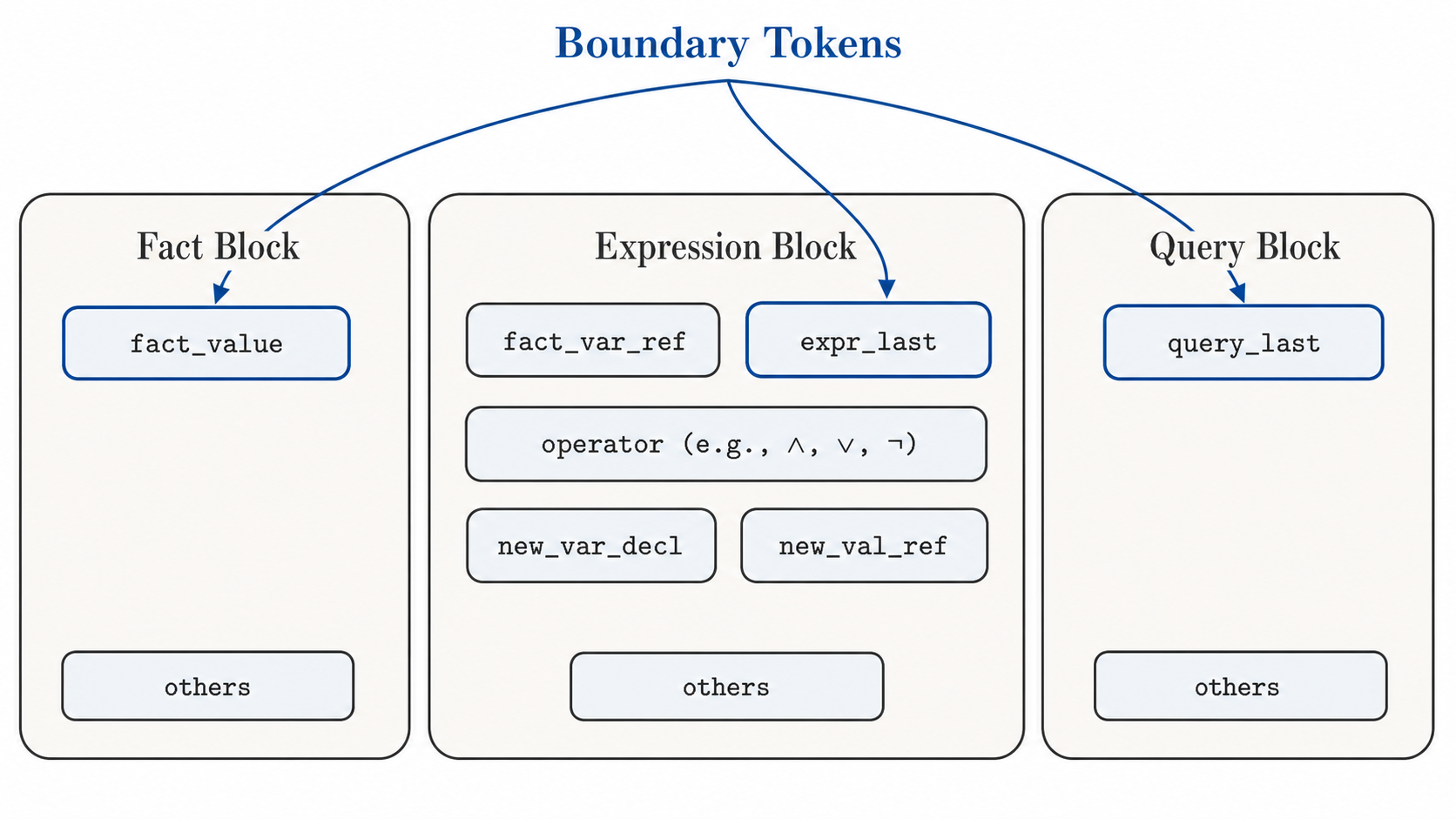}
\caption{\textbf{Two token-categorization granularities.} Coarse logical blocks (\texttt{Fact}, \texttt{Expression}, \texttt{Query}) nest the fine structural categories; arrows mark the boundary tokens (\texttt{fact\_value}, \texttt{expr\_last}, \texttt{query\_last}). Two-hop prompts additionally introduce \texttt{new\_var\_decl} and \texttt{new\_val\_ref} inside the Expression Block.}
\label{fig:granularity}
\end{figure}

\paragraph{Semantic Aggregation.}
To localize computation along the prompt, we take two semantically aggregated granularities (Figure~\ref{fig:granularity}):
(1) \textit{Coarse logical blocks} partition each prompt into three contiguous, role-defined spans --- the \texttt{Fact Block} (\textit{e.g.}, \texttt{A is True, B is False}), the \texttt{Expression Block} (\textit{e.g.}, \texttt{C = (¬A or ¬B)}), and the \texttt{Query Block} (\textit{e.g.}, \texttt{C is}) --- and are intervened on as a whole (used in §\ref{sec:staged}).
(2) \textit{Fine structural categories} further subdivide each block into syntactically meaningful token classes --- the boundary tokens \texttt{fact\_value}, \texttt{expr\_last}, and \texttt{query\_last} that close each block, vs.\ interior tokens such as \texttt{fact\_var\_ref} or \texttt{operator} --- and are scored at the token level (used in §\ref{sec:info_and_retrieval}).
Both granularities share a single per-layer aggregation. Let $\mathcal{T}_c$ denote the units belonging to category $c$; the mean $\mathrm{dPD}$ at layer $l$ is
\begin{equation}
  \overline{\mathrm{dPD}}^{(l)}_{c} = \frac{1}{|\mathcal{T}_c|} \sum_{t \in \mathcal{T}_c} \mathrm{dPD}^{(l)}(t),
  \label{eq:aggregated mean dPD}
\end{equation}
where the unit $t$ is set by the granularity: under (1), $t$ is an entire logical block and $|\mathcal{T}_c|=1$, so $\overline{\mathrm{dPD}}^{(l)}_{c}$ reduces to the block-level $\mathrm{dPD}$; under (2), $t$ ranges over token positions inside structural category $c$ and $|\mathcal{T}_c|$ is the number of such tokens.

\input{experiments}
\section{Discussion}

The most consequential finding concerns \textit{Fact Retrospection}: how the model keeps fact tokens causally active beyond their initial encoding, in two sub-patterns. (1) A \emph{binding-flow lookup} in middle layers that transfers from \texttt{fact\_value} to expression-side variable references, with Fact-Retrieval heads routing from query and expression positions to the bound truth-value tokens. This sub-pattern is universal across our models and hops, and is the propositional-logic counterpart of the binding-ID mechanism of \citet{feng2024languagemodelsrepresent}. (2) A \emph{sustained late-layer access} that emerges only as the task nears the model's capacity limit (\textit{e.g.}, on two-hop setting), where \texttt{fact\_value} stays causally active in parallel with forward routing toward the answer, showing the model holding onto premise tokens when it cannot fully compress them into a single forward path.

The cross-family pattern further ties this mechanism to model capacity. The staged template recurs without exception across families, but how cleanly the three head roles separate from the \textit{Random-Outside} control varies considerably: \textit{Qwen3} shows the cleanest separation, \textit{Mistral} is globally sensitive (both role-targeted and random ablations damage accuracy), and \textit{Llama} saturates at small $k$. The same weaker models also produce the strongest late-layer sub-pattern of Fact Retrospection on multi-hop. Combined, we suppose that the mechanism we describe runs in all four models, but it is implemented more cleanly in models with capacity headroom for the task. While the link between the mechanism and task difficulty is established in §\ref{sec:retro}, a more fine-grained analysis remains necessary, particularly regarding individual logical rules.


\section{Conclusion}
In this paper, we present a mechanistic analysis of propositional logical reasoning and adopt a top-down analytical strategy from macroscopic patterns
to microscopic implementations.
Our investigation uncovers a coherent computational architecture: these LLMs organizes reasoning into temporally distinct phases,
routes information through semantic boundary tokens,
maintains active fact retrieval,
and implements these strategies through specialized attention heads.
These four phenomena form an interlocking system where each mechanism supports and validates the others.
Furthermore, these mechanisms generalize across models, rule categories, and reasoning hops, indicating a robust property of LLM propositional reasoning rather than model- or task-specific artifacts. By characterizing \textit{how} models reason rather than \textit{which components} are necessary, we contribute a strategy-level mechanistic understanding of propositional reasoning in LLMs.

\section*{Limitations}
Our analysis has three main limitations. First, while we causally validate attention-head specialization in detail, the internal computations within MLP layers remain largely uncharacterized. Our MLP patching reveals \emph{when} MLPs are causally important, but not \emph{what} transformation they implement at each stage --- whether they encode fact values, execute rule-specific Boolean operations, or integrate bound truth values at the answer position --- nor how these transformations emerge during training.


Second, the task setting is deliberately controlled: prompts are short, symbolic, and limited to propositional rules with one- and two-hop chains; our four models span three families (\textit{Qwen3}, \textit{Llama-3.1}, \textit{Mistral}). Whether the same staged organization persists with longer chains, natural-language phrasings, richer logical systems, or explicit intermediate-step formats remains an open empirical question, and generality to other LLM families remains to be established.

Third, all mechanistic analyses are restricted to the \emph{dual-correct} pool, namely prompts that each model answers correctly on both the clean and corrupted runs. Retention rates differ across families (about 75\% for \textit{Qwen3} versus 35--40\% for \textit{Llama-3.1} and \textit{Mistral}), so our claims describe the computation underlying \emph{successful} propositional reasoning on these matched pools, not failure-case behavior or absolute cross-family difficulty.

\section*{Acknowledgments}
This work was supported in part by the Beijing Major Science and Technology Project under Contract No.~Z251100008125054. This work was supported by the Beijing Academy of Artificial Intelligence (BAAI). This work was supported by Yixin Group Limited. We gratefully acknowledge their provision of the essential computing resources required for our experiments.

\bibliography{custom}


\appendix
\section*{Appendix Roadmap}
The appendix supplies (i) full dataset specification and prompt templates (§\ref{sec:appendix_dataset}), (ii) implementation details for the causal probes (§\ref{sec:appendix_setup}), (iii) hop-wise and rule-wise behavioral baselines (§\ref{sec:appendix_performance}), and three results sections that supplement the body experiments along the same backbone: staged computation (§\ref{sec:appendix_stage}), information routing (§\ref{sec:appendix_info_converge}), and specialized attention heads (§\ref{appendix_heads}). Each results section pools all models, prompt orders, and hops; the model-by-model comparison summary is in Table~\ref{tab:cross_family_summary}.

\section{Dataset Details}
\label{sec:appendix_dataset}

\textit{PropLogic-MI} is released in two prompt-order variants ($\pi_{\mathrm{FF}}$ and $\pi_{\mathrm{EF}}$) generated from the same seed and the same underlying logical rows. Each file contains $20{,}000$ unique clean/corrupted pairs and ends with the answer suffix \texttt{"Answer with one word only: True or False."} Table~\ref{tab:rule_coverage} lists the eleven rule categories and Table~\ref{tab:reasoning_depth} illustrates the one-hop and two-hop prompt templates.

\begin{table*}[!htbp]
\centering
\small
\renewcommand{\arraystretch}{0.9}
\setlength{\tabcolsep}{6pt}
\begin{tabularx}{\textwidth}{@{} l c X @{}}
\toprule
\textbf{Category} & \textbf{Formal Definitions} & \textbf{Example Prompt Templates (One-Hop)} \\
\midrule
Identity
    & $P \land \top \equiv P, \quad P \lor \bot \equiv P$
    & \texttt{"A is T, A and T is"}, \quad \texttt{"A is F, A or F is"} \\
Domination
    & $P \land \bot \equiv \bot, \quad P \lor \top \equiv \top$
    & \texttt{"A is T, A and F is"}, \quad \texttt{"A is F, A or T is"} \\
Idempotent
    & $P \land P \equiv P, \quad P \lor P \equiv P$
    & \texttt{"A is T, A and A is"}, \quad \texttt{"A is F, A or A is"} \\
\midrule
Dbl. Negation
    & $\lnot(\lnot P) \equiv P$
    & \texttt{"A is T, ($\lnot$($\lnot$A)) is"} \\
Excluded Mid.
    & $P \lor \lnot P \equiv \top$
    & \texttt{"A is T, A or $\lnot$A is"} \\
Contradiction
    & $P \land \lnot P \equiv \bot$
    & \texttt{"A is F, A and $\lnot$A is"} \\
\midrule
Commutative
    & $P \land Q \equiv Q \land P$
    & \texttt{"A is T, B is F, A and B is"} \\
Associative
    & $(P \land Q) \land R \equiv P \land (Q \land R)$
    & \texttt{"A is T, B is T, C is F, (A and B) and C is"} \\
Distributive
    & $P \land (Q \lor R) \equiv (P \land Q) \lor (P \land R)$
    & \texttt{"A is T, B is F, C is T, A and (B or C) is"} \\
De Morgan
    & $\lnot(P \land Q) \equiv \lnot P \lor \lnot Q$
    & \texttt{"A is T, B is F, ($\lnot$A or $\lnot$B) is"} \\
Absorption
    & $P \land (P \lor Q) \equiv P$
    & \texttt{"A is T, B is F, A and (A or B) is"} \\
\bottomrule
\end{tabularx}
\caption{\textbf{Rule coverage.} The 11 rule categories in \textit{PropLogic-MI}, each with its formal Boolean-algebra definition and an illustrative one-hop template. Placeholder letters \{\texttt{A,B,C}\} are compact stand-ins; in the actual corpus, variables are randomly remapped to fresh uppercase symbols (excluding \texttt{T}/\texttt{F}).}
\label{tab:rule_coverage}
\end{table*}

\begin{table}[!htbp]
\centering
\small
\renewcommand{\arraystretch}{0.9}
\setlength{\tabcolsep}{4pt}
\begin{tabularx}{\columnwidth}{@{}lX@{}}
\toprule
\textbf{Depth} & \textbf{Prompt Template \& Reasoning Process} \\
\midrule
\multirow{3.5}{*}{\textbf{One-Hop}}
& \texttt{"A is True, B is False, (A and B) is"} \newline
  \color{gray}\textit{\footnotesize $\hookrightarrow$ Logic: $A \land B$} \\
\addlinespace[0.4em]
& \texttt{"A is True, ($\lnot$($\lnot$A)) is"} \newline
  \color{gray}\textit{\footnotesize $\hookrightarrow$ Logic: $\lnot(\lnot A) \equiv A$} \\
\addlinespace[0.4em]
& \texttt{"A is True, B is False, ($\lnot$A or $\lnot$B) is"} \newline
  \color{gray}\textit{\footnotesize $\hookrightarrow$ Logic: De Morgan ($\lnot A \lor \lnot B$)} \\
\midrule
\multirow{5}{*}{\textbf{Two-Hop}}
& \texttt{"A is T, B is A and T, C is F, B and C is"} \newline
  \color{gray}\textit{ \footnotesize $\hookrightarrow$ Chain: $(A \land \top) \to B; \quad (B \land C) \to \text{Ans}$} \\
\addlinespace[0.4em]
& \texttt{"A is T, B is F, C is $\lnot$(A and B), D is T, C or D is"} \newline
  \color{gray}\textit{\footnotesize $\hookrightarrow$ Chain: $\text{NAND}(A,B) \to C; \quad (C \lor D) \to \text{Ans}$} \\
\addlinespace[0.4em]
& \texttt{"A is T, B is F, C is A and B, D is T, C or D is"} \newline
  \color{gray}\textit{\footnotesize $\hookrightarrow$ Chain: $(A \land B) \to C; \quad (C \lor D) \to \text{Ans}$} \\
\bottomrule
\end{tabularx}
\caption{\textbf{Probing prompts across reasoning depths.} One-hop queries apply a rule directly; two-hop queries require an inline intermediate assignment $M = e_{1}$. Each logical row is realized under both $\pi_{\mathrm{FF}}$ and $\pi_{\mathrm{EF}}$.}
\label{tab:reasoning_depth}
\end{table}

\paragraph{Generation, corruption, and filtering.}
For one-hop rows, we instantiate a symbolic rule template, randomly remap the template variables to fresh uppercase symbols (excluding \texttt{F}/\texttt{T}), sample a Boolean assignment, and evaluate the resulting expression. For two-hop rows, we additionally introduce an intermediate variable $M = e_{1}$ and a gate variable $G$, with the final query being either $M \land G$ or $M \lor G$. The generation budget targets a 50/50 hop split, but the uniqueness constraint under our corruption semantics yields a final $9{,}523$ one-hop / $10{,}477$ two-hop allocation that is deterministic across the two prompt-order variants. Corruption performs an exact label-flipping search over the factual variables only ($19{,}808/20{,}000$ rows); the remaining $192$ one-hop rows whose expressions are constant under the available facts fall back to a label-flipping expression substitution. The raw dataset is not filtered by model behavior. Each mechanistic experiment uses the model-specific \emph{dual-correct} subset, namely rows that the model answers correctly on both the clean and corrupted runs (Table~\ref{tab:dataset_baselines}). All rows carry rule, hop, prompt-order, and corruption metadata so that subsets can be reproduced from the released JSONL.

\section{Methodology and Implementation Details}
\label{sec:appendix_setup}

\subsection{Hardware and Software}
\label{sec:appendix_hardware}
All experiments run on NVIDIA H100 GPUs with 96\,GB memory, using PyTorch~2.0 with CUDA~11.8. Causal probes are implemented with the TransformerLens library~\citep{nanda2022transformerlens}. All forward passes use \texttt{torch.no\_grad()} with \texttt{float16} weights and activations. A single residual-stream token-by-layer patching pass takes about $30$\,s on \textit{Qwen3-8B} ($36$ layers $\times {\approx}16$ token positions); a single head-by-layer mean-ablation pass takes about $2$\,min ($36$ layers $\times 32$ heads). The full paper-scale pipeline (all four models, both prompt orders, all interventions) takes roughly $50$ GPU-hours, excluding replotting from saved CSV/JSON/PKL artifacts.

\subsection{Probability-Difference Metric}
\label{sec:appendix_metric}
All three intervention branches score with the same probability-difference metric introduced in §\ref{sec:methodology}. For an answer label $y\in\{\texttt{True},\texttt{False}\}$ and a softmax restricted to those two answer tokens, we define
\begin{equation}
\mathrm{PD}(x) = p_{\theta}(\texttt{True}\mid x) - p_{\theta}(\texttt{False}\mid x),
\end{equation}
and its label-aligned form $\mathrm{PD}_{\mathrm{label}}(x,y) = \mathrm{PD}(x)$ if $y=\texttt{True}$, else $-\mathrm{PD}(x)$, so larger values always indicate stronger support for the correct answer. For an intervention producing $\tilde{x}$, the shift is
\begin{equation}
\mathrm{dPD} = \mathrm{PD}_{\mathrm{label}}(\tilde{x}, y) - \mathrm{PD}_{\mathrm{label}}(x_{\mathrm{baseline}}, y),
\end{equation}
where $x_{\mathrm{baseline}}$ is the corrupted prompt for activation patching and the clean prompt for mean ablation. Throughout the paper, we report this shift rather than raw logit differences.

\subsection{Layer Partition and Sample Budgets}
\label{sec:appendix_layer_partition}
For Early/Middle/Late summaries we use fixed token-level boundaries Early $=[0,13]$, Middle $=[14,23]$, Late $=[24,L-1]$ for the \textit{Qwen3} family and \textit{Mistral-7B} (with $L\in\{36,40,32\}$). For \textit{Llama-3.1-8B} ($L=32$), whose staged computation peaks earlier in depth, we use Early $=[0,8]$, Middle $=[9,15]$, Late $=[16,L-1]$ so that the labels remain aligned with the model's empirical phases. The logical-block analyses split the full depth into three approximately equal contiguous spans with \texttt{numpy.array\_split}. Sample budgets per setting are: 3000 rows for logical-block ablation, 1000 rows for token patching, 2000 rows for head discovery / classification, 1000 rows for the head-validation $\mathrm{dPD}$ curves, and 500 rows for the head-validation accuracy tables.

\subsection{Intervention Scopes}
\label{sec:appendix_interventions}

\paragraph{Logical-block ablation (MLP / Attention / Joint).}
For a logical block $R\in\{\text{Facts},\text{Expression},\text{Query}\}$ and layer $l$, the \textbf{MLP block ablation} runs the clean prompt and replaces \texttt{mlp\_out} for all positions $i\in R$ with the within-sequence mean MLP vector at that layer:
\begin{equation}
\tilde{m}^{(l)}_i = \frac{1}{T}\sum_{j=1}^{T} m^{(l)}_j, \qquad i \in R,
\end{equation}
where $T$ is the sequence length. The \textbf{attention block ablation} applies the same within-sequence mean replacement to \texttt{attn\_out} instead. The \textbf{joint attention+MLP block ablation} replaces both \texttt{attn\_out} and \texttt{mlp\_out} inside $R$ during one forward pass. All three branches reuse the same three-block parser (logical blocks are defined by character-span alignment and then mapped back to token indices) and the same $\mathrm{dPD}$ scoring.

\paragraph{Token-wise residual patching.}
For each clean/corrupted pair we patch \texttt{resid\_pre} one layer and one token at a time from the clean prompt into the corrupted prompt, recording the layer$\times$token matrix $\mathrm{dPD}_{l,i}$. Raw prompt positions are first aligned to model-input positions using tokenizer offset mappings; when offset-based alignment fails, the implementation falls back to a contiguous body-span mapping inside the model input. Tokens are then assigned to a prompt-order-sensitive taxonomy: \texttt{fact\_value}, \texttt{fact\_var\_ref}, \texttt{query\_last}, \texttt{operator}, \texttt{expr\_last}, \texttt{new\_var\_decl}, \texttt{new\_val\_ref}, \texttt{others}. The classifier is prompt-order-sensitive because the same lexical token plays different structural roles under $\pi_{\mathrm{FF}}$ vs.\ $\pi_{\mathrm{EF}}$ (\textit{e.g.}, the query token is the first \texttt{is} in $\pi_{\mathrm{EF}}$ and the last \texttt{is} in $\pi_{\mathrm{FF}}$). For each category $c$ and stage $s\in\{\text{Early},\text{Middle},\text{Late}\}$, the per-token causal density reported in figures is
\begin{equation}
M_{c}^{(s)} = \frac{1}{|c|}\sum_{i \in c} \frac{1}{|s|}\sum_{l \in s} |\mathrm{dPD}_{l,i}|,
\label{eq:per_token_mean}
\end{equation}
averaged across samples. The $|c|$ normalization controls for category size, so frequent token classes do not dominate merely because they occupy more positions.

\paragraph{Attention-head mean ablation and role classification.}
We replace the output $z_{l,h}$ of head $h$ at layer $l$ by its dataset mean computed over a balanced clean sample set, at all token positions. Heads are ranked by
\begin{equation}
\mathrm{Sel}(l,h) = -\mathbb{E}[\mathrm{dPD}_{\mathrm{label}}(l,h)],
\label{eq:head_selectivity}
\end{equation}
so larger values correspond to heads whose ablation most strongly reduces correct-answer support. We first run a small probe over all heads, keep the top candidates globally together with the top heads from each layer, and then re-evaluate the candidate pool on a larger balanced set.

Retained heads are classified into Splitting (SP), Transmission (TR), and Fact-Retrieval (FR) using six attention-pattern statistics measured on clean prompts: fact-focus ratio $f$, boundary-focus ratio $b$, intra-clause ratio $i$, attention entropy $e$, top-$1$ mass $t$, and split-half stability $s$. After feature standardization, each head receives a rule-based score for each role,
\begin{equation}
\begin{split}
\mathrm{Score}_{\mathrm{FR}}
&= f - 0.20\,i - 0.15\,b + 0.20\,t \\
&\quad - 0.10\,e + 0.08\,s,
\end{split}
\end{equation}
\begin{equation}
\begin{split}
\mathrm{Score}_{\mathrm{SP}}
&= b - 0.15\,i - 0.10\,f + 0.15\,t \\
&\quad - 0.08\,e + 0.10\,s,
\end{split}
\end{equation}
\begin{equation}
\begin{split}
\mathrm{Score}_{\mathrm{TR}}
&= i - 0.20\,b - 0.10\,f + 0.18\,t \\
&\quad - 0.08\,e + 0.12\,s,
\end{split}
\end{equation}
and the winning label is assigned. The Splitting / Transmission / Fact-Retrieval taxonomy used throughout the paper is the output of this classifier.

For validation, we mean-ablate the top-$k$ heads of each role on held-out clean prompts and report both post-ablation accuracy and the mean $\mathrm{dPD}$ shift. The \textit{Random-Outside} control samples heads from layers used in the validation stage but outside the selected head pool; \textit{Mixed-All} ablates the union of the top heads from all three roles.

\subsection{Reproducibility}
\label{sec:appendix_reproducibility}
Dataset construction is handled by \texttt{src.data.generate\_dataset}; inference, dual-correct filtering, and metrics by \texttt{src.eval}; block ablations by \texttt{src.mlp\_analysis}, \texttt{src.attn\_analysis}, and \texttt{src.attn\_mlp\_analysis}; token patching by \texttt{src.token\_analysis}; head discovery and validation by \texttt{src.heads\_analysis}. Wrapper scripts in \texttt{scripts/} run the per-model pipelines, while \texttt{scripts/replot\_all\_figures.sh} regenerates figures from saved report files. All dataset generation and sampling entry points expose a \texttt{--seed} argument with default $42$. Generated datasets, predictions, filtered pools, reports, figures, and logs are written to ignored output directories. Reproducing the \textit{Llama-3.1} runs requires the usual Hugging Face access to the gated model weights.

\section{Behavioral Performance on PropLogic-MI}
\label{sec:appendix_performance}

Table~\ref{tab:appendix_performance_summary} reports overall clean and corrupted accuracy, dual-correct rate, and hop-wise clean accuracy for the two \textit{Qwen3} models that anchor the main text. Table~\ref{tab:appendix_rule_performance} breaks down clean accuracy by rule. The same statistics for \textit{Llama-3.1-8B-Instruct} and \textit{Mistral-7B-Instruct} are reported alongside their mechanistic profile in Table~\ref{tab:cross_family_summary}.

\begin{table*}[!htbp]
\centering
\small
\renewcommand{\arraystretch}{0.95}
\setlength{\tabcolsep}{6pt}
\begin{tabularx}{\textwidth}{@{} l l c c X c c @{} }
\toprule
\textbf{Model} & \textbf{Prompt Order} & \textbf{Clean} & \textbf{Corrupted} & \textbf{Dual-Correct Rate} & \textbf{One-Hop} & \textbf{Two-Hop} \\
\midrule
\textit{Qwen3-8B} & $\pi_{\mathrm{FF}}$ & 86.98 & 88.17 & 75.44\% (15087/20000) & 86.35 & 87.55 \\
\textit{Qwen3-14B} & $\pi_{\mathrm{FF}}$ & 87.42 & 89.11 & 76.55\% (15310/20000) & 92.64 & 82.68 \\
\textit{Qwen3-8B} & $\pi_{\mathrm{EF}}$ & 74.22 & 79.25 & 53.89\% (10777/20000) & 79.09 & 69.79 \\
\textit{Qwen3-14B} & $\pi_{\mathrm{EF}}$ & 74.39 & 77.91 & 52.80\% (10561/20000) & 88.34 & 61.72 \\
\bottomrule
\end{tabularx}
\caption{\textbf{Overall zero-shot performance of Qwen3 on \textit{PropLogic-MI}.} All values except the dual-correct column are accuracies (\%). The dual-correct subset defines the candidate pool for the mechanistic analyses.}
\label{tab:appendix_performance_summary}
\end{table*}

\begin{table}[!htbp]
\centering
\small
\renewcommand{\arraystretch}{0.95}
\setlength{\tabcolsep}{5pt}
\begin{tabularx}{\columnwidth}{@{} X c c c c @{} }
\toprule
\textbf{Rule} & \textbf{8B FF} & \textbf{14B FF} & \textbf{8B EF} & \textbf{14B EF} \\
\midrule
Absorption & 84.8 & 95.0 & 84.4 & 85.4 \\
Associative & 92.1 & 99.0 & 61.8 & 76.0 \\
Commutative & 99.8 & 98.1 & 92.0 & 94.4 \\
Contradiction & 64.0 & 54.1 & 49.4 & 48.1 \\
De Morgan & 80.4 & 100.0 & 80.9 & 100.0 \\
Distributive & 72.8 & 71.5 & 76.8 & 88.8 \\
Domination & 95.9 & 88.6 & 62.5 & 50.3 \\
Double Negation & 84.8 & 87.1 & 72.7 & 70.3 \\
Excluded Middle & 98.7 & 86.1 & 72.9 & 51.4 \\
Idempotent & 92.1 & 85.6 & 83.5 & 76.5 \\
Identity & 91.4 & 96.7 & 79.4 & 77.2 \\
\bottomrule
\end{tabularx}
\caption{\textbf{Rule-wise clean accuracy (\%) of Qwen3 on \textit{PropLogic-MI}.} Prompt order has a large impact on several rules (Associative, Domination, Excluded Middle), while Commutative remains consistently easy and Contradiction consistently difficult.}
\label{tab:appendix_rule_performance}
\end{table}

\begin{table*}[!htbp]
\centering
\small
\renewcommand{\arraystretch}{1.05}
\setlength{\tabcolsep}{4pt}
\begin{tabularx}{\textwidth}{@{} l c c X X @{}}
\toprule
\textbf{Model} & \textbf{FF Clean/Dual} & \textbf{EF Clean/Dual} & \textbf{Token-level profile} & \textbf{Head-ablation regime} \\
\midrule
\textit{Qwen3-8B} & 87.0 / 75.4 & 74.2 / 53.9 & Most fact-centered at 8B scale: \texttt{fact\_value} competitive with or above \texttt{query\_last} in the per-token mean view. & Cleanly selective: role-targeted ablations are strongly harmful while \textit{Random-Outside} stays near baseline through moderate $k$. \\
\textit{Qwen3-14B} & 87.4 / 76.6 & 74.4 / 52.8 & Preserves the \textit{Qwen3} fact-centered profile with cleaner middle-layer binding-flow exchange under $\pi_{\mathrm{EF}}$ two-hop. & Cleanly selective; the strongest Fact-Retrieval + Transmission coupling under $\pi_{\mathrm{EF}}$ two-hop. \\
\textit{Llama-3.1-8B} & 63.9 / 38.4 & 60.3 / 37.0 & More query-centered and diffuse; \texttt{query\_last} dominates \texttt{fact\_value} under both prompt orders. & Weakly selective: role labels recoverable, but role-targeted ablations collapse to a common accuracy floor by moderate $k$. \\
\textit{Mistral-7B} & 65.1 / 35.1 & 66.7 / 40.0 & Intermediate between \textit{Qwen3} and \textit{Llama}; less fact-centered than \textit{Qwen3} but not as uniformly query-dominant as \textit{Llama}. & Globally sensitive: role-targeted ablations are severe, but \textit{Random-Outside} is also substantially damaging. \\
\bottomrule
\end{tabularx}
\caption{\textbf{Cross-family summary.} Clean/Dual columns report clean accuracy and dual-correct rate (\%) under $\pi_{\mathrm{FF}}$ (FF) and $\pi_{\mathrm{EF}}$ (EF). The mechanistic columns are detailed in §\ref{sec:appendix_stage}--§\ref{appendix_heads}.}
\label{tab:cross_family_summary}
\end{table*}

\section{Supplementary Results: Staged Computation}
\label{sec:appendix_stage}

This section supplements §\ref{sec:staged} along the three ablation pathways used in the body. Within each pathway we show all available model, prompt-order, and hop conditions so that the staged template can be read off the same pathway across settings, rather than per family.

\subsection{MLP-Block Ablation}
\label{sec:appendix_stage_mlp}

\begin{figure*}[!htbp]
    \centering
    \includegraphics[width=0.32\textwidth]{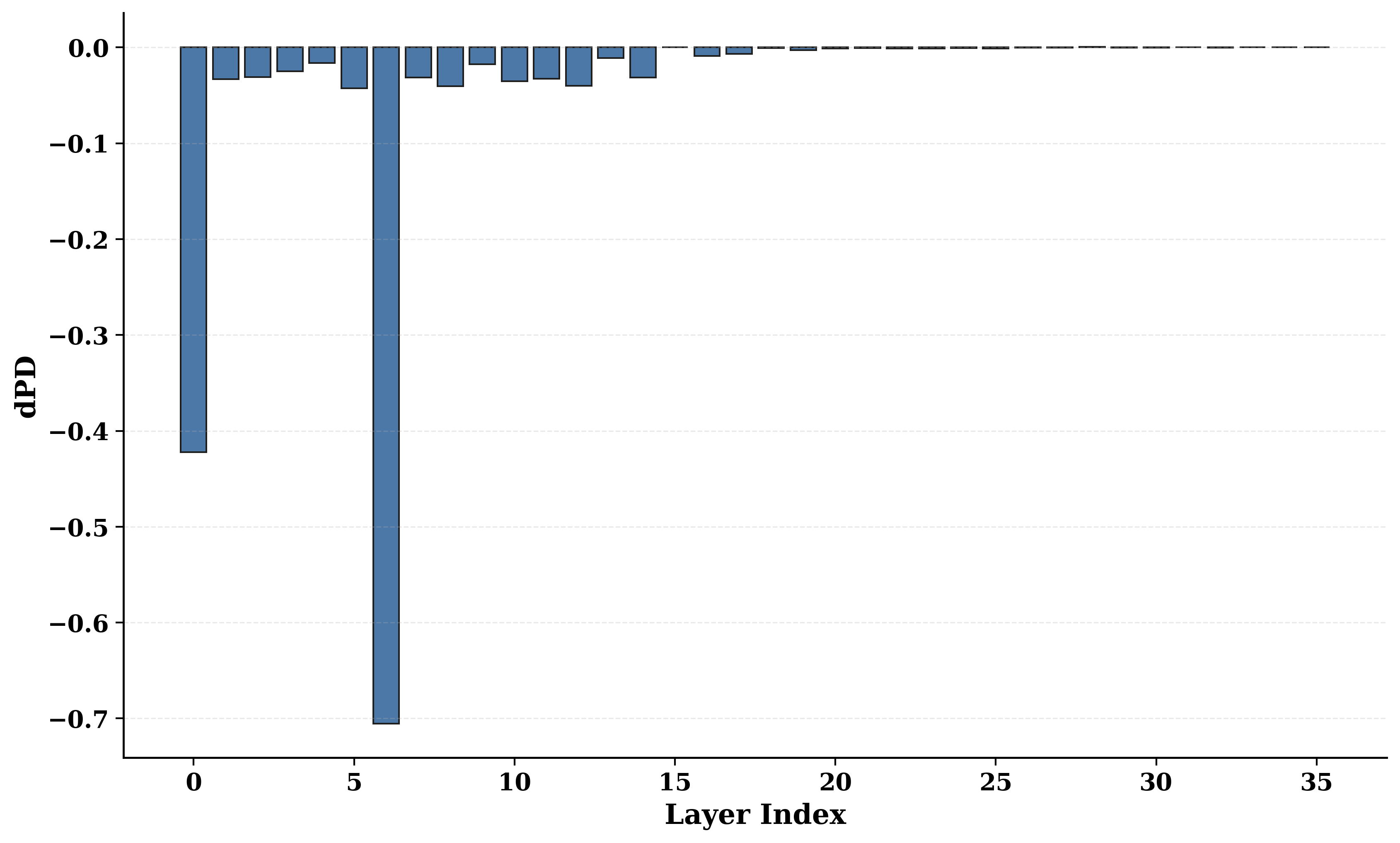}\hfill
    \includegraphics[width=0.32\textwidth]{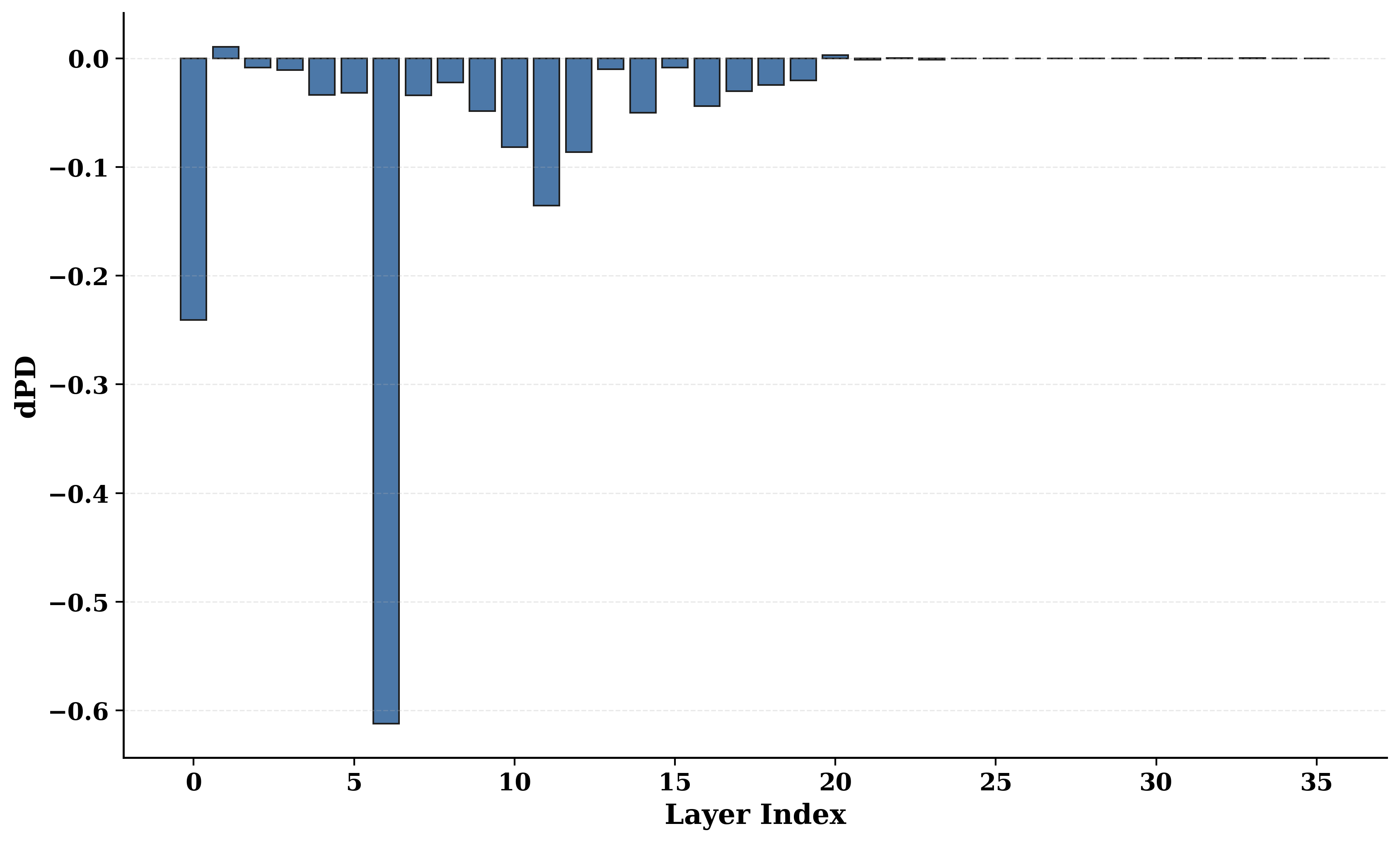}\hfill
    \includegraphics[width=0.32\textwidth]{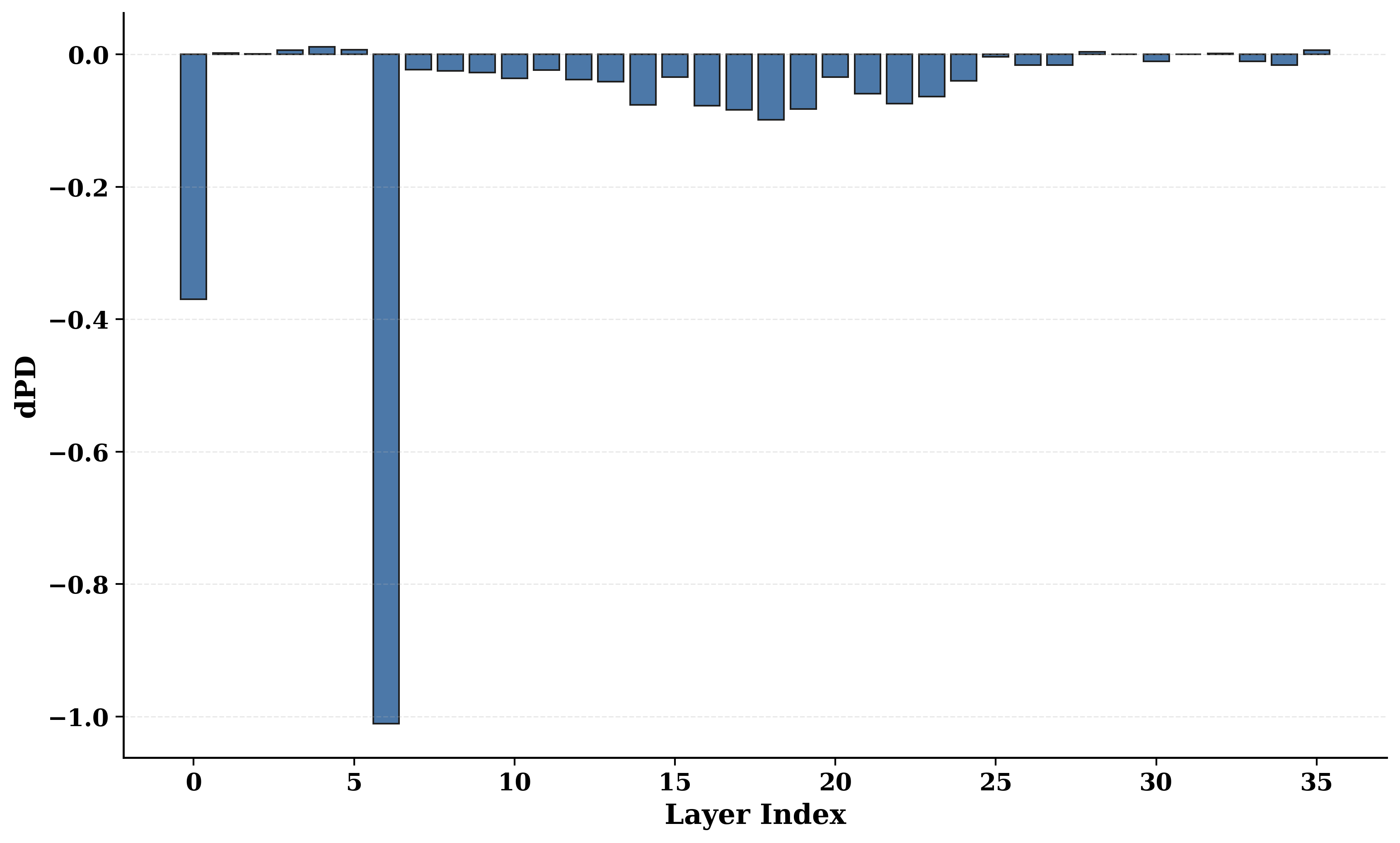}

    \includegraphics[width=0.32\textwidth]{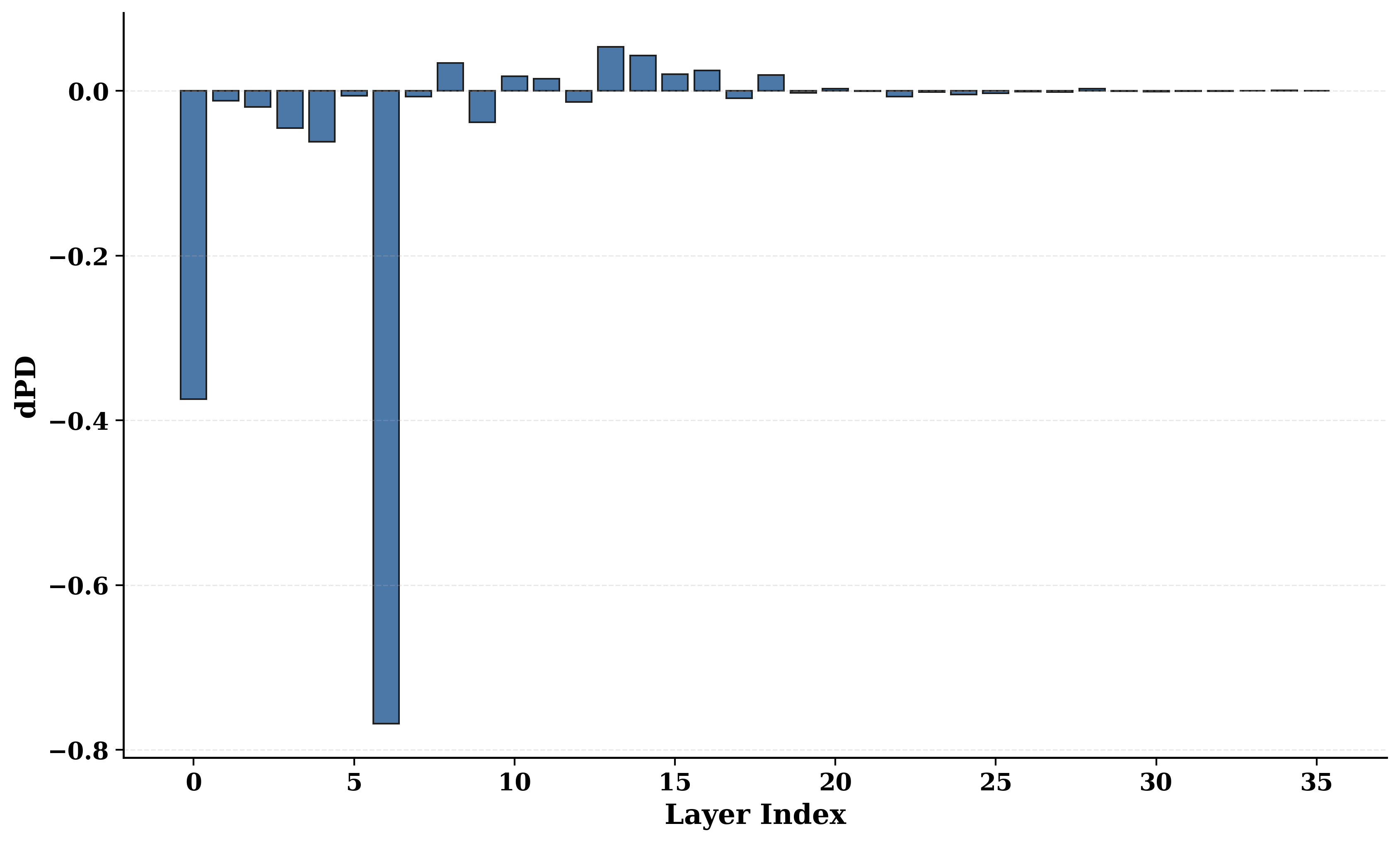}\hfill
    \includegraphics[width=0.32\textwidth]{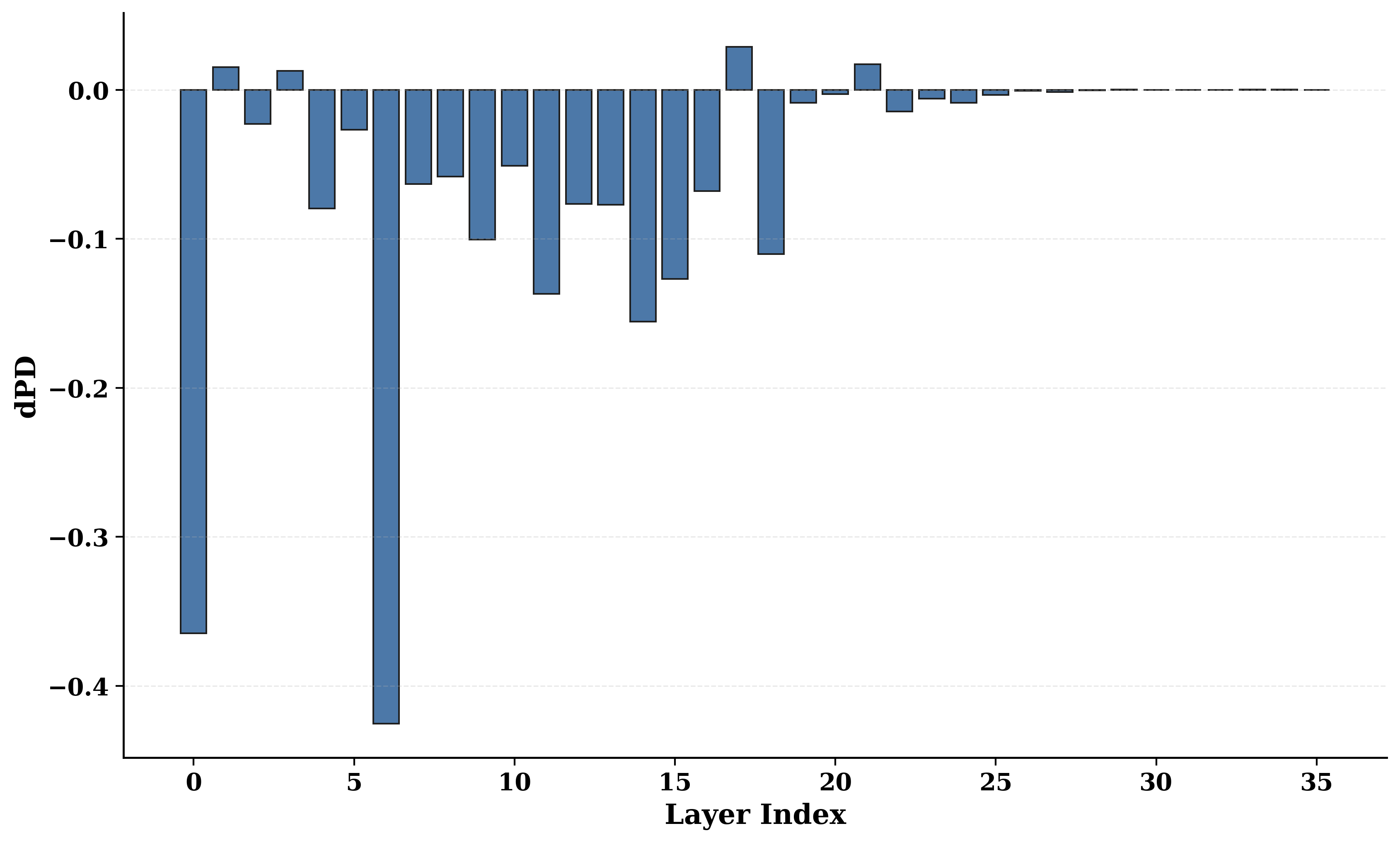}\hfill
    \includegraphics[width=0.32\textwidth]{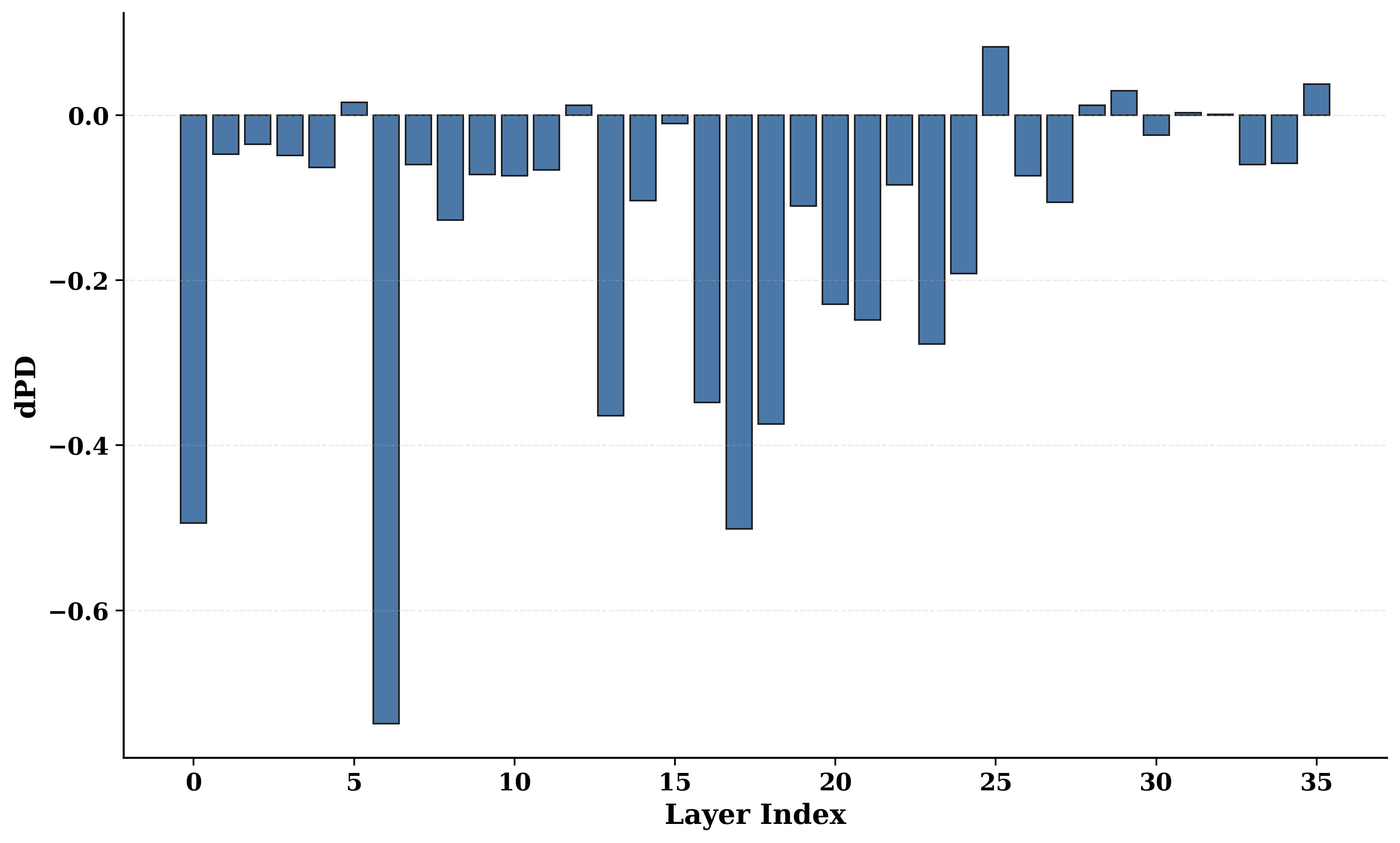}
    \caption{\textbf{MLP-block mean ablation on \textit{Qwen3-8B} under $\pi_{\mathrm{FF}}$.} Per-layer $\mathrm{dPD}$ when each logical block is mean-ablated. Top row: one-hop. Bottom row: two-hop. Columns: Facts / Expression / Query.}
    \label{fig:stage_mlp_q8_ff}
\end{figure*}

\begin{figure*}[!htbp]
    \centering
    \includegraphics[width=0.32\textwidth]{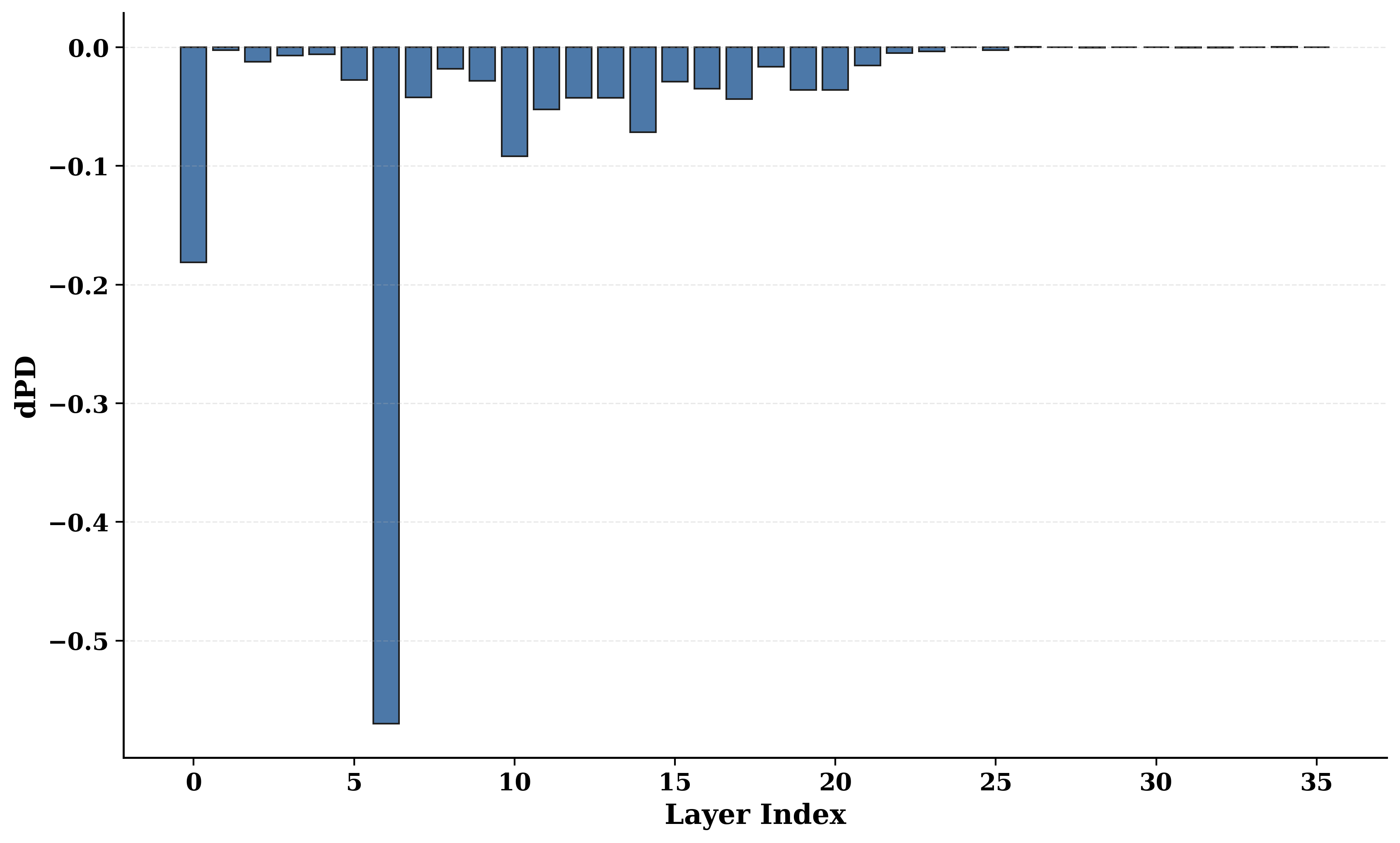}\hfill
    \includegraphics[width=0.32\textwidth]{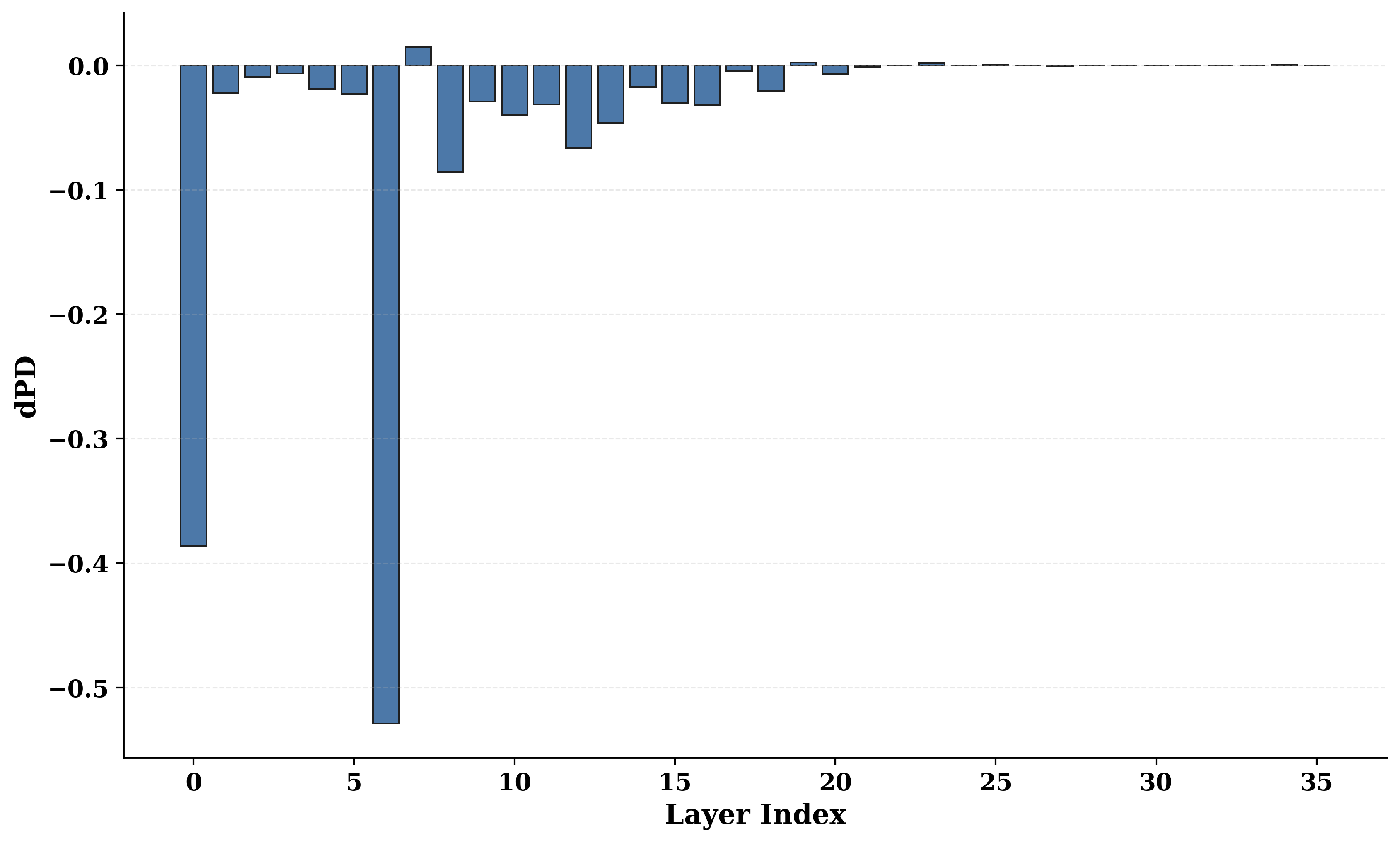}\hfill
    \includegraphics[width=0.32\textwidth]{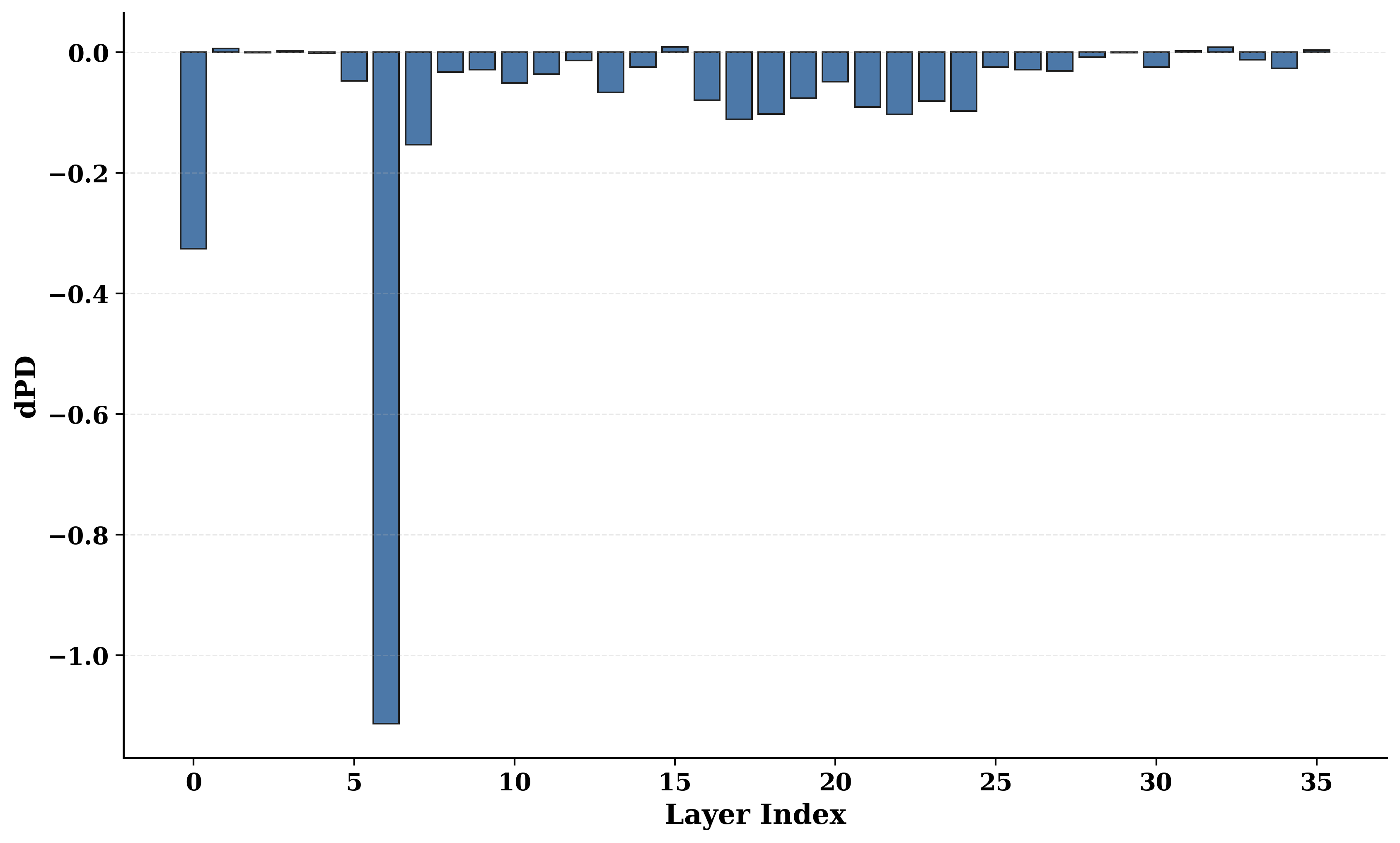}

    \includegraphics[width=0.32\textwidth]{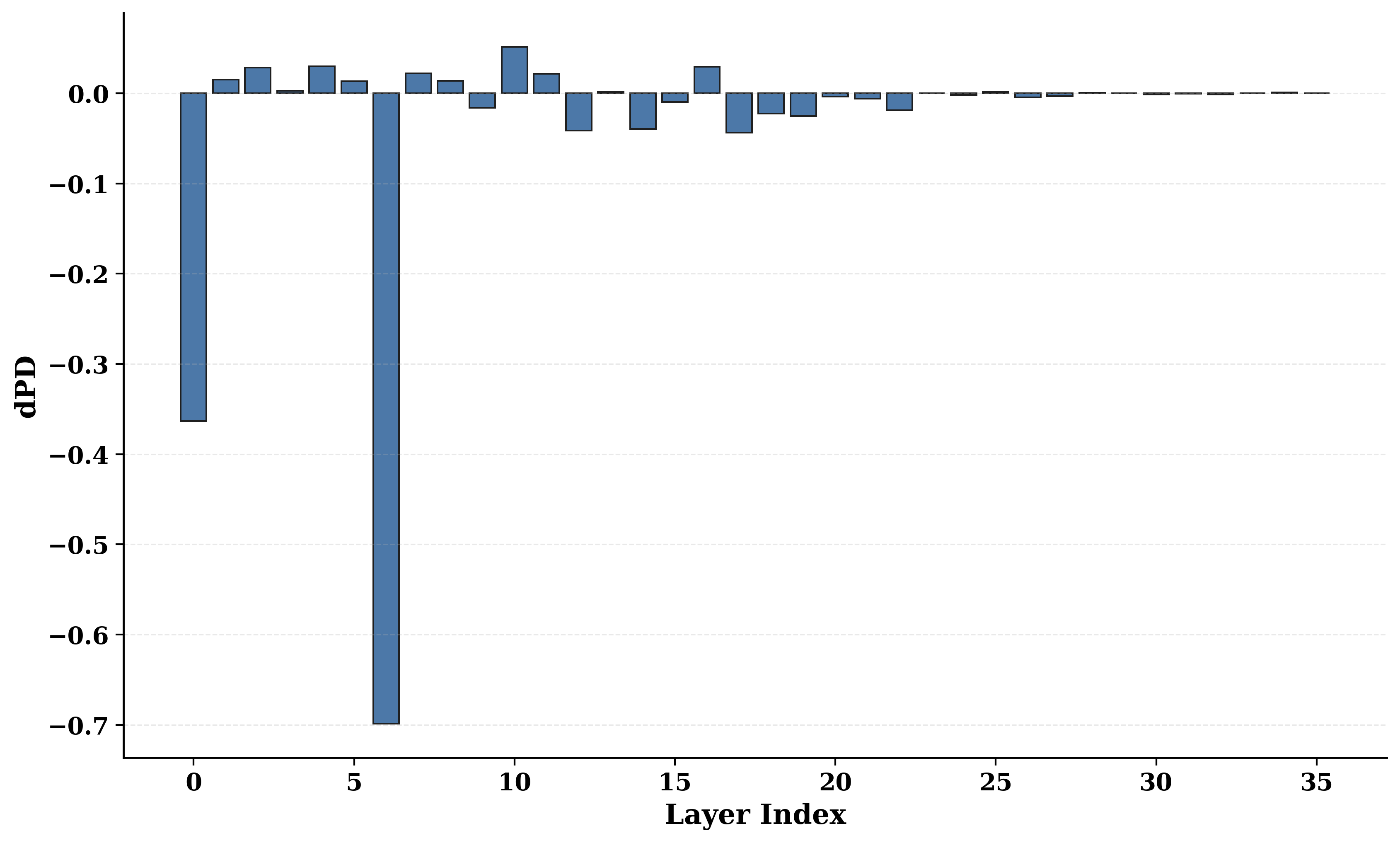}\hfill
    \includegraphics[width=0.32\textwidth]{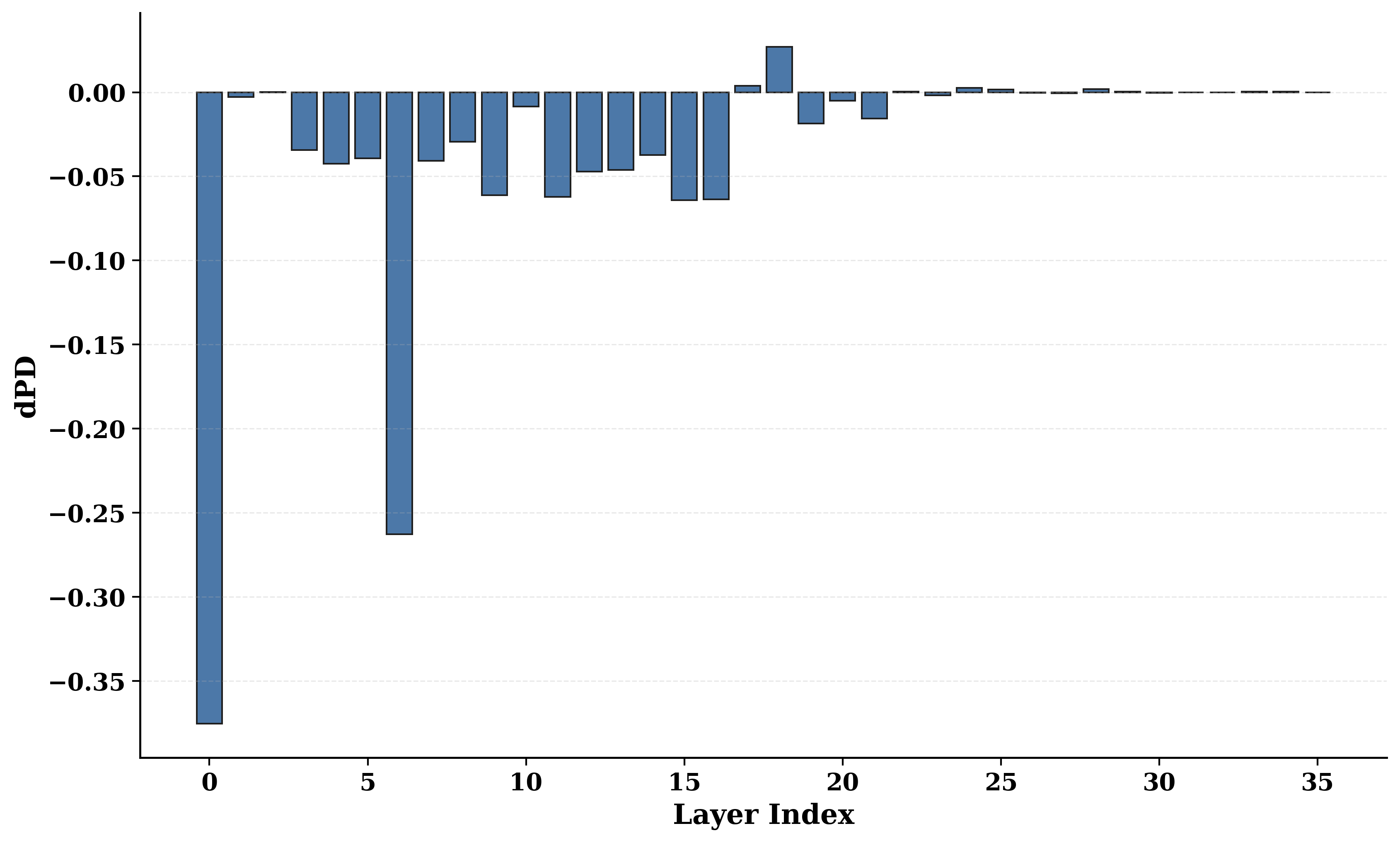}\hfill
    \includegraphics[width=0.32\textwidth]{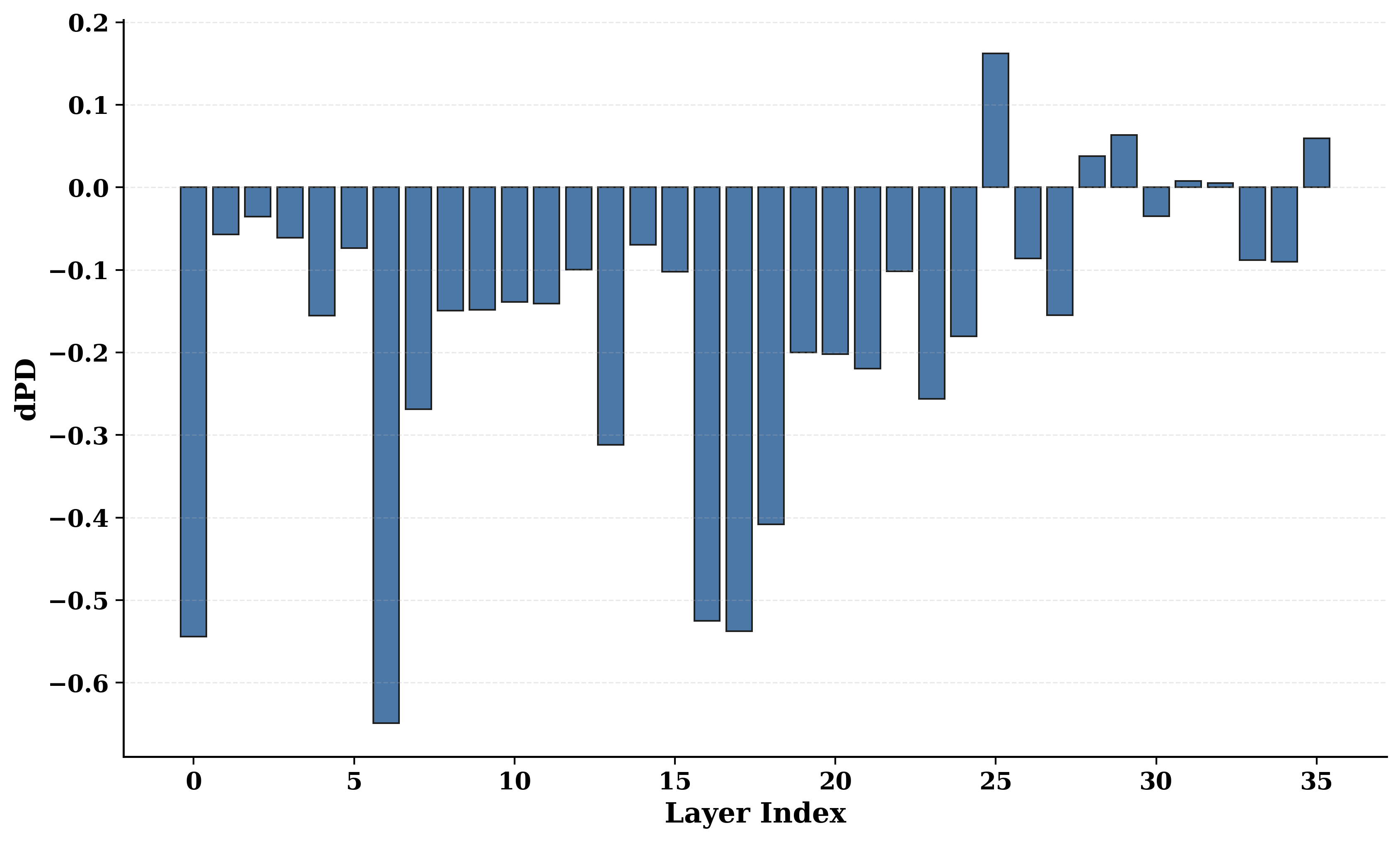}
    \caption{\textbf{MLP-block mean ablation on \textit{Qwen3-8B} under $\pi_{\mathrm{EF}}$.} Per-layer $\mathrm{dPD}$ when each logical block is mean-ablated. Top row: one-hop. Bottom row: two-hop. Columns: Facts / Expression / Query.}
    \label{fig:stage_mlp_q8_ef}
\end{figure*}

\begin{figure*}[!htbp]
    \centering
    \includegraphics[width=0.32\textwidth]{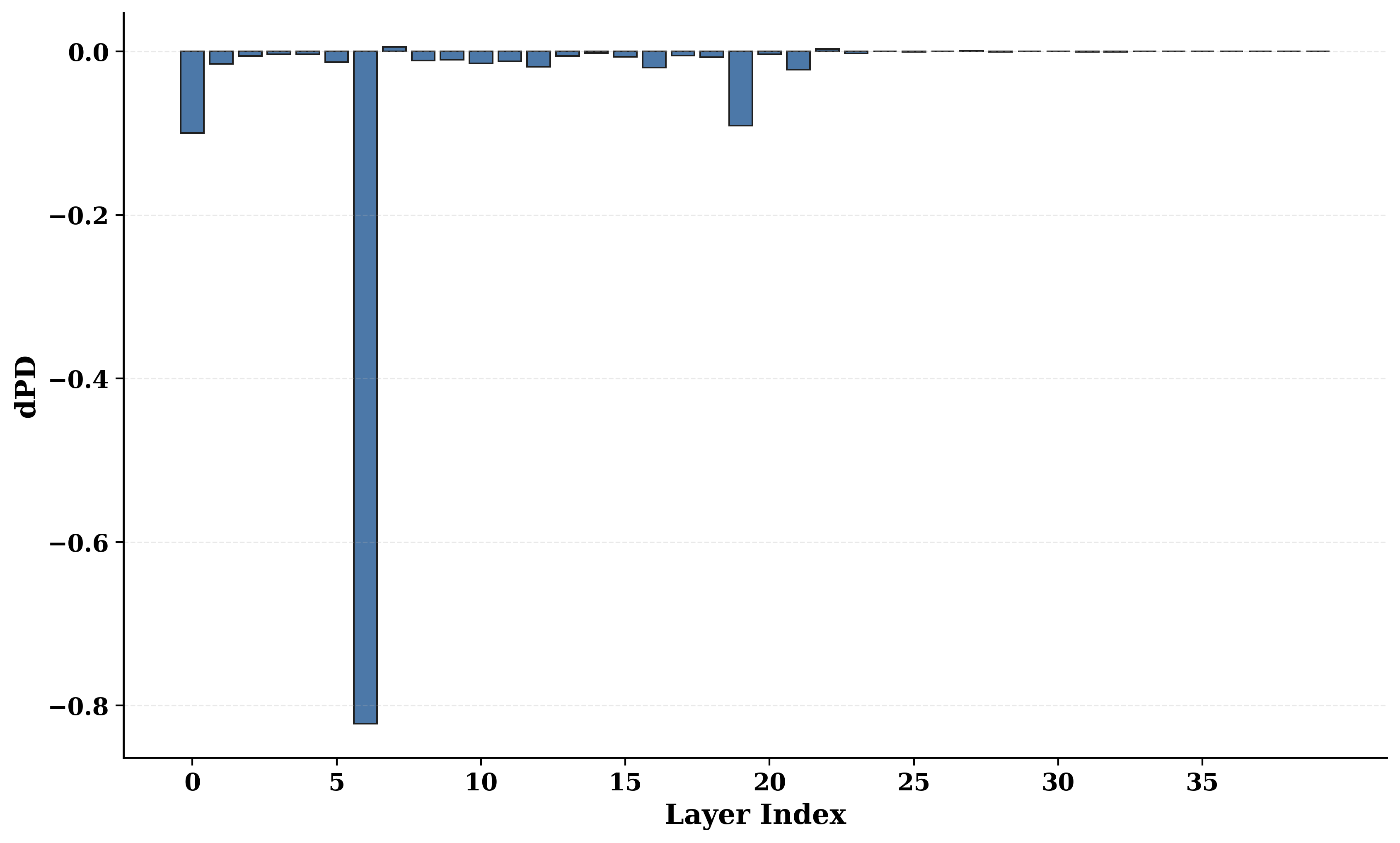}\hfill
    \includegraphics[width=0.32\textwidth]{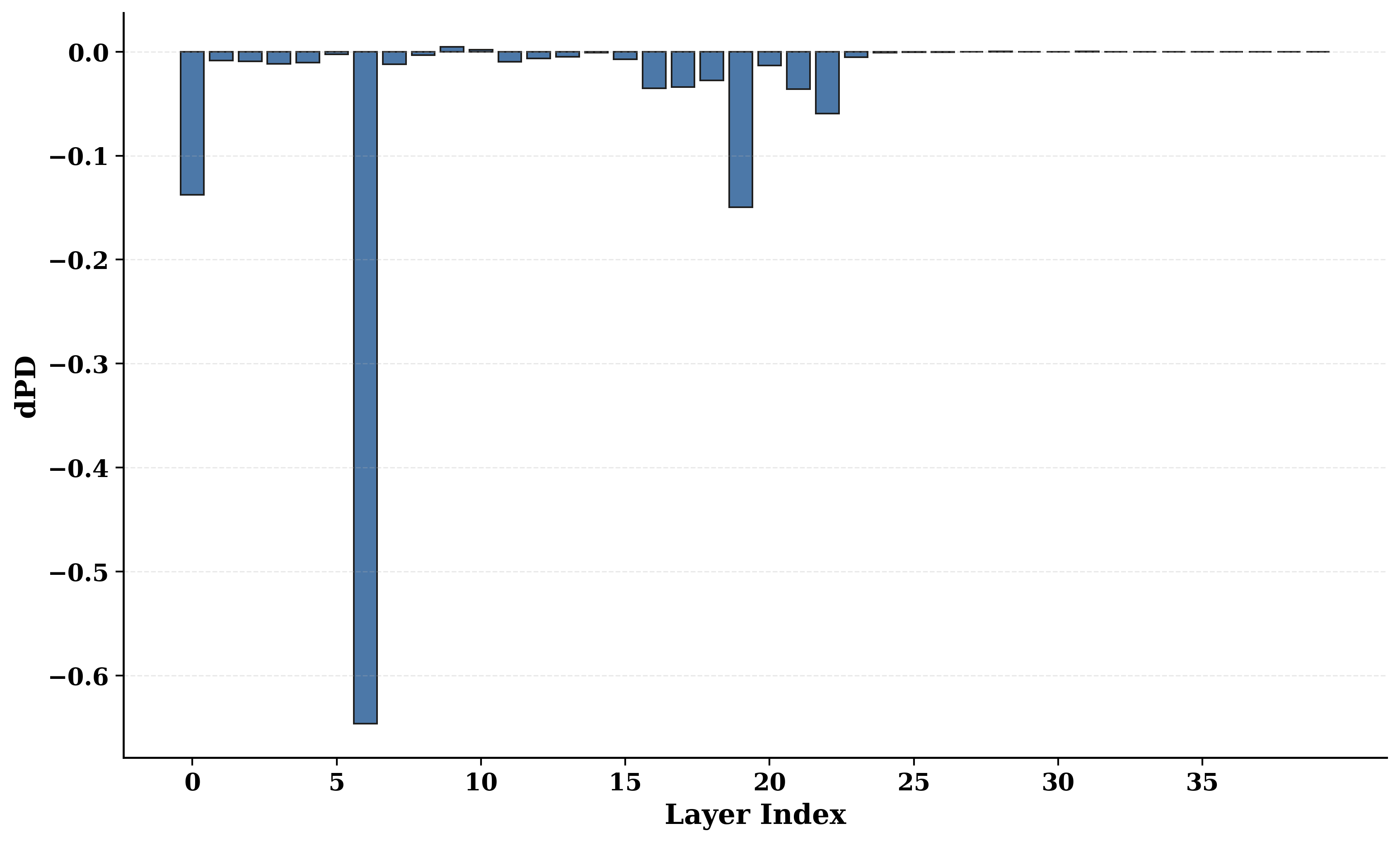}\hfill
    \includegraphics[width=0.32\textwidth]{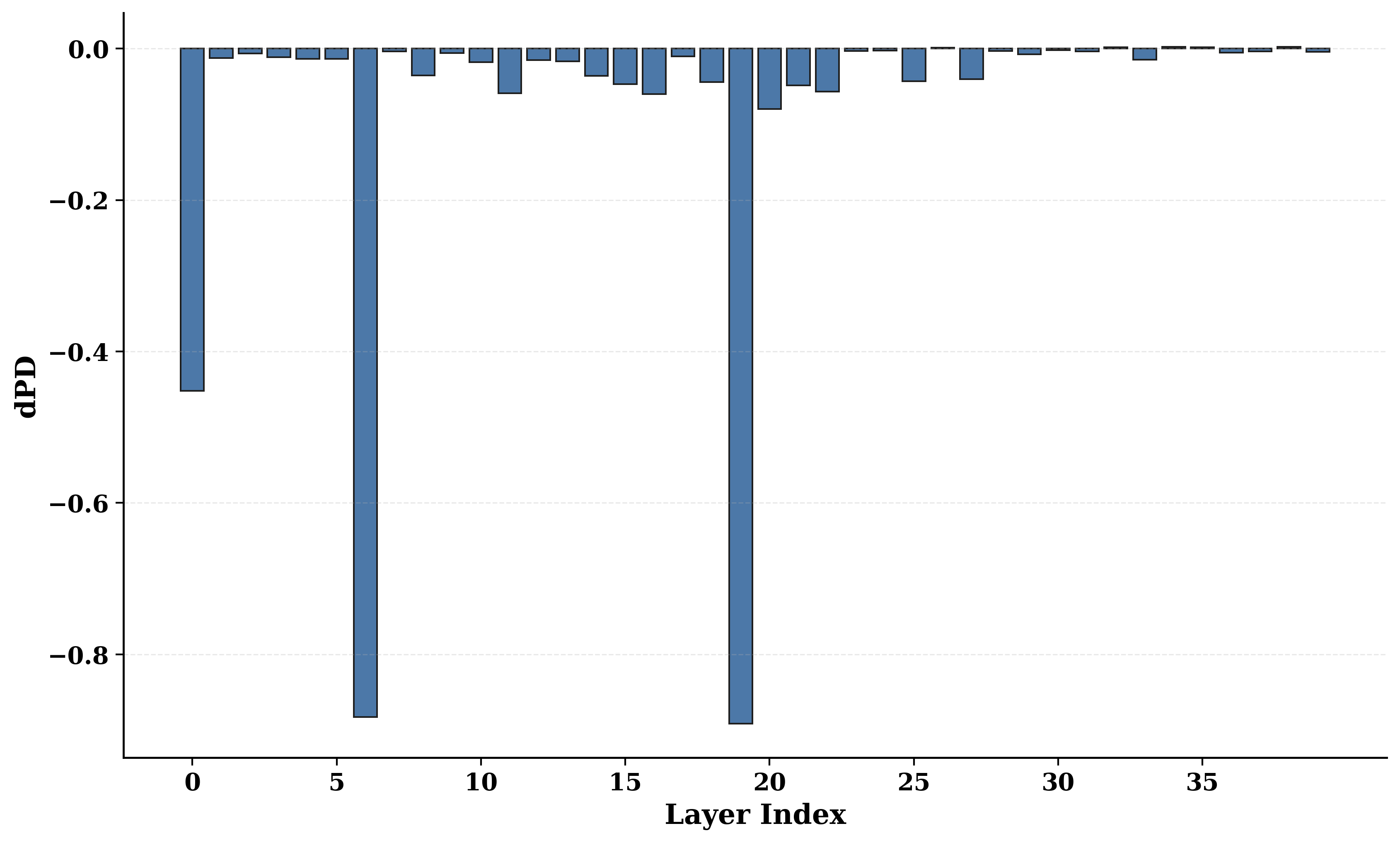}

    \includegraphics[width=0.32\textwidth]{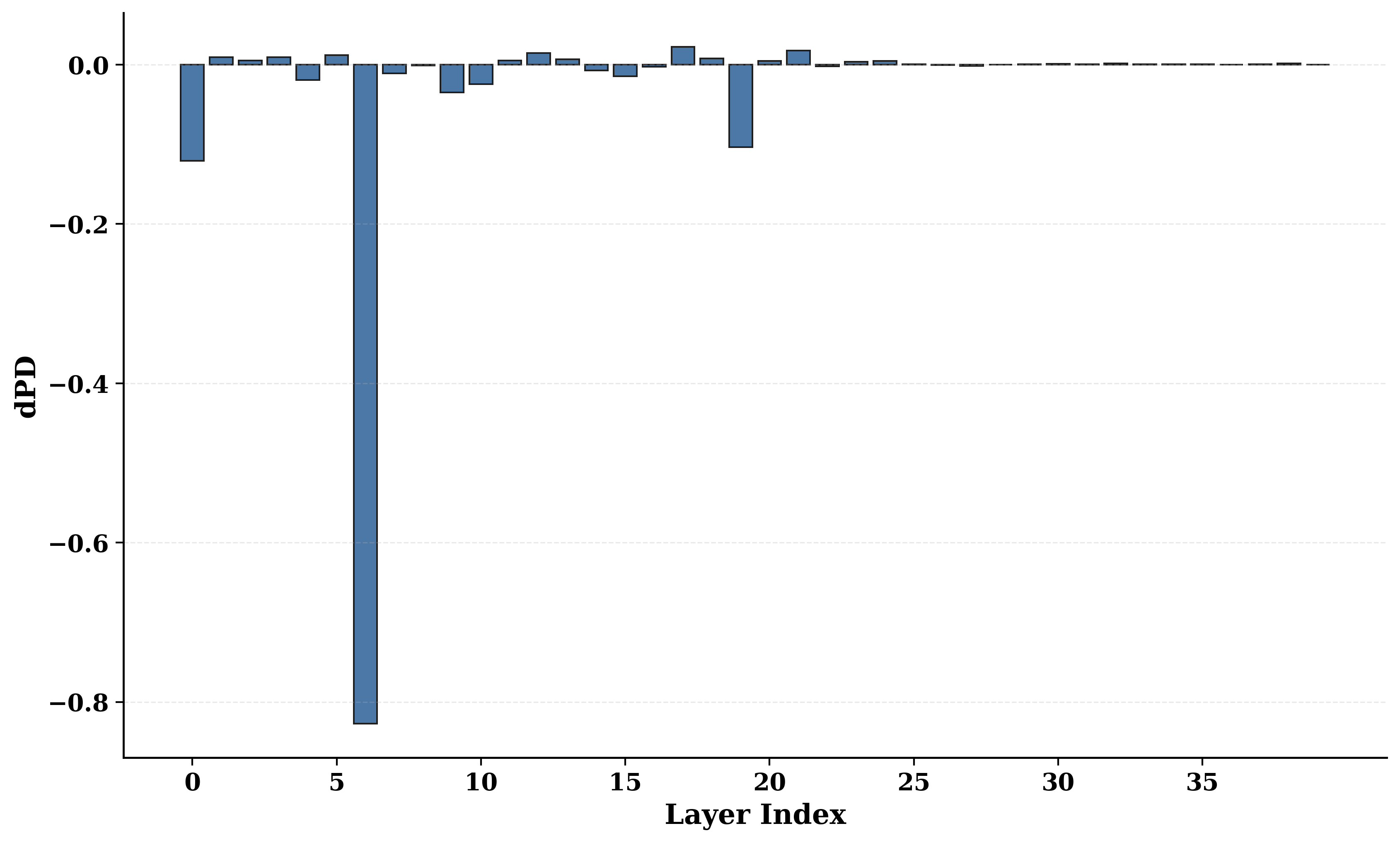}\hfill
    \includegraphics[width=0.32\textwidth]{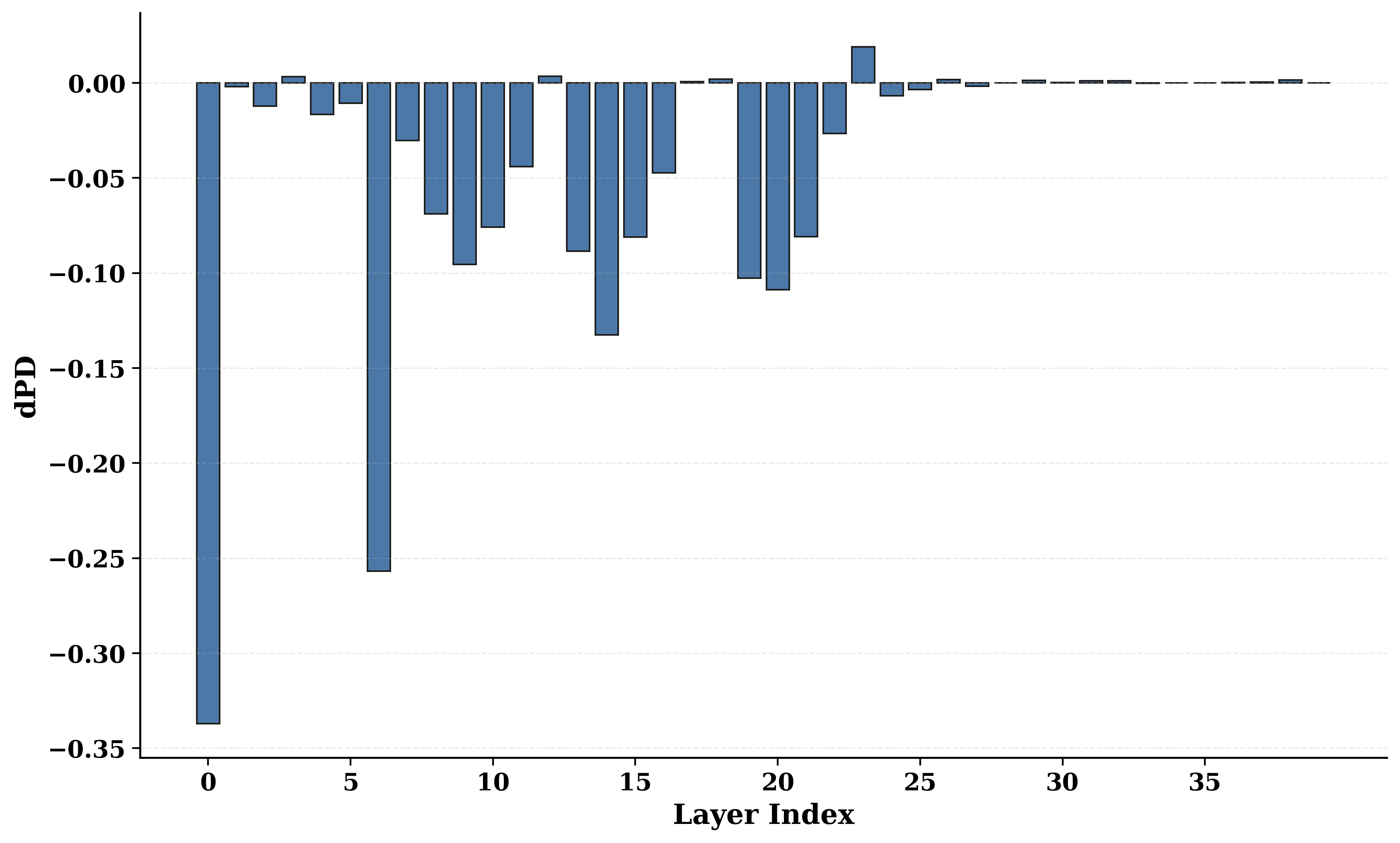}\hfill
    \includegraphics[width=0.32\textwidth]{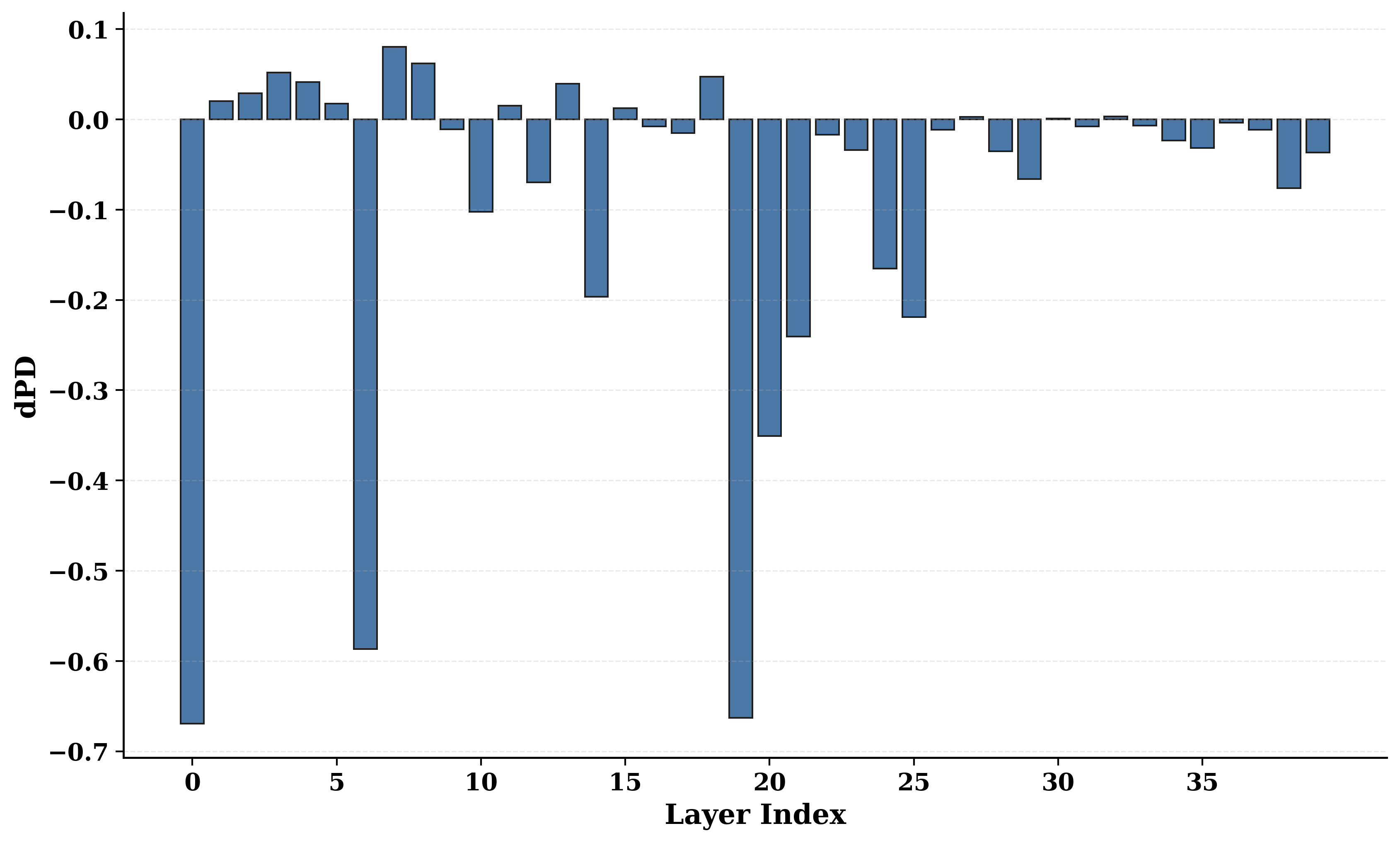}
    \caption{\textbf{MLP-block mean ablation on \textit{Qwen3-14B} under $\pi_{\mathrm{FF}}$.} Per-layer $\mathrm{dPD}$ when each logical block is mean-ablated. Top row: one-hop. Bottom row: two-hop. Columns: Facts / Expression / Query.}
    \label{fig:stage_mlp_q14_ff}
    \label{fig:staged_computation_14B}
\end{figure*}

\begin{figure*}[!htbp]
    \centering
    \includegraphics[width=0.32\textwidth]{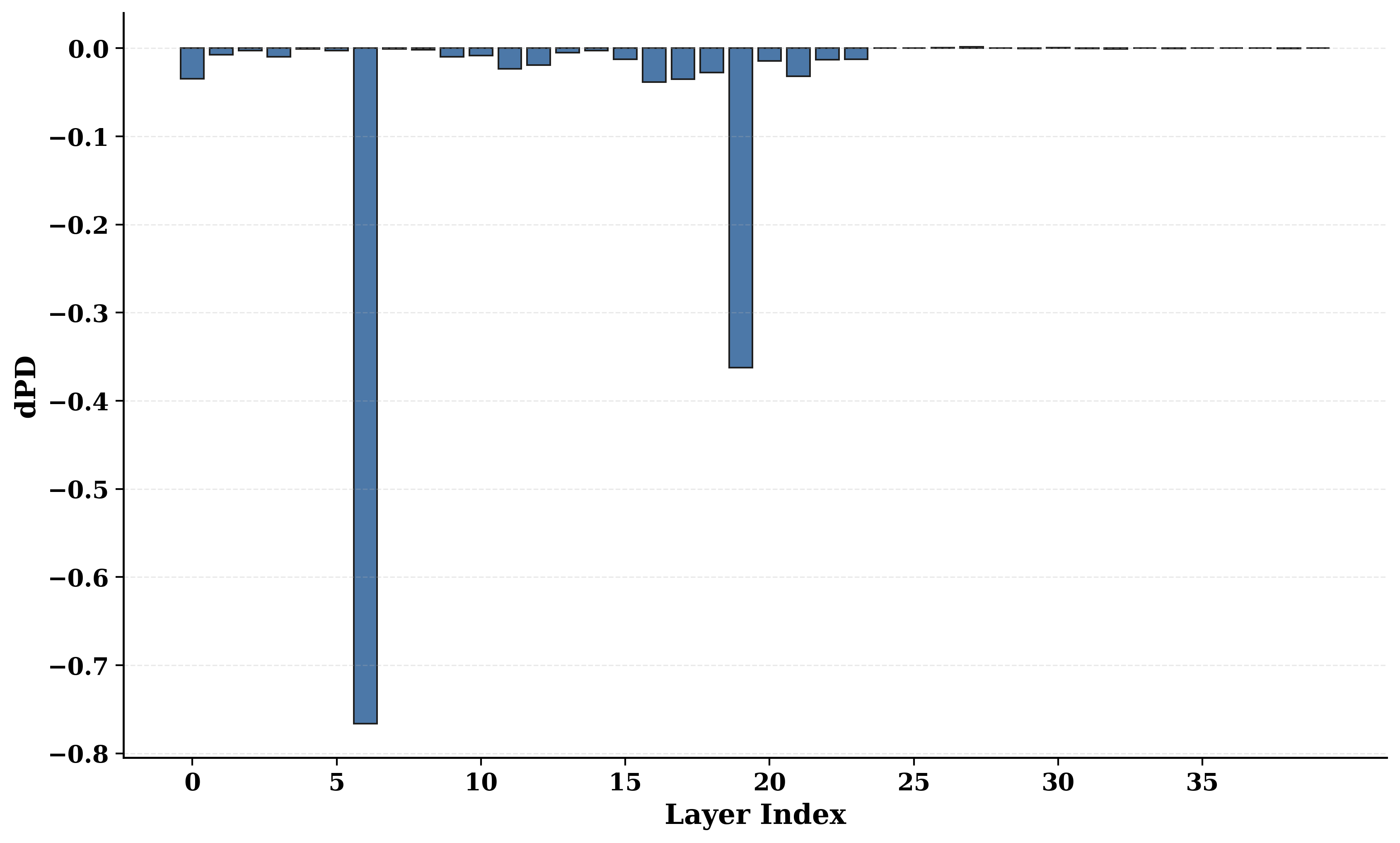}\hfill
    \includegraphics[width=0.32\textwidth]{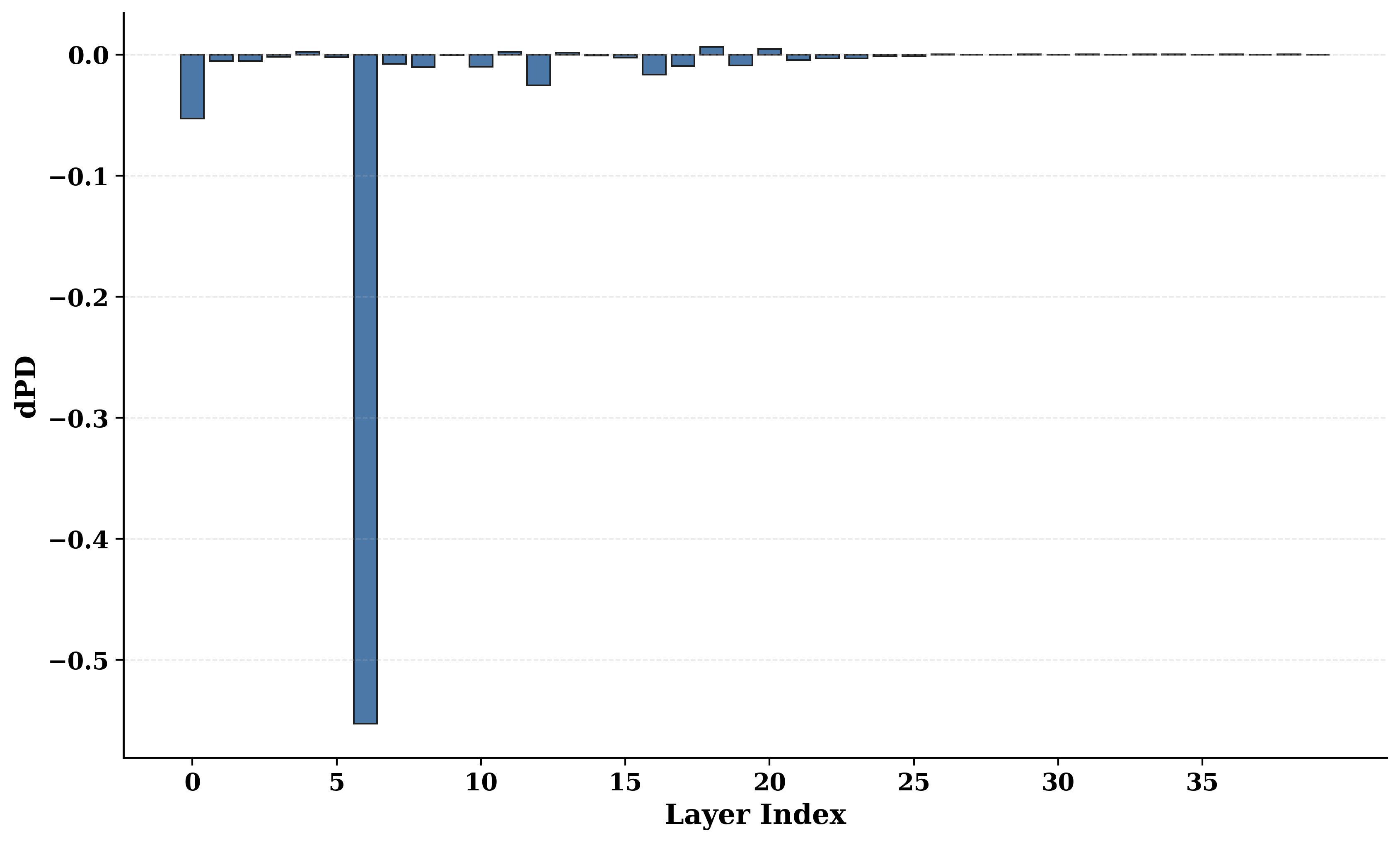}\hfill
    \includegraphics[width=0.32\textwidth]{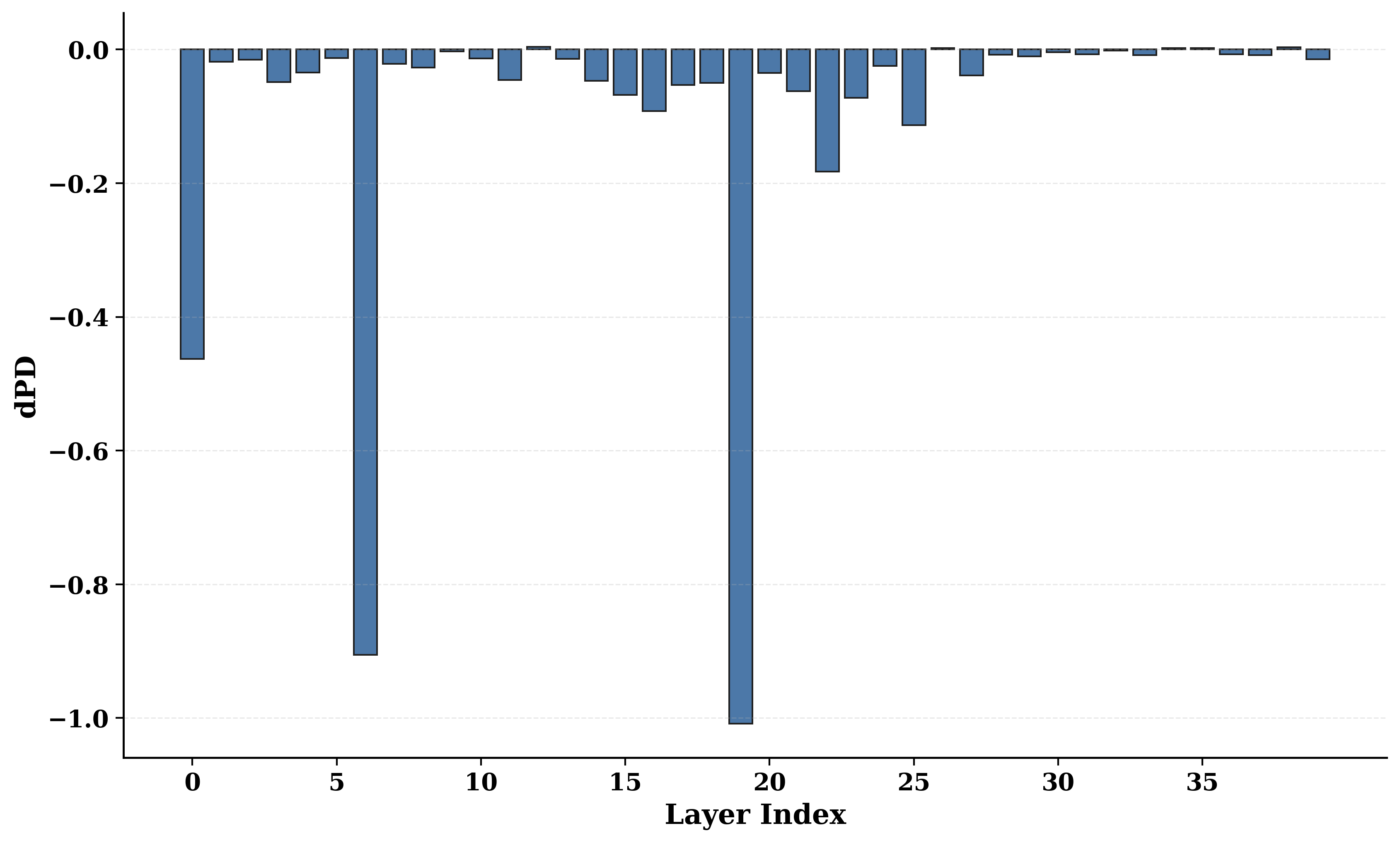}

    \includegraphics[width=0.32\textwidth]{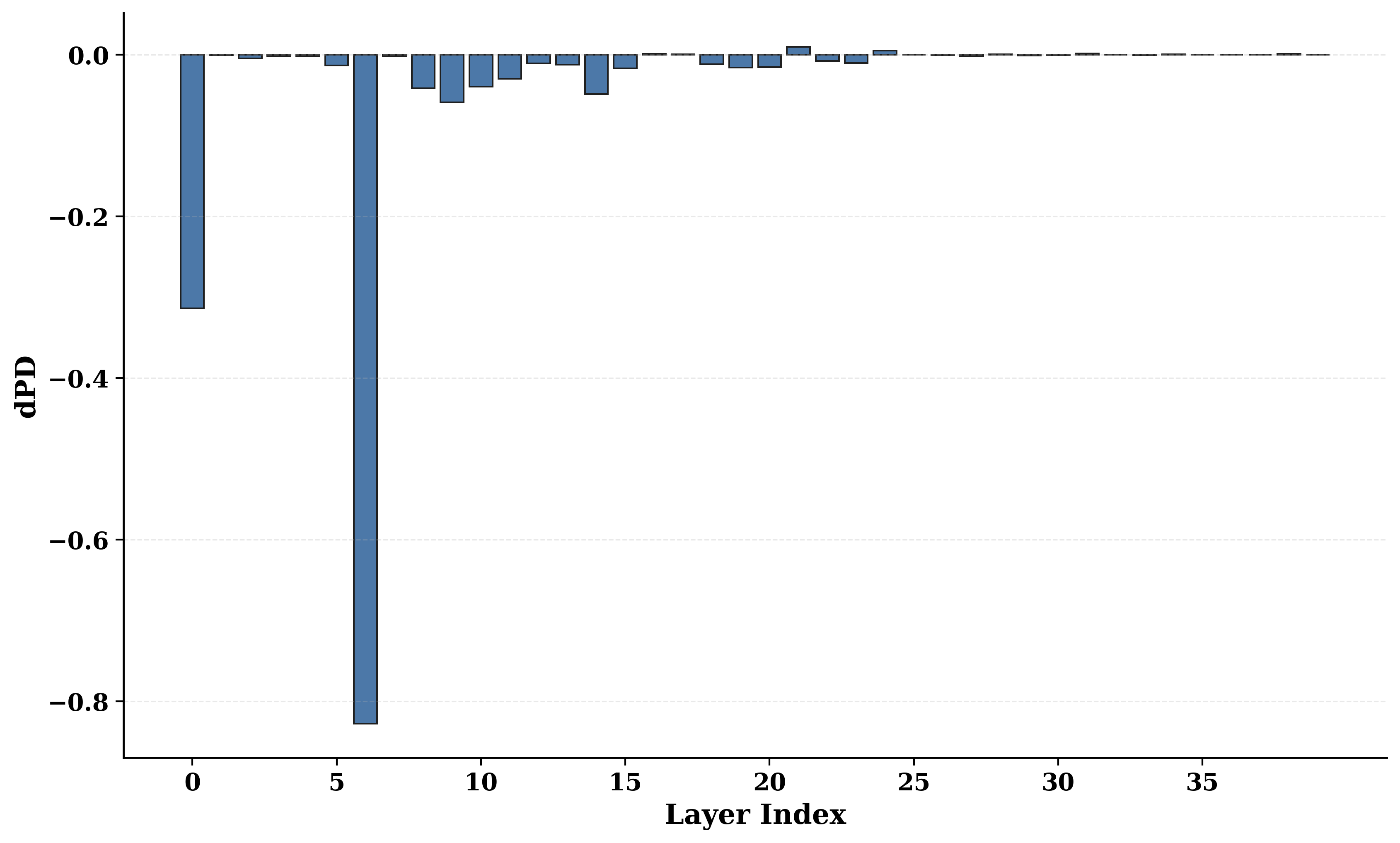}\hfill
    \includegraphics[width=0.32\textwidth]{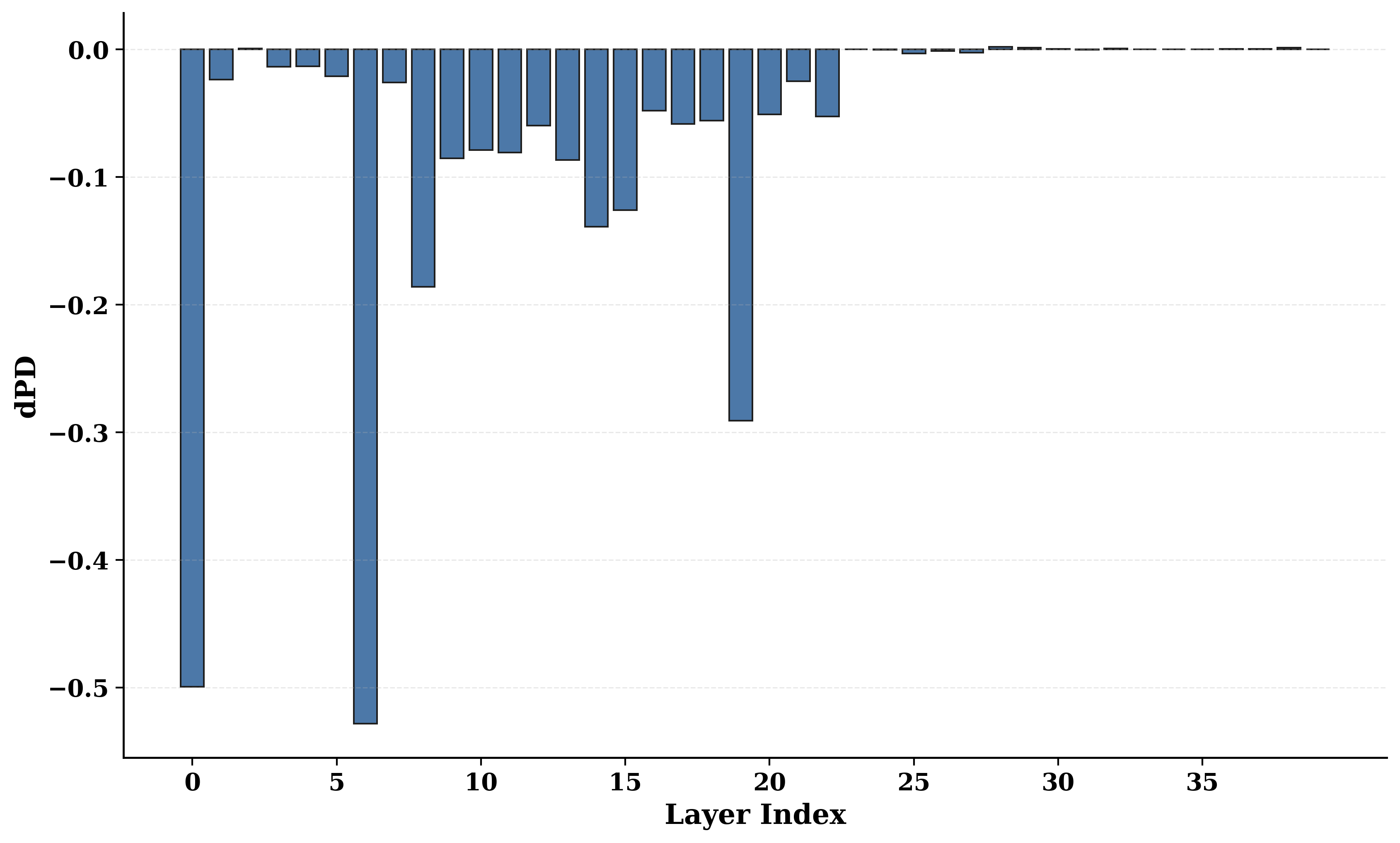}\hfill
    \includegraphics[width=0.32\textwidth]{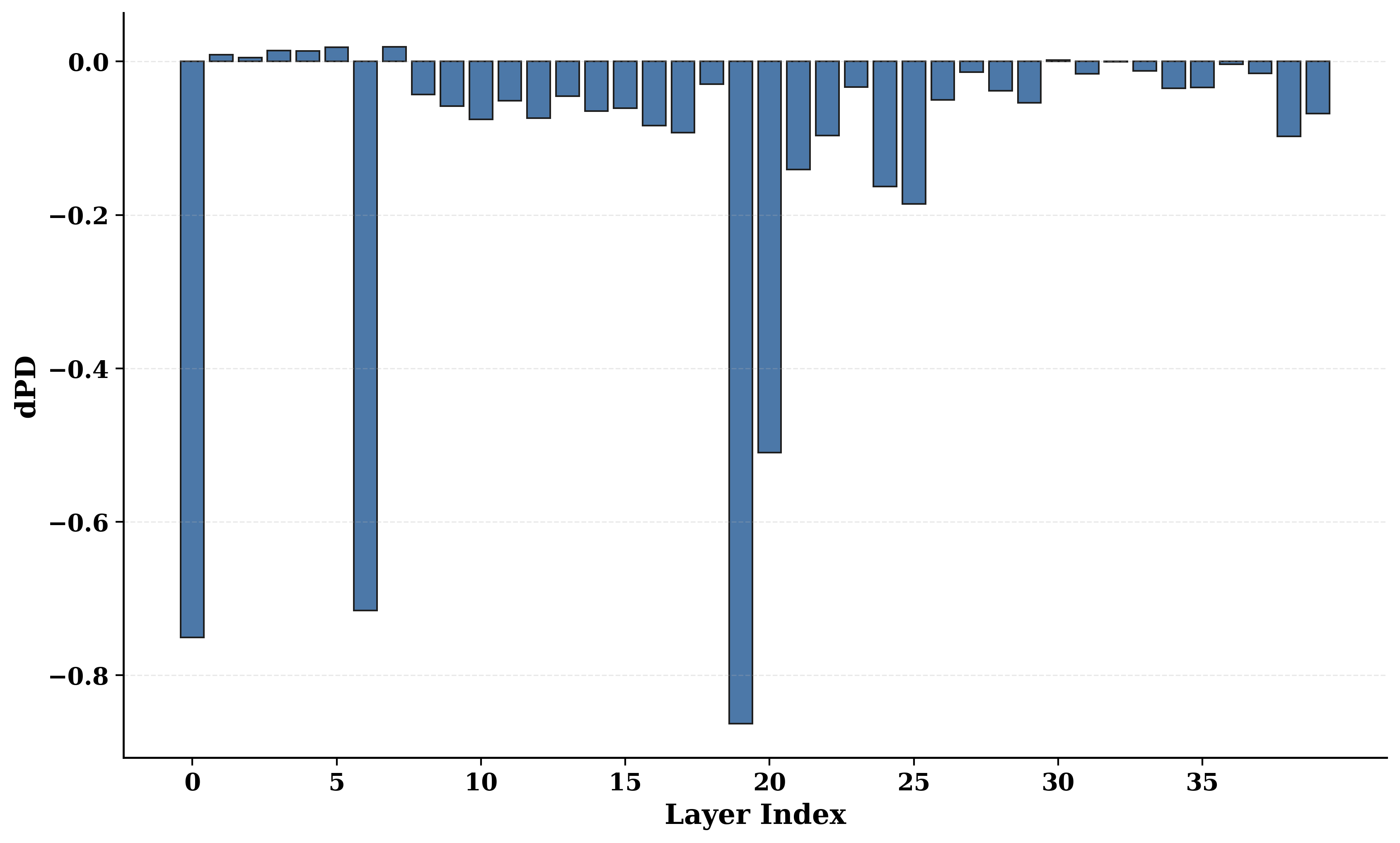}
    \caption{\textbf{MLP-block mean ablation on \textit{Qwen3-14B} under $\pi_{\mathrm{EF}}$.} Per-layer $\mathrm{dPD}$ when each logical block is mean-ablated. Top row: one-hop. Bottom row: two-hop. Columns: Facts / Expression / Query.}
    \label{fig:stage_mlp_q14_ef}
\end{figure*}

\begin{figure*}[!htbp]
    \centering
    \includegraphics[width=0.32\textwidth]{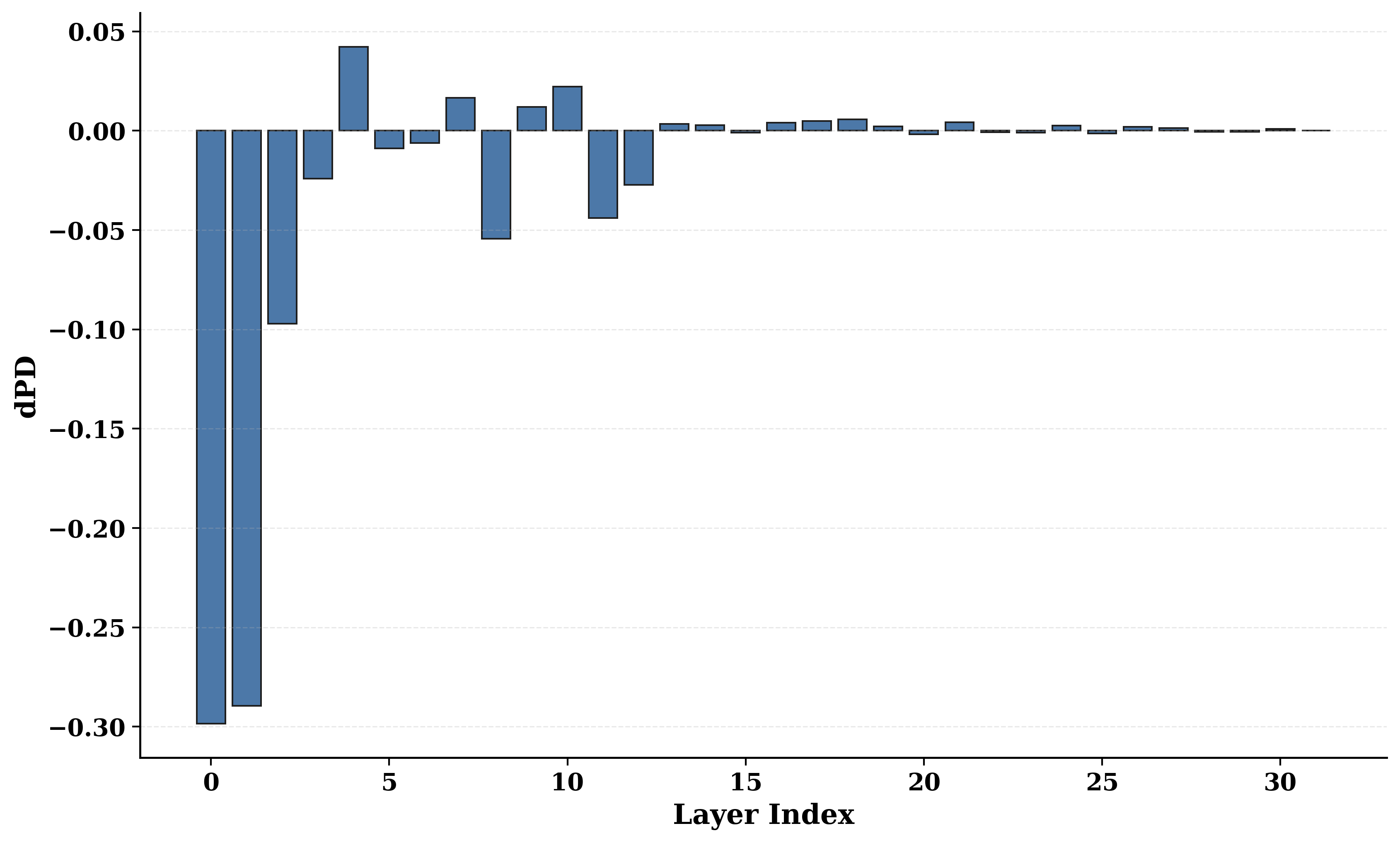}\hfill
    \includegraphics[width=0.32\textwidth]{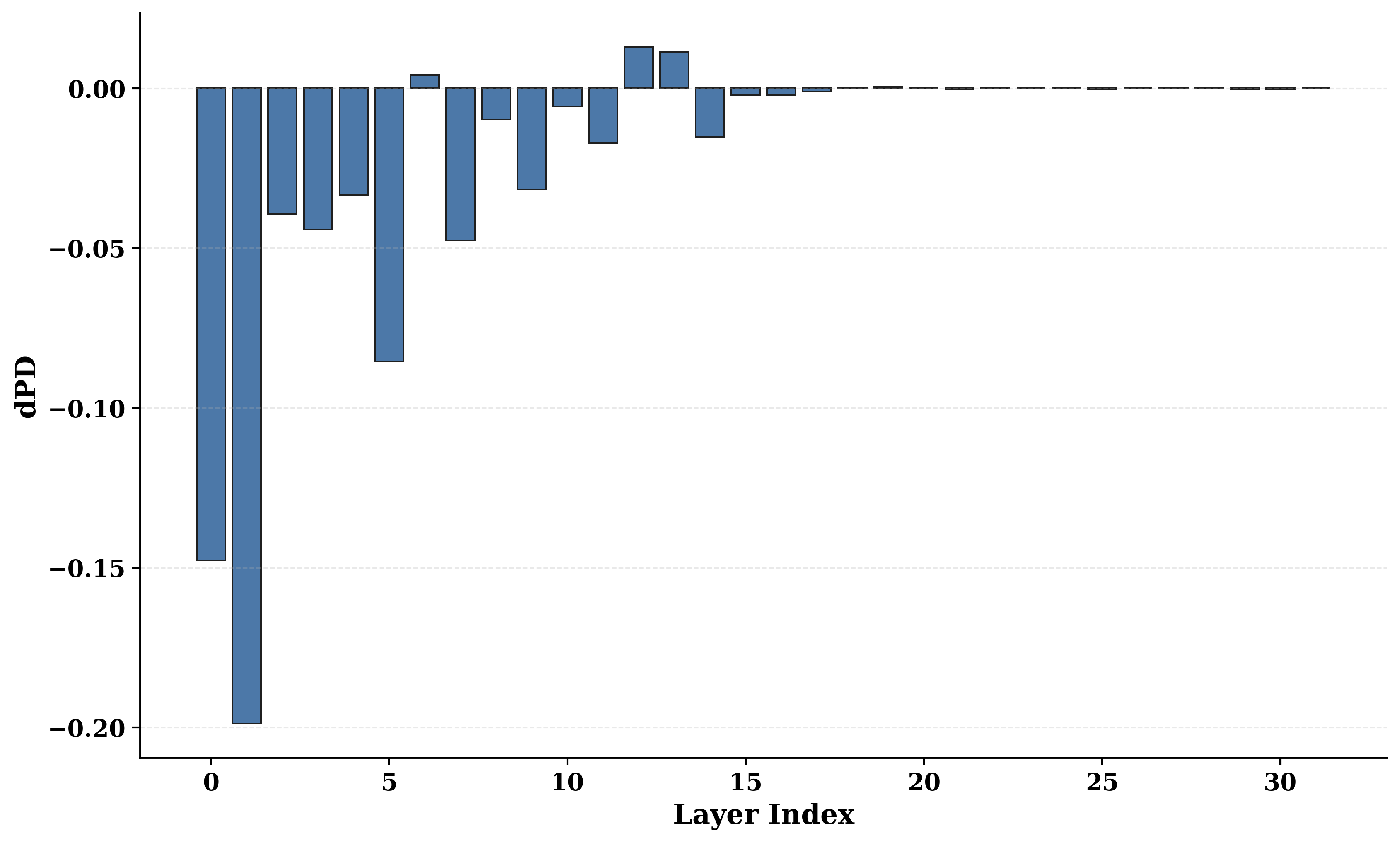}\hfill
    \includegraphics[width=0.32\textwidth]{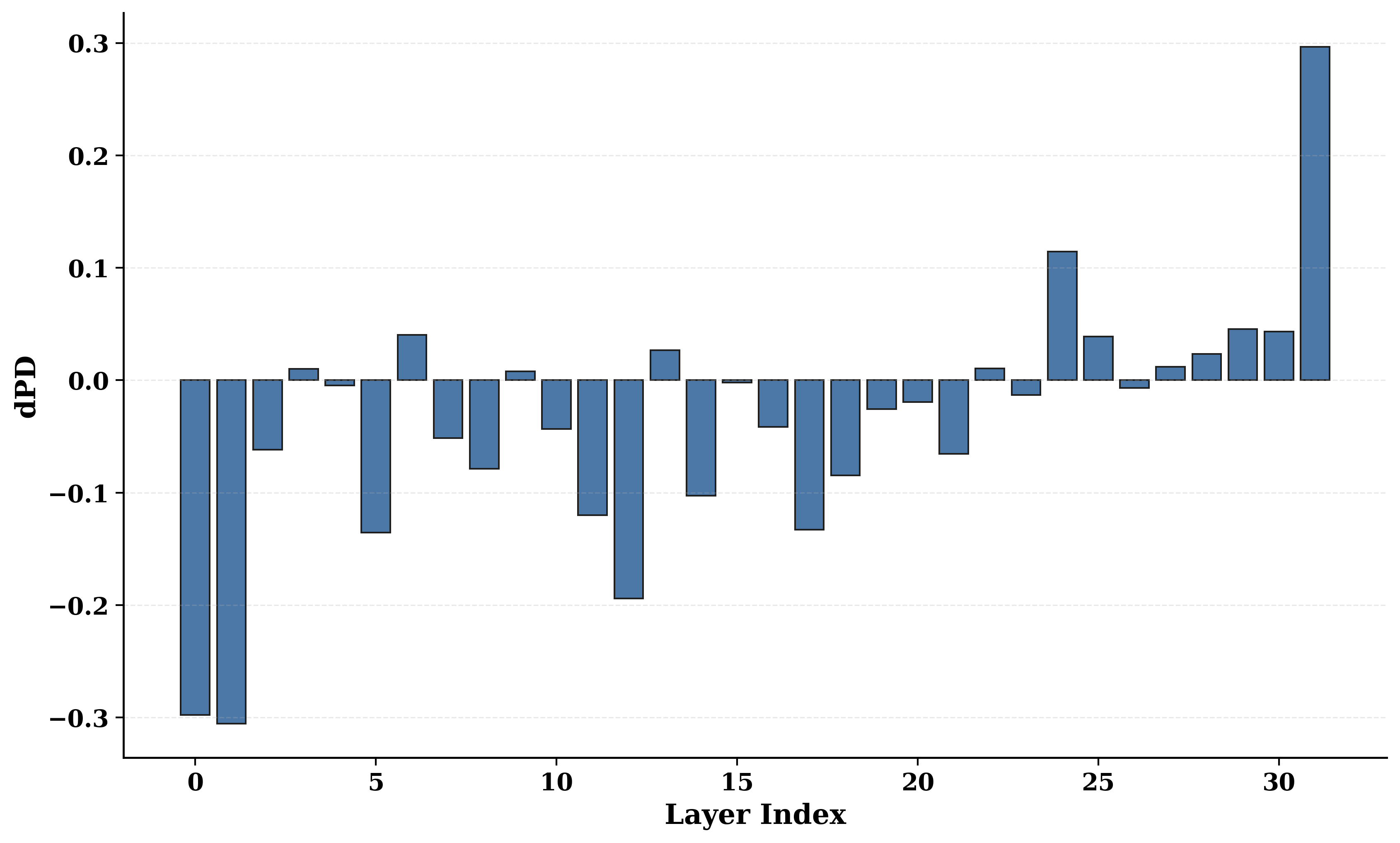}

    \includegraphics[width=0.32\textwidth]{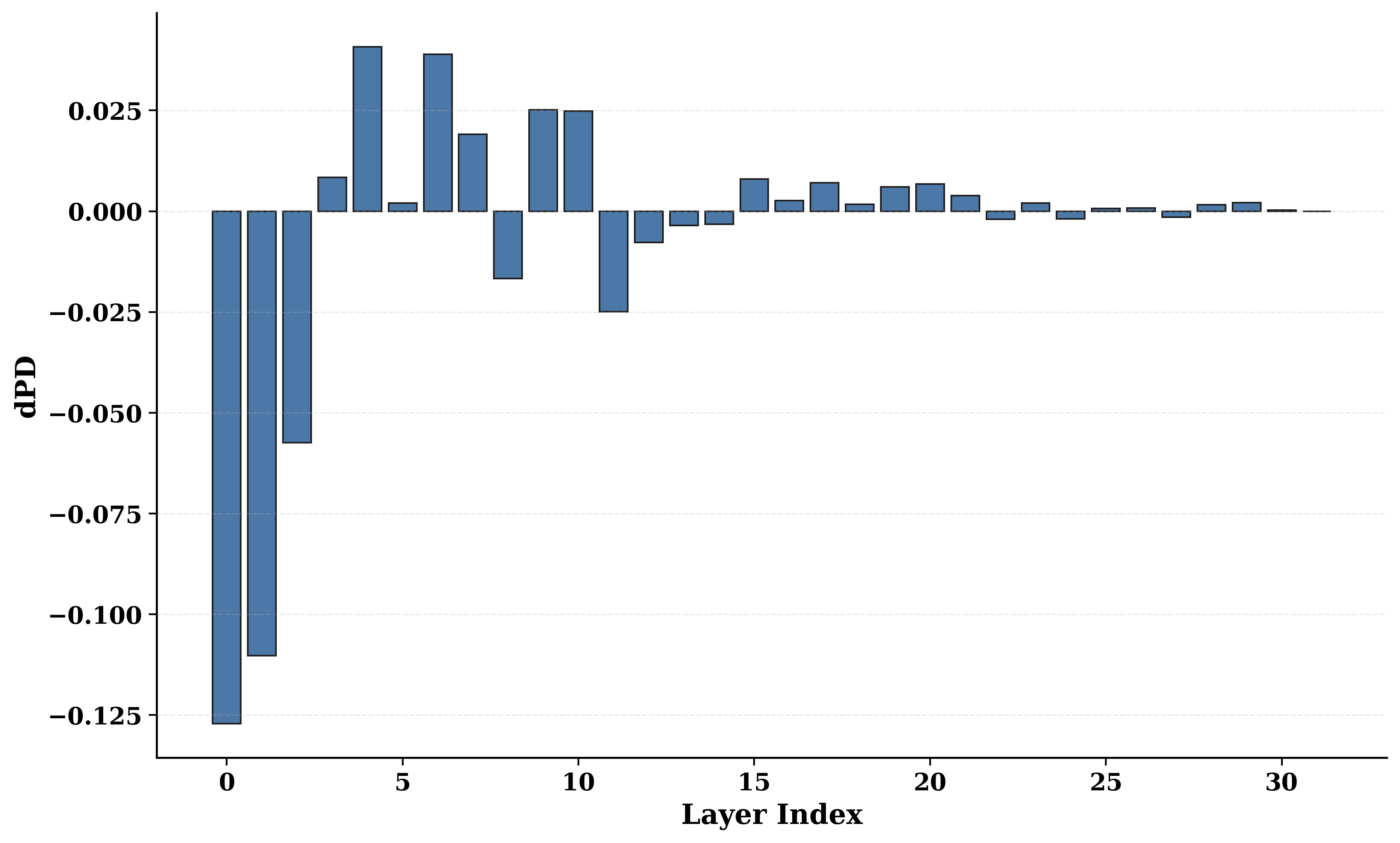}\hfill
    \includegraphics[width=0.32\textwidth]{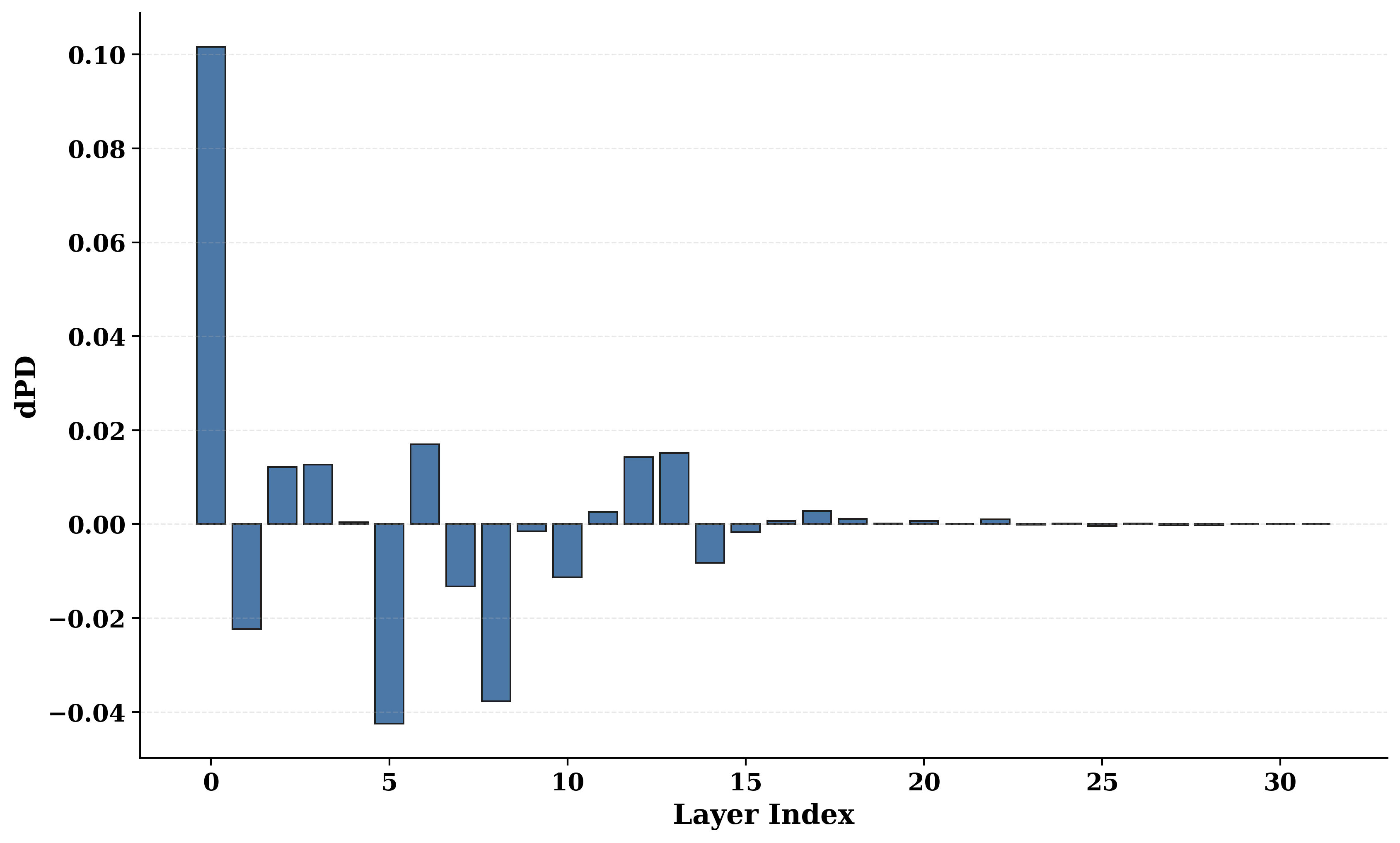}\hfill
    \includegraphics[width=0.32\textwidth]{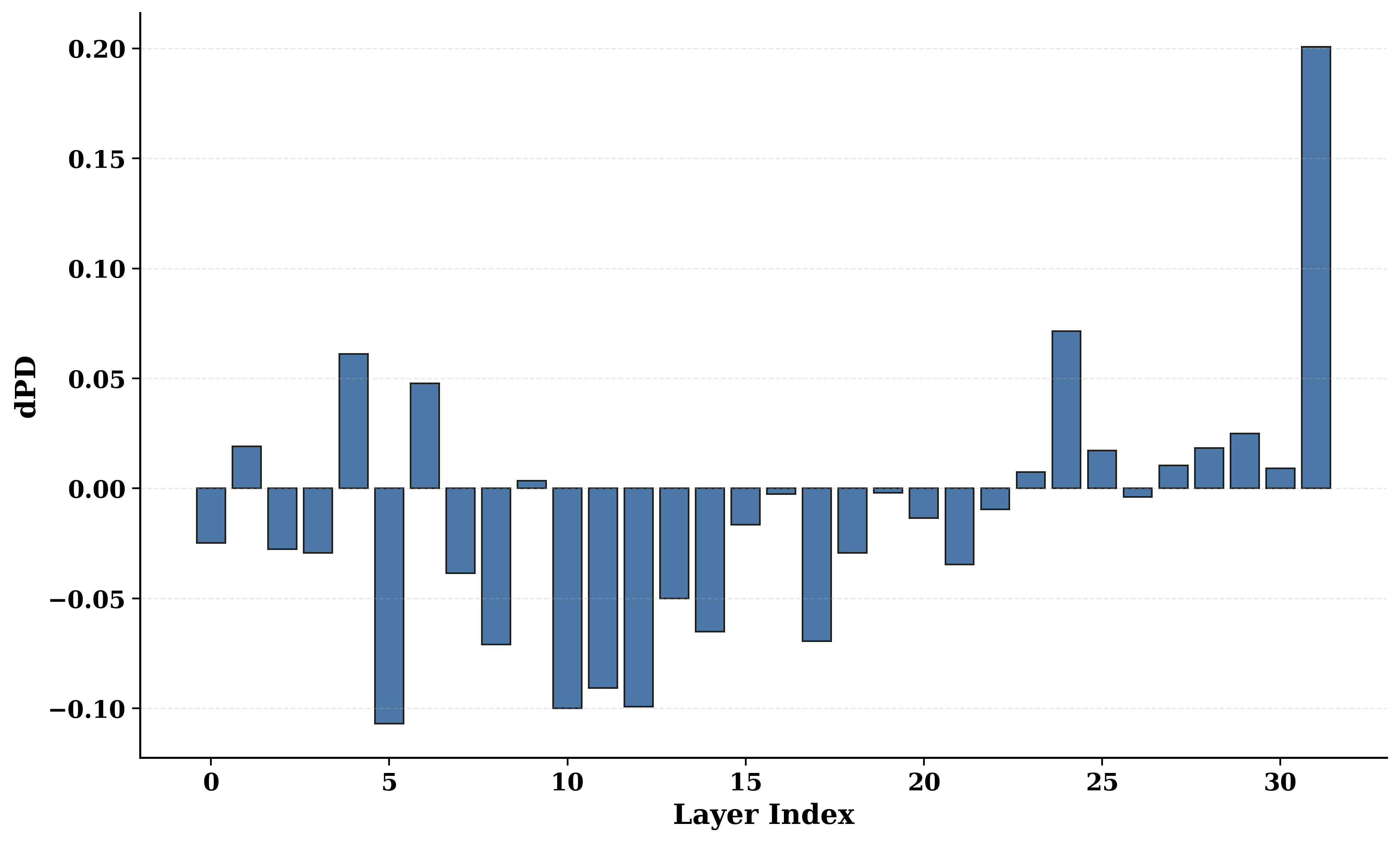}
    \caption{\textbf{MLP-block mean ablation on \textit{Llama-3.1-8B-Instruct} under $\pi_{\mathrm{FF}}$.} Per-layer $\mathrm{dPD}$ when each logical block is mean-ablated. Top row: one-hop. Bottom row: two-hop. Columns: Facts / Expression / Query.}
    \label{fig:stage_mlp_llama_ff}
    \label{fig:llama_mlp_facts_gallery}
\end{figure*}

\begin{figure*}[!htbp]
    \centering
    \includegraphics[width=0.32\textwidth]{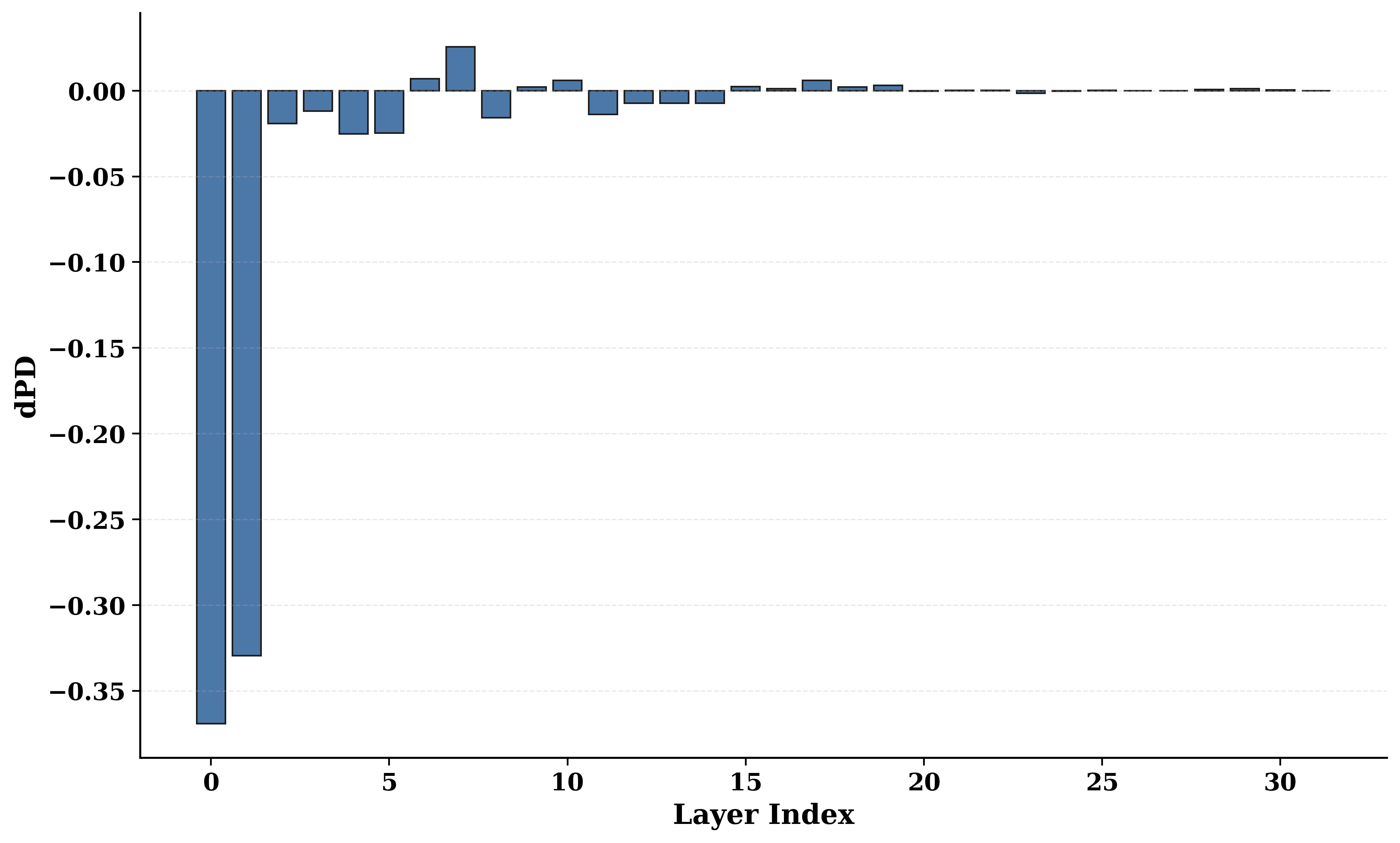}\hfill
    \includegraphics[width=0.32\textwidth]{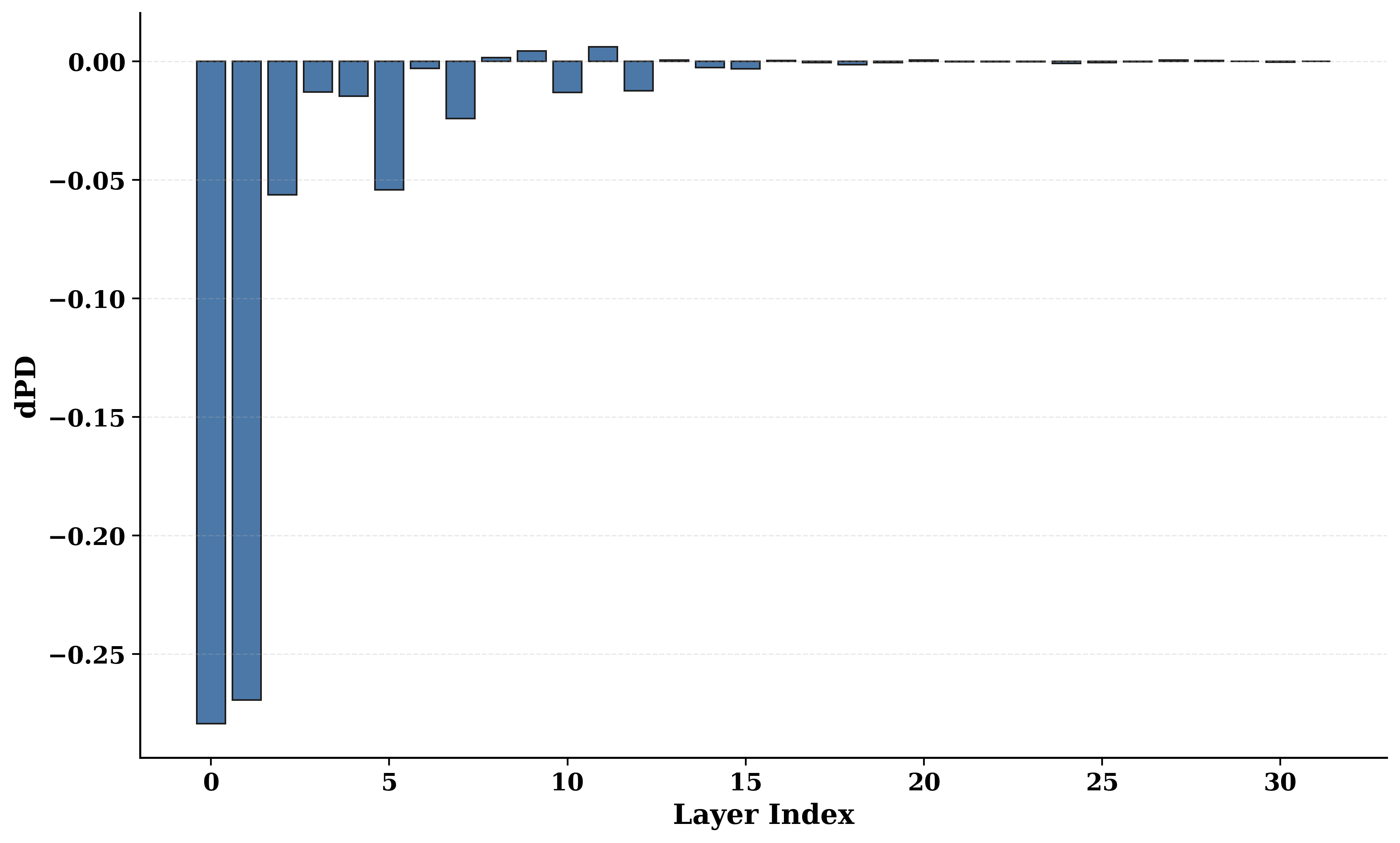}\hfill
    \includegraphics[width=0.32\textwidth]{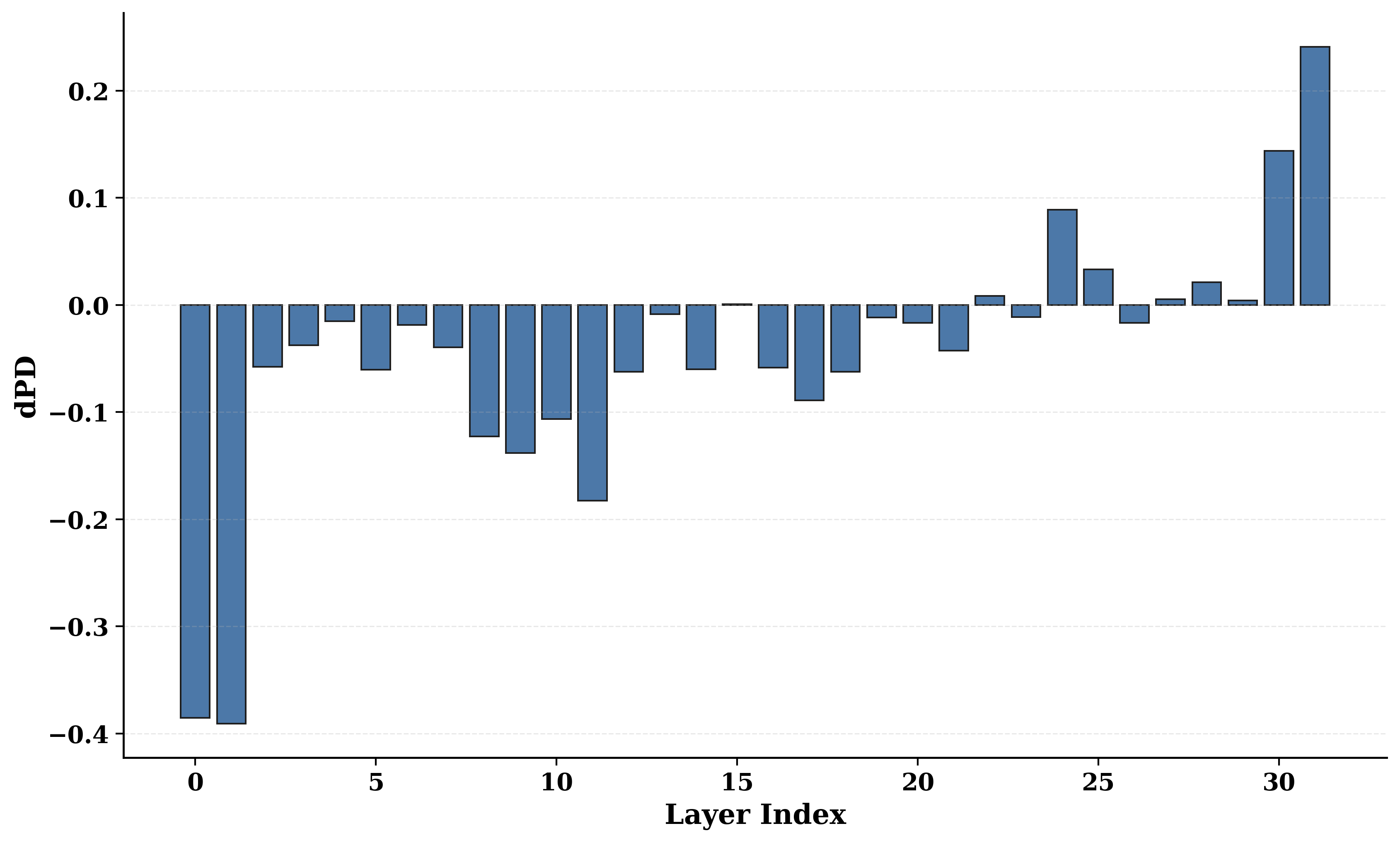}

    \includegraphics[width=0.32\textwidth]{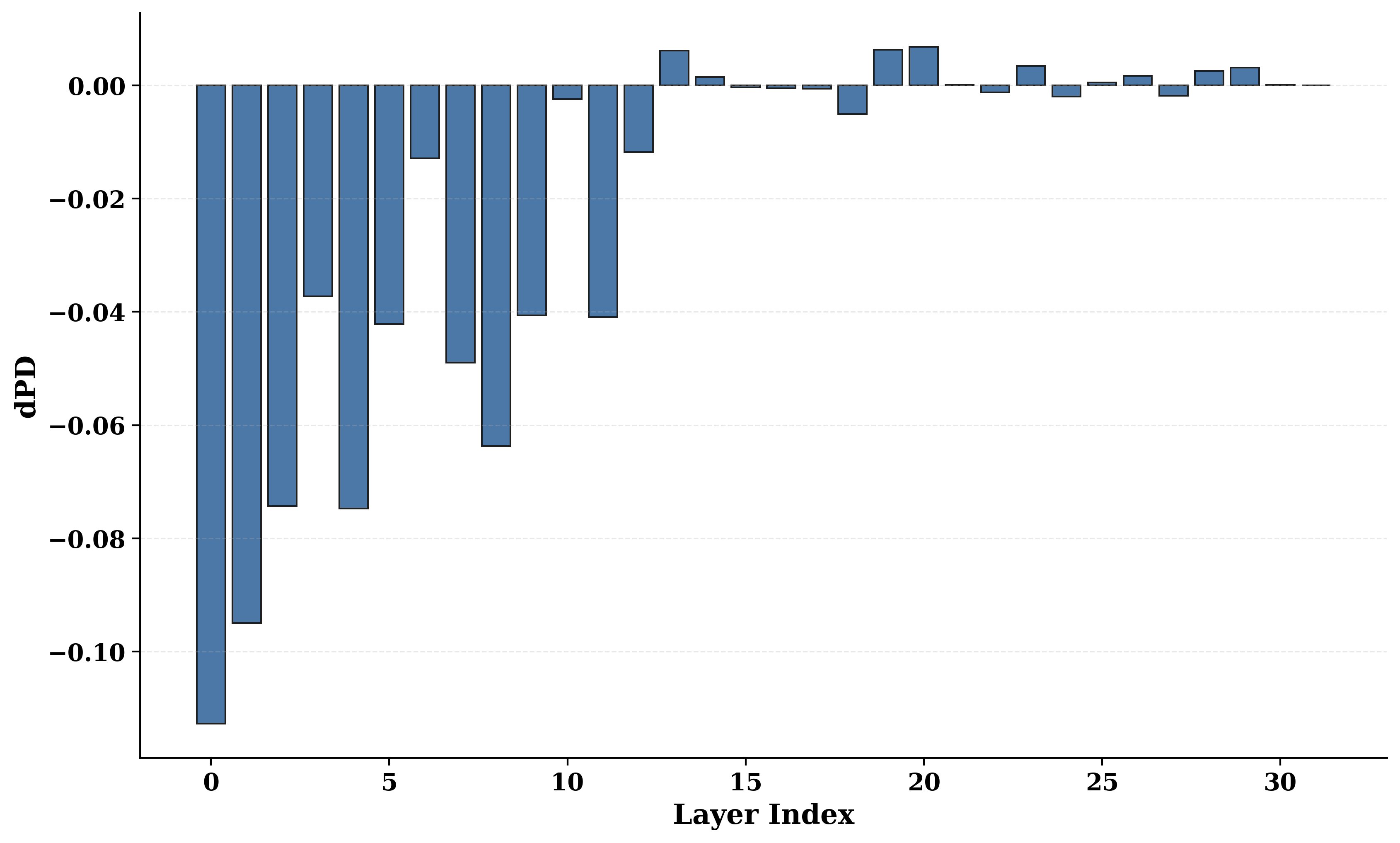}\hfill
    \includegraphics[width=0.32\textwidth]{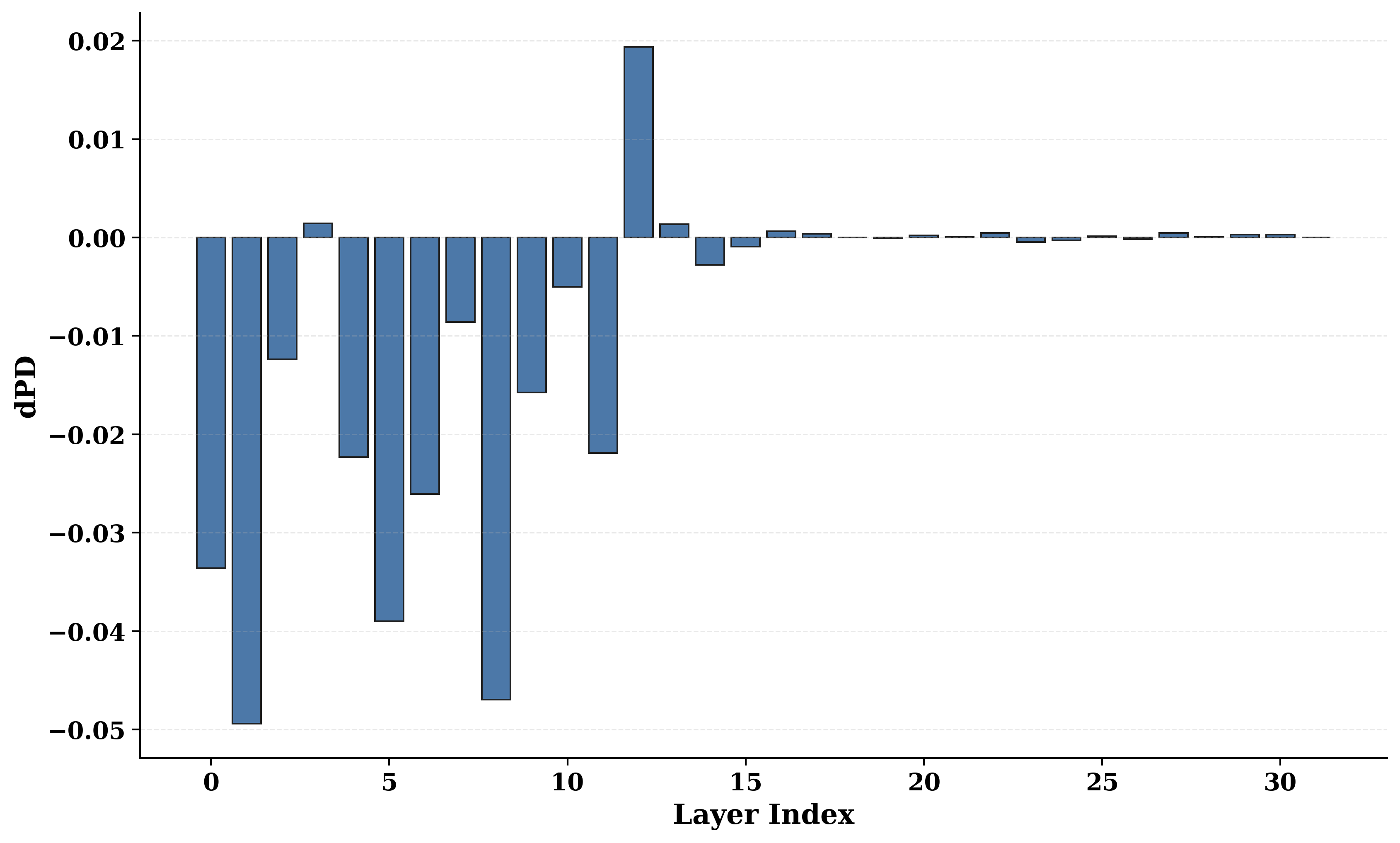}\hfill
    \includegraphics[width=0.32\textwidth]{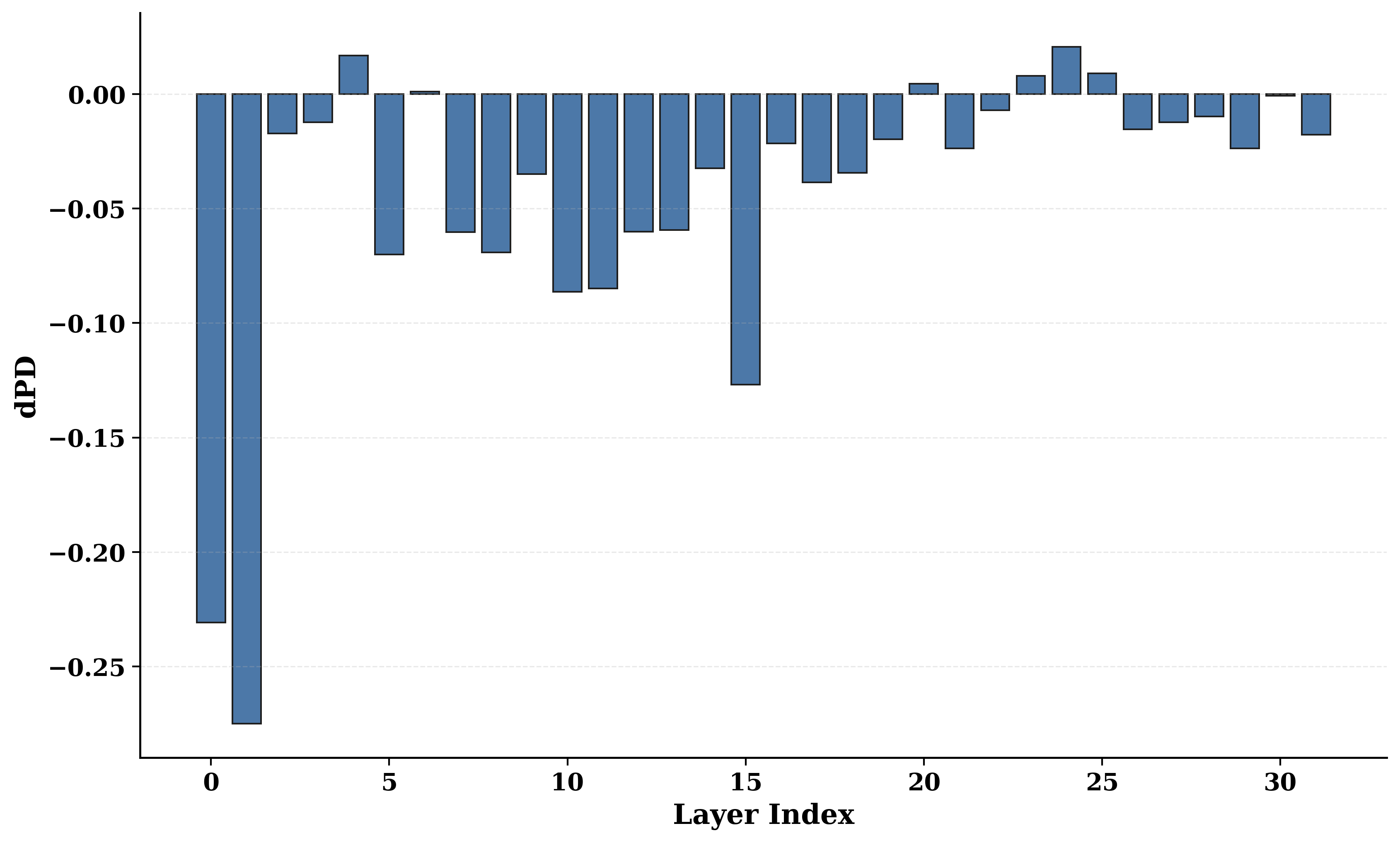}
    \caption{\textbf{MLP-block mean ablation on \textit{Llama-3.1-8B-Instruct} under $\pi_{\mathrm{EF}}$.} Per-layer $\mathrm{dPD}$ when each logical block is mean-ablated. Top row: one-hop. Bottom row: two-hop. Columns: Facts / Expression / Query.}
    \label{fig:stage_mlp_llama_ef}
\end{figure*}

\begin{figure*}[!htbp]
    \centering
    \includegraphics[width=0.32\textwidth]{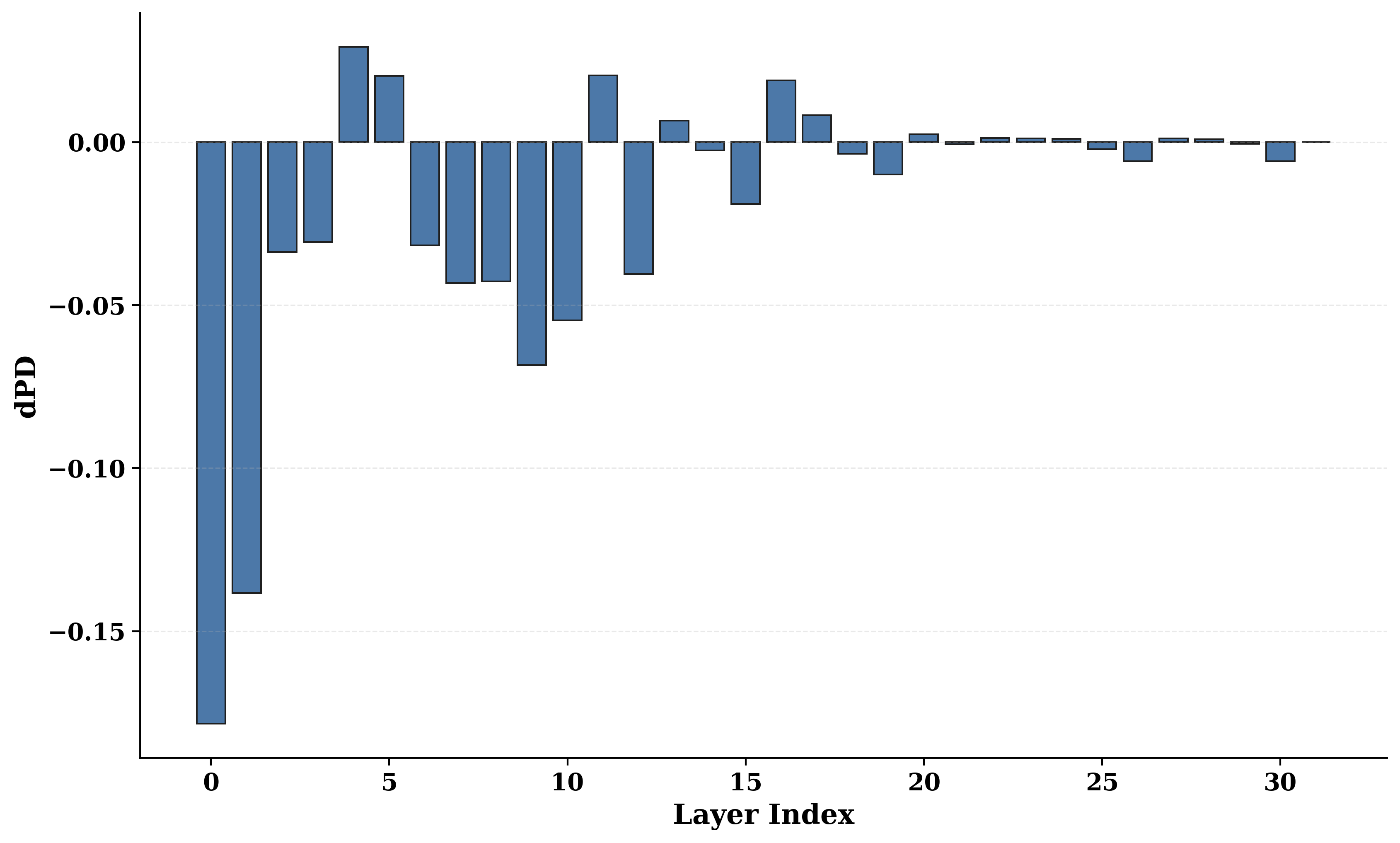}\hfill
    \includegraphics[width=0.32\textwidth]{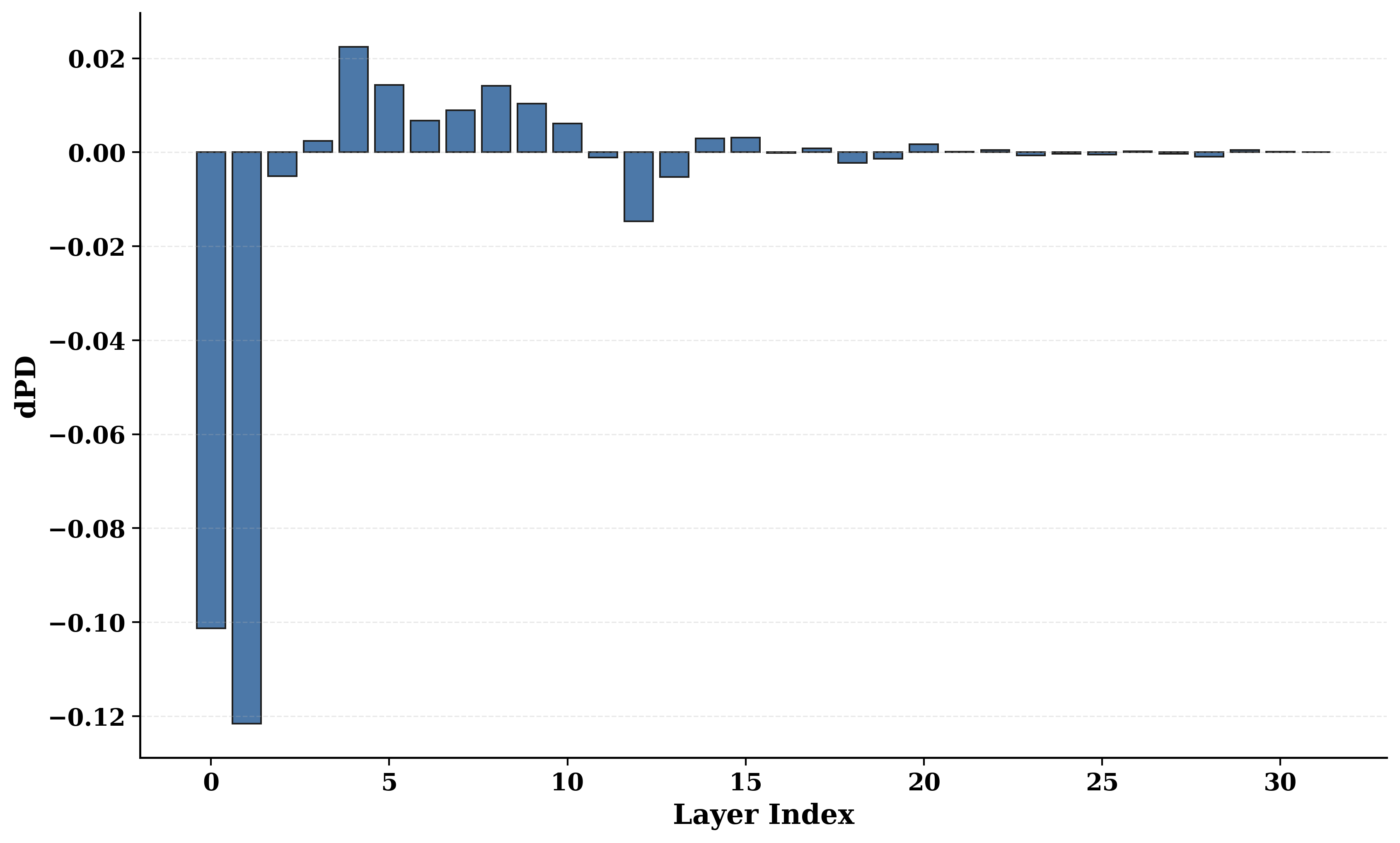}\hfill
    \includegraphics[width=0.32\textwidth]{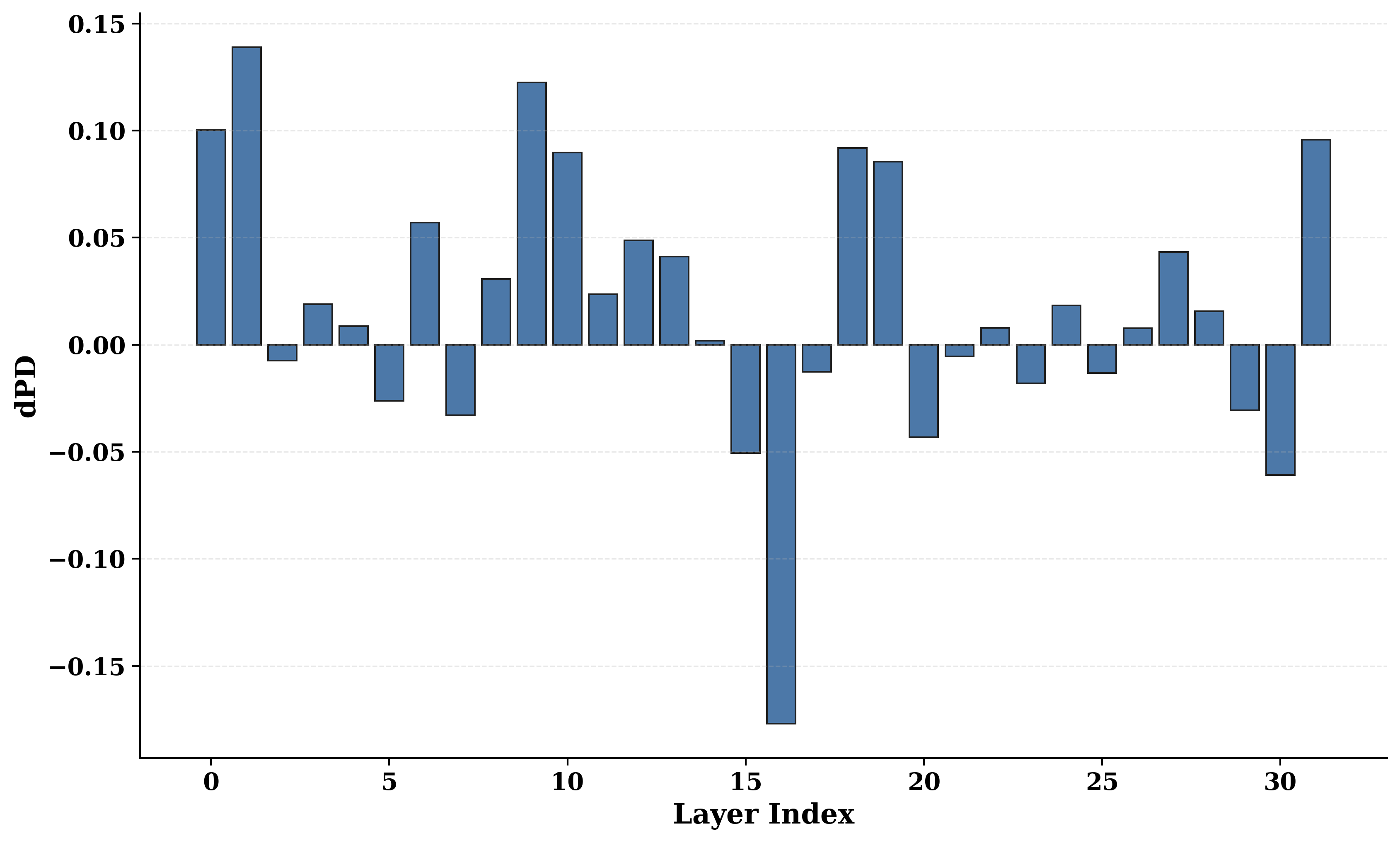}

    \includegraphics[width=0.32\textwidth]{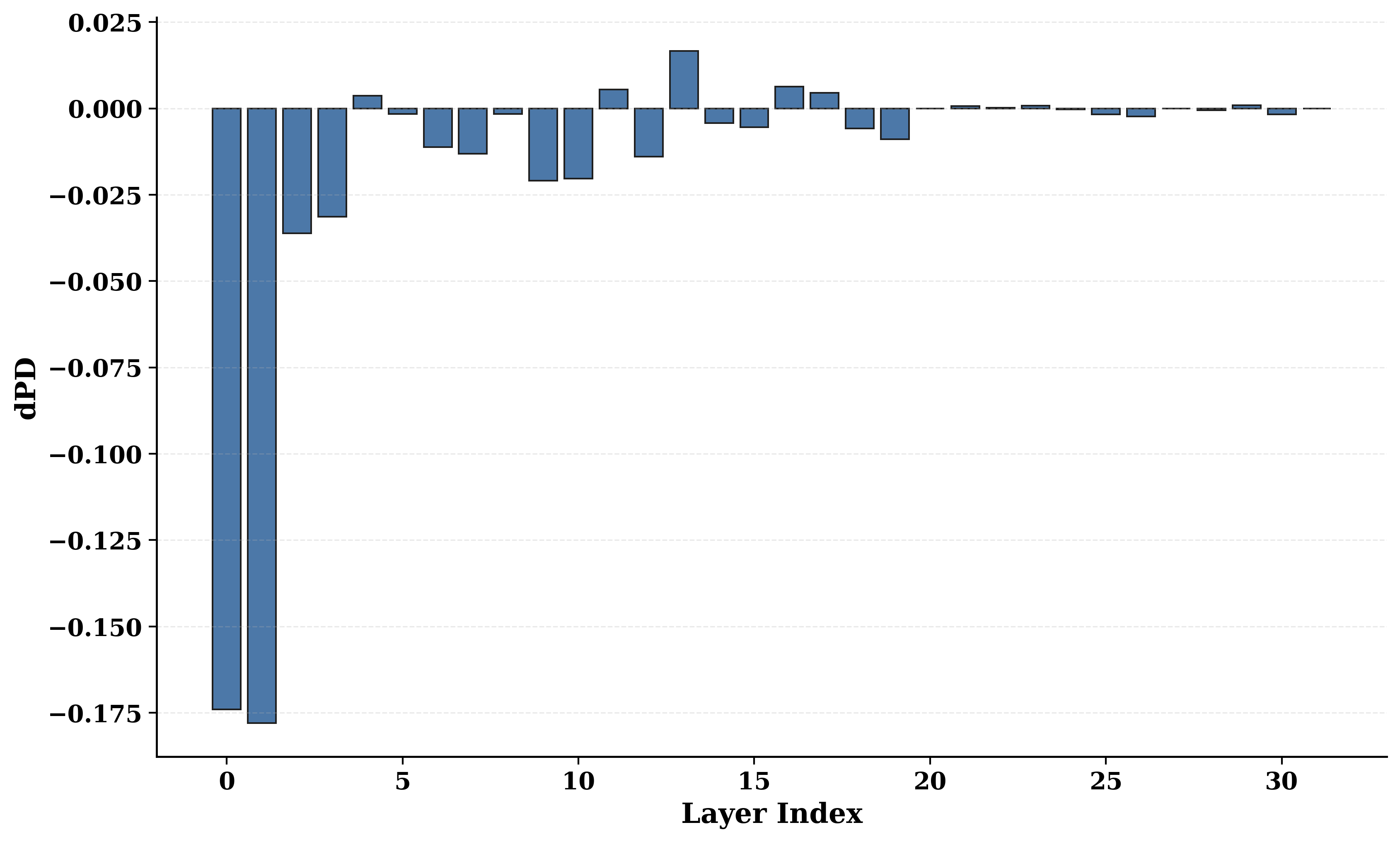}\hfill
    \includegraphics[width=0.32\textwidth]{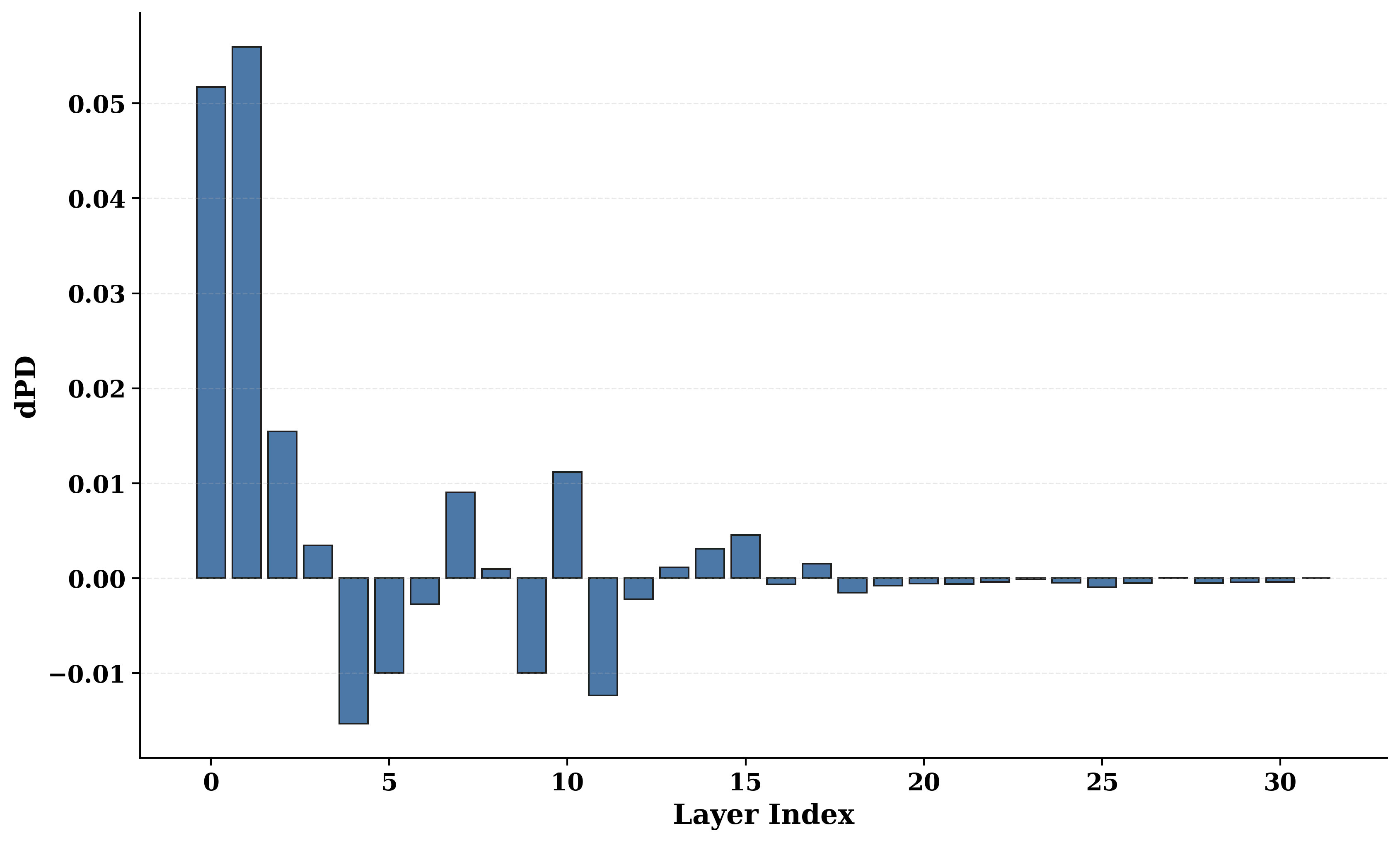}\hfill
    \includegraphics[width=0.32\textwidth]{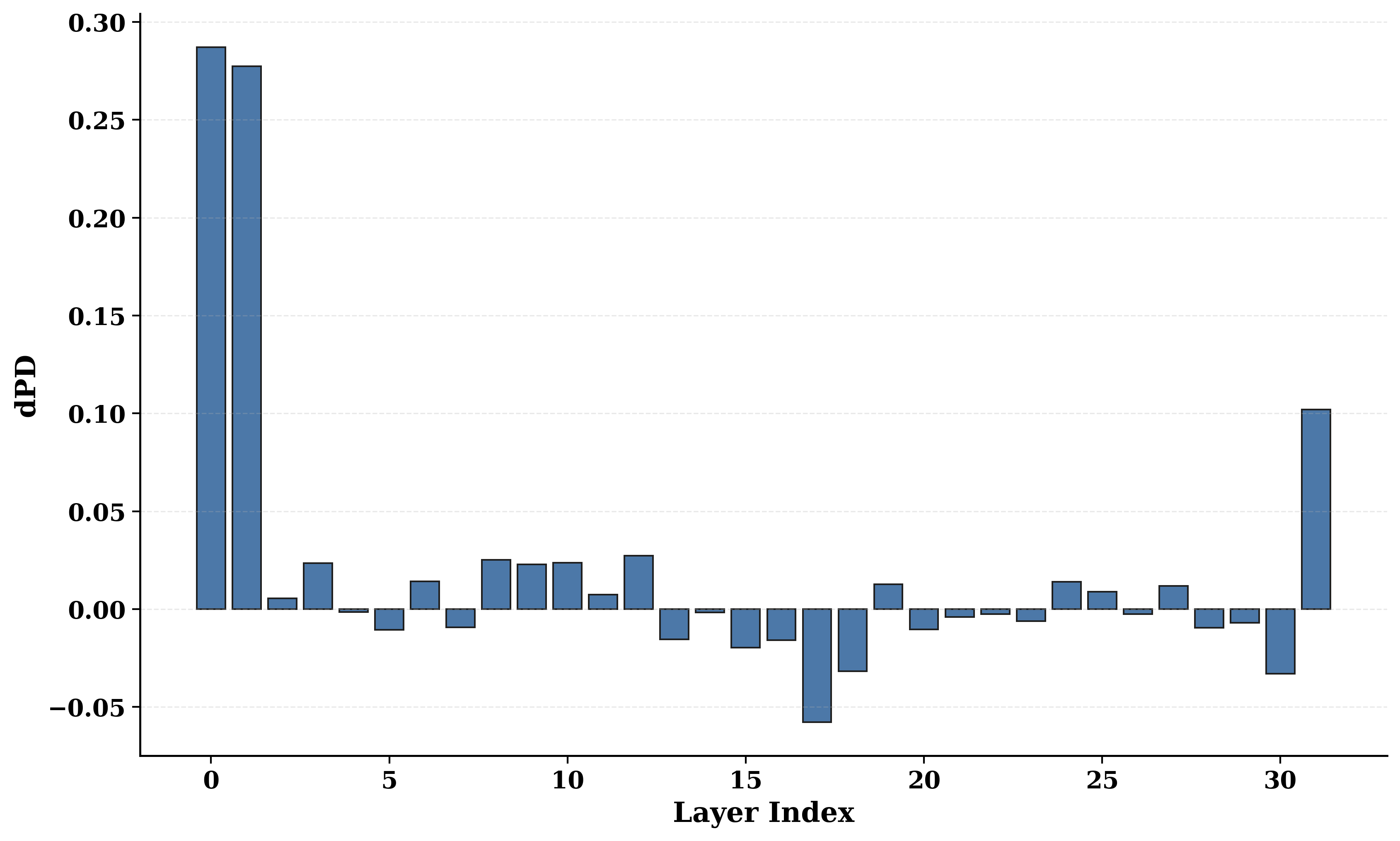}
    \caption{\textbf{MLP-block mean ablation on \textit{Mistral-7B-Instruct} under $\pi_{\mathrm{FF}}$.} Per-layer $\mathrm{dPD}$ when each logical block is mean-ablated. Top row: one-hop. Bottom row: two-hop. Columns: Facts / Expression / Query.}
    \label{fig:stage_mlp_mistral_ff}
    \label{fig:mistral_mlp_facts_gallery}
\end{figure*}

\begin{figure*}[!htbp]
    \centering
    \includegraphics[width=0.32\textwidth]{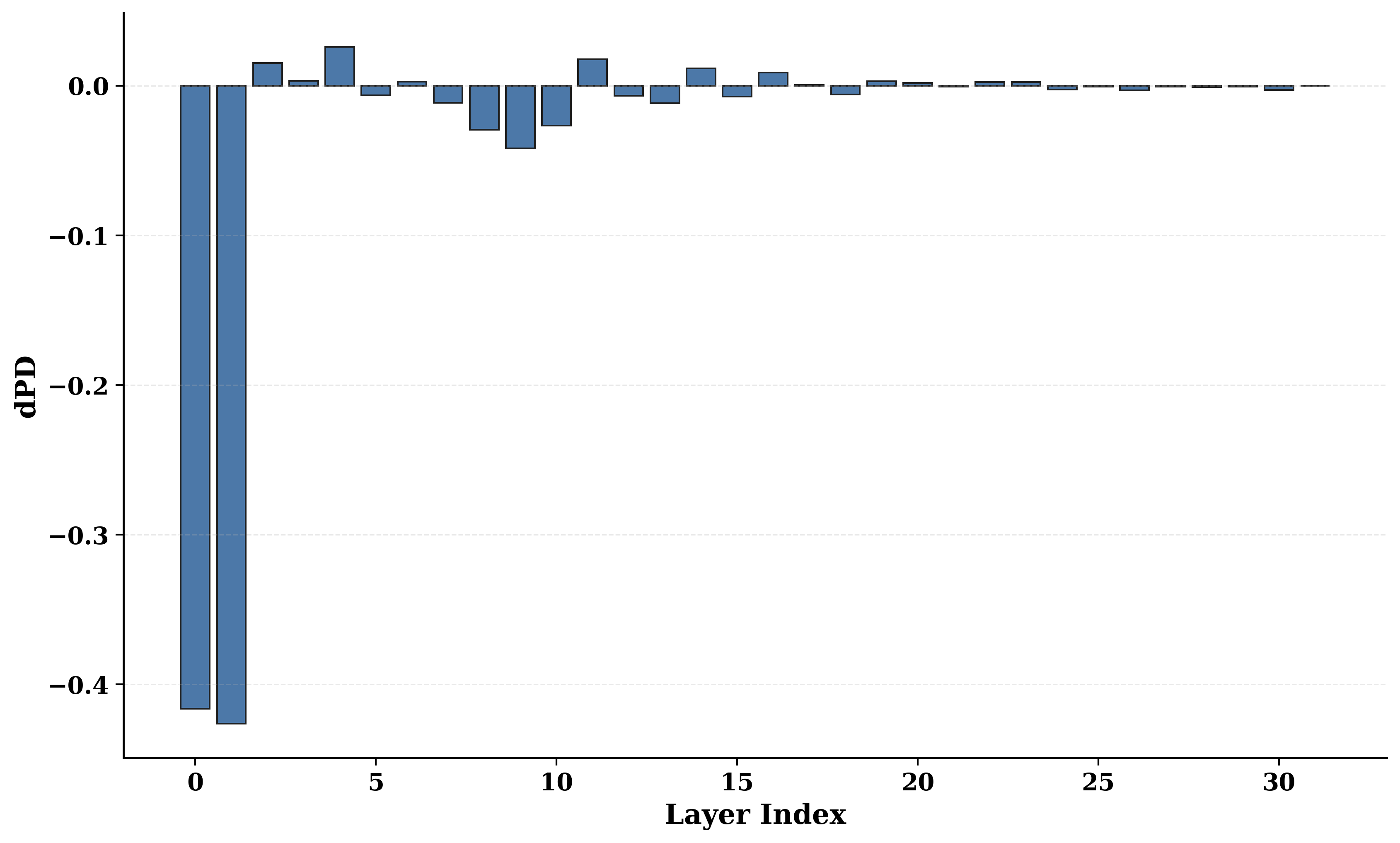}\hfill
    \includegraphics[width=0.32\textwidth]{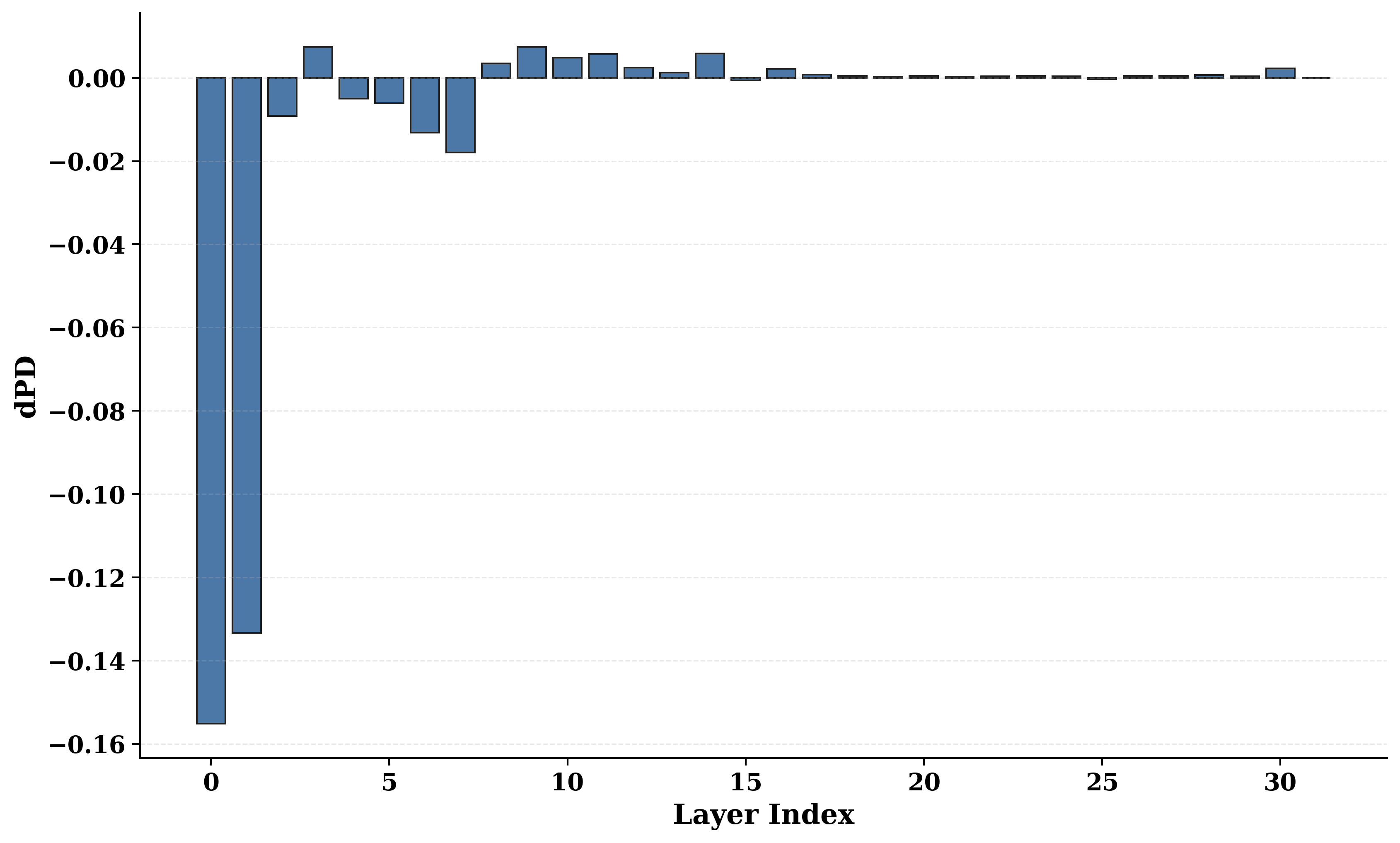}\hfill
    \includegraphics[width=0.32\textwidth]{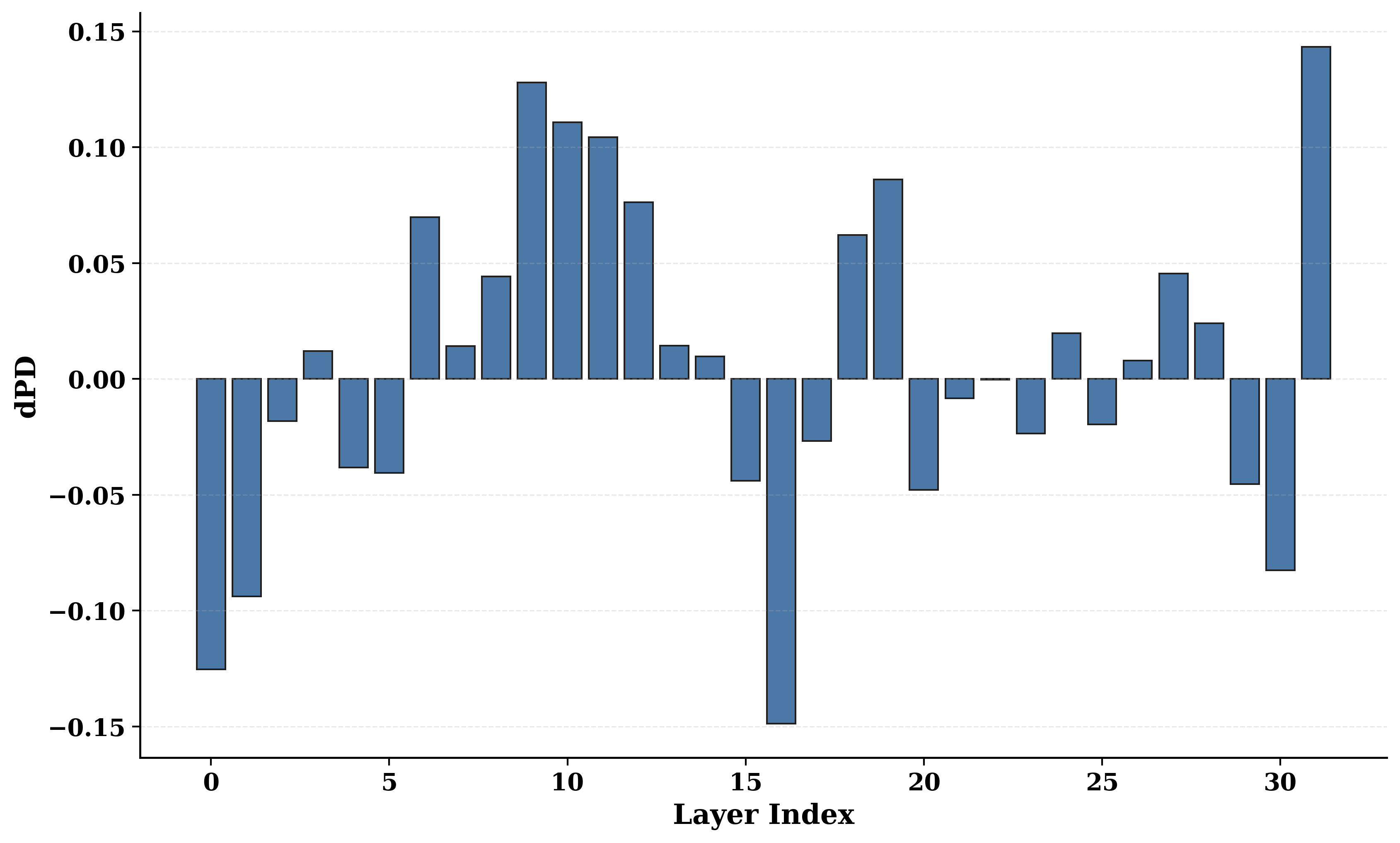}

    \includegraphics[width=0.32\textwidth]{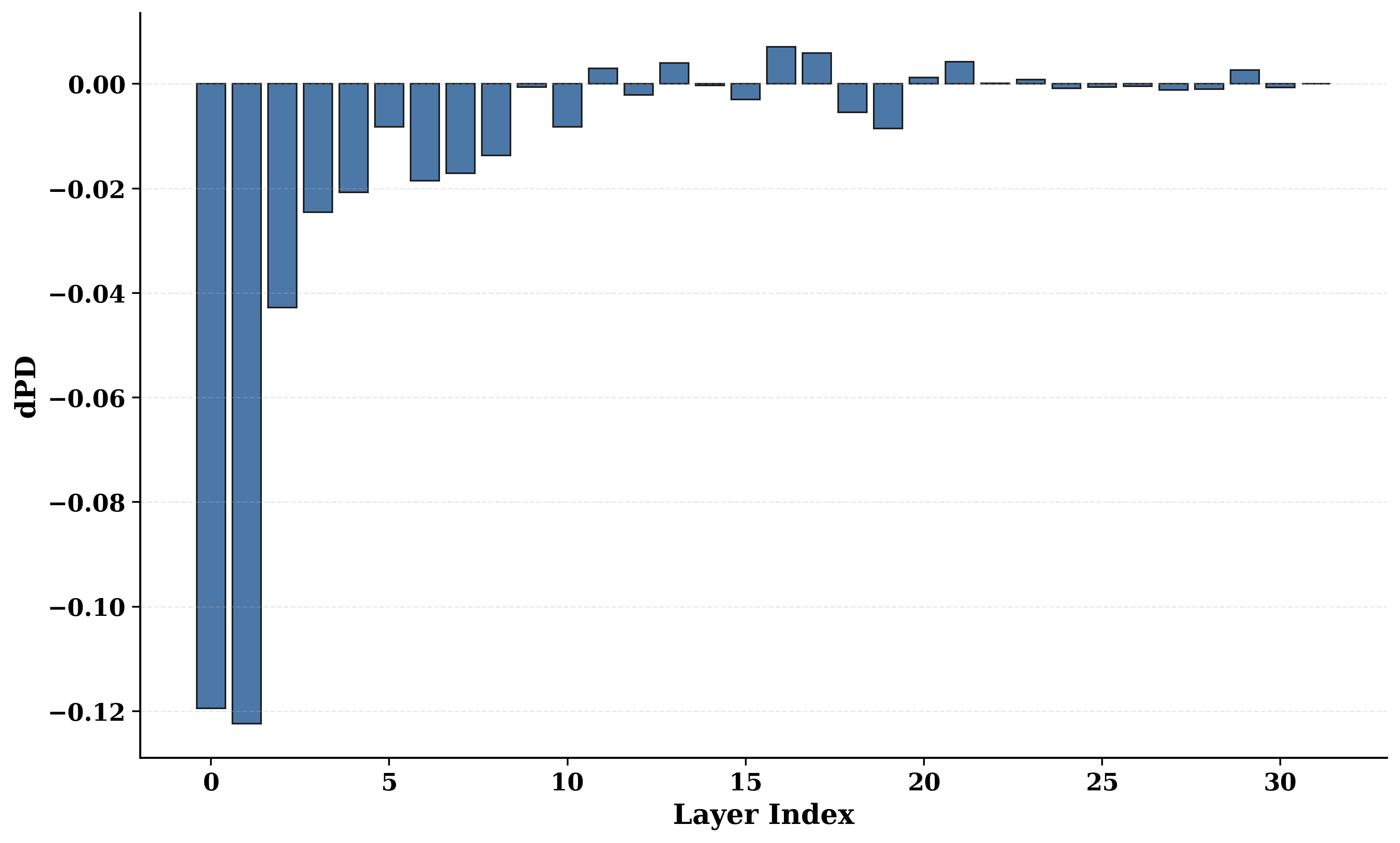}\hfill
    \includegraphics[width=0.32\textwidth]{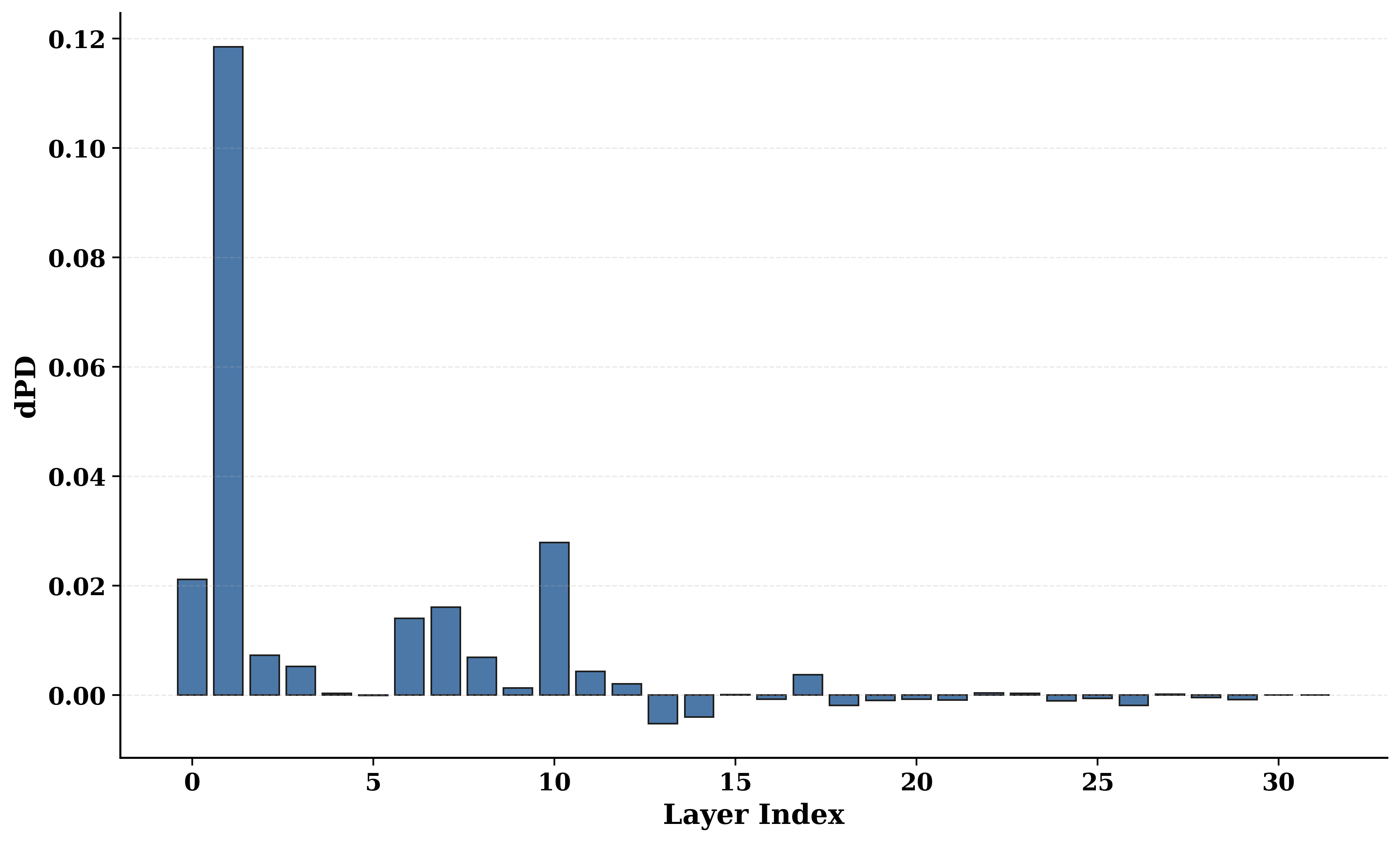}\hfill
    \includegraphics[width=0.32\textwidth]{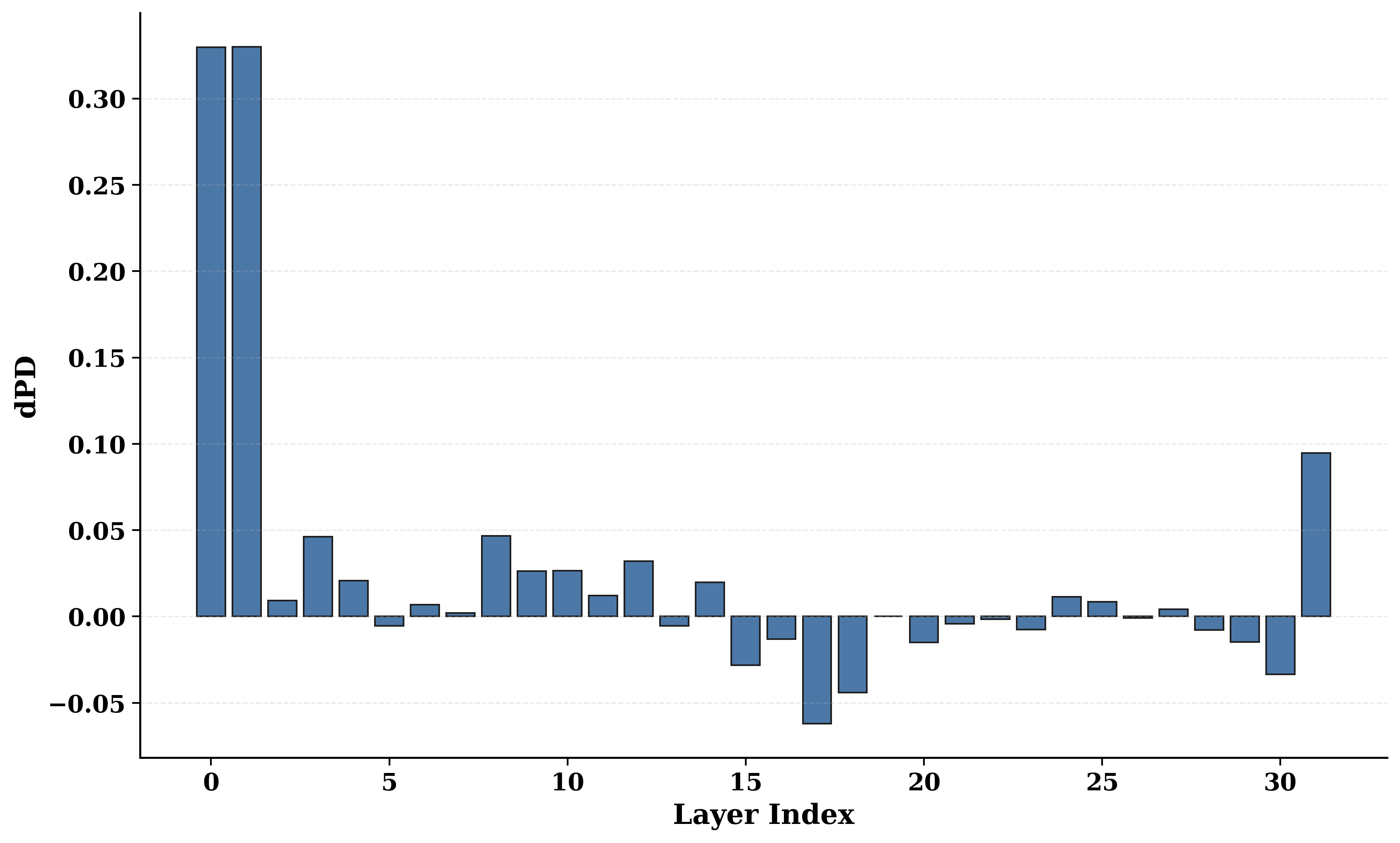}
    \caption{\textbf{MLP-block mean ablation on \textit{Mistral-7B-Instruct} under $\pi_{\mathrm{EF}}$.} Per-layer $\mathrm{dPD}$ when each logical block is mean-ablated. Top row: one-hop. Bottom row: two-hop. Columns: Facts / Expression / Query.}
    \label{fig:stage_mlp_mistral_ef}
\end{figure*}

Figures~\ref{fig:stage_mlp_q8_ff}--\ref{fig:stage_mlp_mistral_ef} show MLP-block mean-ablation $\mathrm{dPD}$ for all four models under both prompt orders, following the body figure template (Figure~\ref{fig:qwen3_8b_attn_mlp_facts_gallery} in \S\ref{sec:staged}). The early-Facts / middle-Expression / middle-to-late-Query order persists across scales, prompt orders, reasoning depths, and model families. \textit{Qwen3} produces the sharpest stage boundaries; \textit{Llama} and \textit{Mistral} preserve the same coarse ordering while allocating a heavier Query Block throughout depth.

\subsection{Attention-Block Ablation}
\label{sec:appendix_stage_attn}

\begin{figure*}[!htbp]
    \centering
    \includegraphics[width=0.32\textwidth]{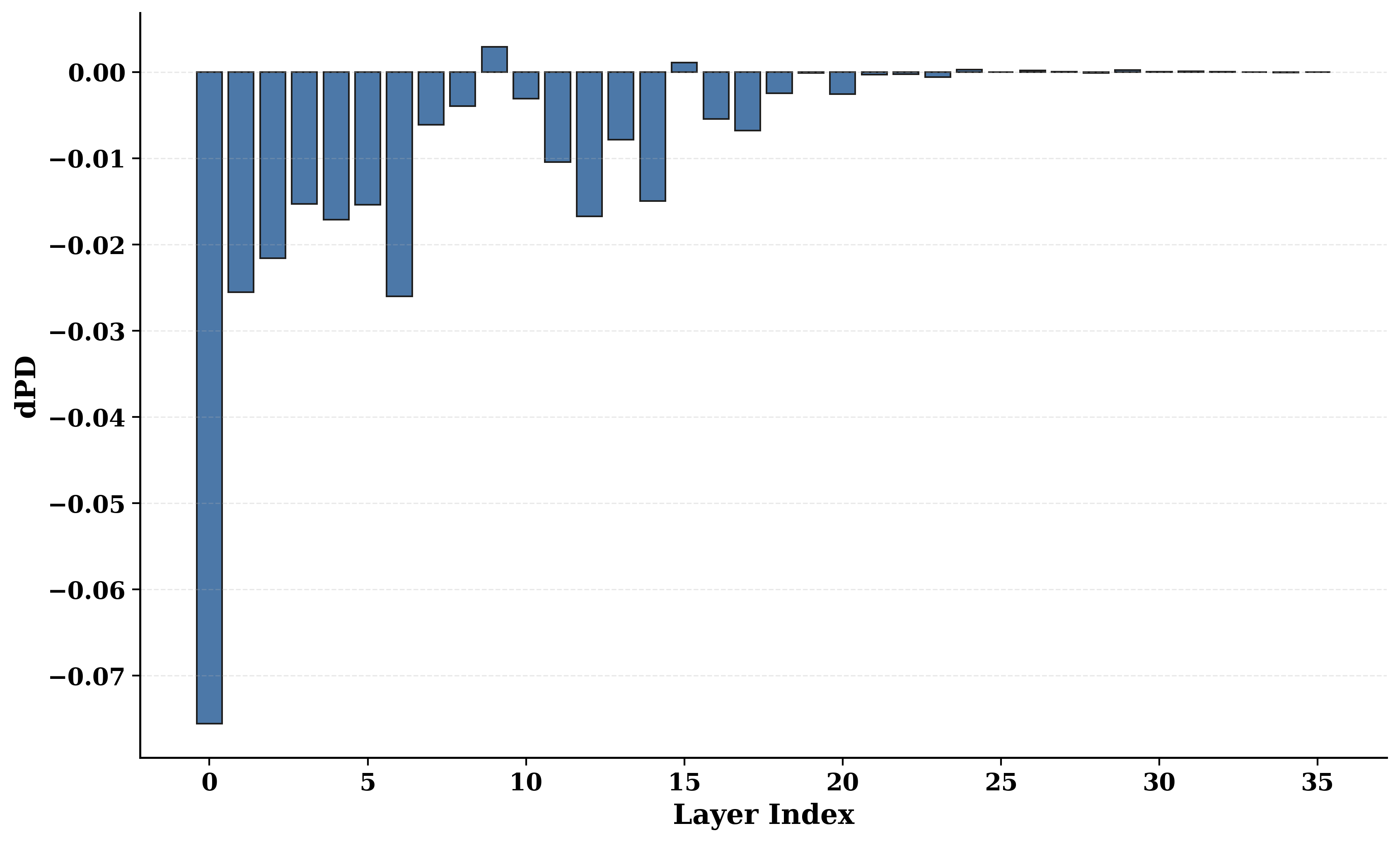}\hfill
    \includegraphics[width=0.32\textwidth]{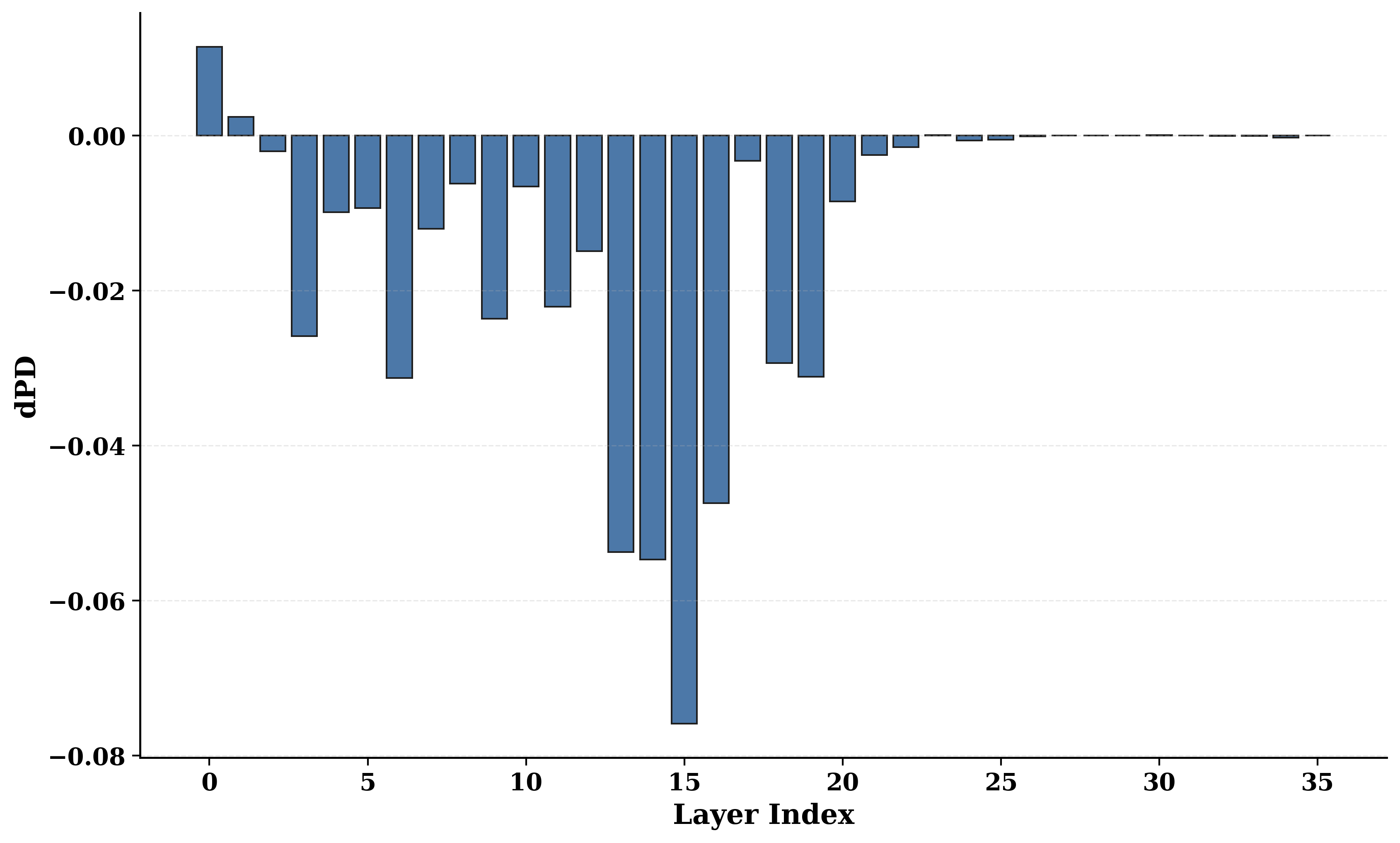}\hfill
    \includegraphics[width=0.32\textwidth]{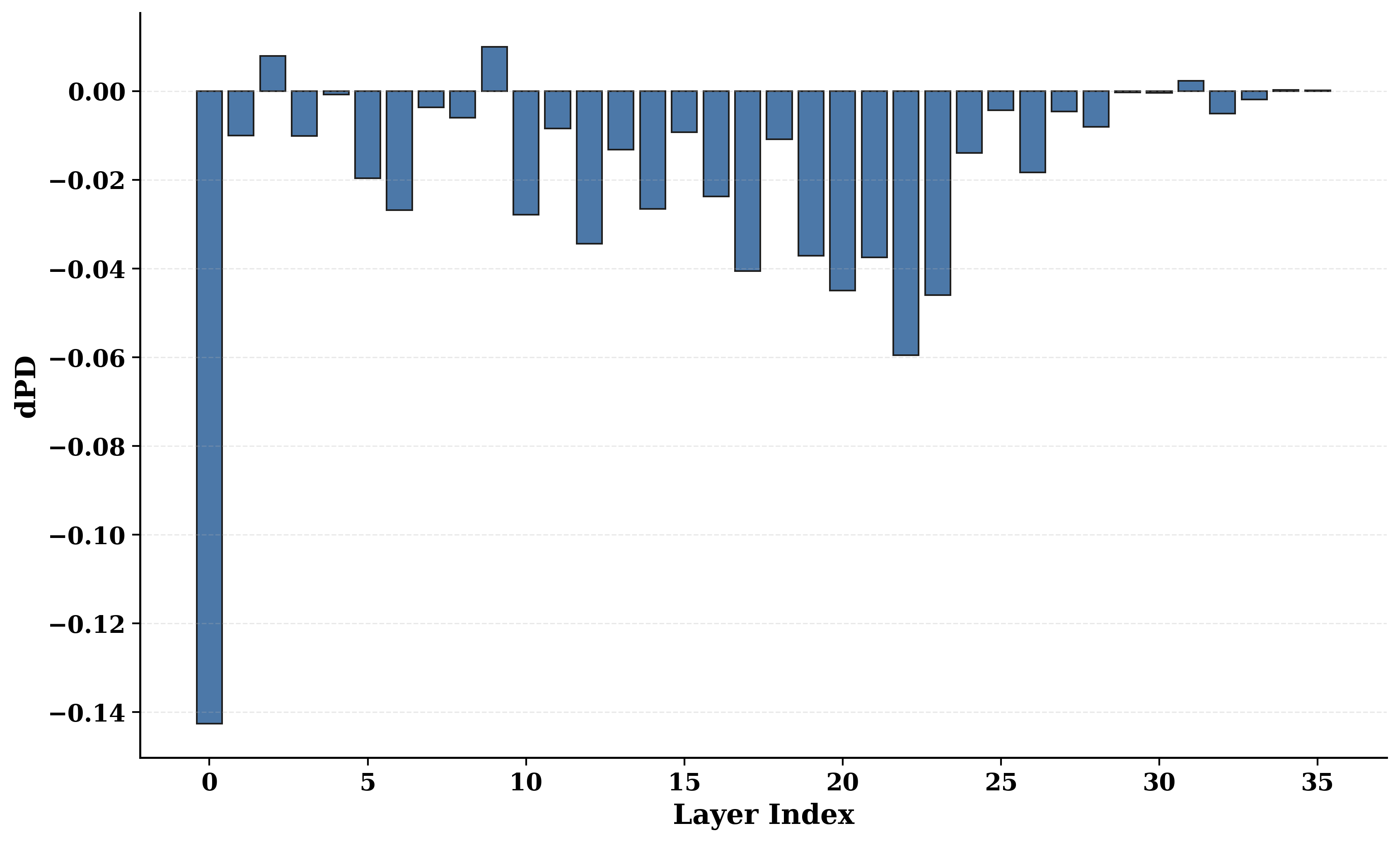}

    \includegraphics[width=0.32\textwidth]{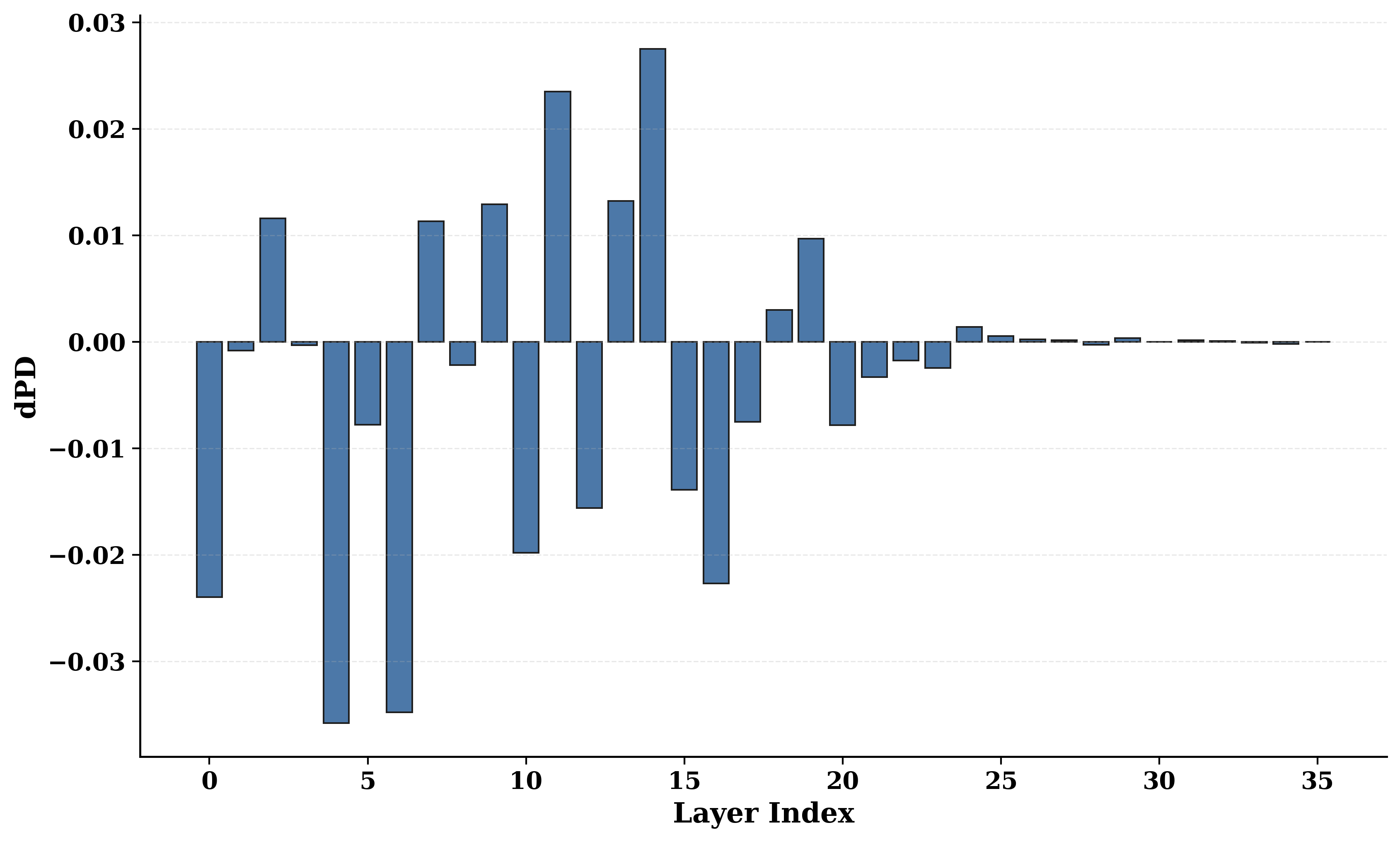}\hfill
    \includegraphics[width=0.32\textwidth]{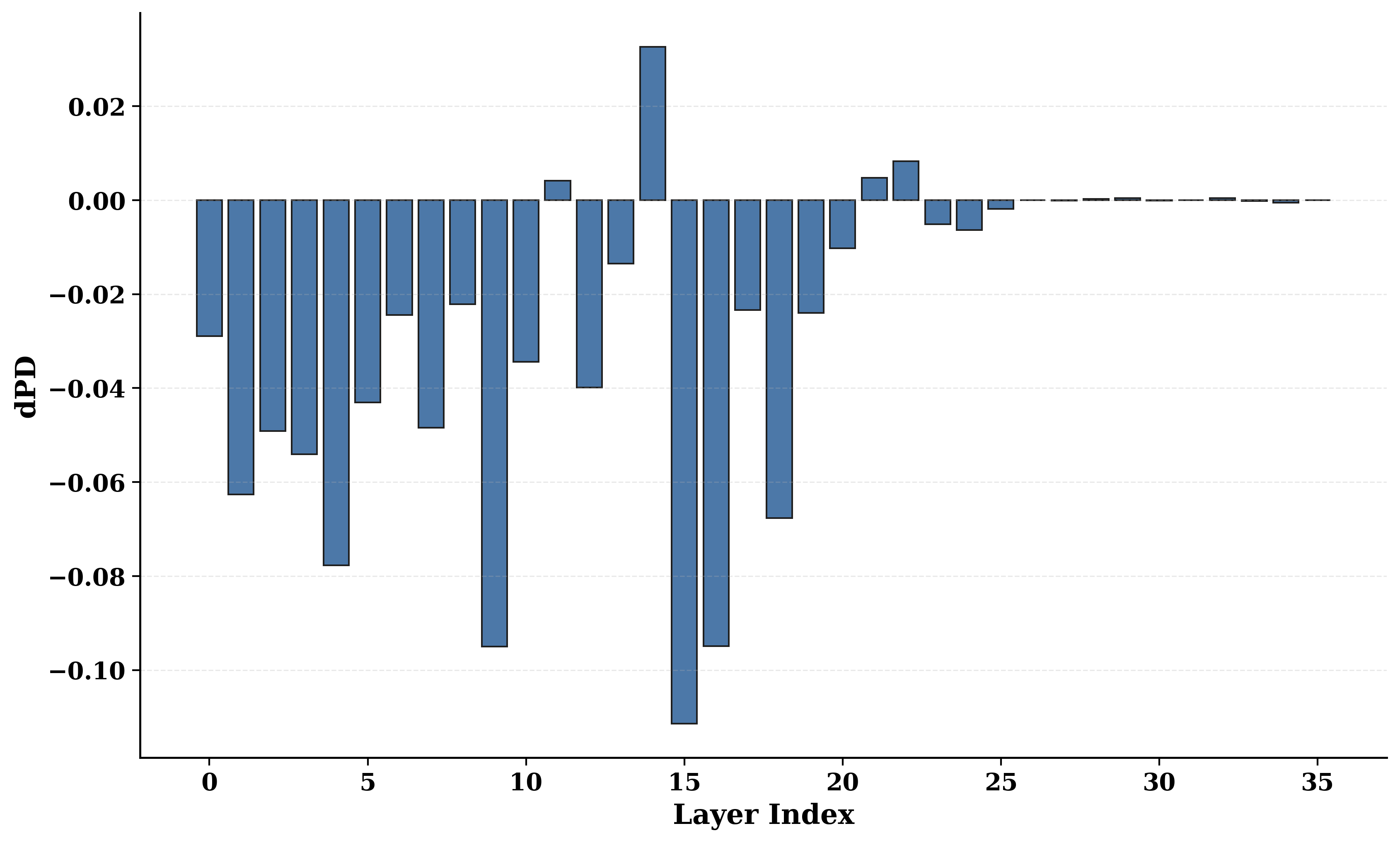}\hfill
    \includegraphics[width=0.32\textwidth]{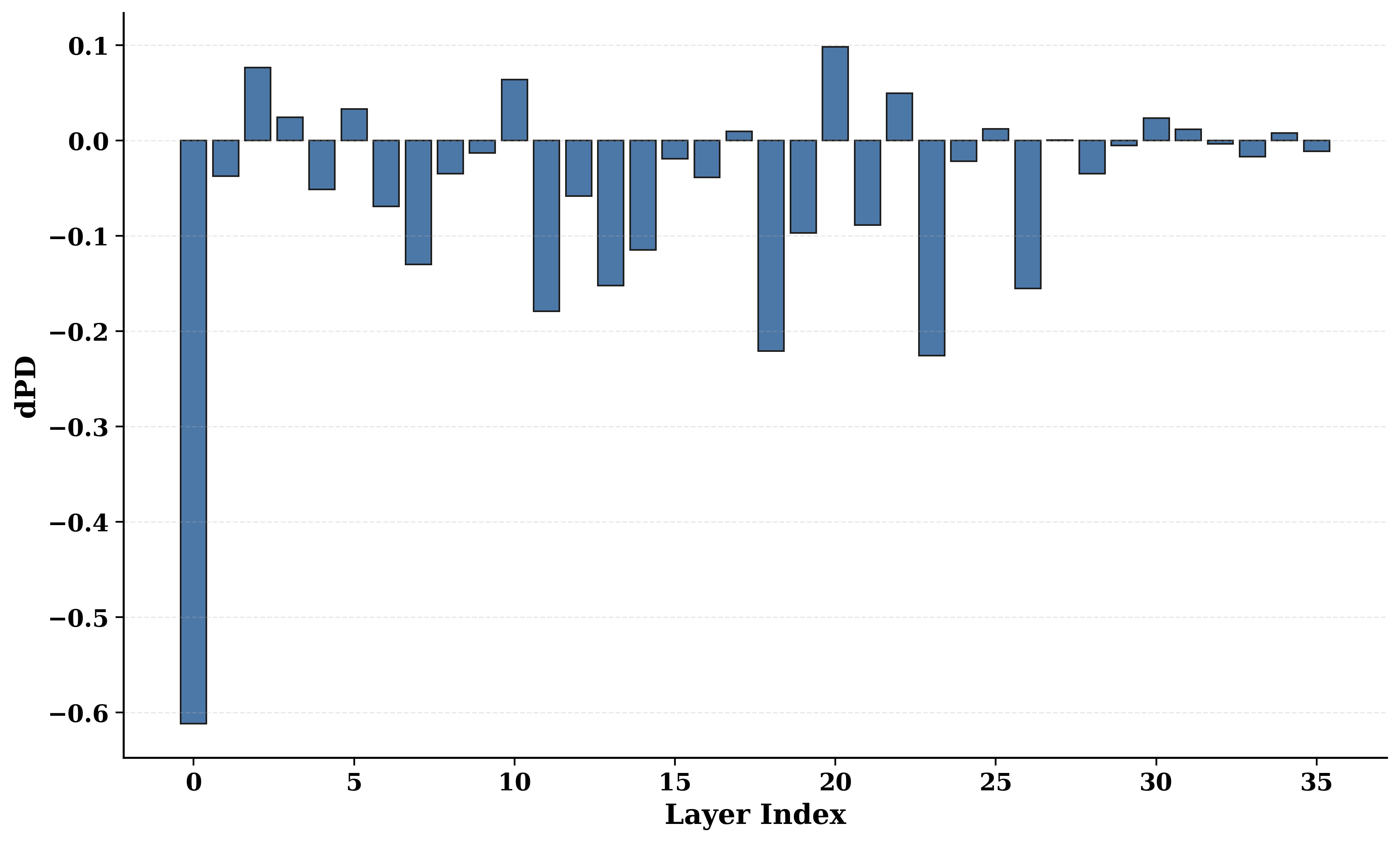}
    \caption{\textbf{Attention-block mean ablation on \textit{Qwen3-8B} under $\pi_{\mathrm{FF}}$.} Per-layer $\mathrm{dPD}$ when each logical block is mean-ablated. Top row: one-hop. Bottom row: two-hop. Columns: Facts / Expression / Query.}
    \label{fig:stage_attn_q8_ff}
\end{figure*}

\begin{figure*}[!htbp]
    \centering
    \includegraphics[width=0.32\textwidth]{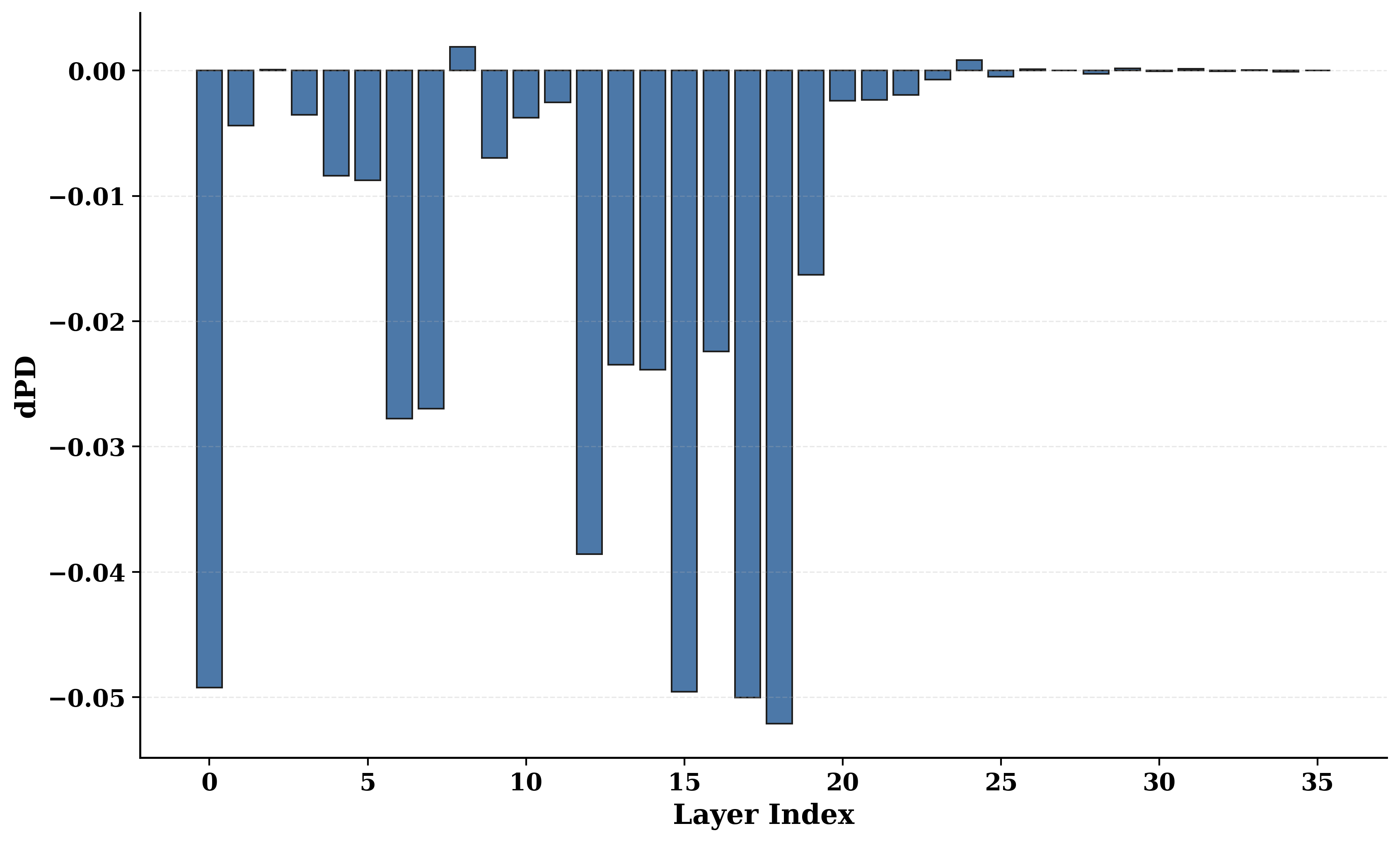}\hfill
    \includegraphics[width=0.32\textwidth]{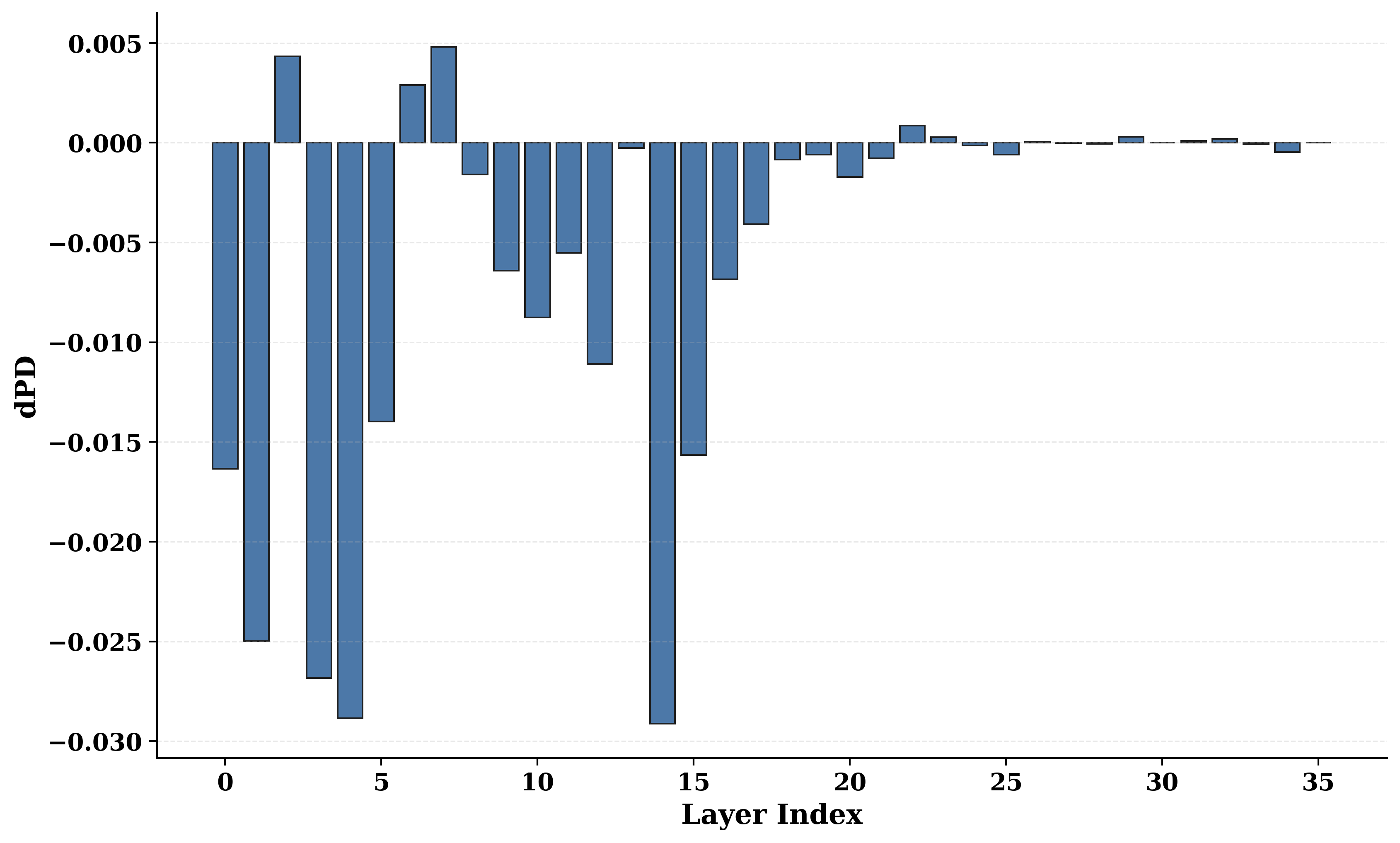}\hfill
    \includegraphics[width=0.32\textwidth]{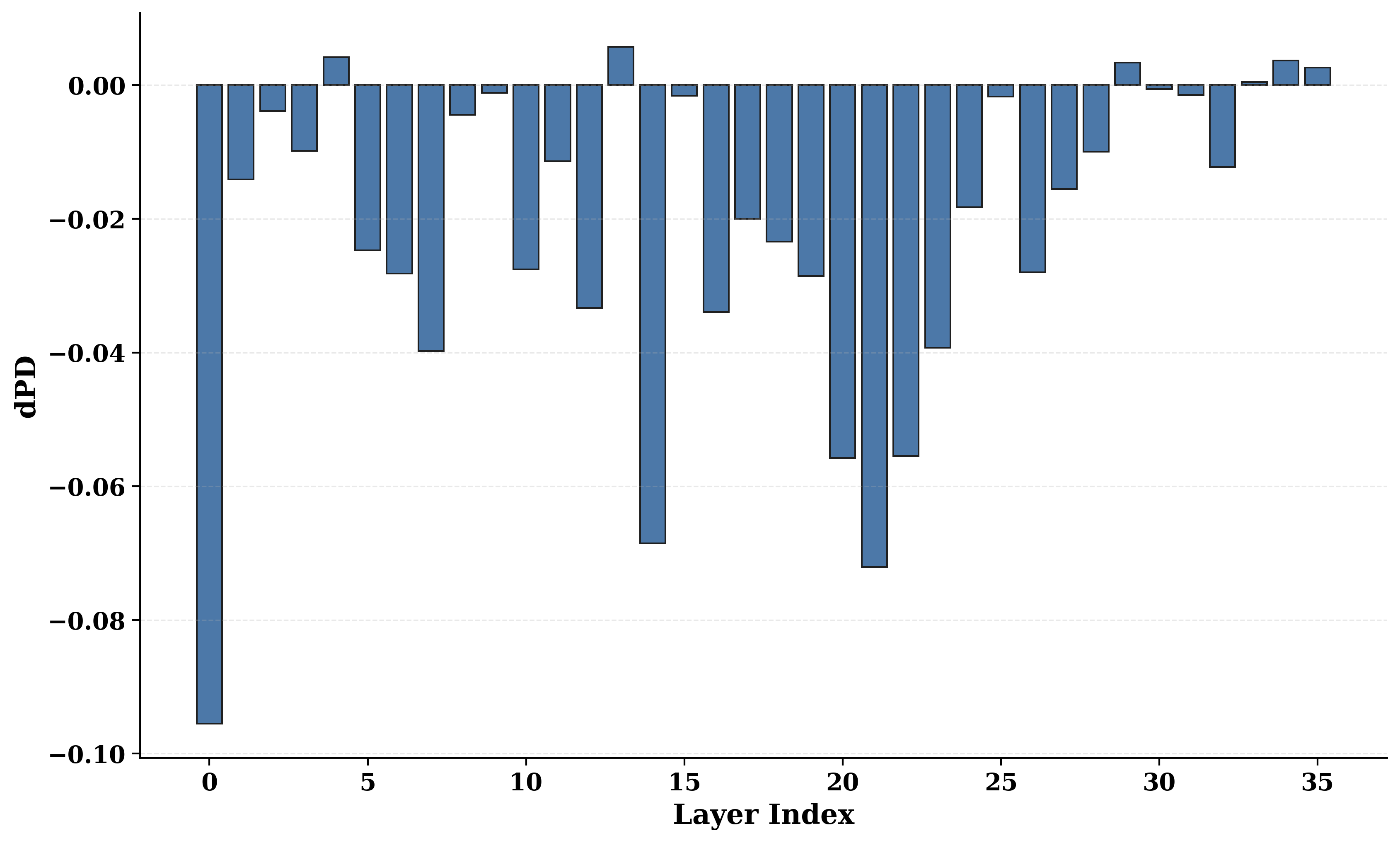}

    \includegraphics[width=0.32\textwidth]{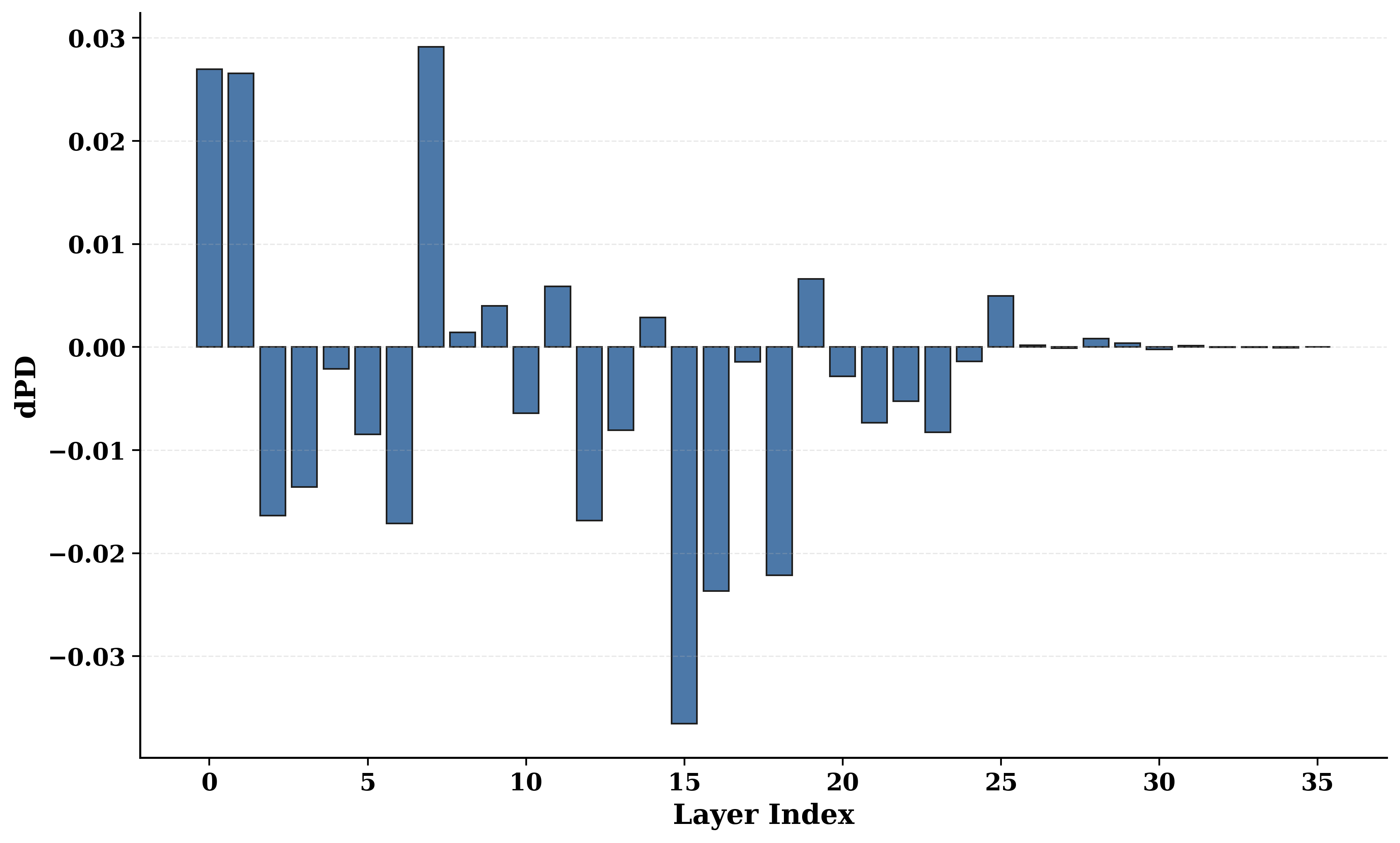}\hfill
    \includegraphics[width=0.32\textwidth]{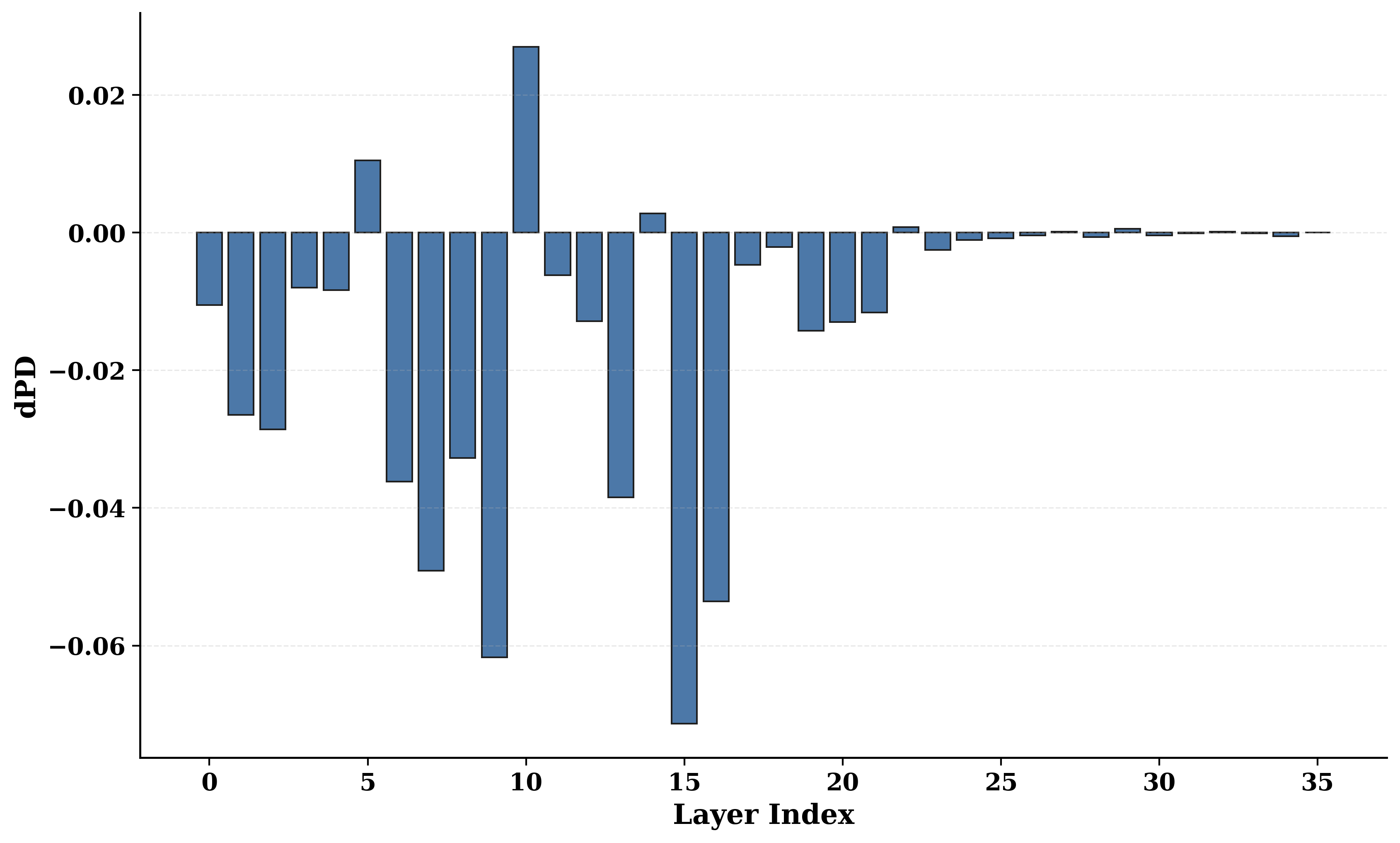}\hfill
    \includegraphics[width=0.32\textwidth]{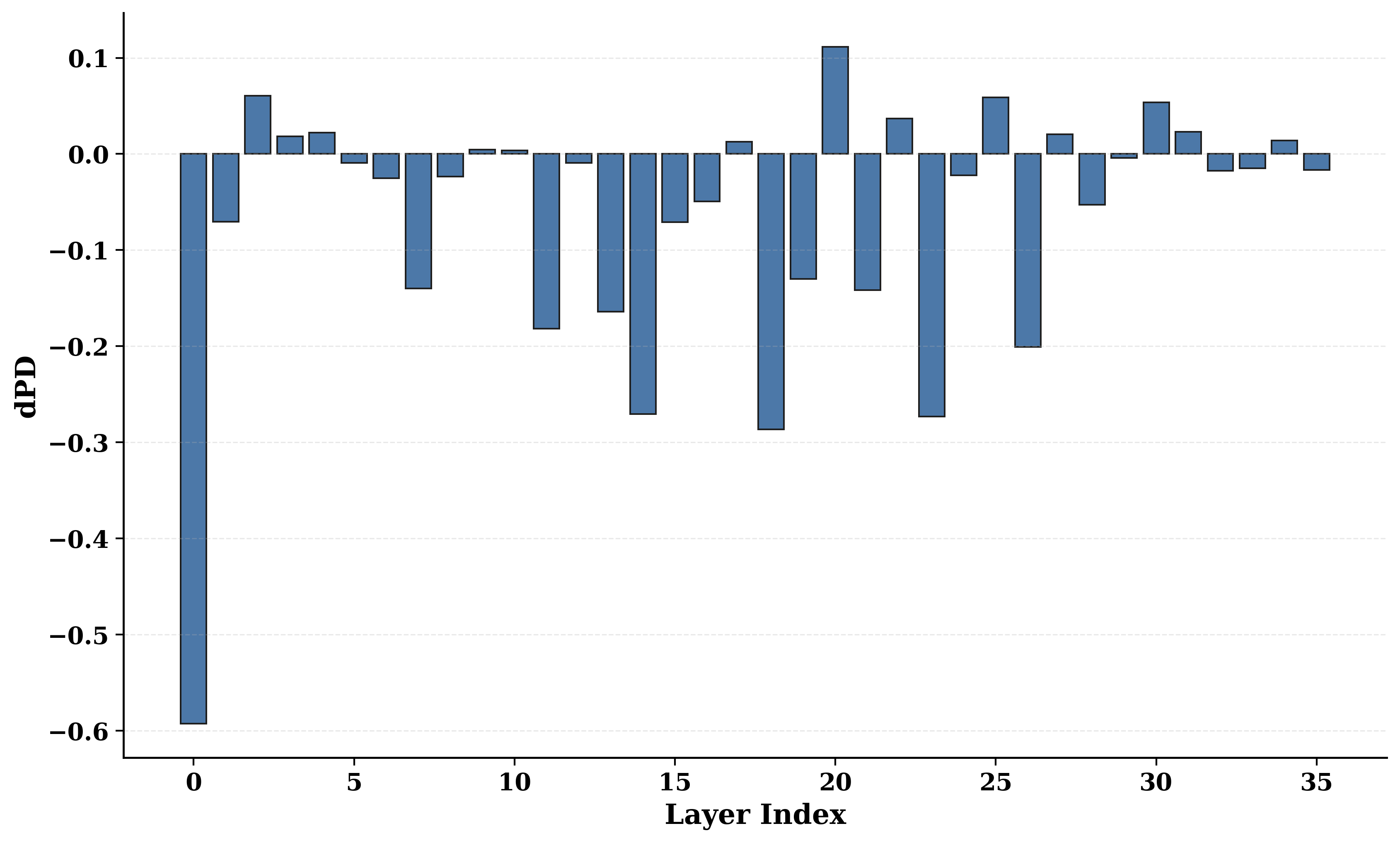}
    \caption{\textbf{Attention-block mean ablation on \textit{Qwen3-8B} under $\pi_{\mathrm{EF}}$.} Per-layer $\mathrm{dPD}$ when each logical block is mean-ablated. Top row: one-hop. Bottom row: two-hop. Columns: Facts / Expression / Query.}
    \label{fig:stage_attn_q8_ef}
\end{figure*}

\begin{figure*}[!htbp]
    \centering
    \includegraphics[width=0.32\textwidth]{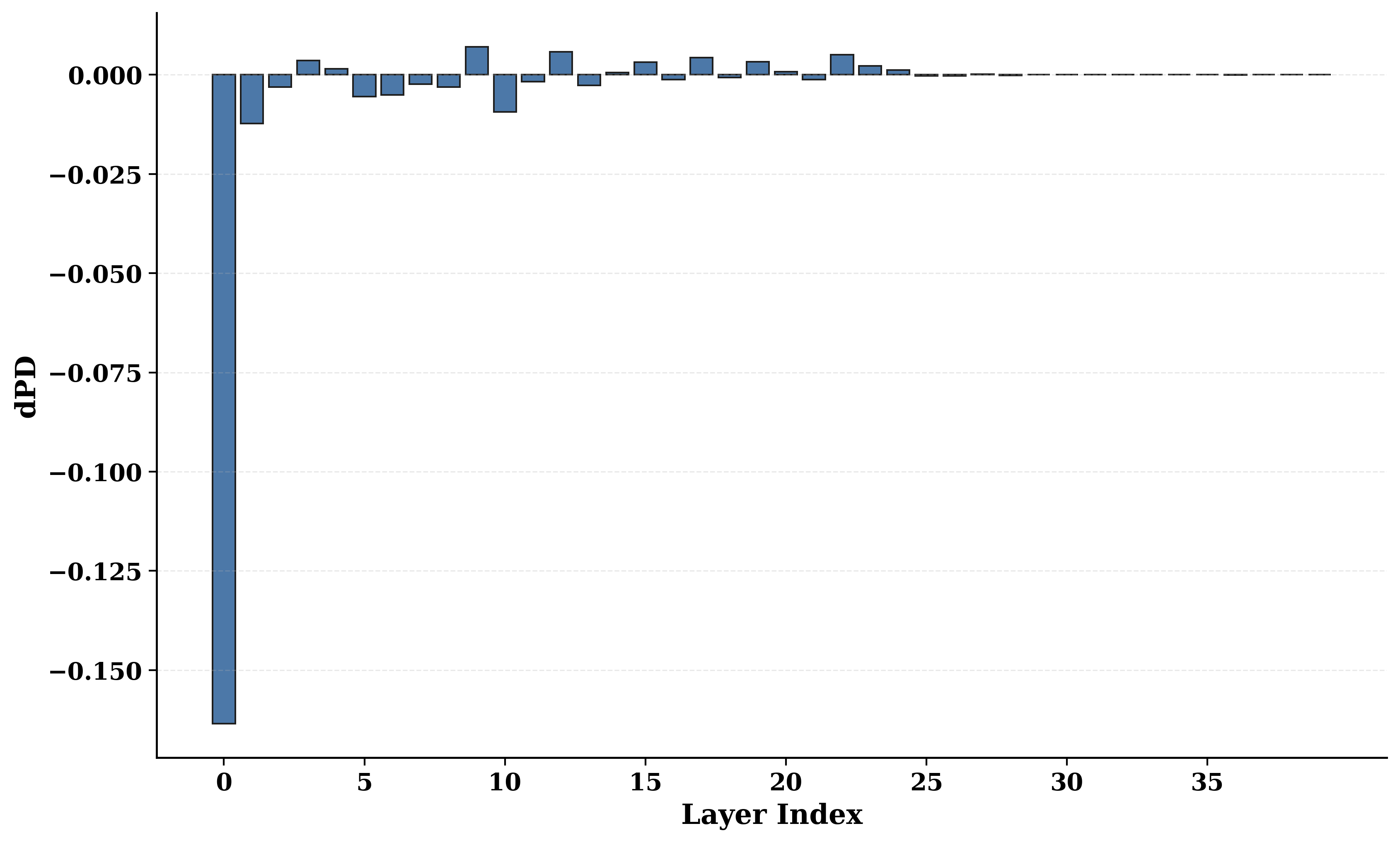}\hfill
    \includegraphics[width=0.32\textwidth]{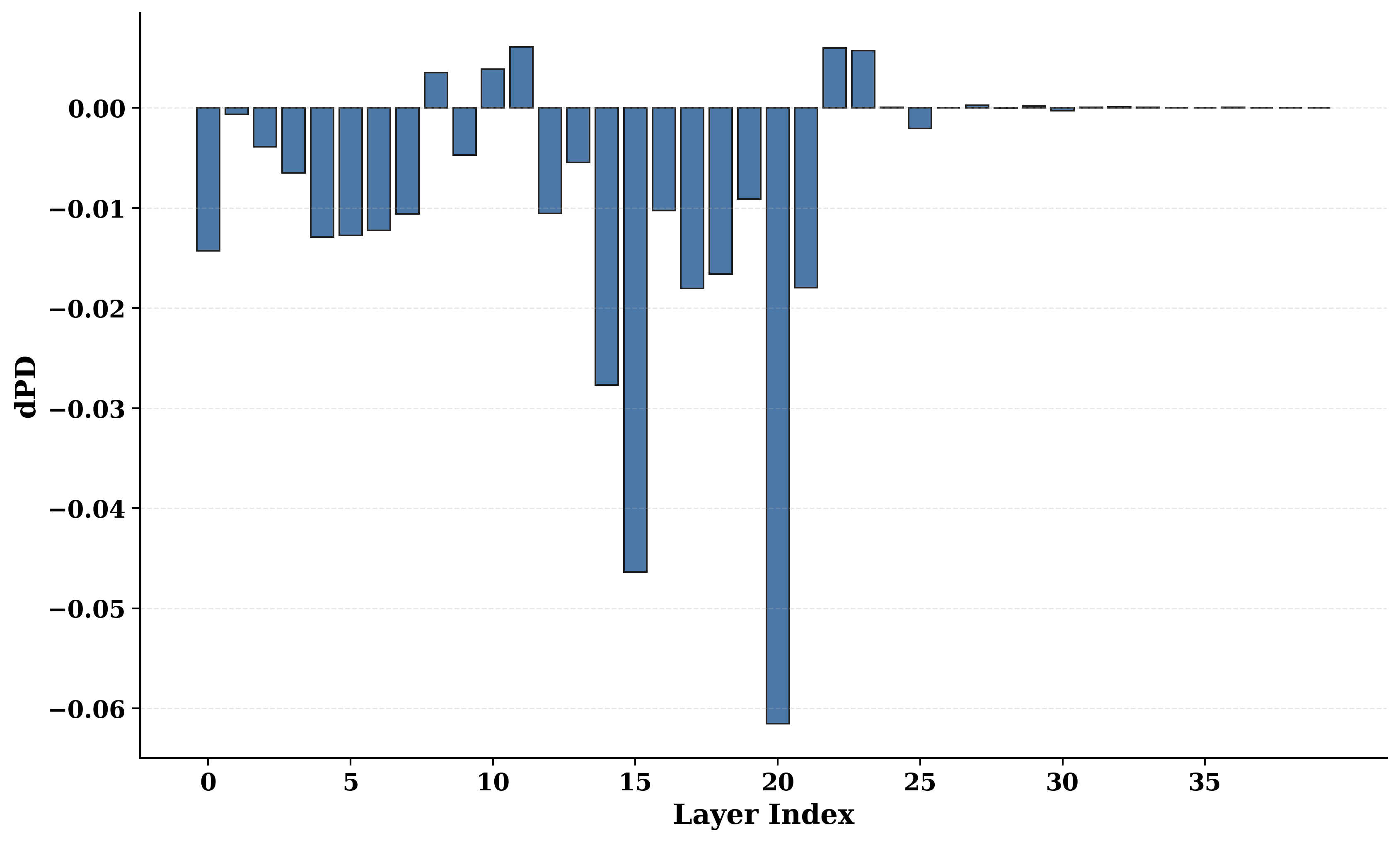}\hfill
    \includegraphics[width=0.32\textwidth]{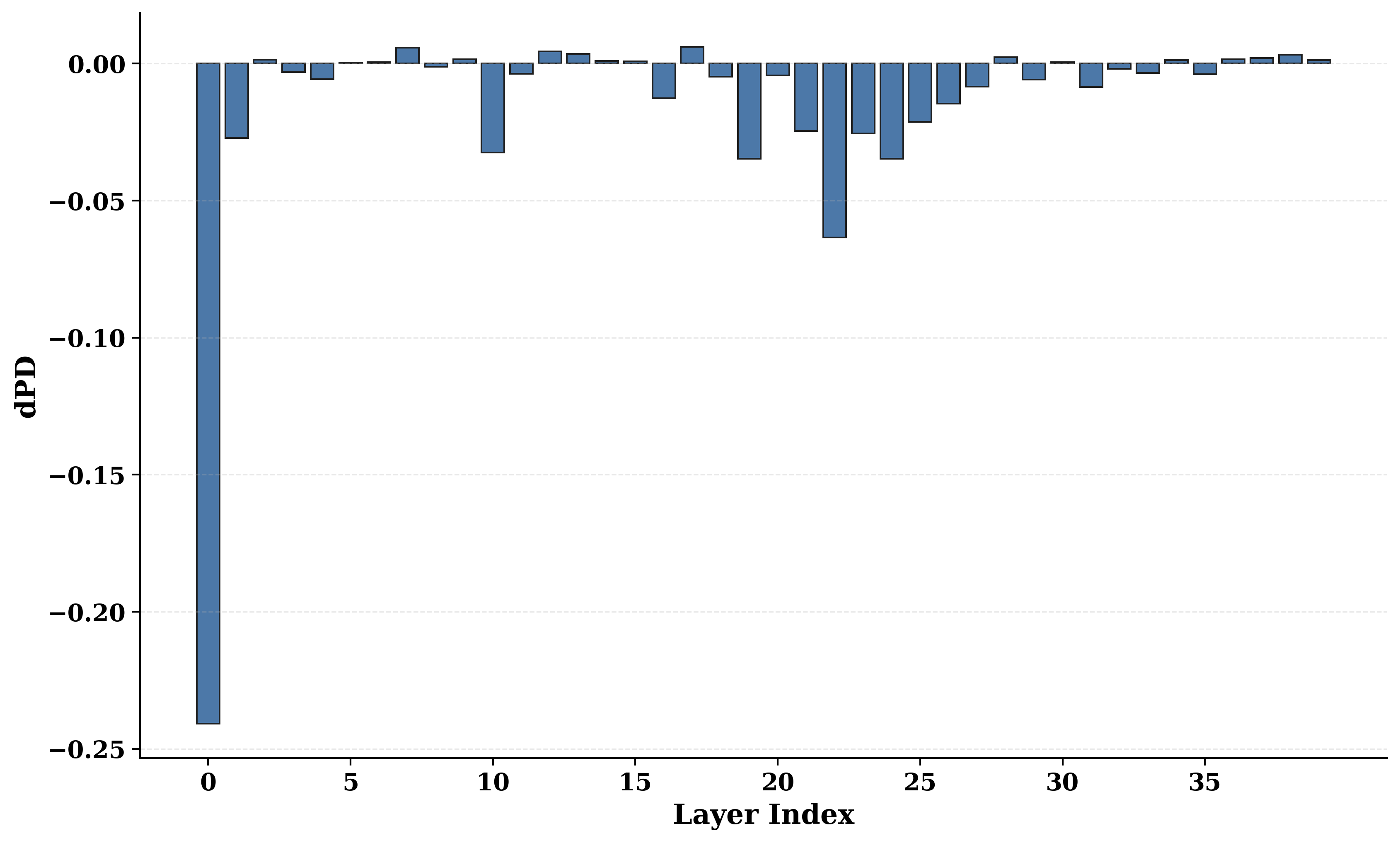}

    \includegraphics[width=0.32\textwidth]{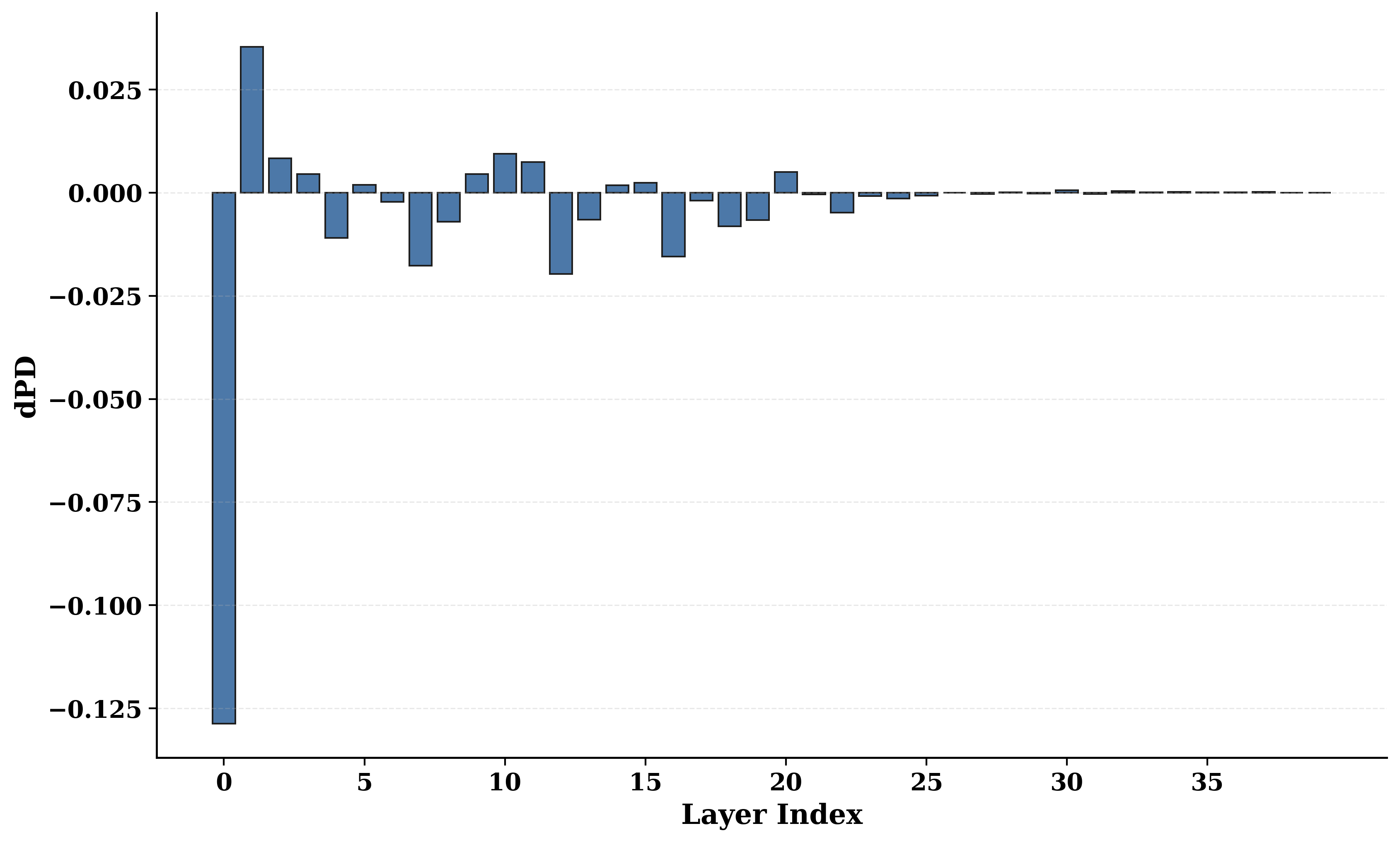}\hfill
    \includegraphics[width=0.32\textwidth]{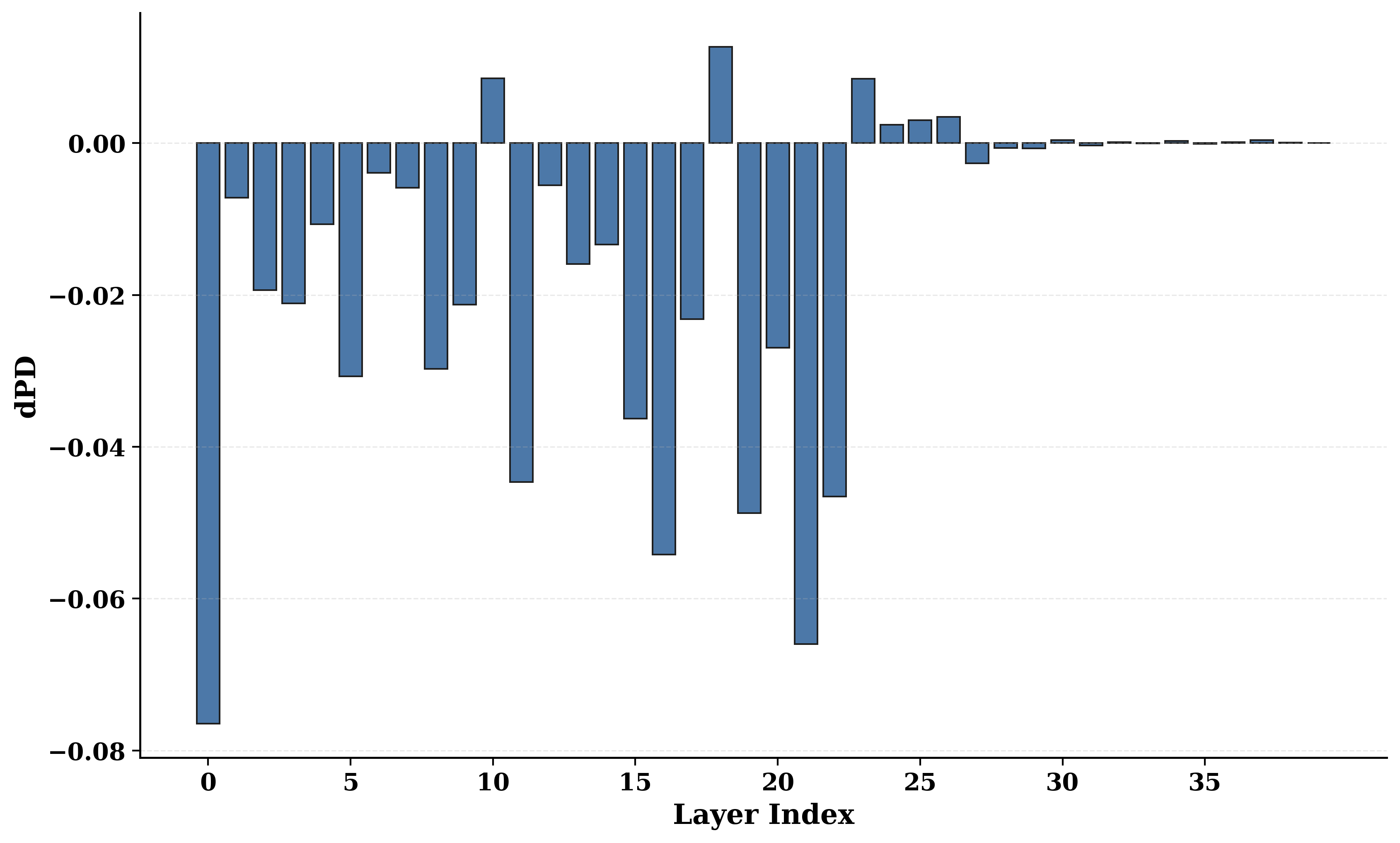}\hfill
    \includegraphics[width=0.32\textwidth]{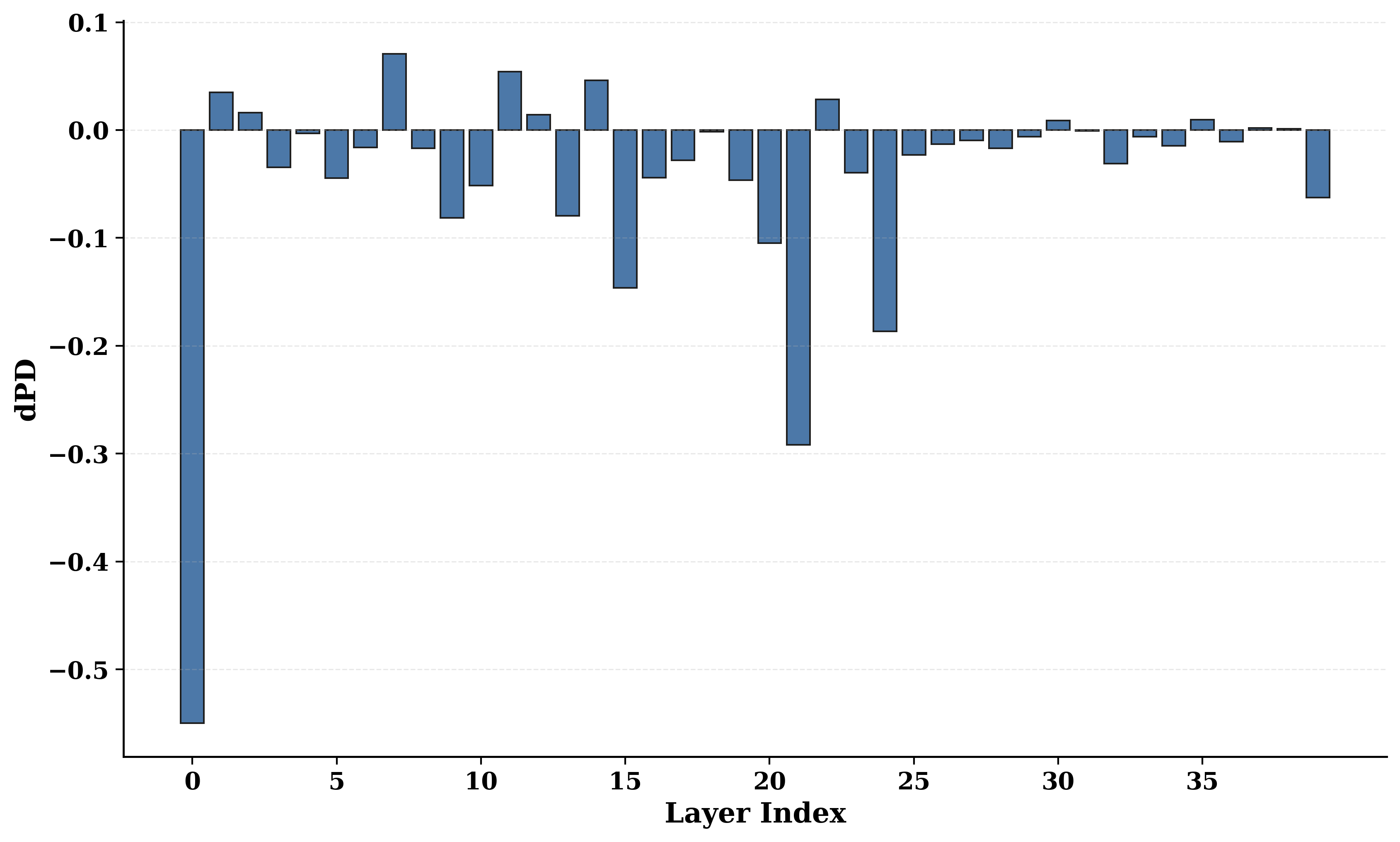}
    \caption{\textbf{Attention-block mean ablation on \textit{Qwen3-14B} under $\pi_{\mathrm{FF}}$.} Per-layer $\mathrm{dPD}$ when each logical block is mean-ablated. Top row: one-hop. Bottom row: two-hop. Columns: Facts / Expression / Query.}
    \label{fig:stage_attn_q14_ff}
    \label{fig:attn_region_triples}
\end{figure*}

\begin{figure*}[!htbp]
    \centering
    \includegraphics[width=0.32\textwidth]{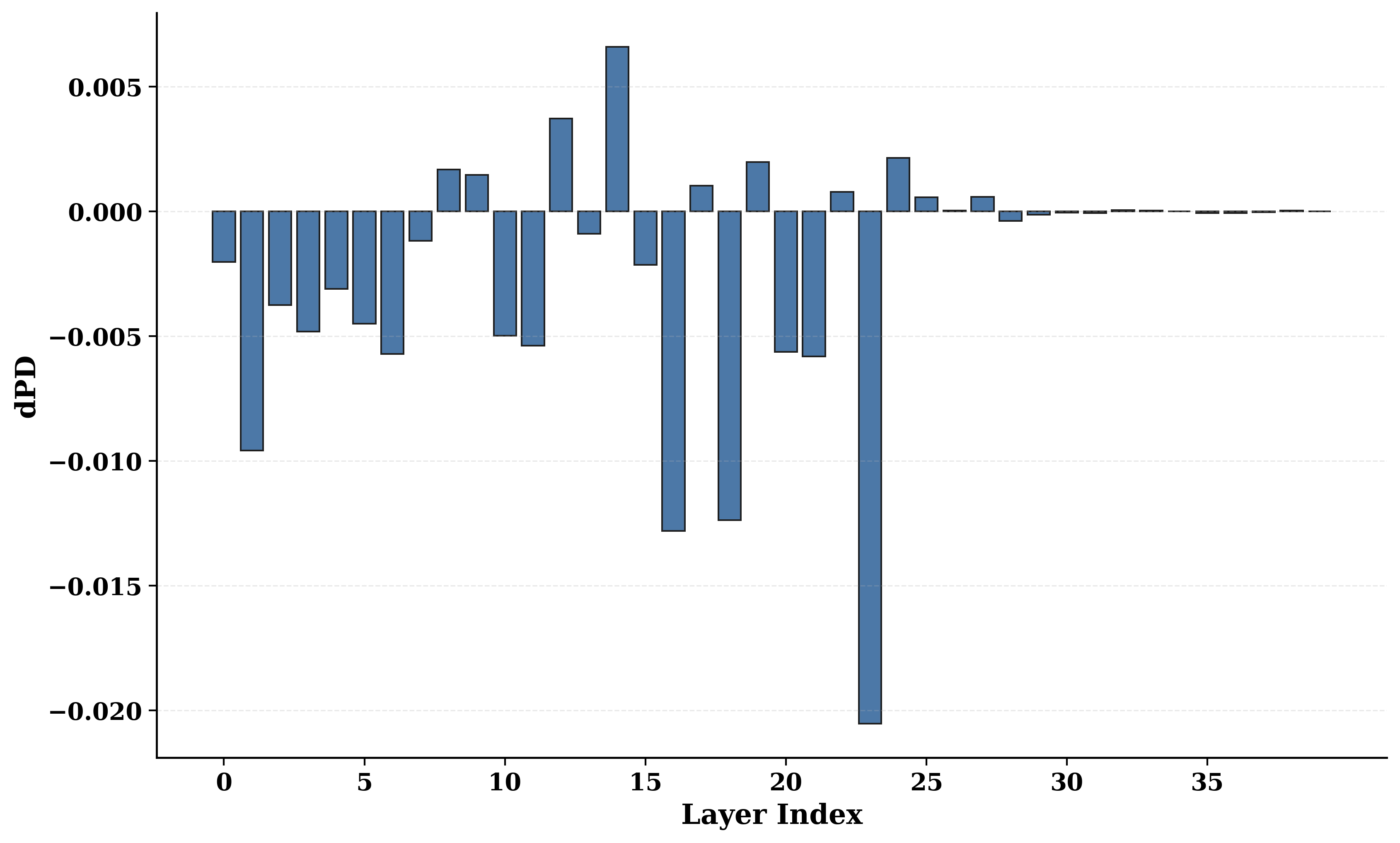}\hfill
    \includegraphics[width=0.32\textwidth]{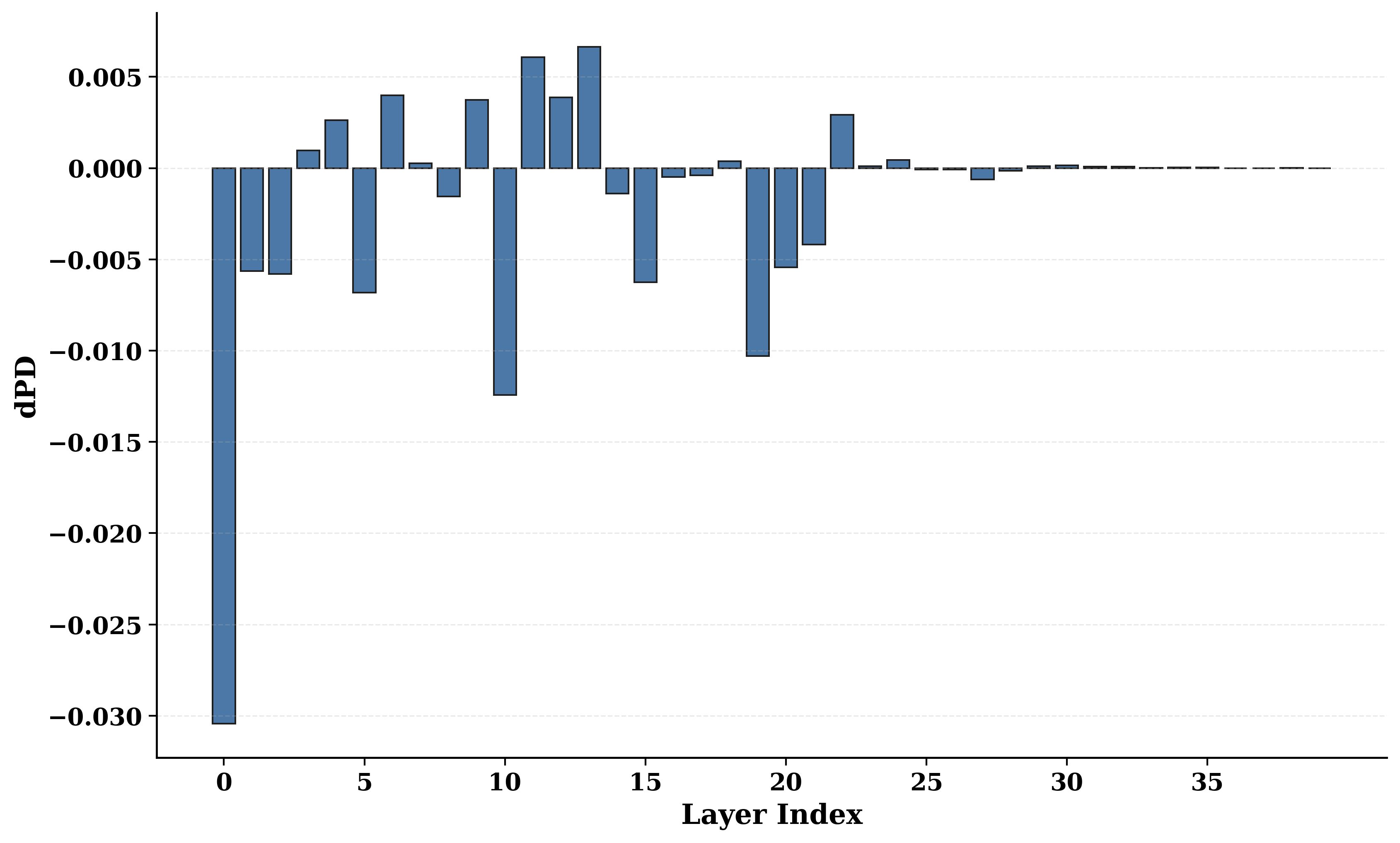}\hfill
    \includegraphics[width=0.32\textwidth]{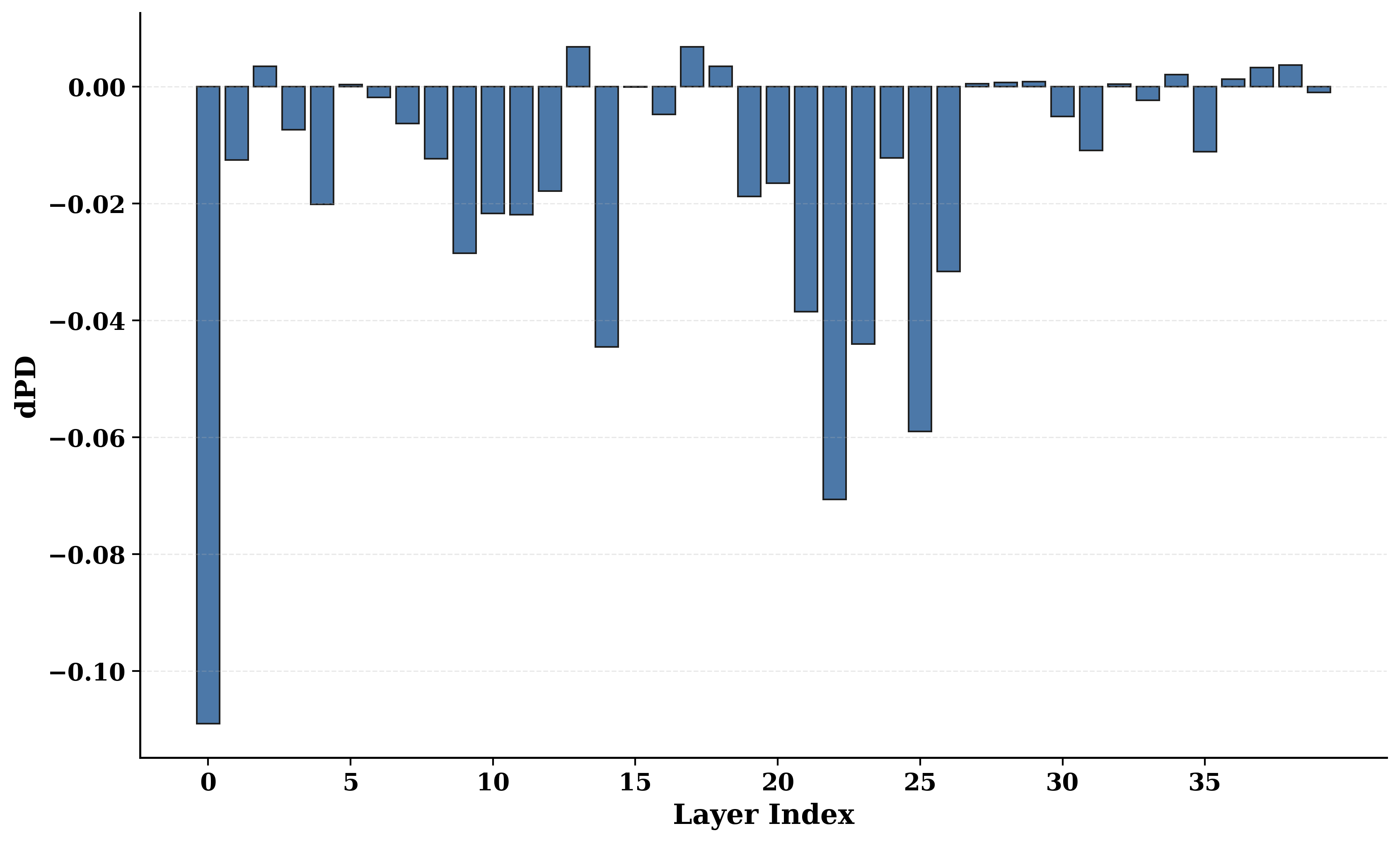}

    \includegraphics[width=0.32\textwidth]{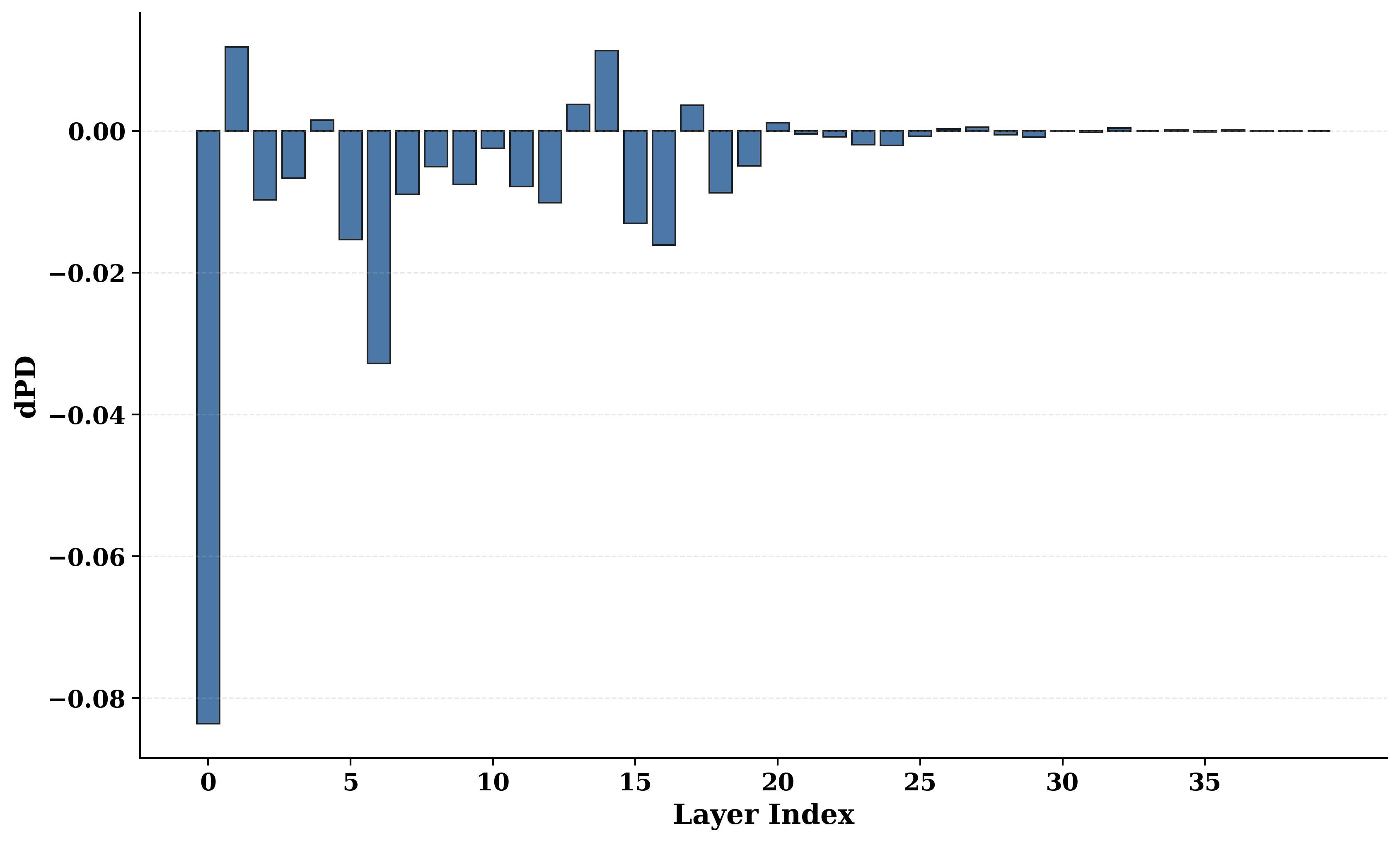}\hfill
    \includegraphics[width=0.32\textwidth]{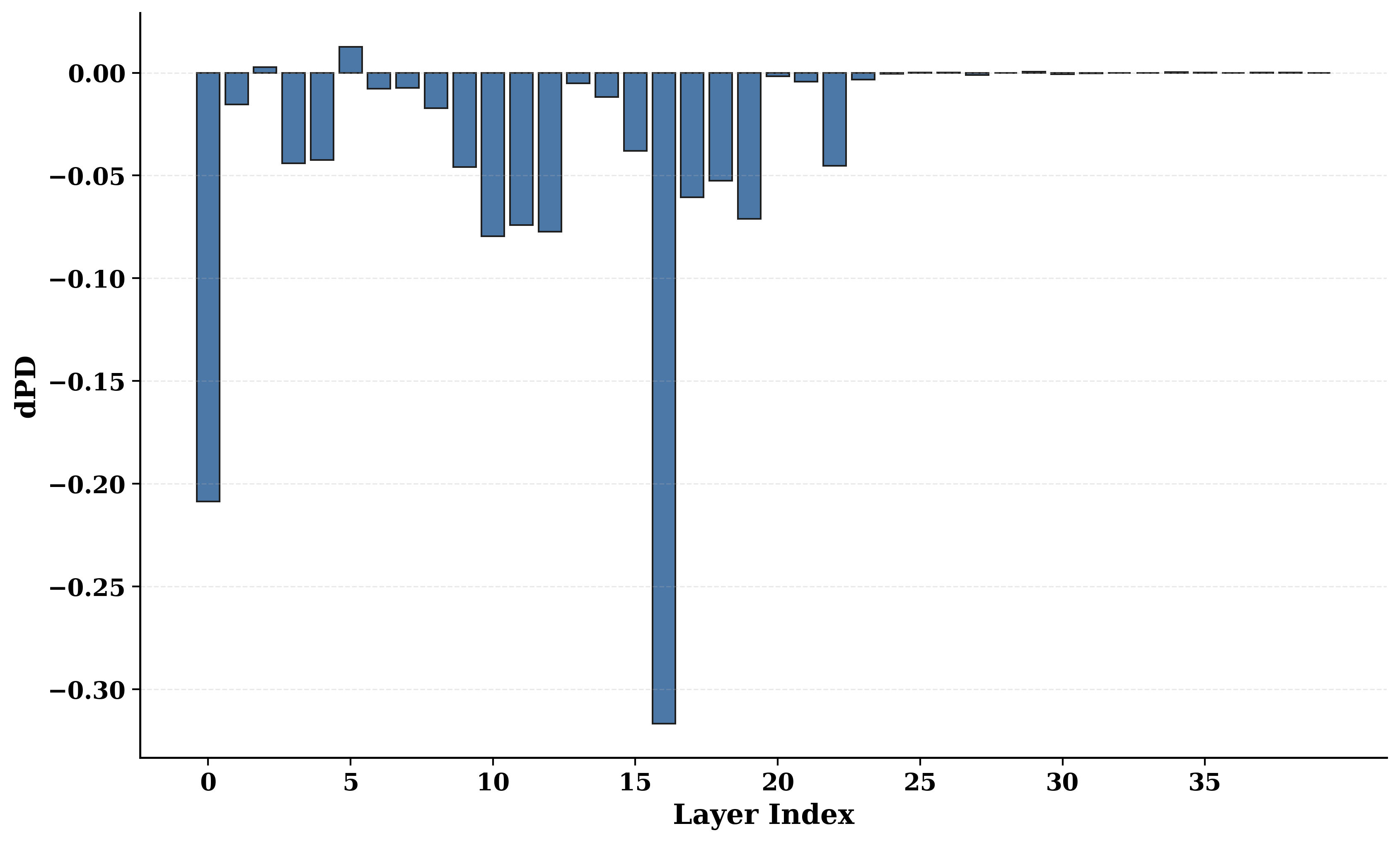}\hfill
    \includegraphics[width=0.32\textwidth]{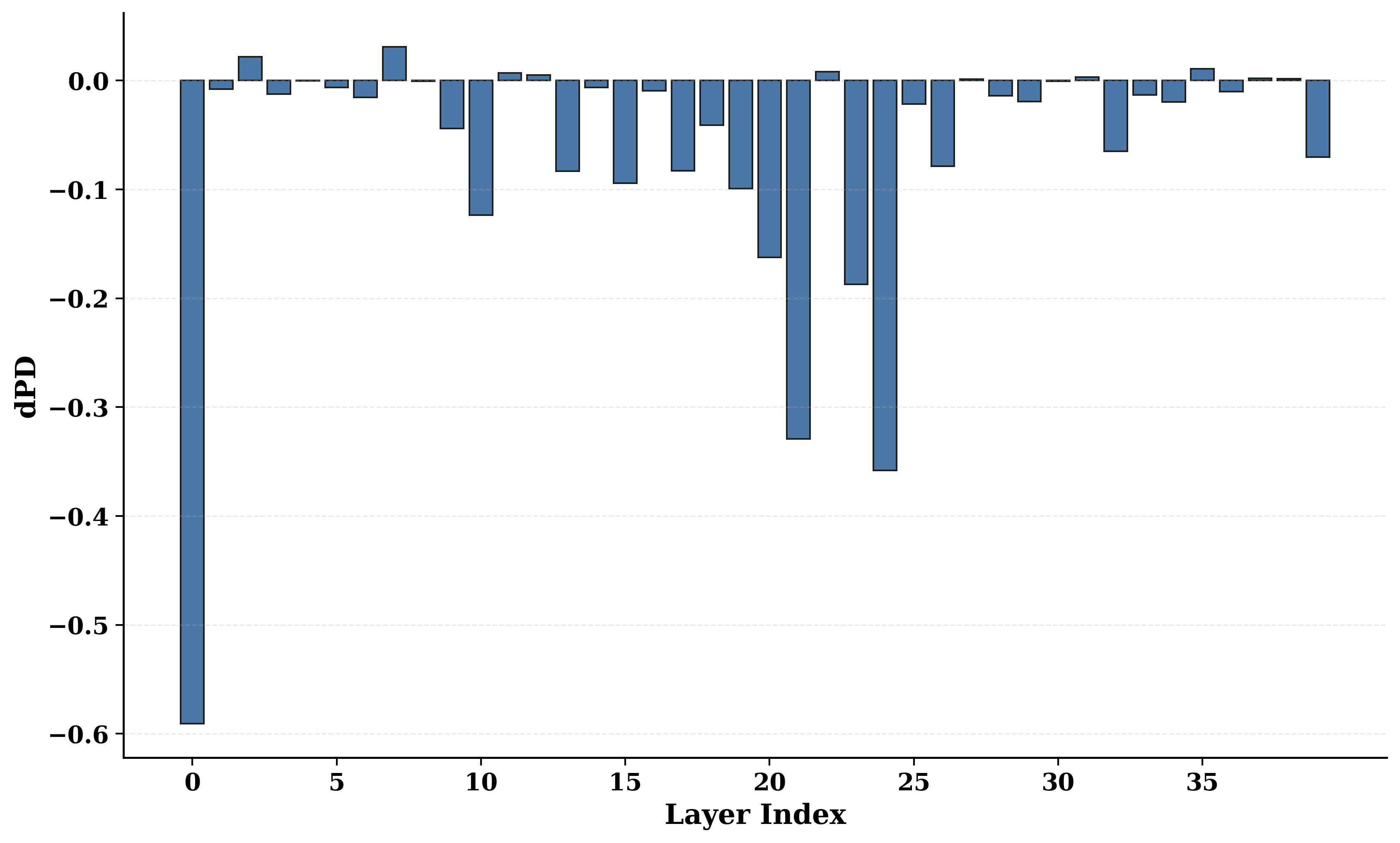}
    \caption{\textbf{Attention-block mean ablation on \textit{Qwen3-14B} under $\pi_{\mathrm{EF}}$.} Per-layer $\mathrm{dPD}$ when each logical block is mean-ablated. Top row: one-hop. Bottom row: two-hop. Columns: Facts / Expression / Query.}
    \label{fig:stage_attn_q14_ef}
\end{figure*}

\begin{figure*}[!htbp]
    \centering
    \includegraphics[width=0.32\textwidth]{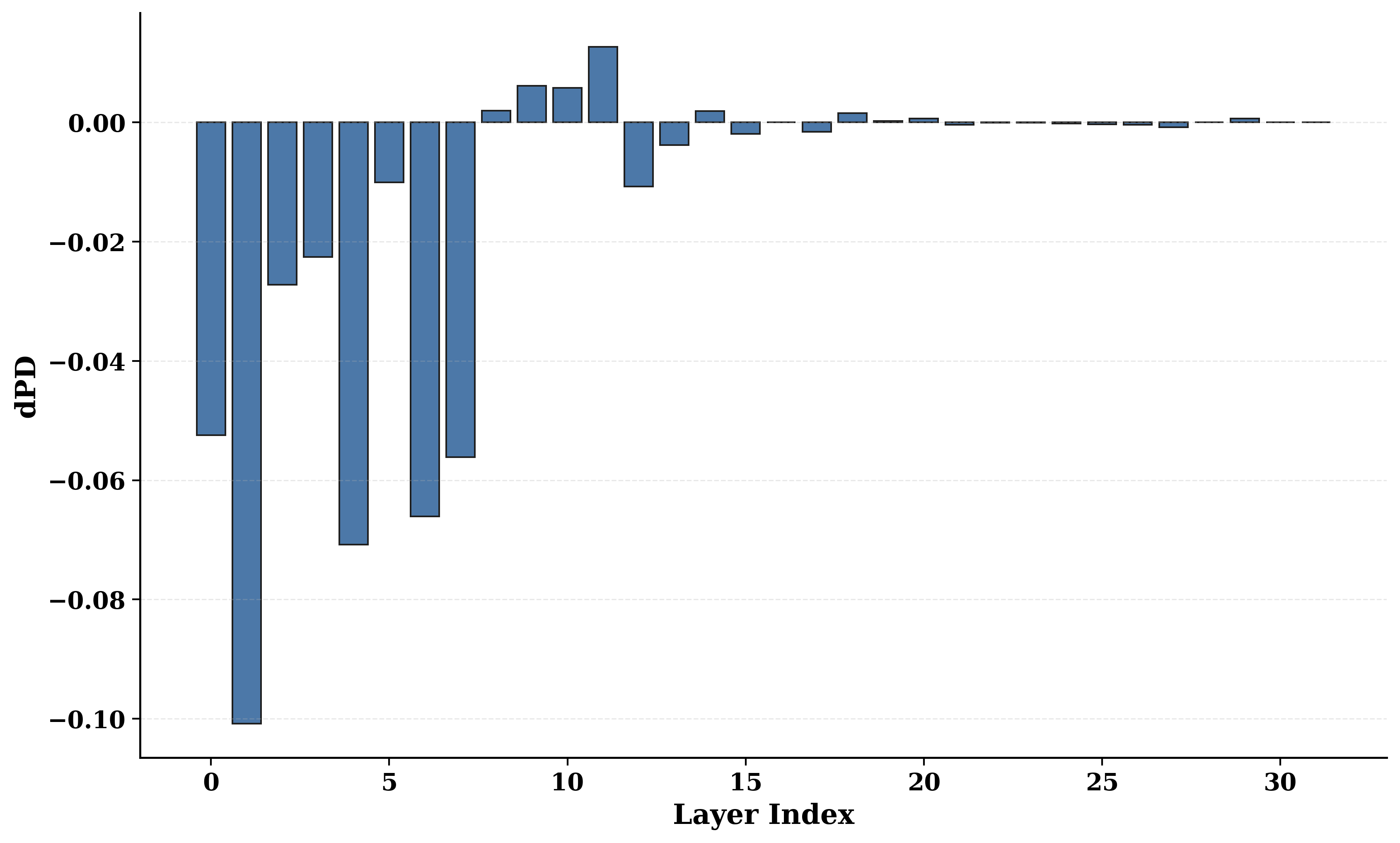}\hfill
    \includegraphics[width=0.32\textwidth]{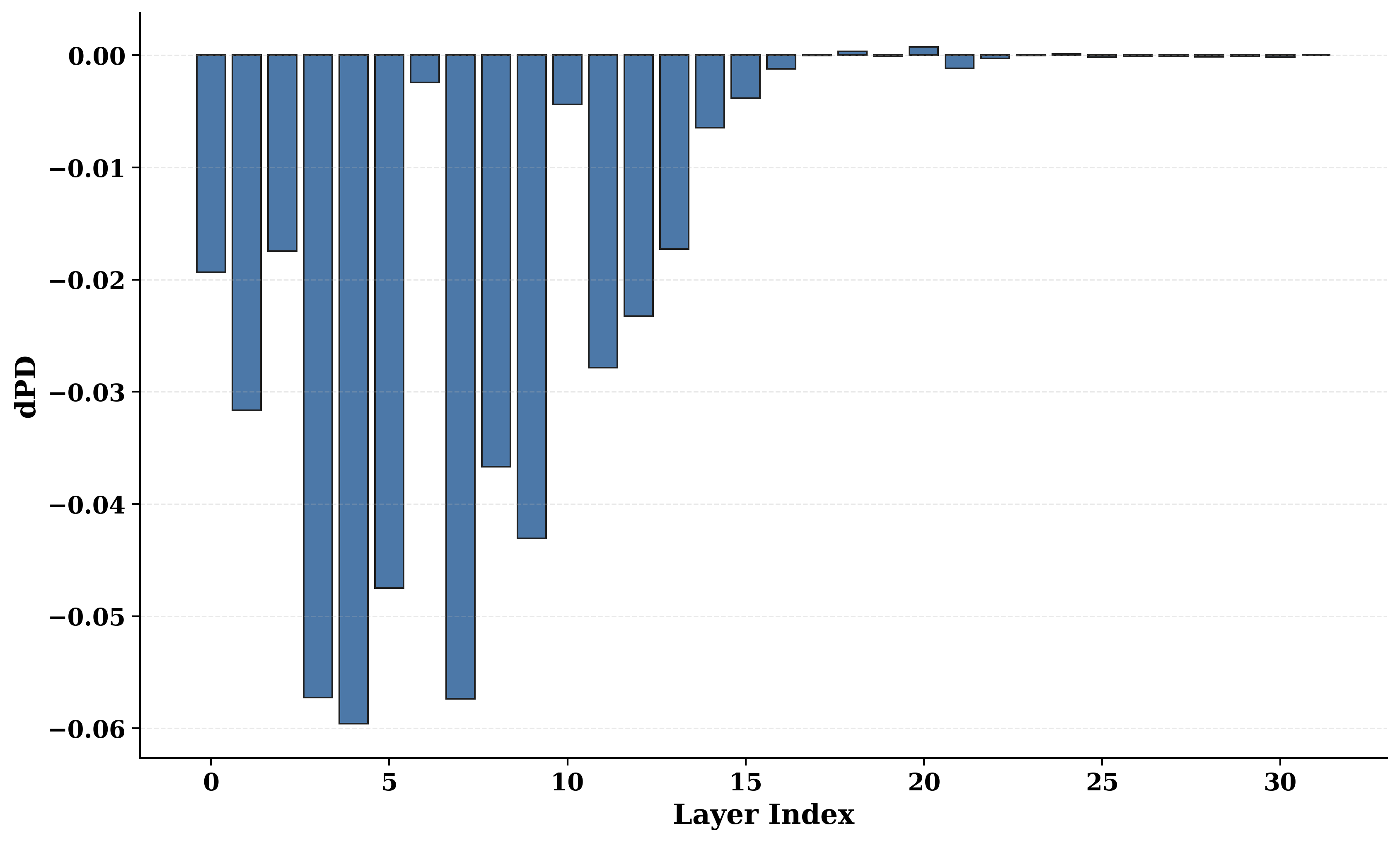}\hfill
    \includegraphics[width=0.32\textwidth]{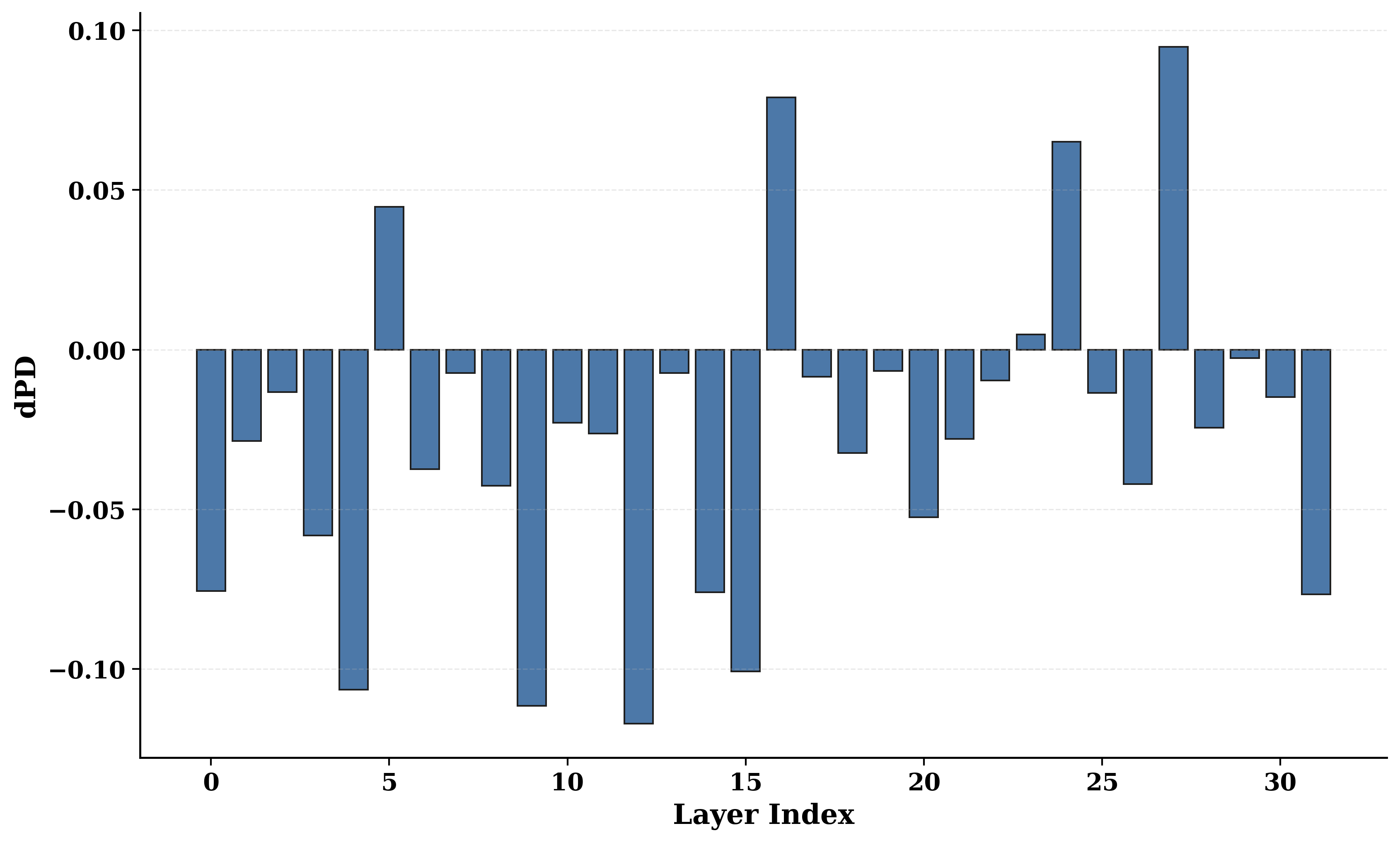}

    \includegraphics[width=0.32\textwidth]{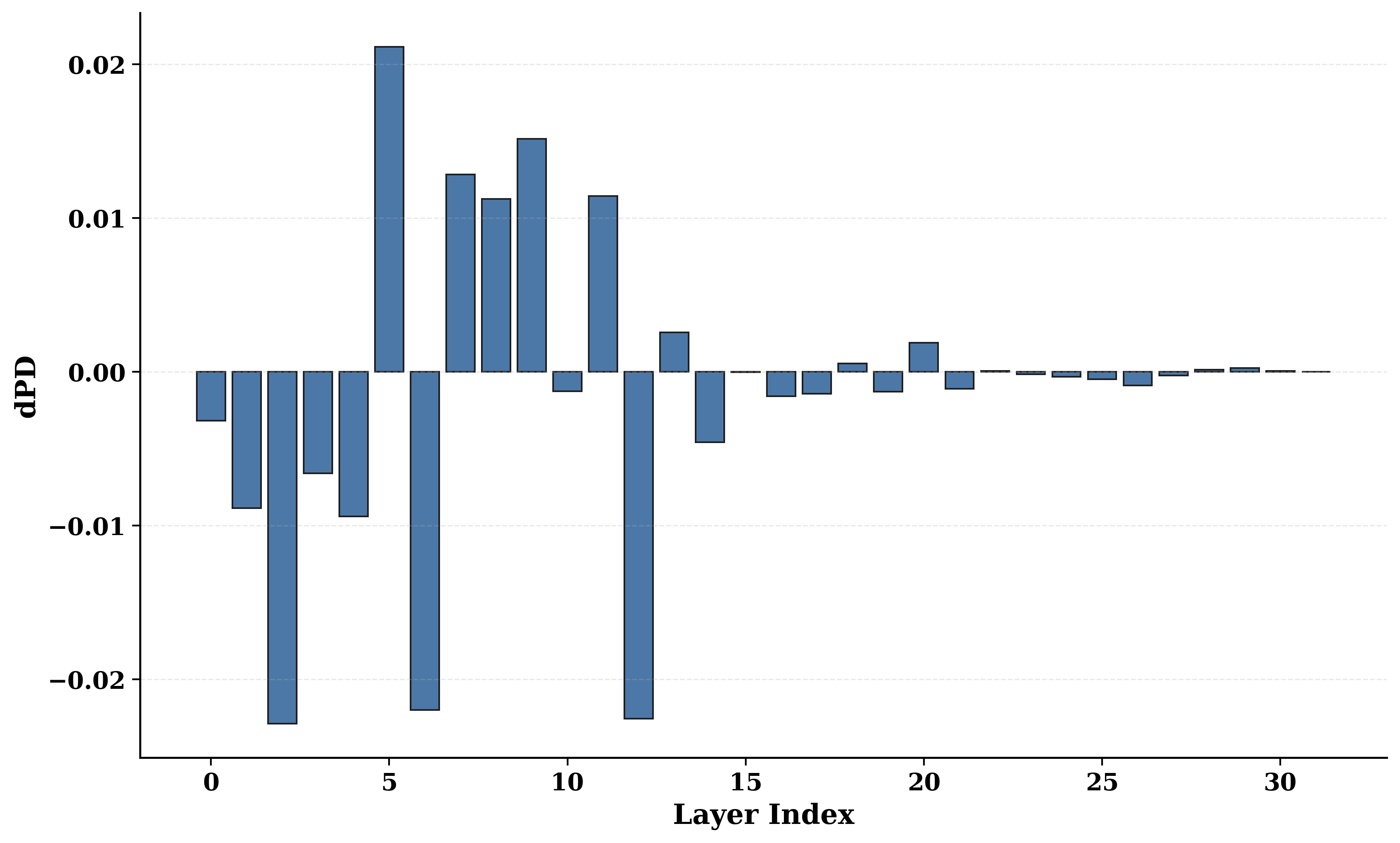}\hfill
    \includegraphics[width=0.32\textwidth]{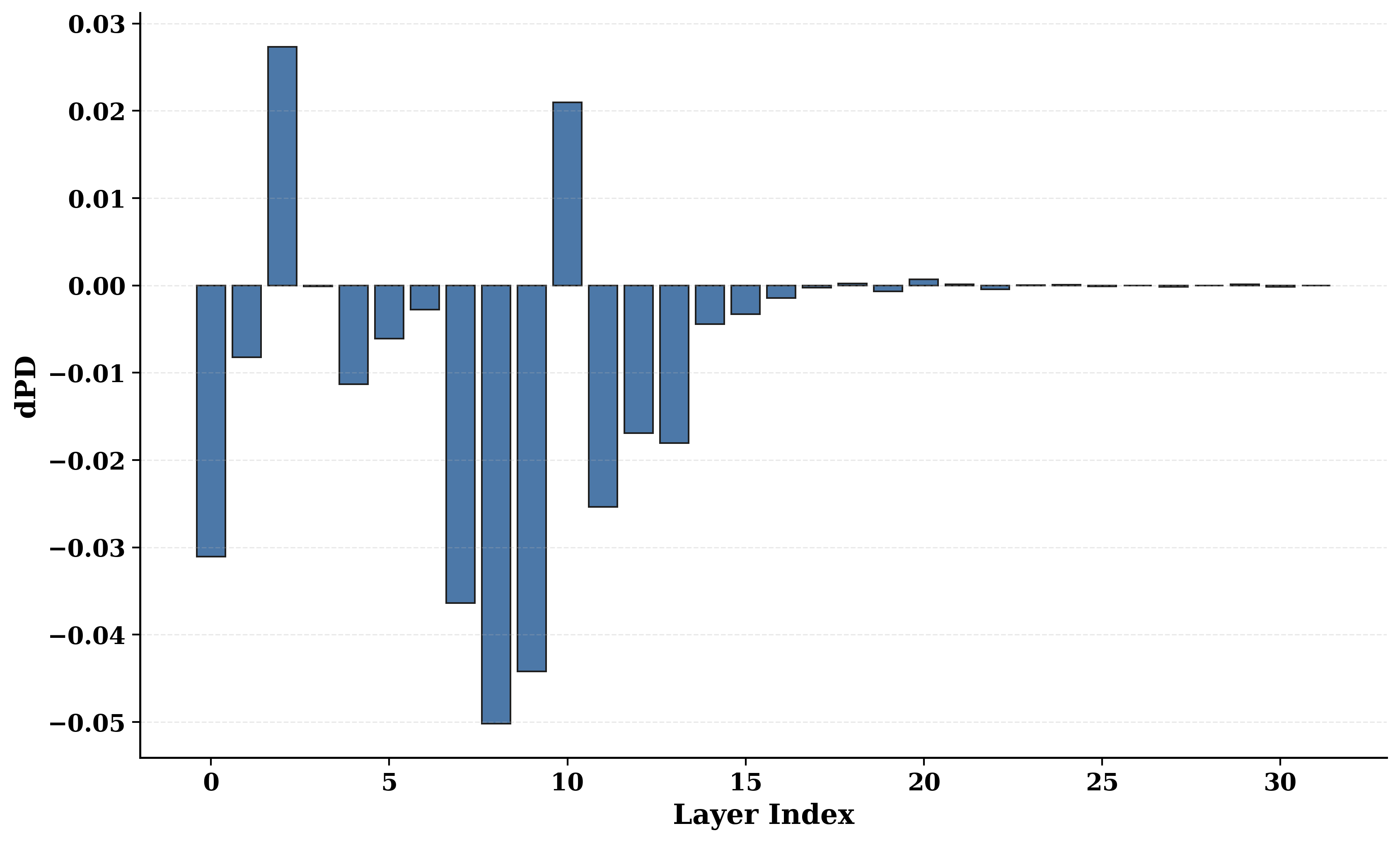}\hfill
    \includegraphics[width=0.32\textwidth]{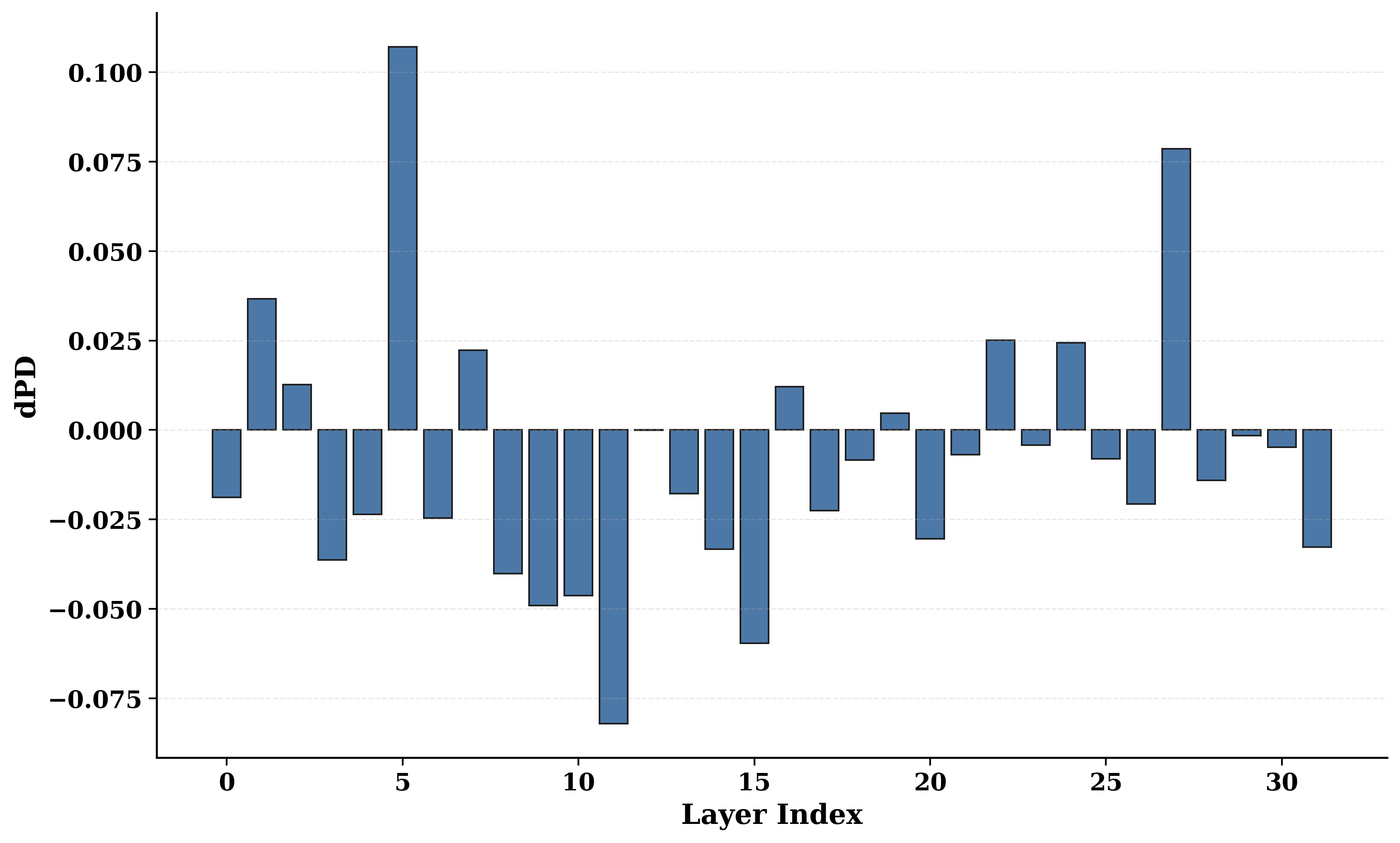}
    \caption{\textbf{Attention-block mean ablation on \textit{Llama-3.1-8B-Instruct} under $\pi_{\mathrm{FF}}$.} Per-layer $\mathrm{dPD}$ when each logical block is mean-ablated. Top row: one-hop. Bottom row: two-hop. Columns: Facts / Expression / Query.}
    \label{fig:stage_attn_llama_ff}
\end{figure*}

\begin{figure*}[!htbp]
    \centering
    \includegraphics[width=0.32\textwidth]{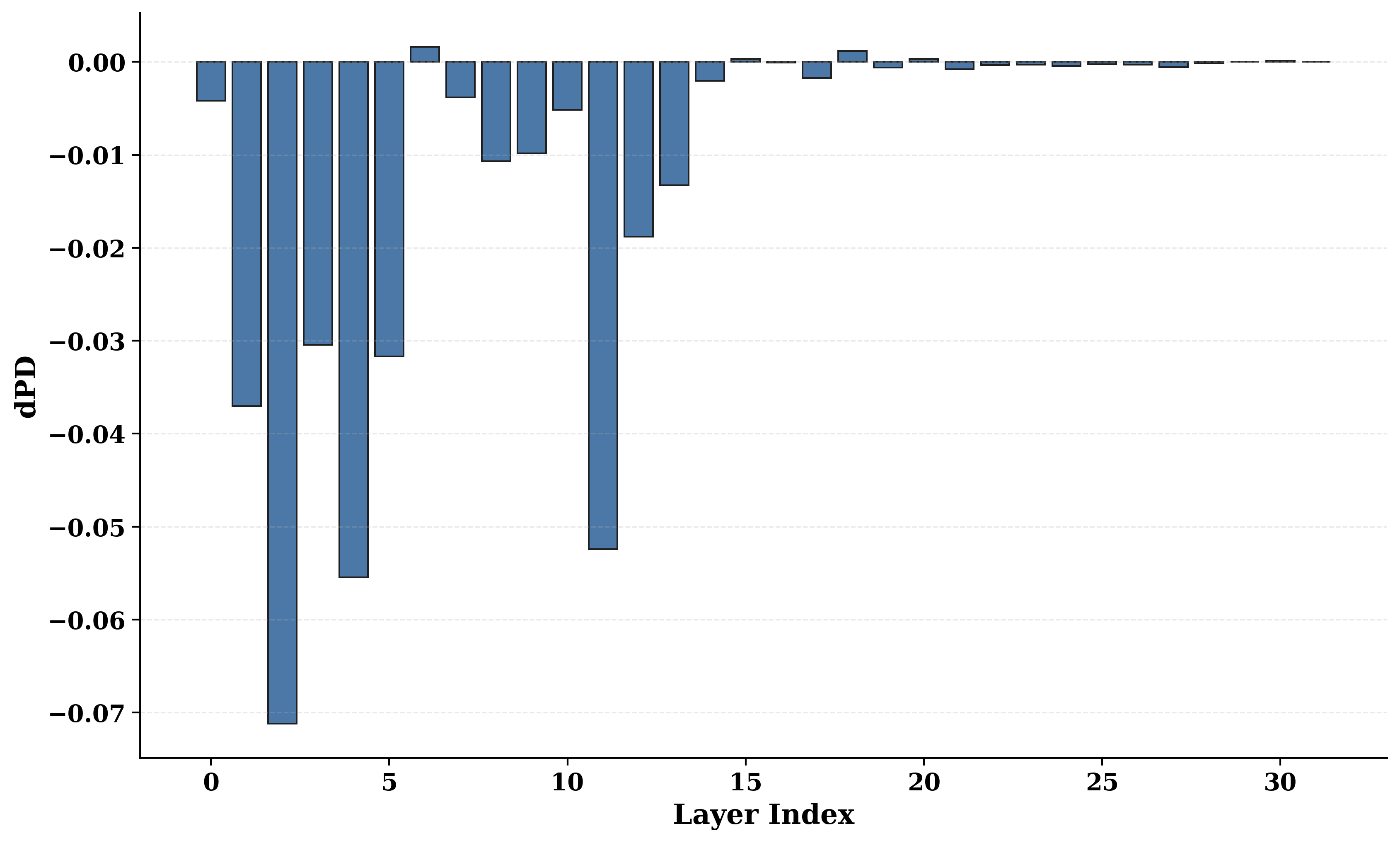}\hfill
    \includegraphics[width=0.32\textwidth]{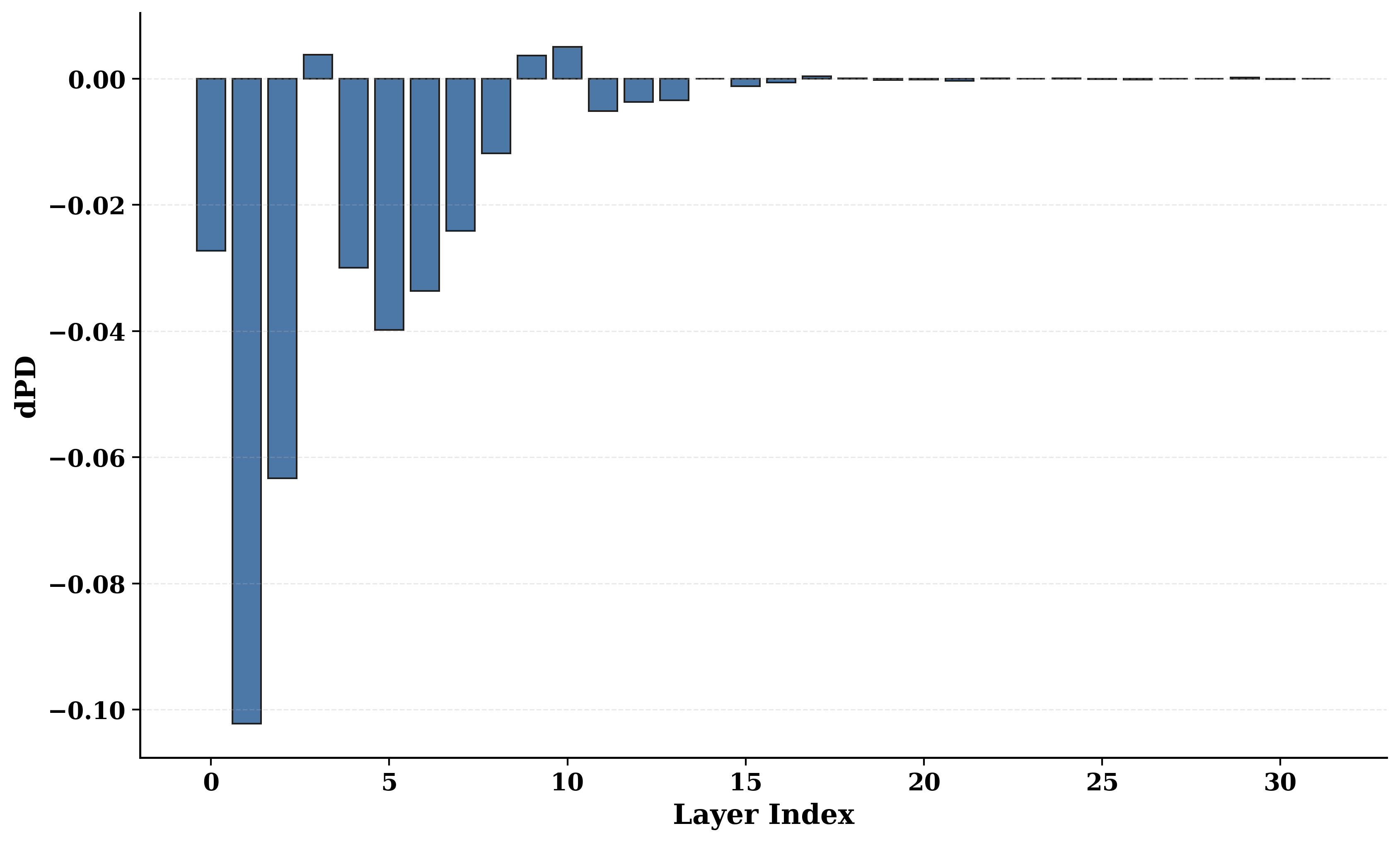}\hfill
    \includegraphics[width=0.32\textwidth]{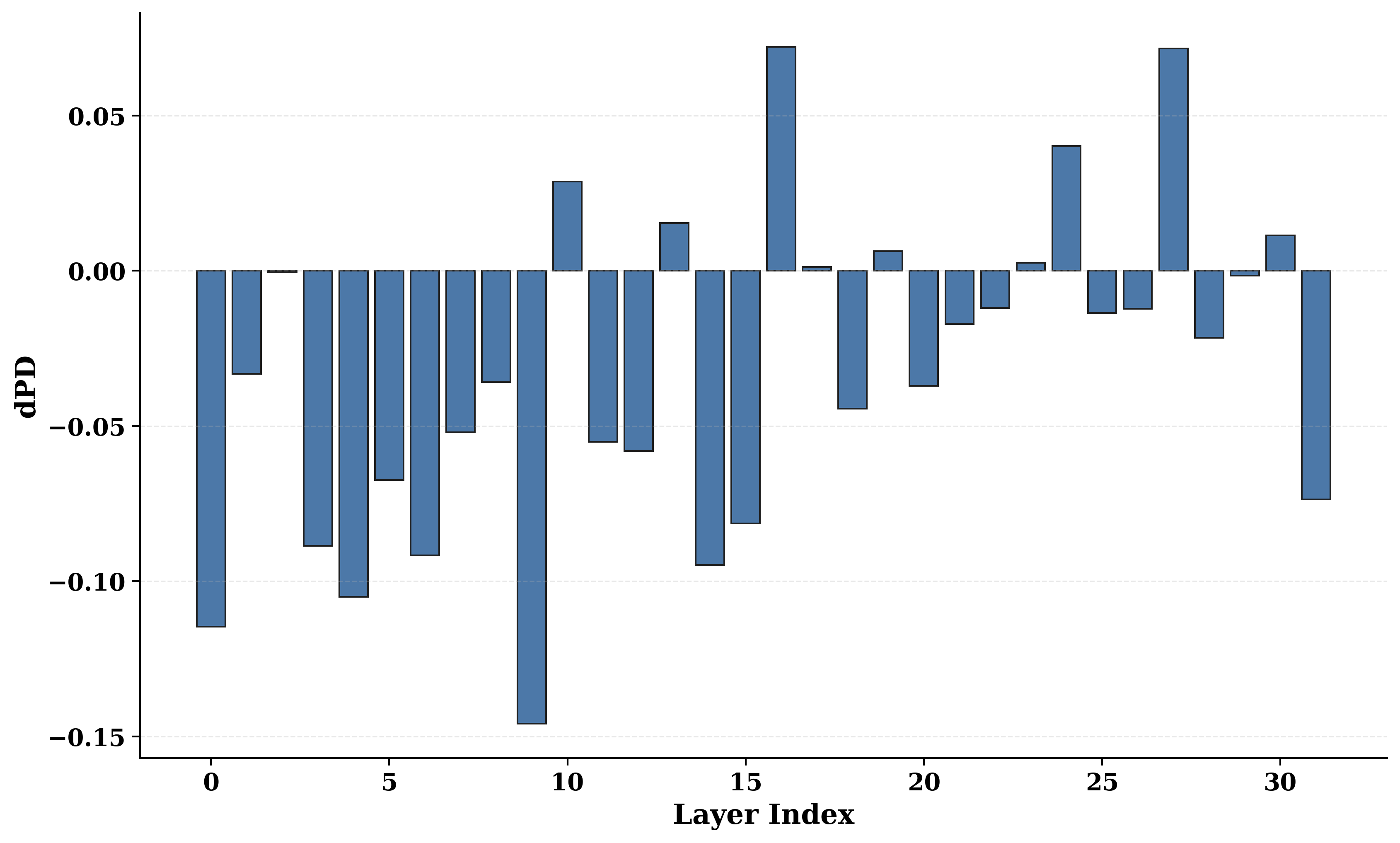}

    \includegraphics[width=0.32\textwidth]{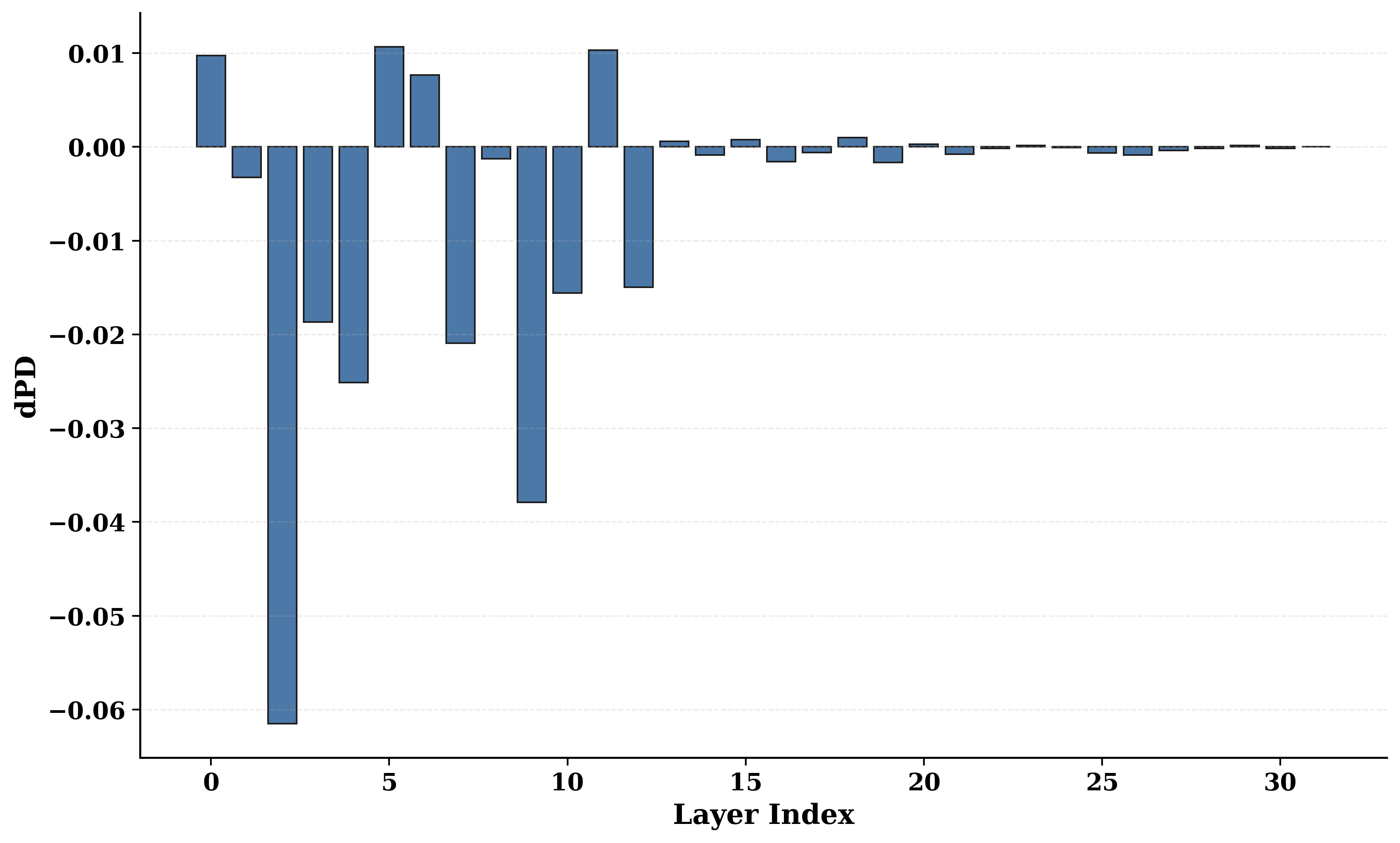}\hfill
    \includegraphics[width=0.32\textwidth]{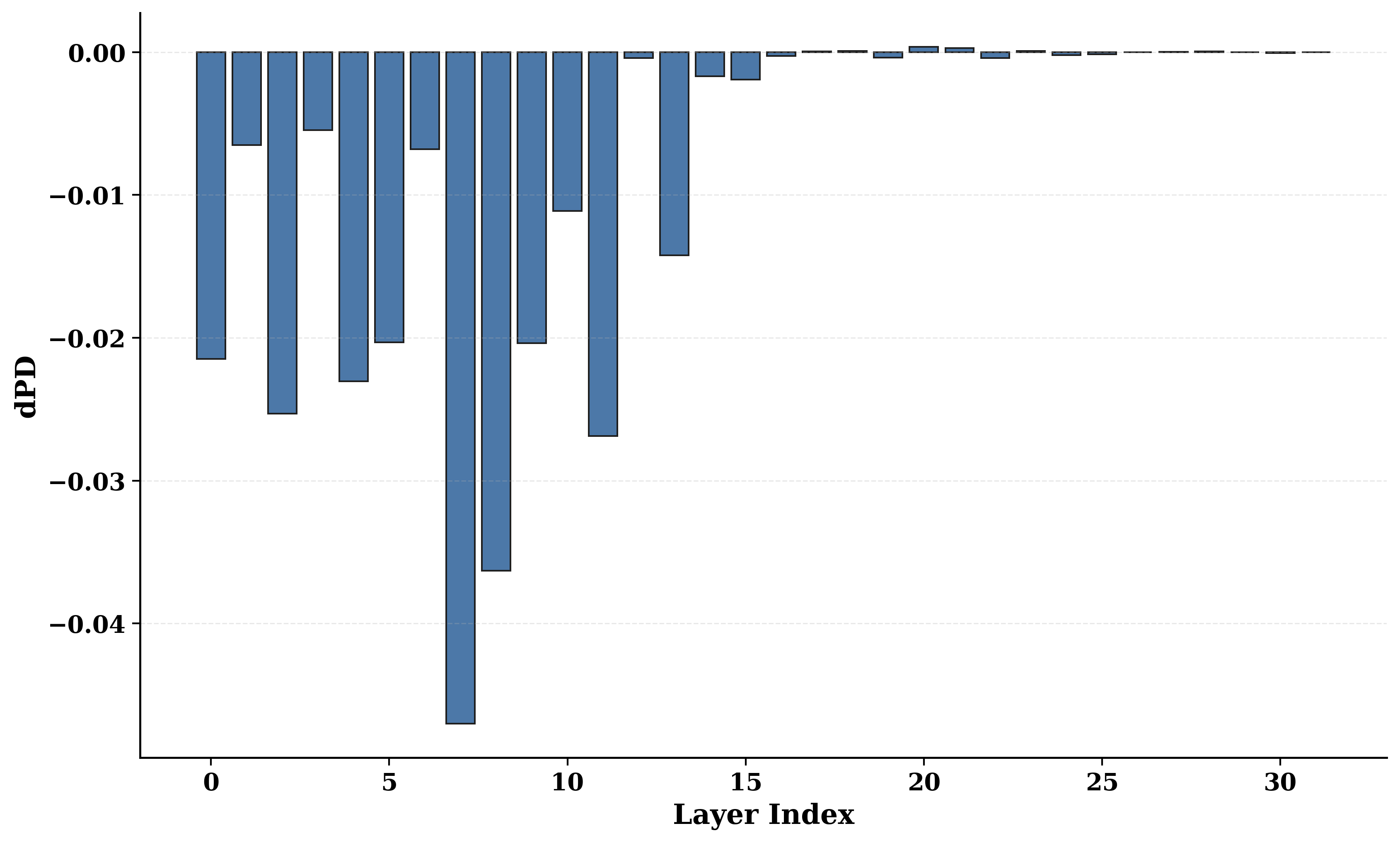}\hfill
    \includegraphics[width=0.32\textwidth]{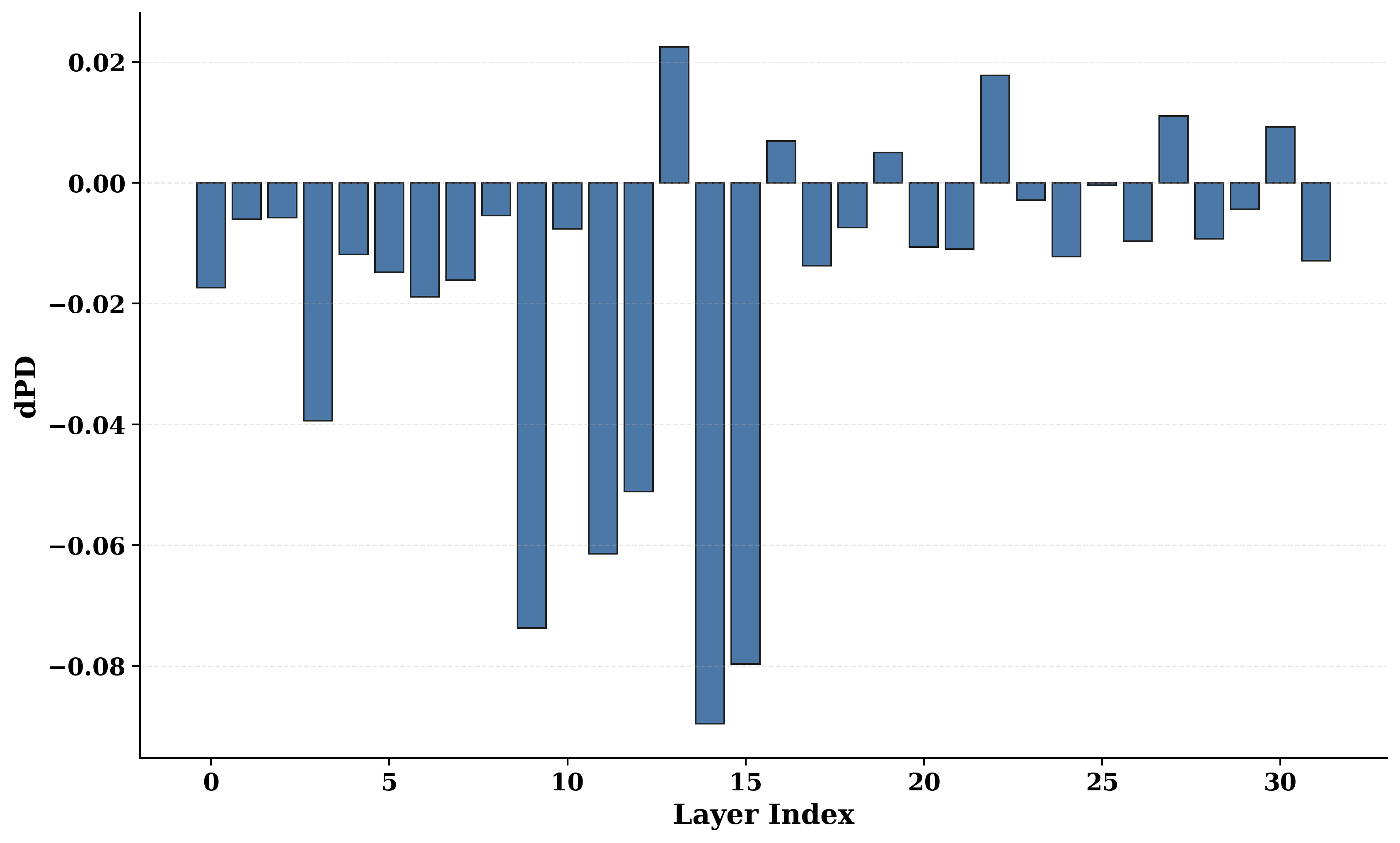}
    \caption{\textbf{Attention-block mean ablation on \textit{Llama-3.1-8B-Instruct} under $\pi_{\mathrm{EF}}$.} Per-layer $\mathrm{dPD}$ when each logical block is mean-ablated. Top row: one-hop. Bottom row: two-hop. Columns: Facts / Expression / Query.}
    \label{fig:stage_attn_llama_ef}
\end{figure*}

\begin{figure*}[!htbp]
    \centering
    \includegraphics[width=0.32\textwidth]{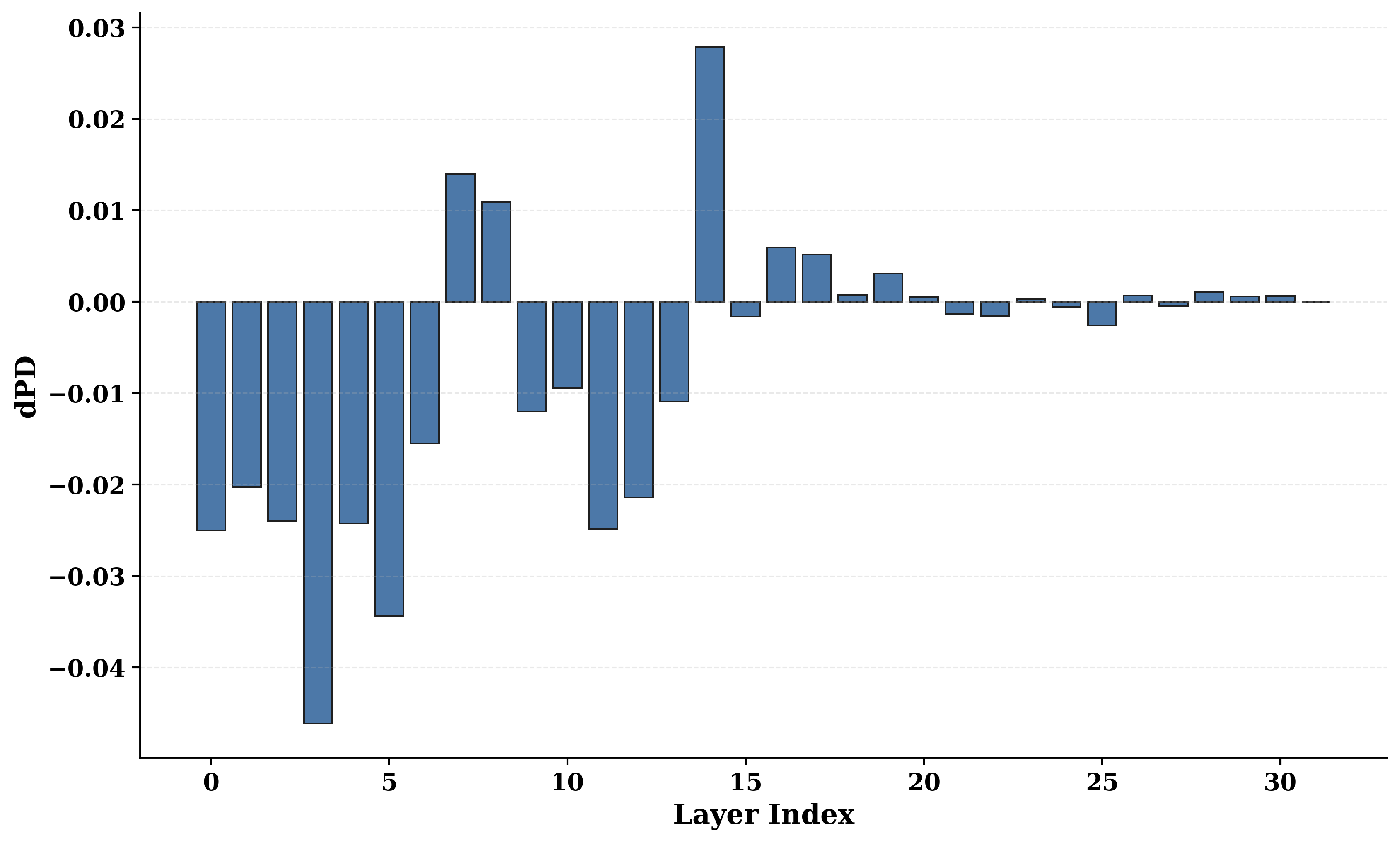}\hfill
    \includegraphics[width=0.32\textwidth]{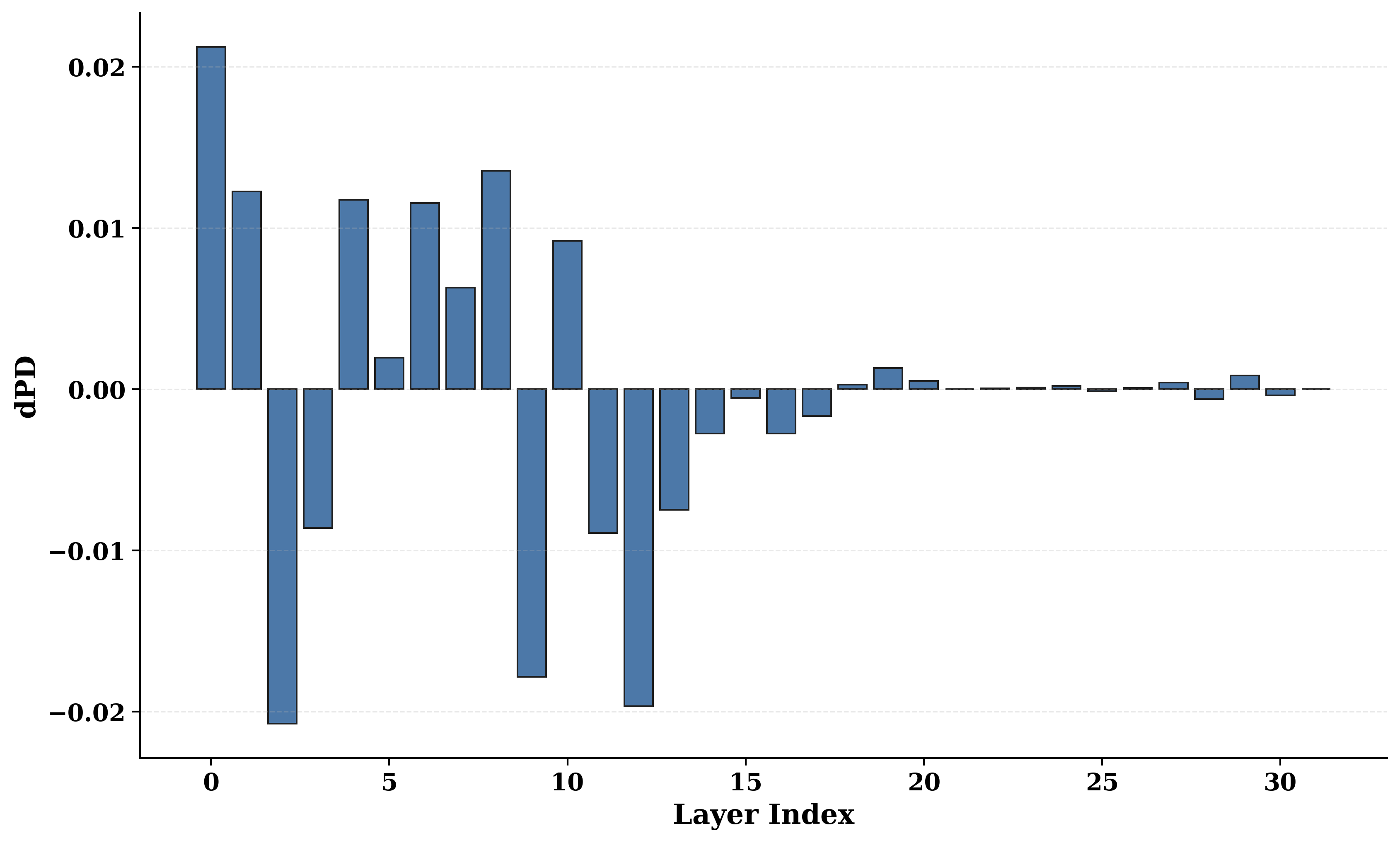}\hfill
    \includegraphics[width=0.32\textwidth]{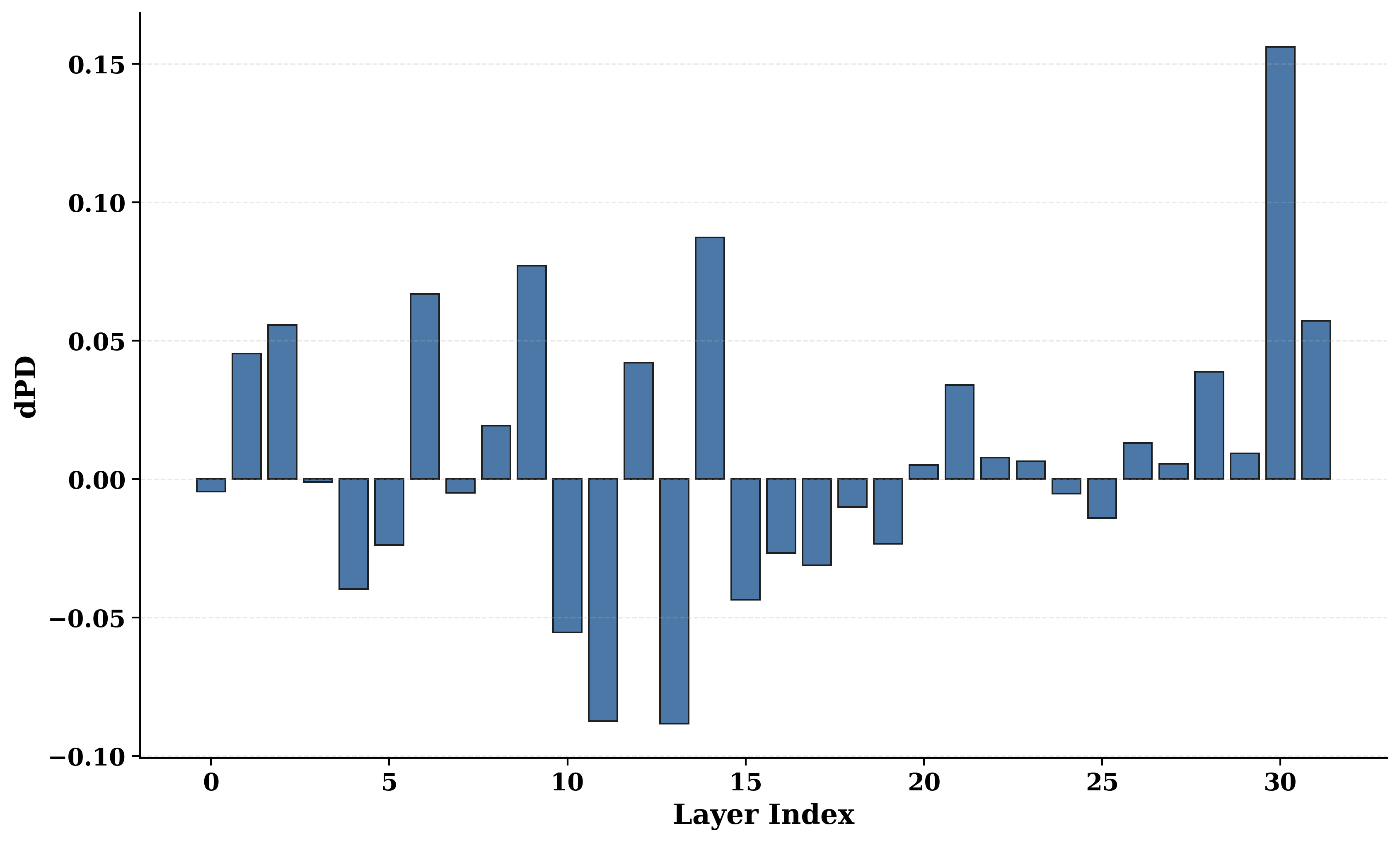}

    \includegraphics[width=0.32\textwidth]{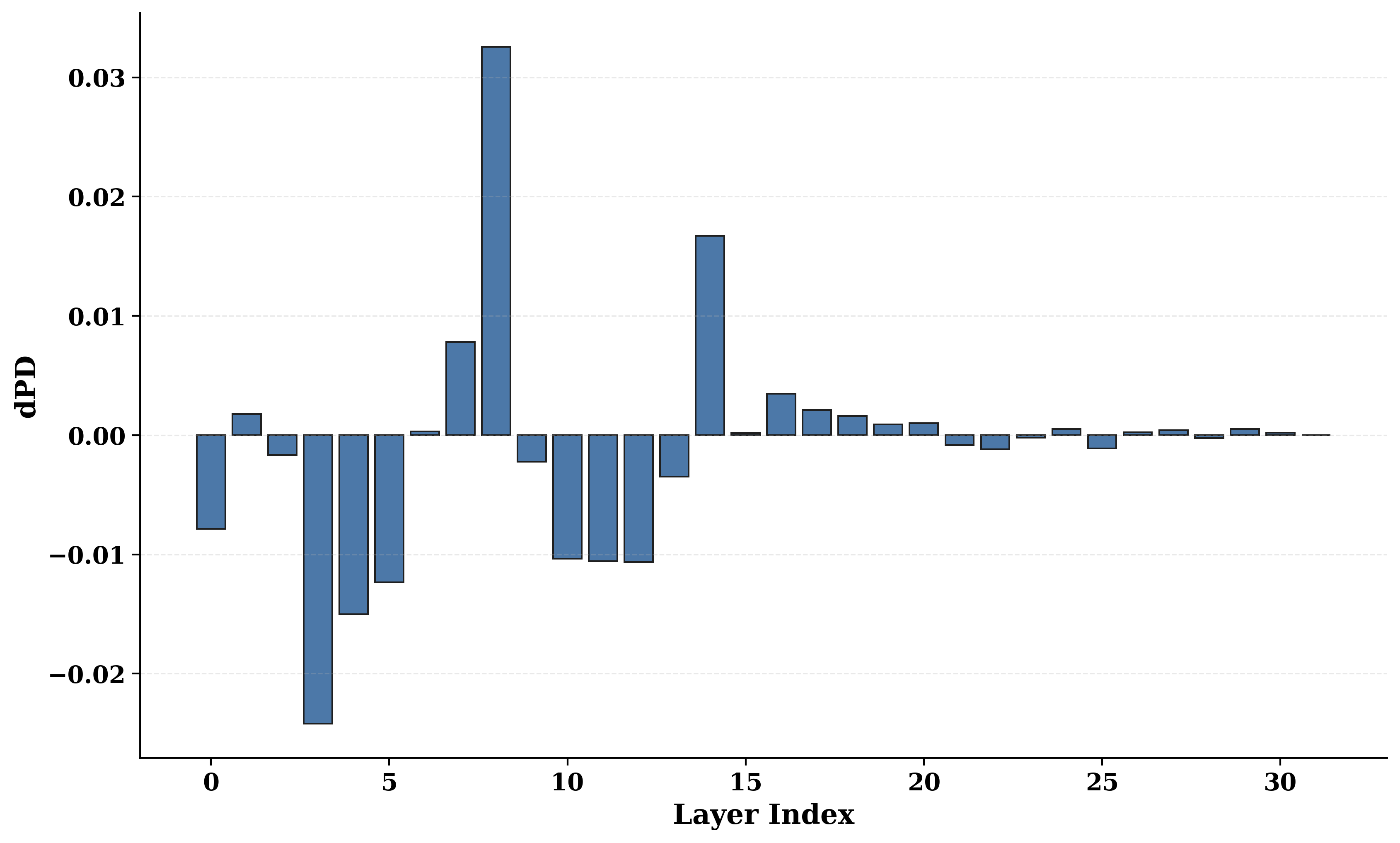}\hfill
    \includegraphics[width=0.32\textwidth]{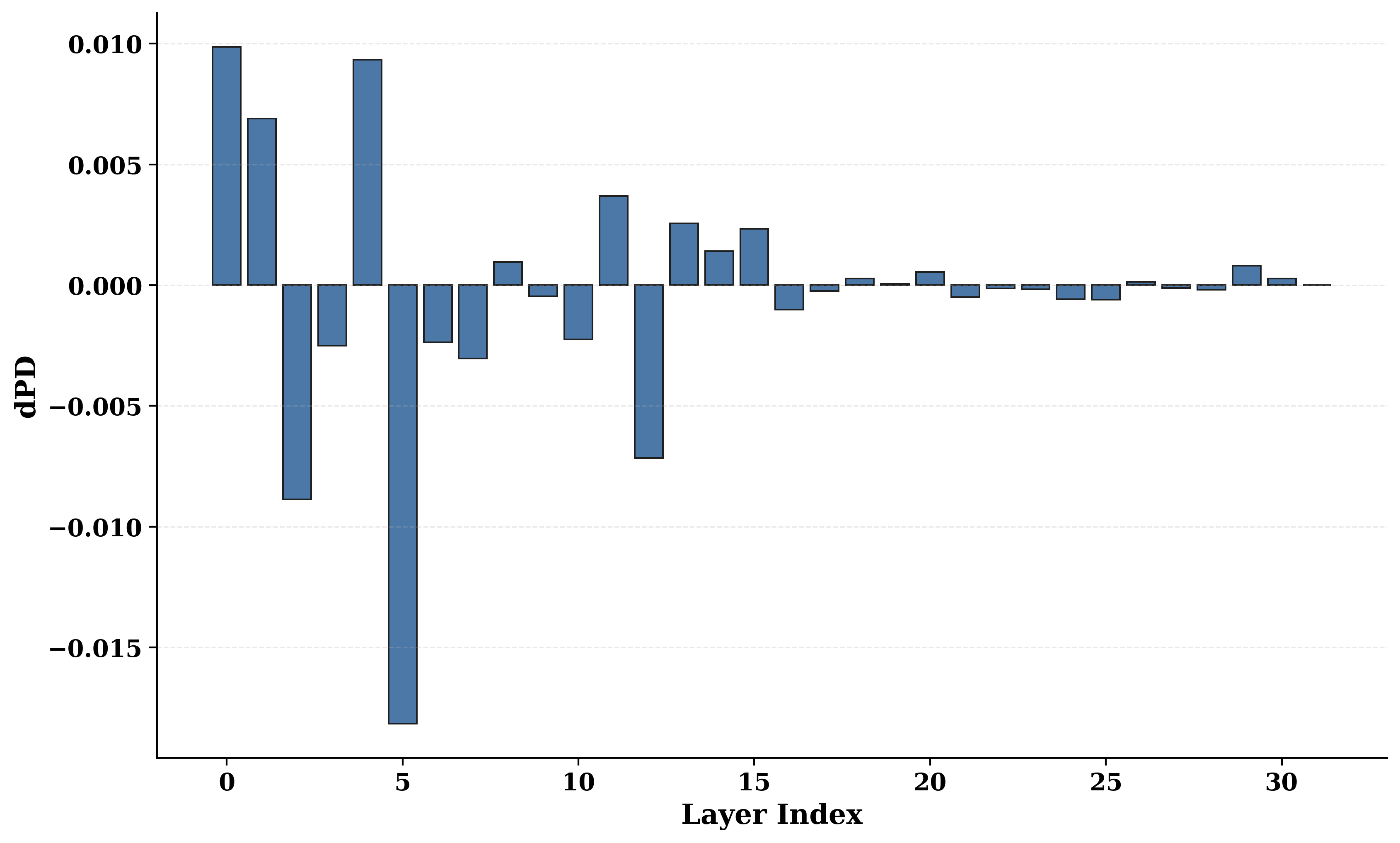}\hfill
    \includegraphics[width=0.32\textwidth]{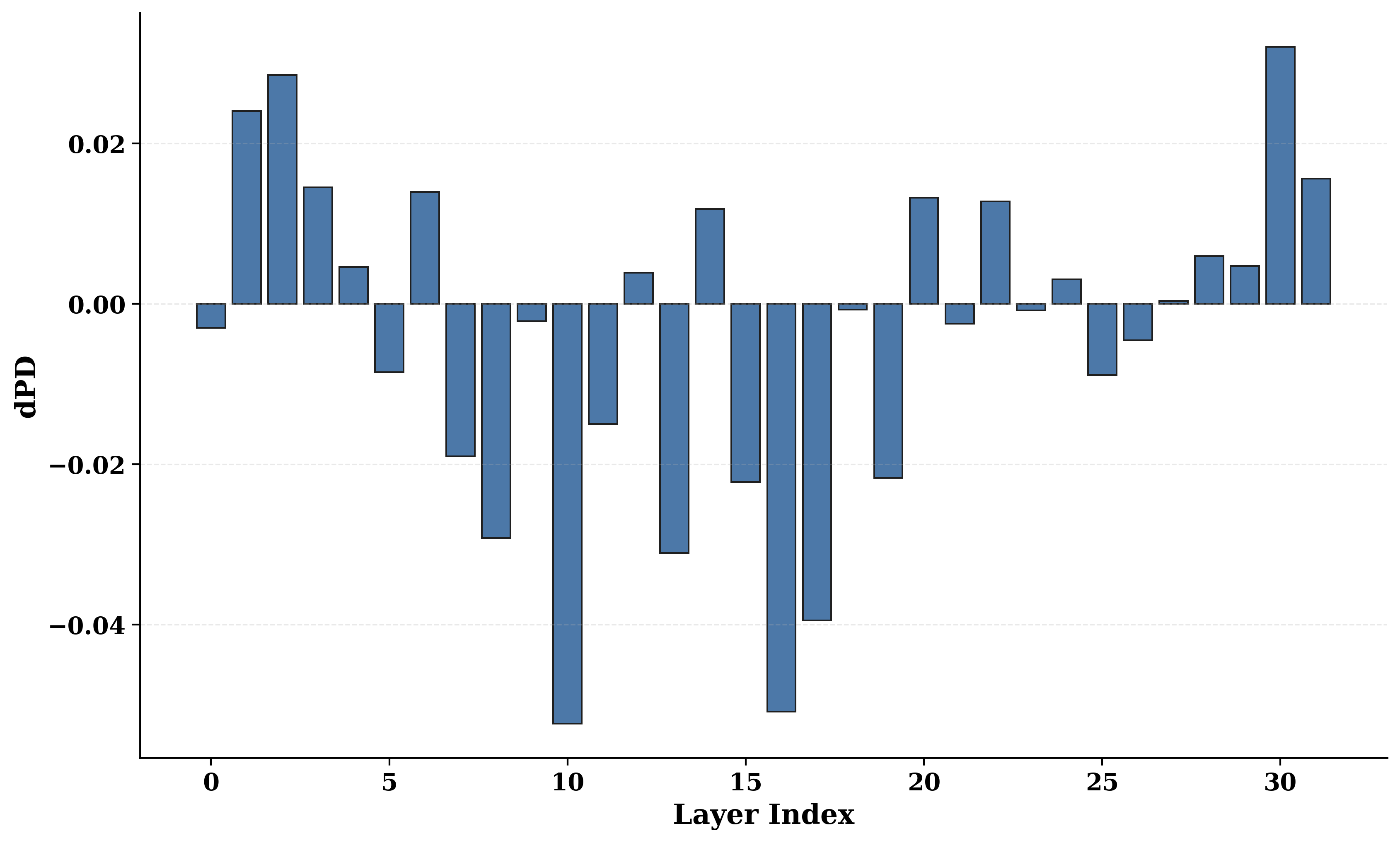}
    \caption{\textbf{Attention-block mean ablation on \textit{Mistral-7B-Instruct} under $\pi_{\mathrm{FF}}$.} Per-layer $\mathrm{dPD}$ when each logical block is mean-ablated. Top row: one-hop. Bottom row: two-hop. Columns: Facts / Expression / Query.}
    \label{fig:stage_attn_mistral_ff}
\end{figure*}

\begin{figure*}[!htbp]
    \centering
    \includegraphics[width=0.32\textwidth]{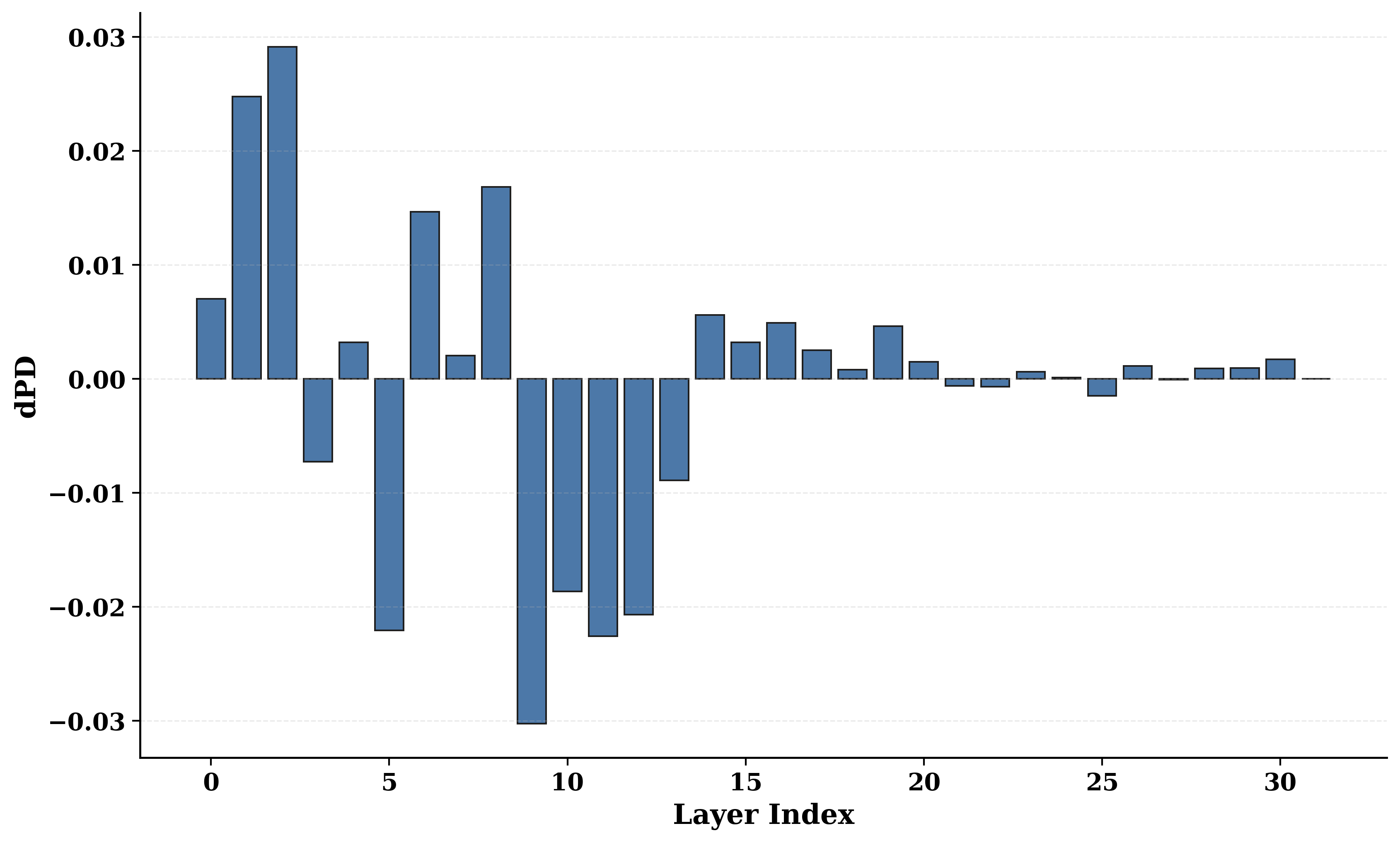}\hfill
    \includegraphics[width=0.32\textwidth]{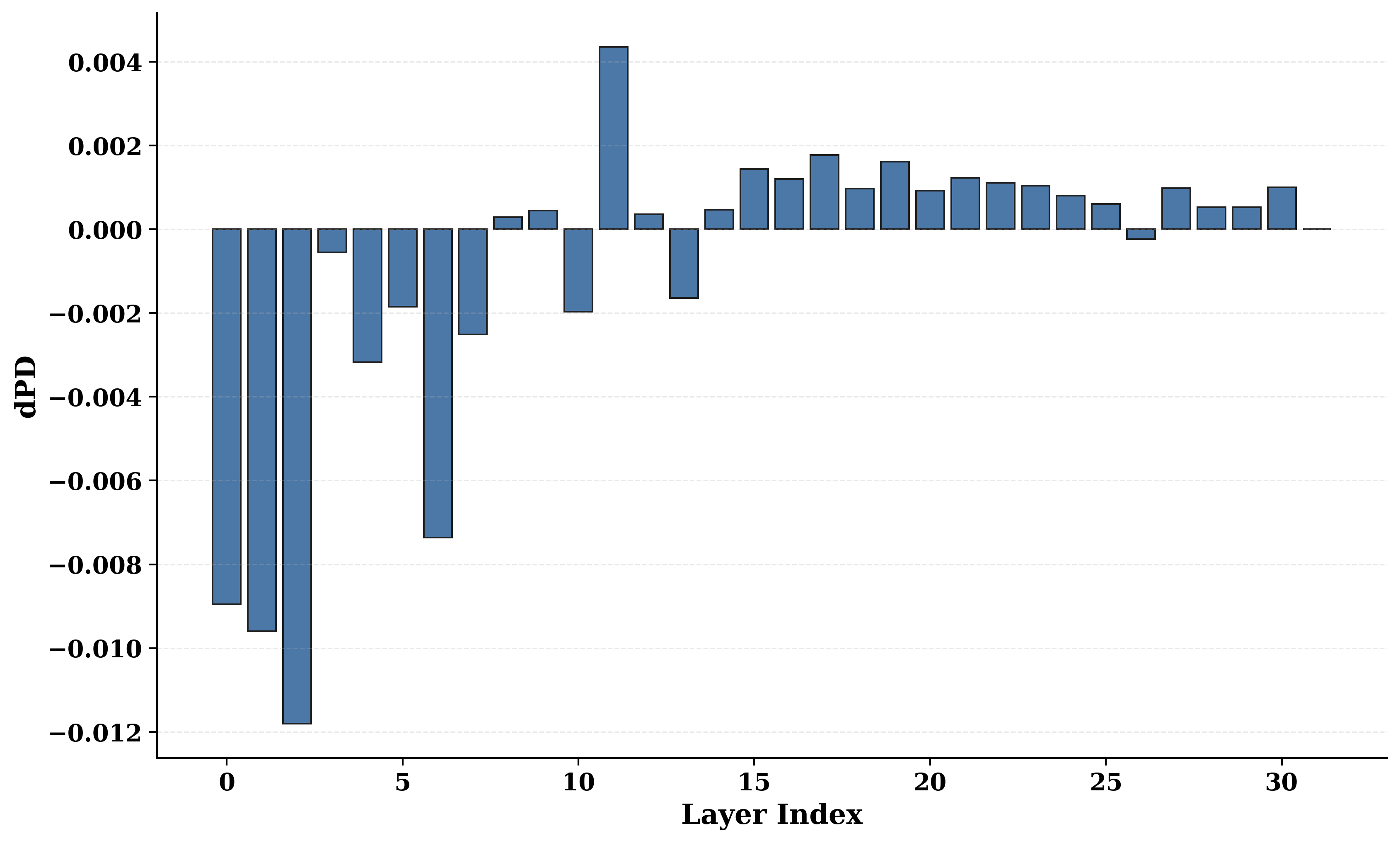}\hfill
    \includegraphics[width=0.32\textwidth]{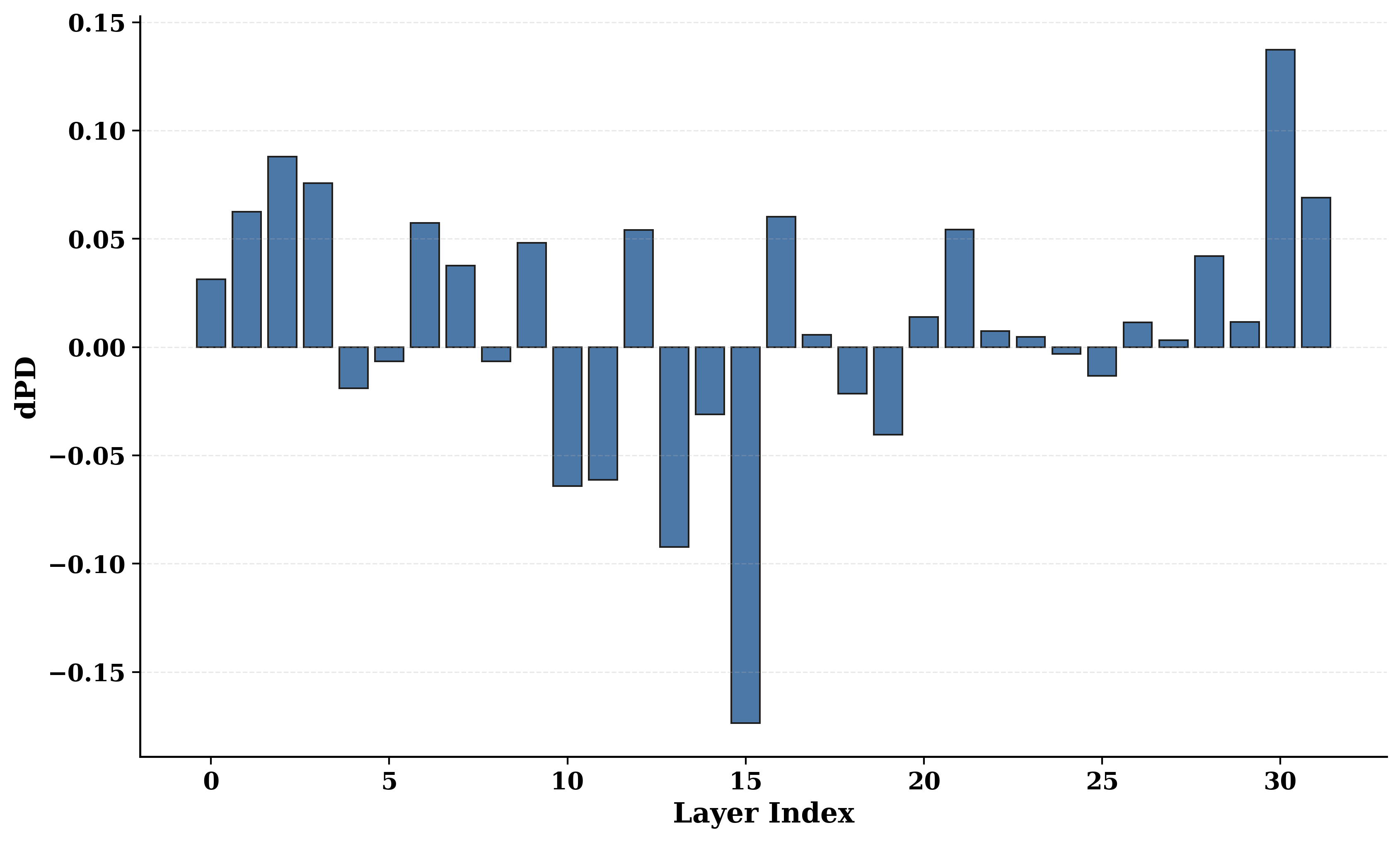}

    \includegraphics[width=0.32\textwidth]{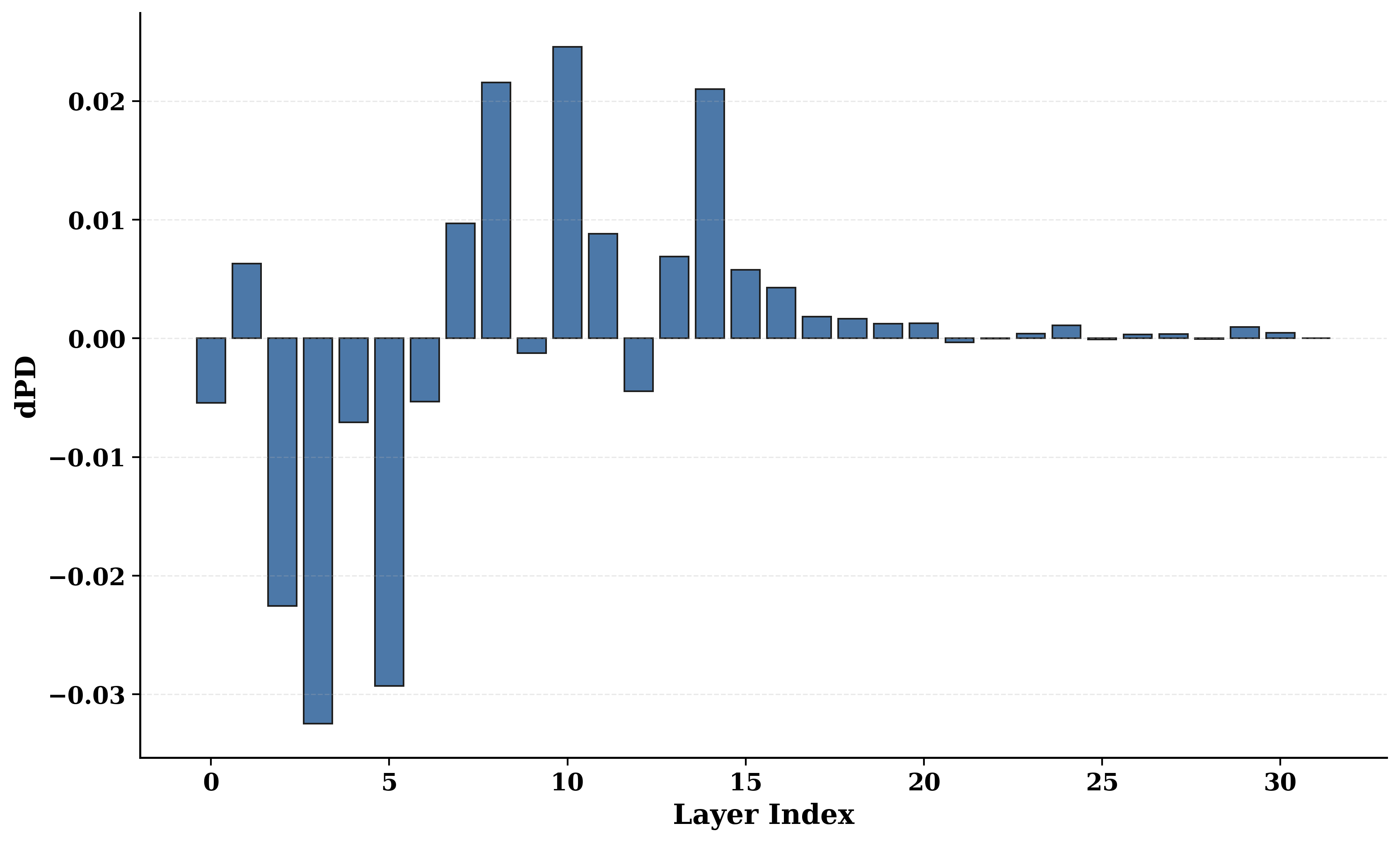}\hfill
    \includegraphics[width=0.32\textwidth]{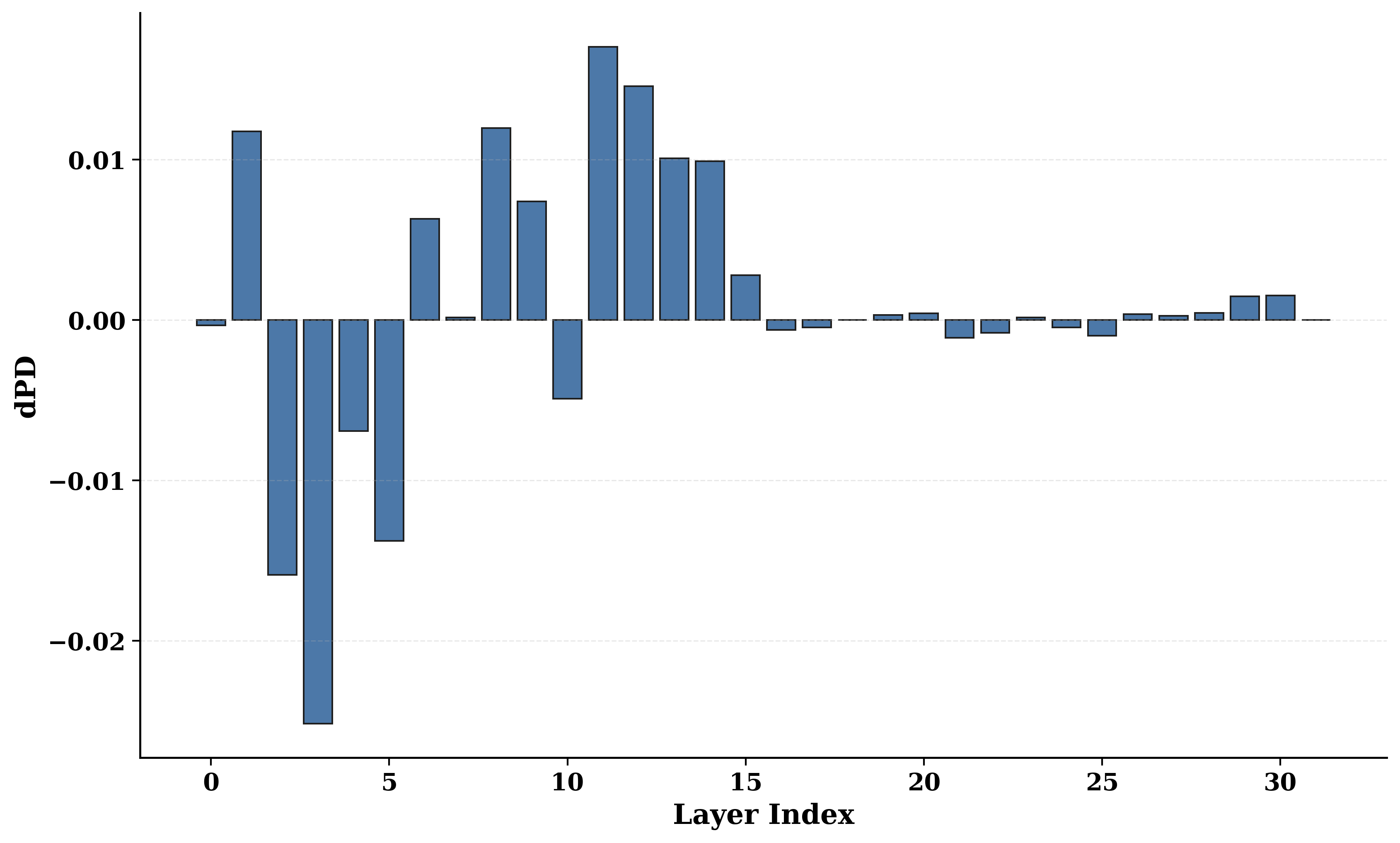}\hfill
    \includegraphics[width=0.32\textwidth]{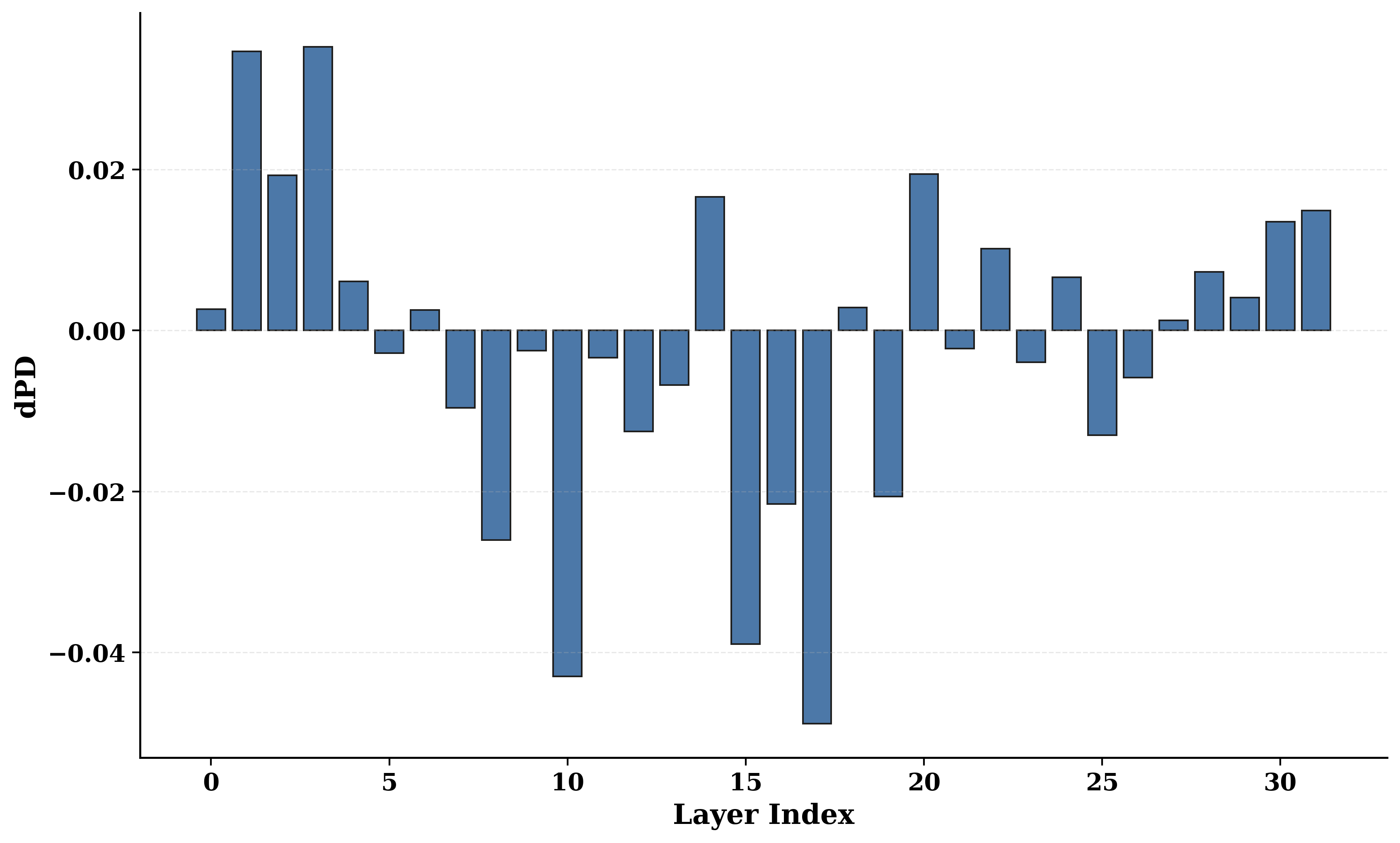}
    \caption{\textbf{Attention-block mean ablation on \textit{Mistral-7B-Instruct} under $\pi_{\mathrm{EF}}$.} Per-layer $\mathrm{dPD}$ when each logical block is mean-ablated. Top row: one-hop. Bottom row: two-hop. Columns: Facts / Expression / Query.}
    \label{fig:stage_attn_mistral_ef}
\end{figure*}

Figures~\ref{fig:stage_attn_q8_ff}--\ref{fig:stage_attn_mistral_ef} show attention-block mean-ablation $\mathrm{dPD}$ in the same format. Compared with the MLP branch, the attention branch shifts more mass into middle-layer Expression and later Query integration, consistent with attention's role in cross-token routing. The coarse staged ordering is preserved across all (model, prompt-order) cells.

\subsection{Joint Attention+MLP Block Ablation}
\label{sec:appendix_region_variants}

\begin{figure*}[!htbp]
    \centering
    \includegraphics[width=0.32\textwidth]{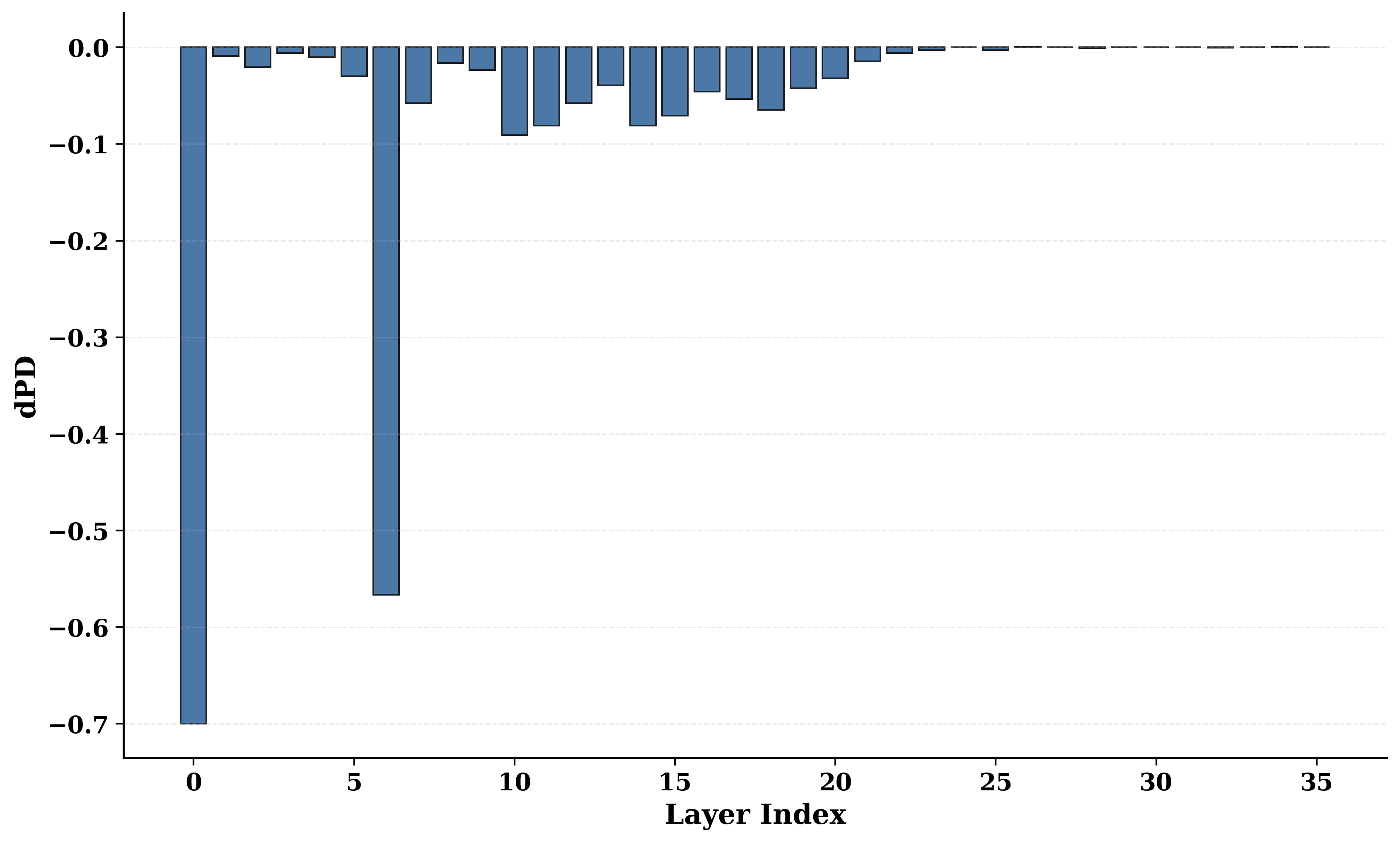}\hfill
    \includegraphics[width=0.32\textwidth]{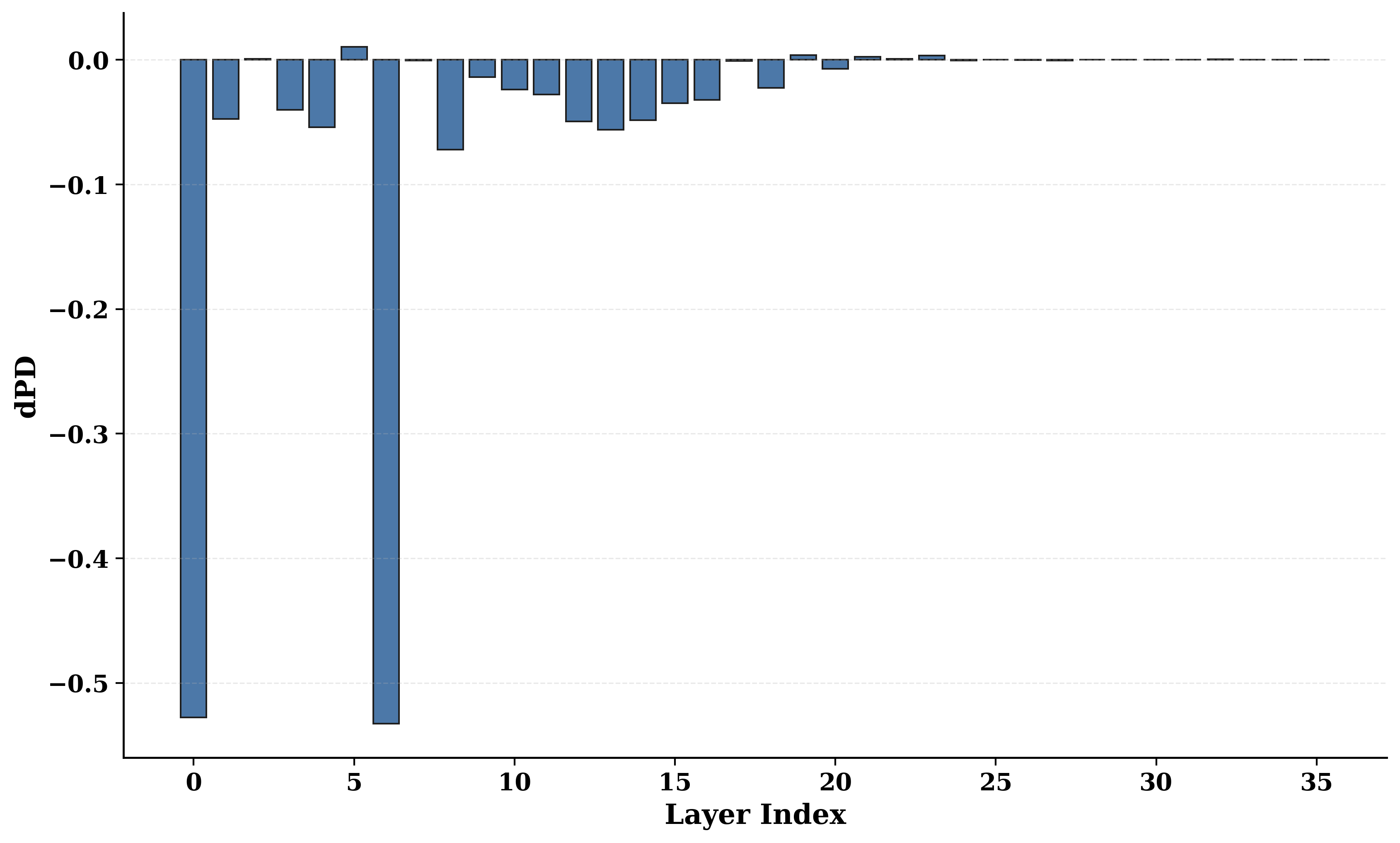}\hfill
    \includegraphics[width=0.32\textwidth]{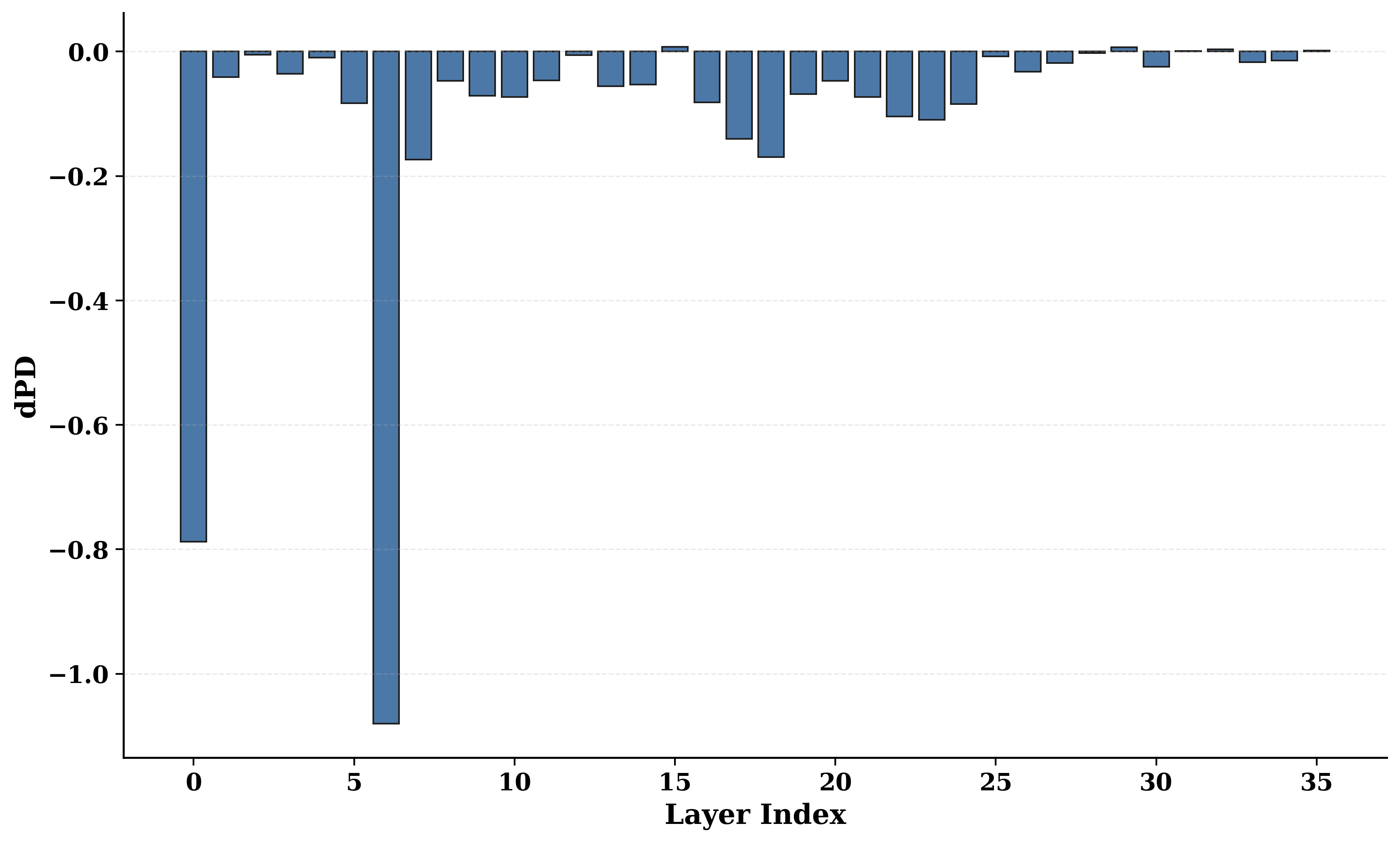}

    \includegraphics[width=0.32\textwidth]{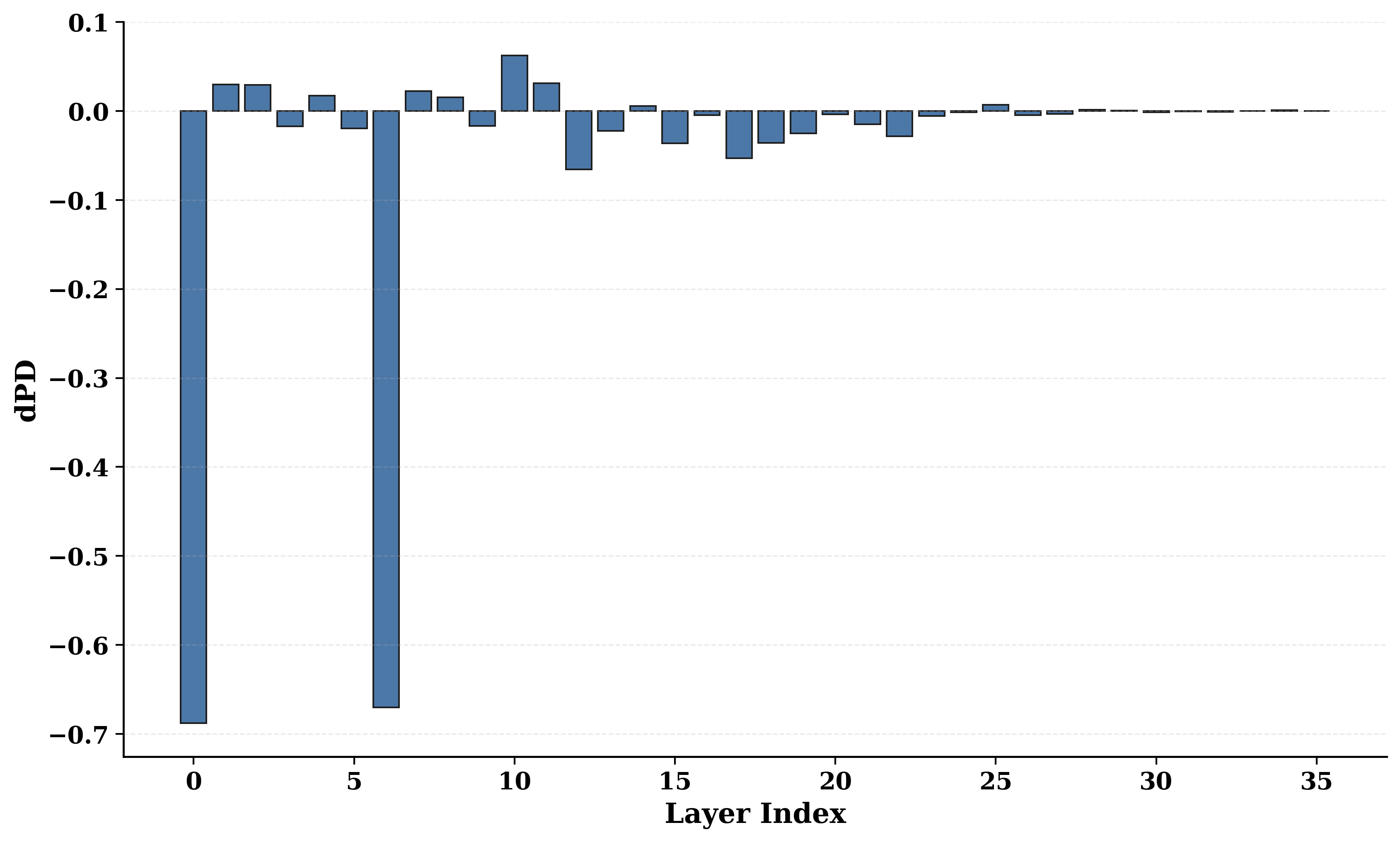}\hfill
    \includegraphics[width=0.32\textwidth]{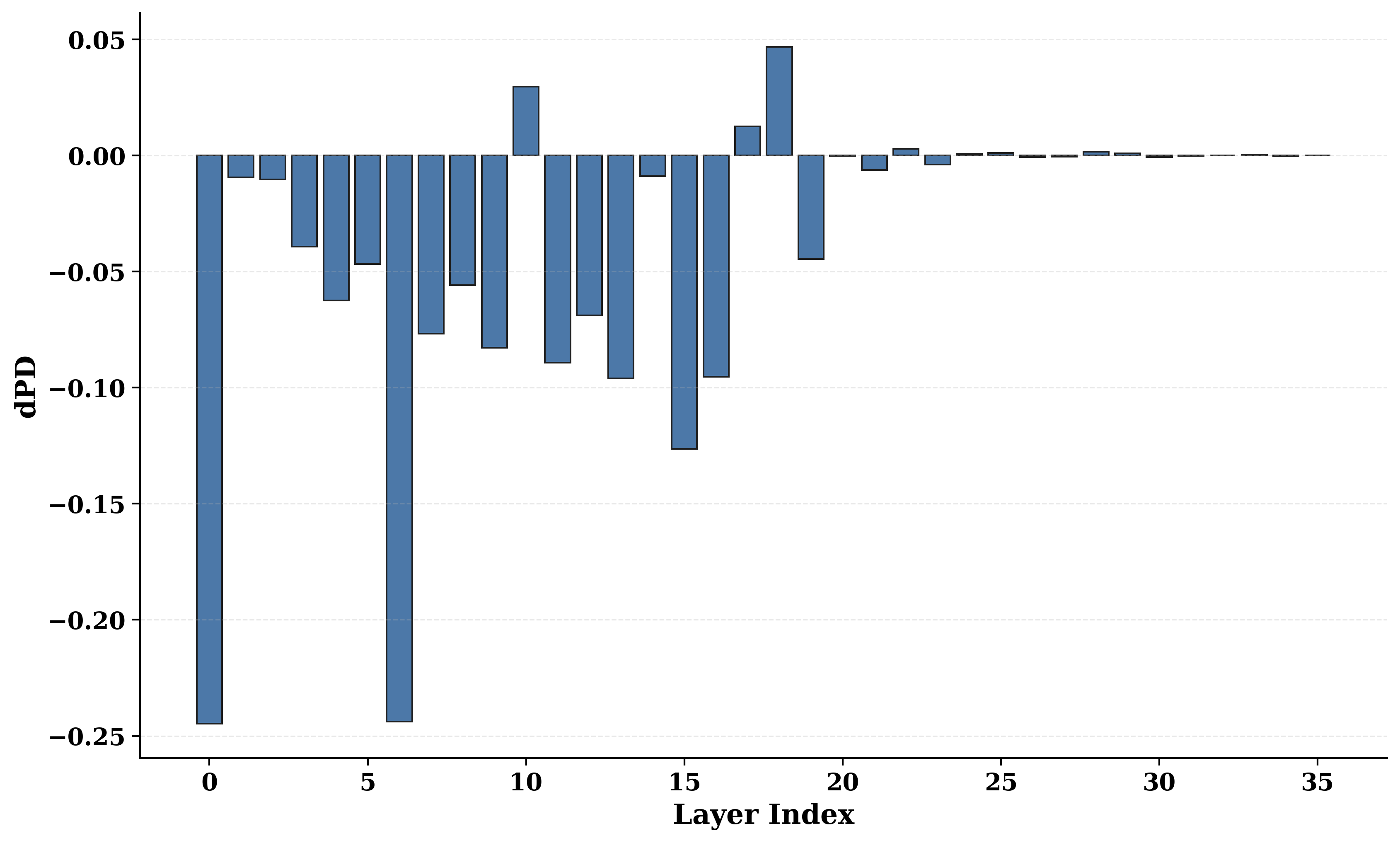}\hfill
    \includegraphics[width=0.32\textwidth]{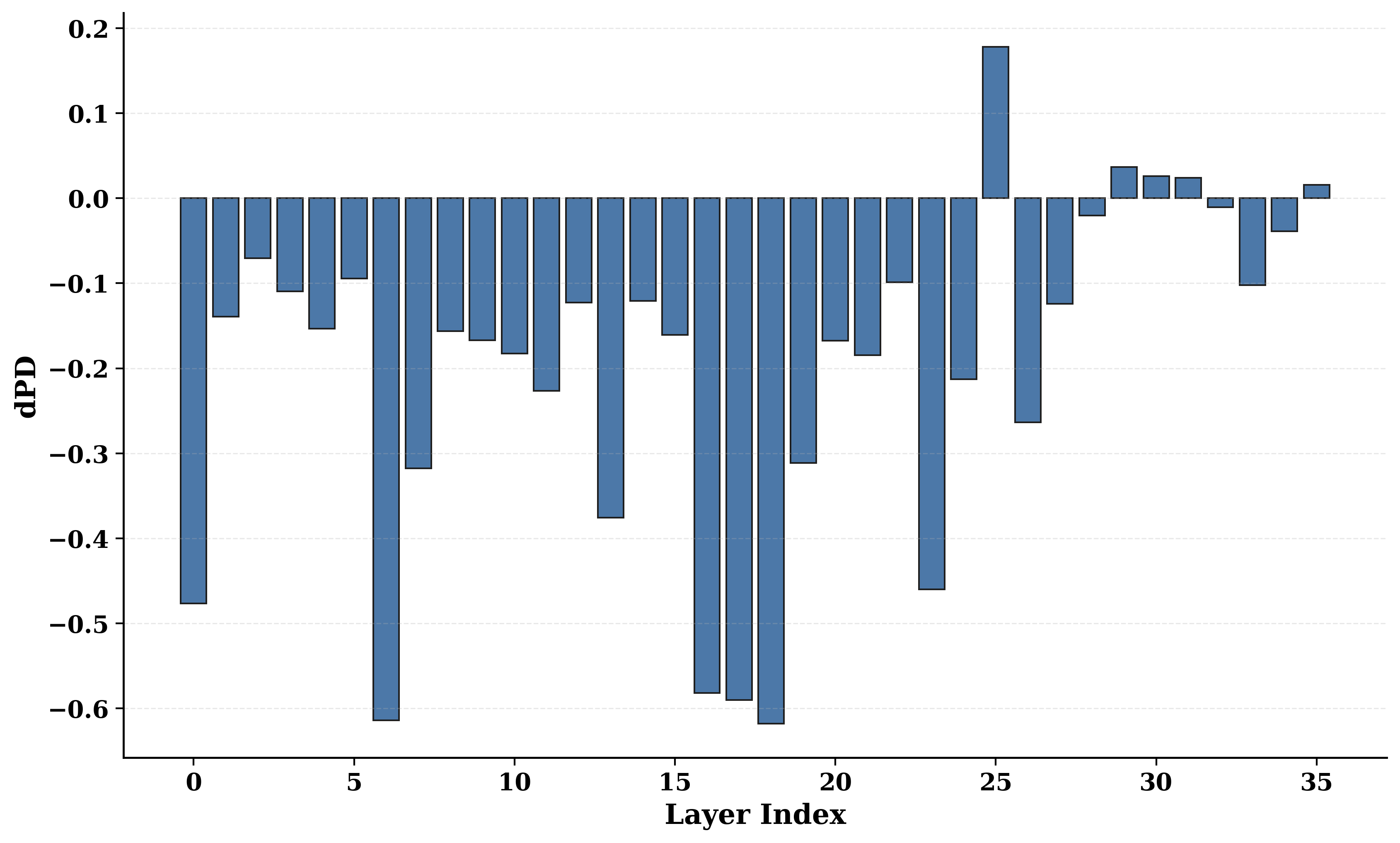}
    \caption{\textbf{Joint attention+MLP-block mean ablation on \textit{Qwen3-8B} under $\pi_{\mathrm{EF}}$.} Per-layer $\mathrm{dPD}$ when each logical block is mean-ablated. Top row: one-hop. Bottom row: two-hop. Columns: Facts / Expression / Query.}
    \label{fig:stage_joint_q8_ef}
\end{figure*}

\begin{figure*}[!htbp]
    \centering
    \includegraphics[width=0.32\textwidth]{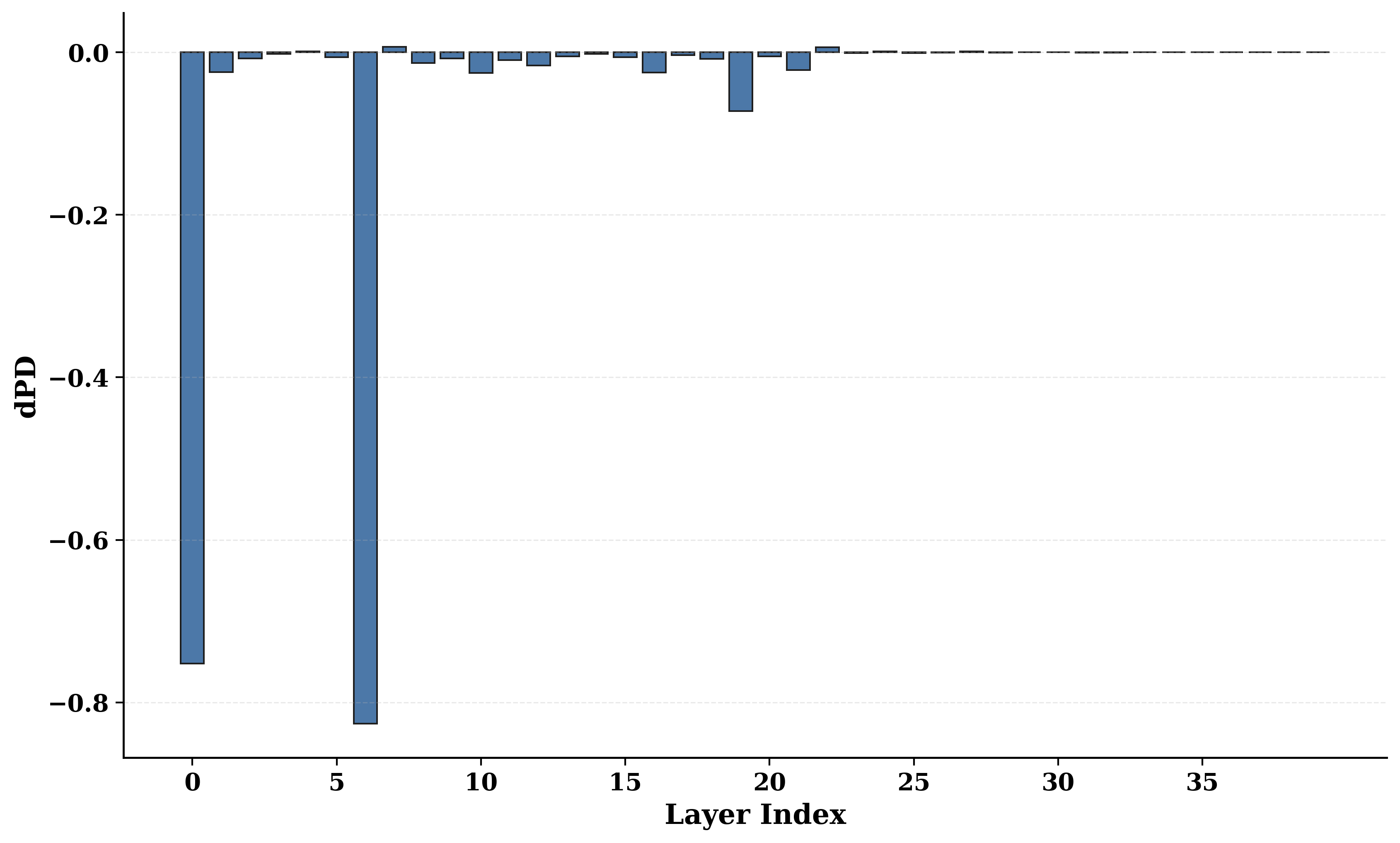}\hfill
    \includegraphics[width=0.32\textwidth]{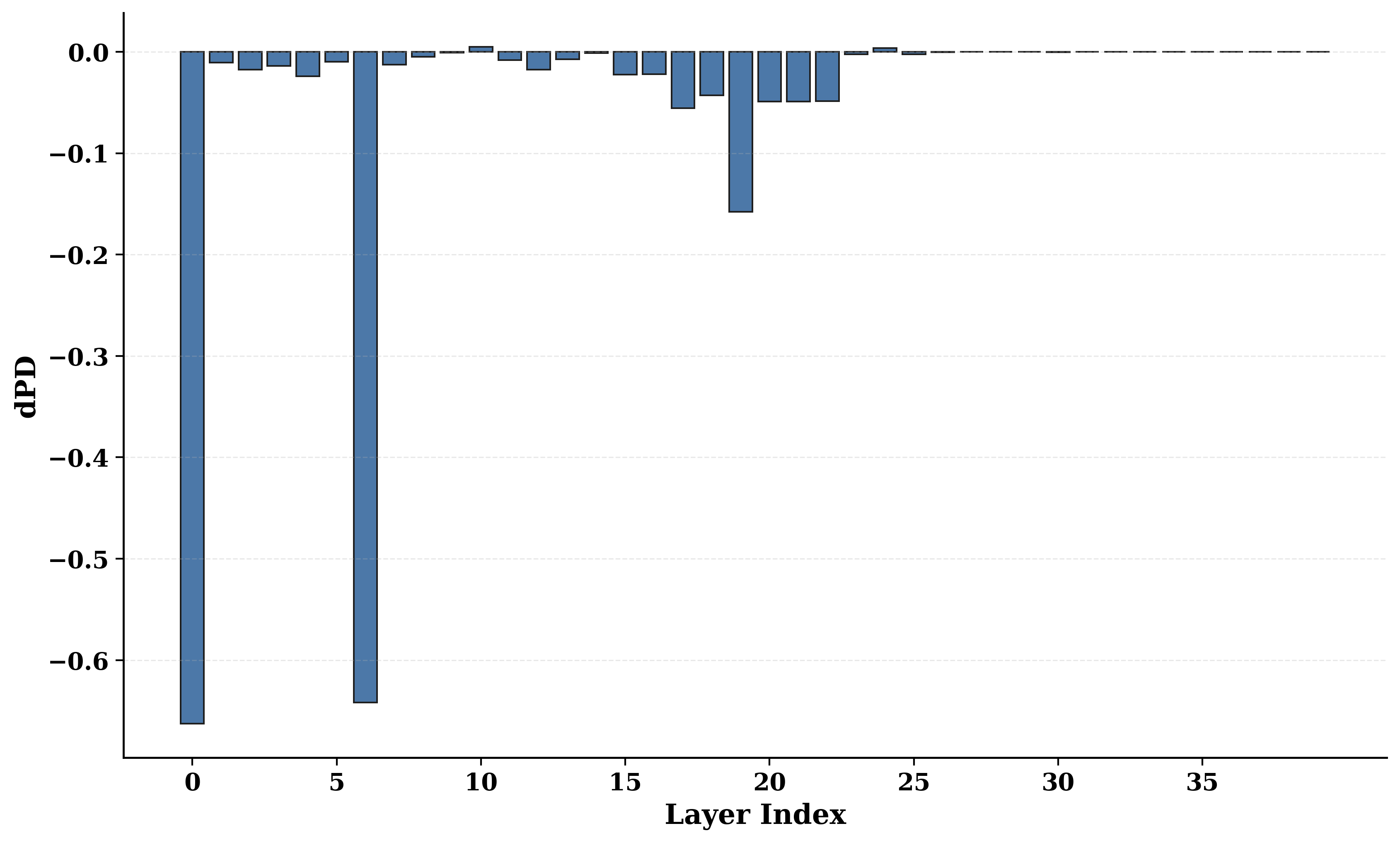}\hfill
    \includegraphics[width=0.32\textwidth]{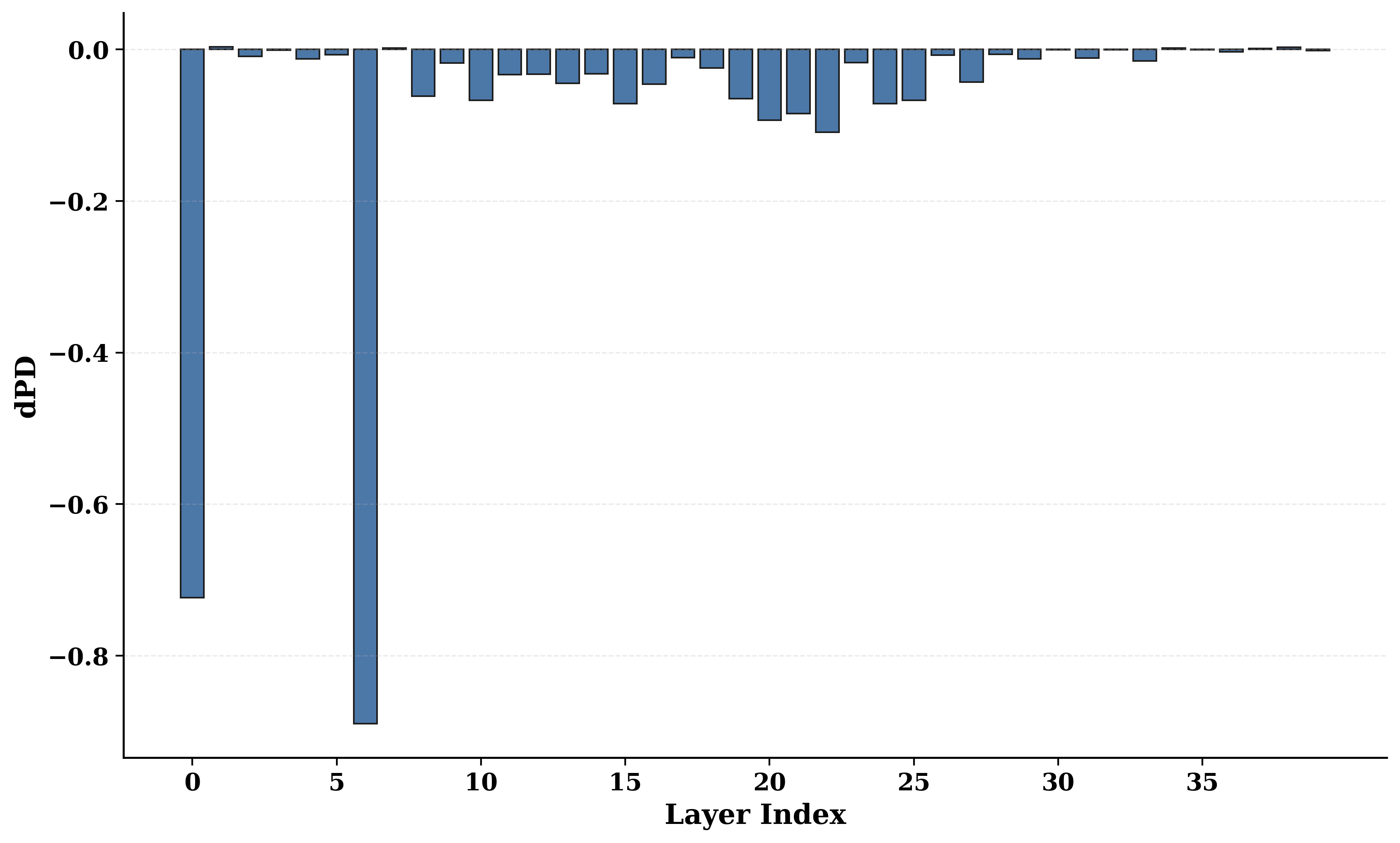}

    \includegraphics[width=0.32\textwidth]{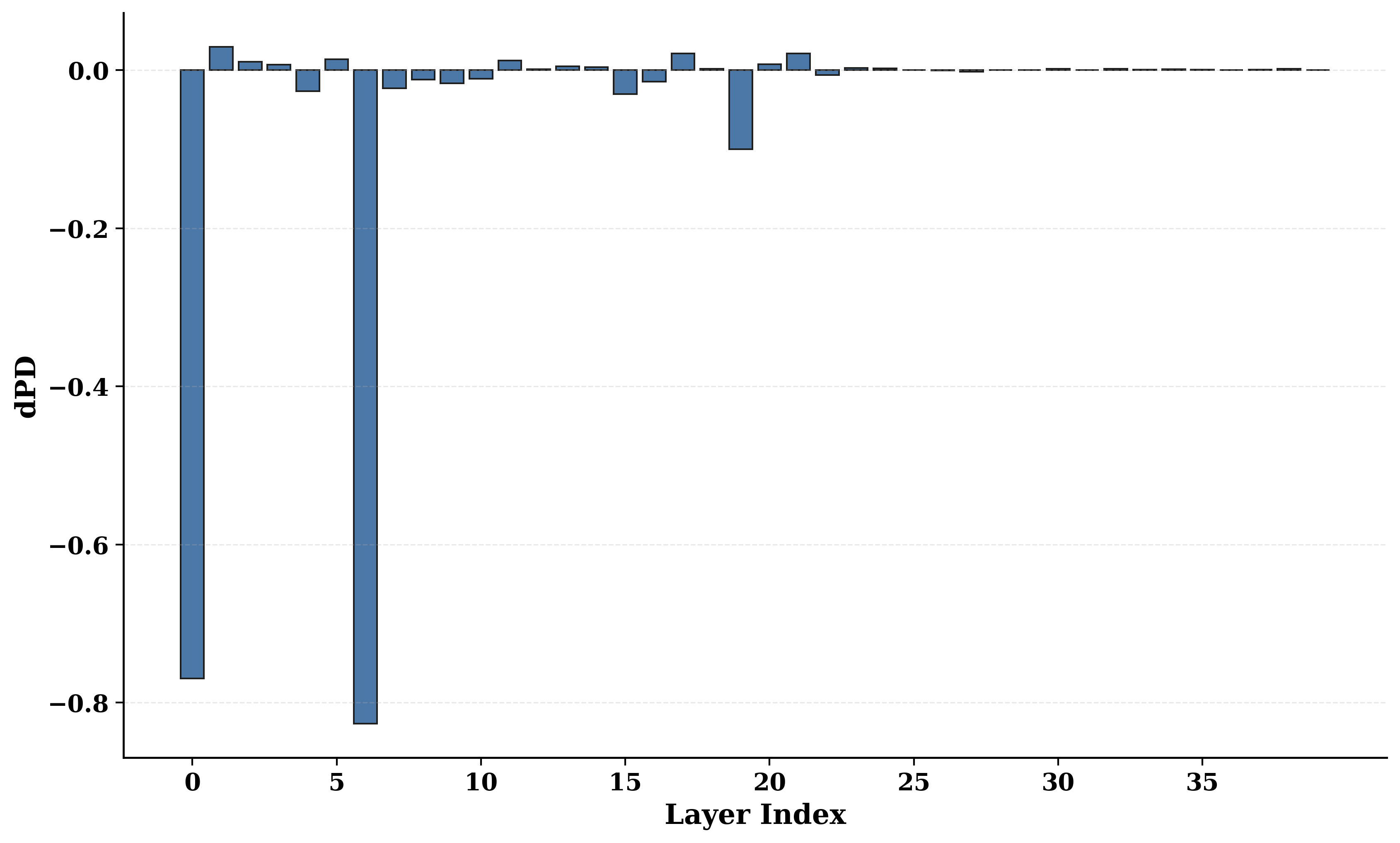}\hfill
    \includegraphics[width=0.32\textwidth]{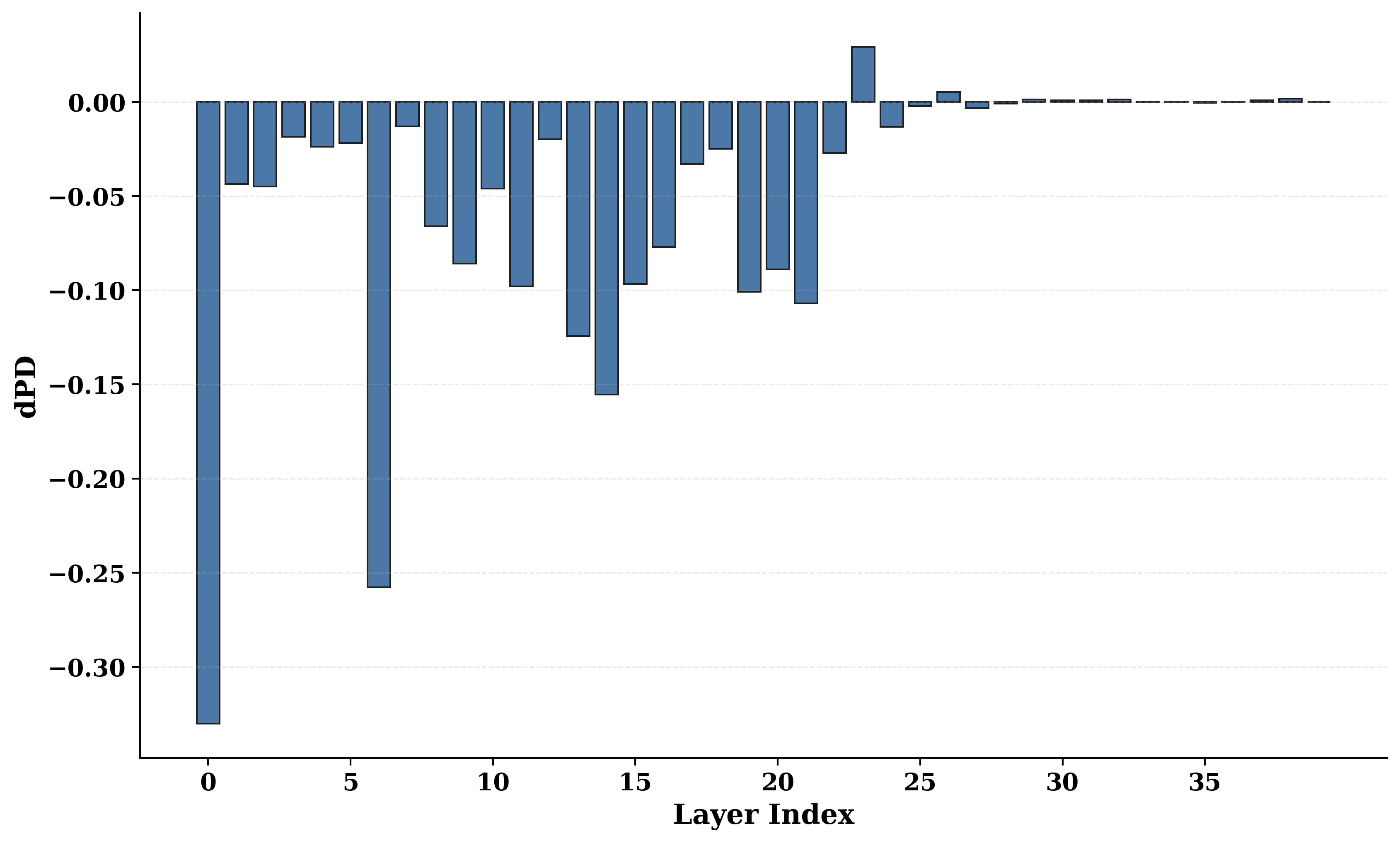}\hfill
    \includegraphics[width=0.32\textwidth]{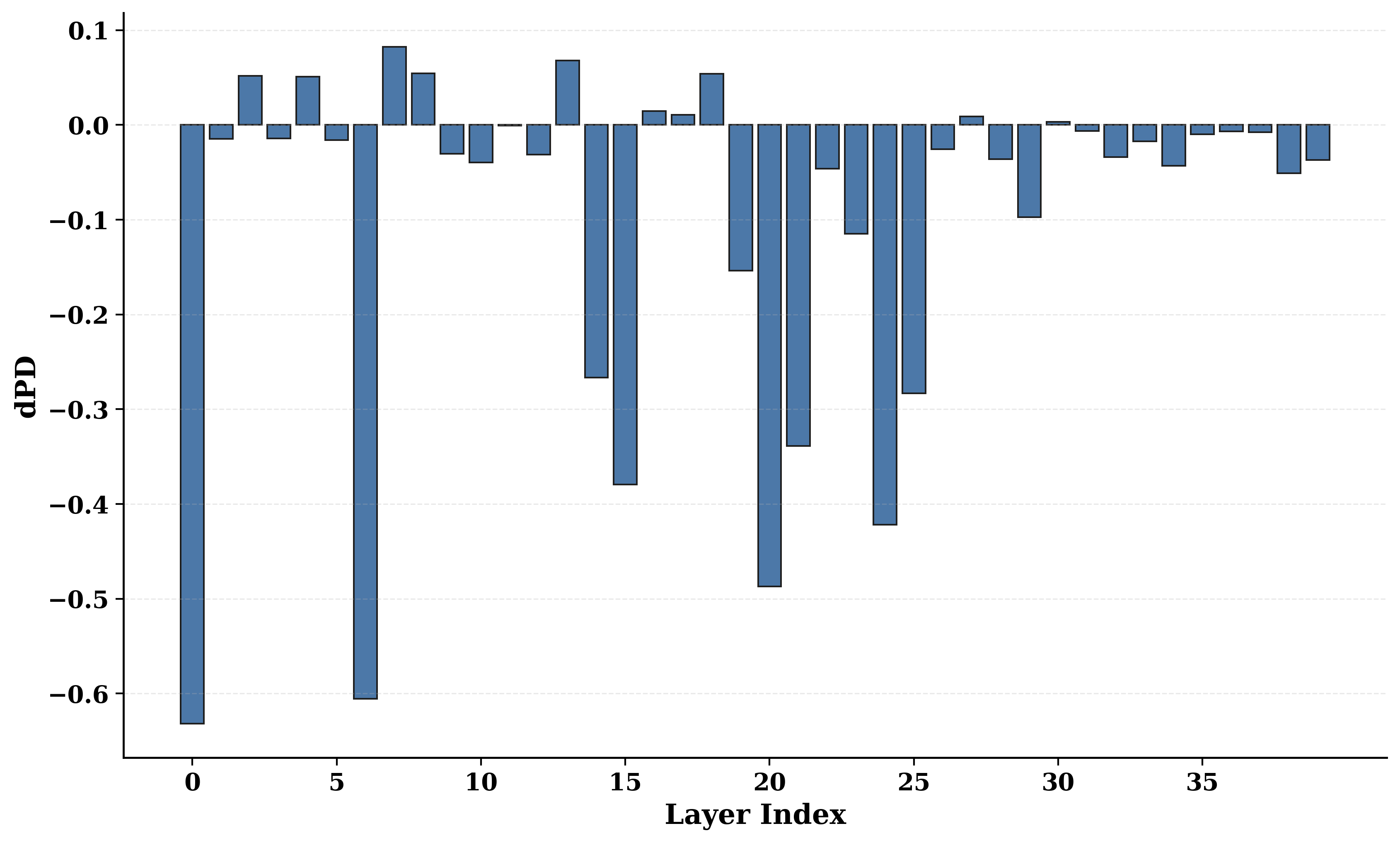}
    \caption{\textbf{Joint attention+MLP-block mean ablation on \textit{Qwen3-14B} under $\pi_{\mathrm{FF}}$.} Per-layer $\mathrm{dPD}$ when each logical block is mean-ablated. Top row: one-hop. Bottom row: two-hop. Columns: Facts / Expression / Query.}
    \label{fig:stage_joint_q14_ff}
    \label{fig:attn_mlp_region_triples}
\end{figure*}

\begin{figure*}[!htbp]
    \centering
    \includegraphics[width=0.32\textwidth]{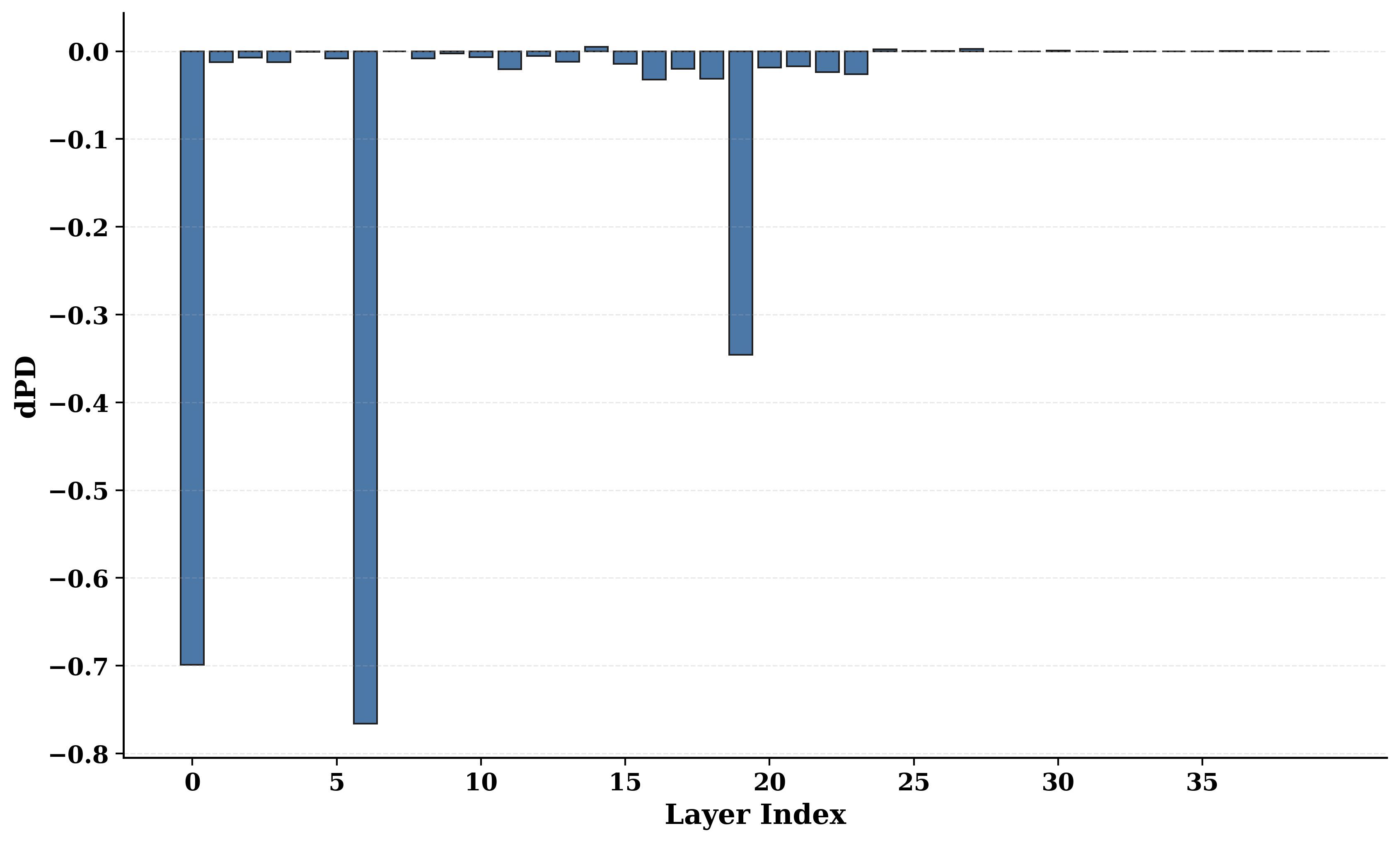}\hfill
    \includegraphics[width=0.32\textwidth]{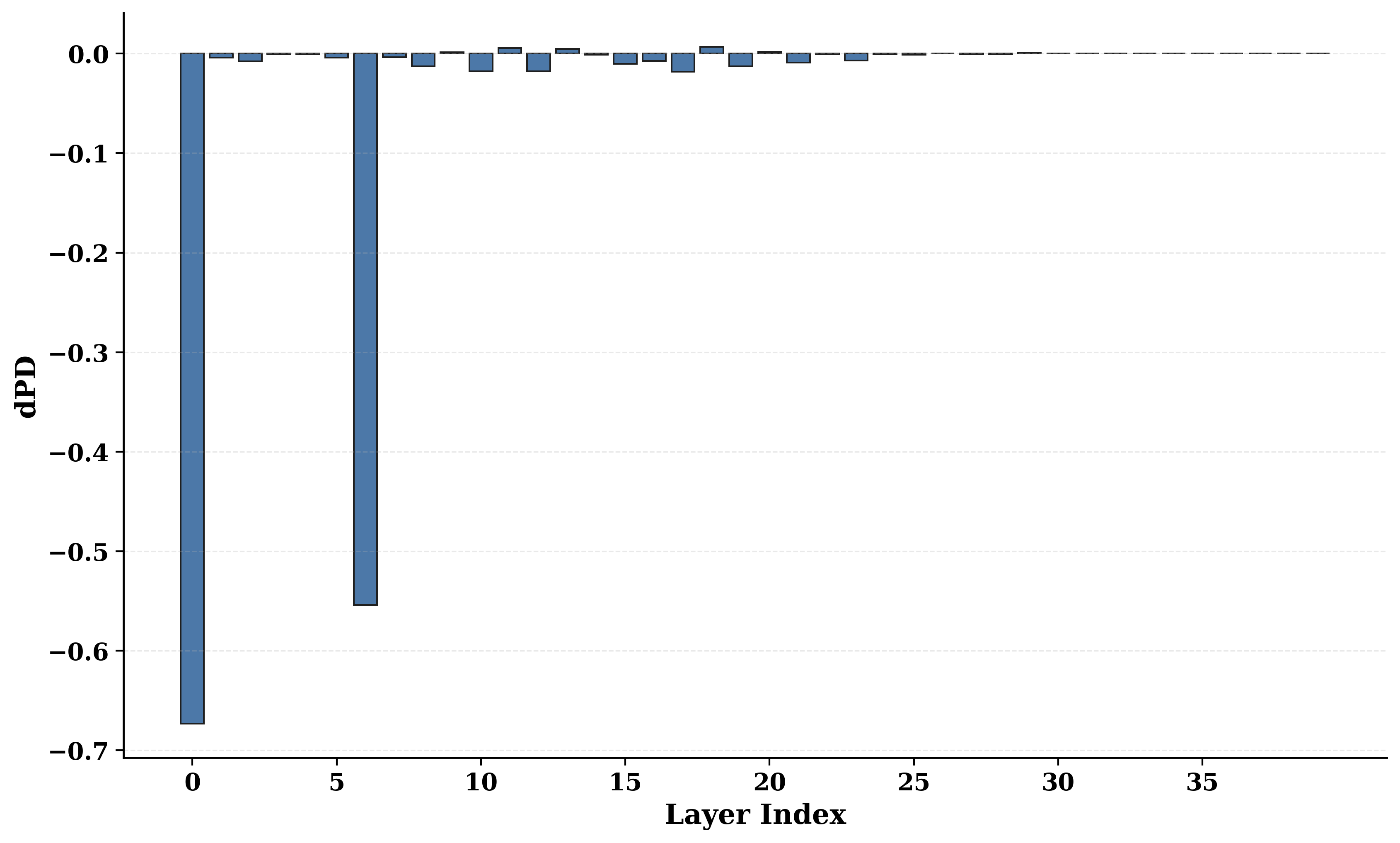}\hfill
    \includegraphics[width=0.32\textwidth]{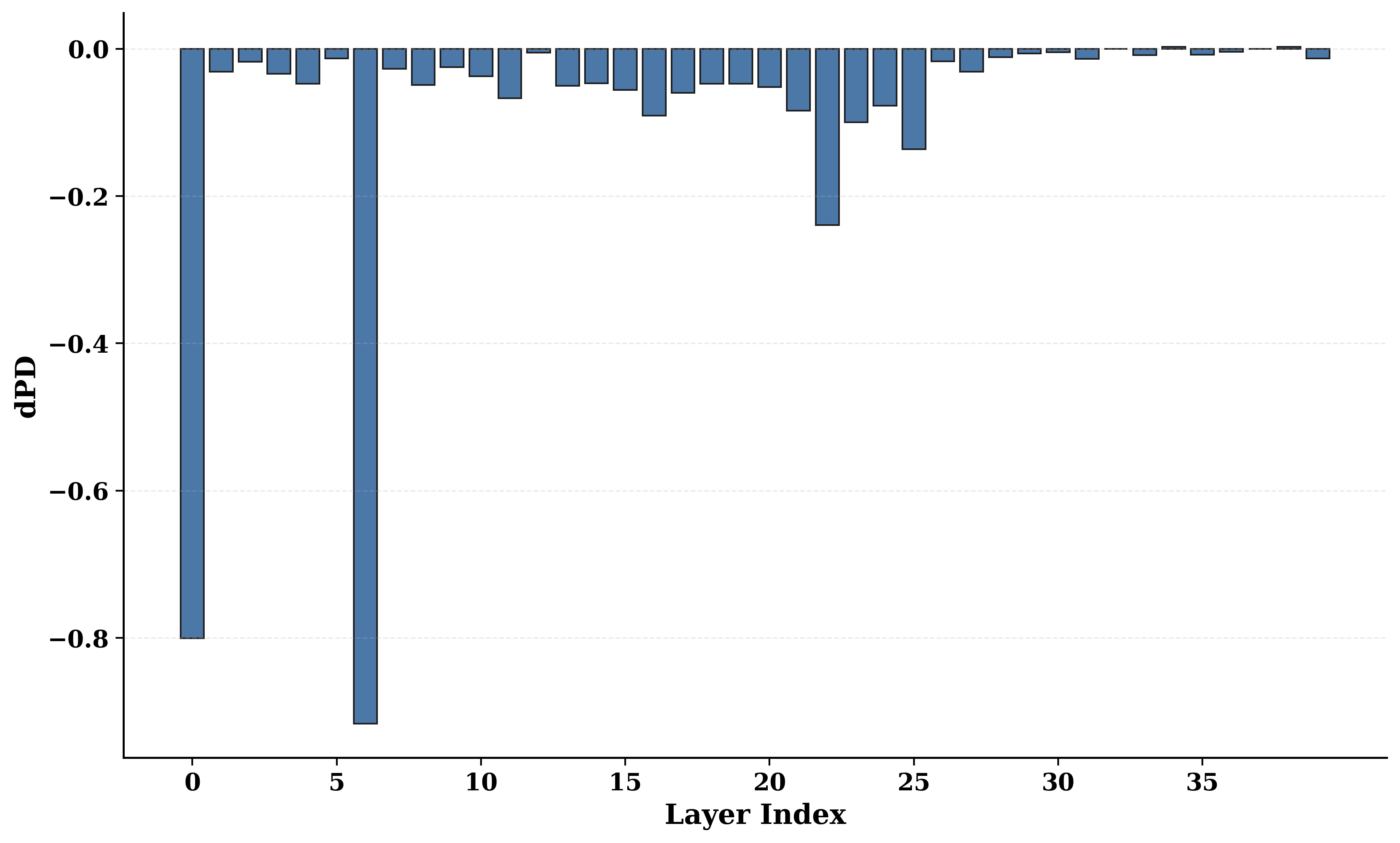}

    \includegraphics[width=0.32\textwidth]{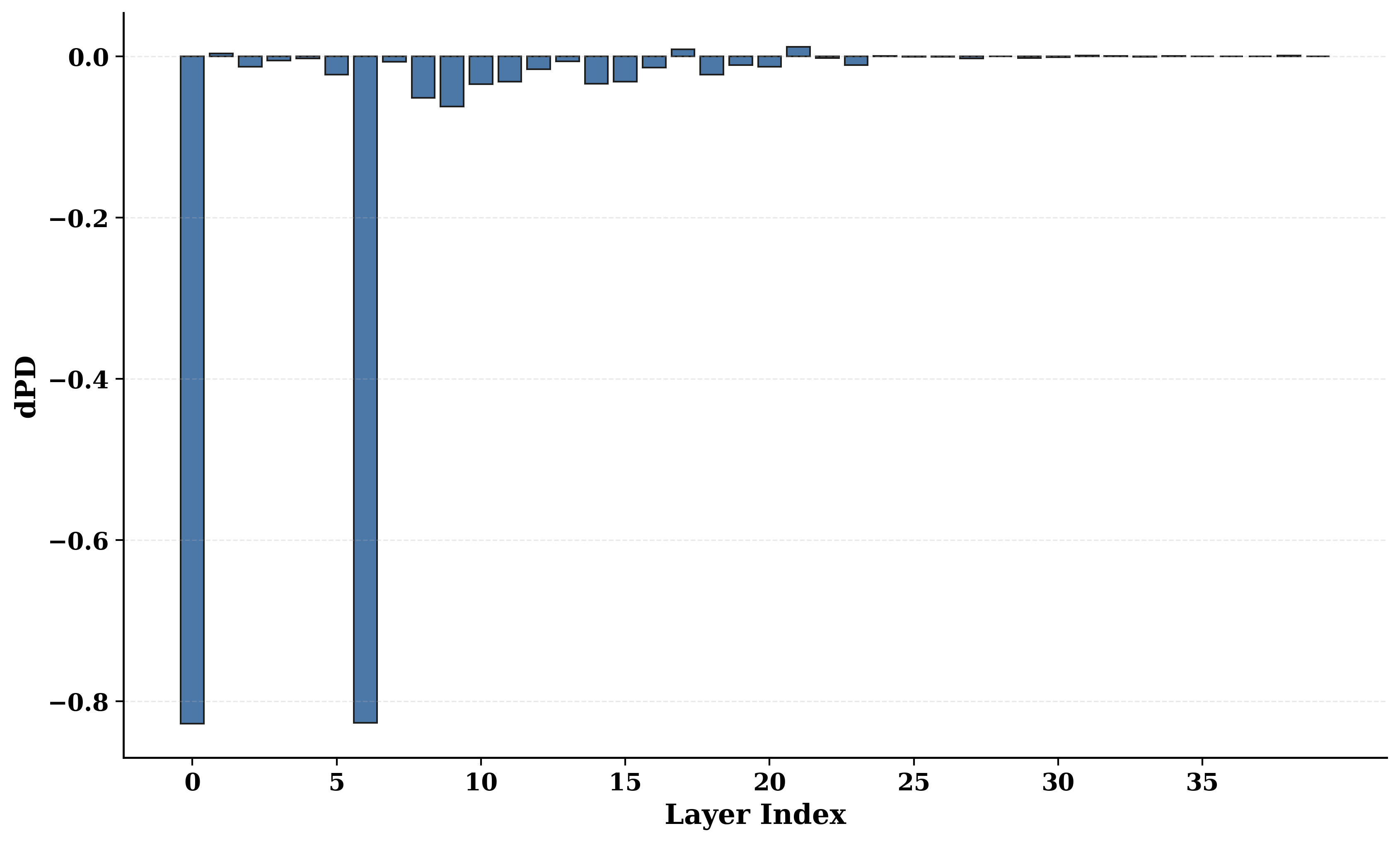}\hfill
    \includegraphics[width=0.32\textwidth]{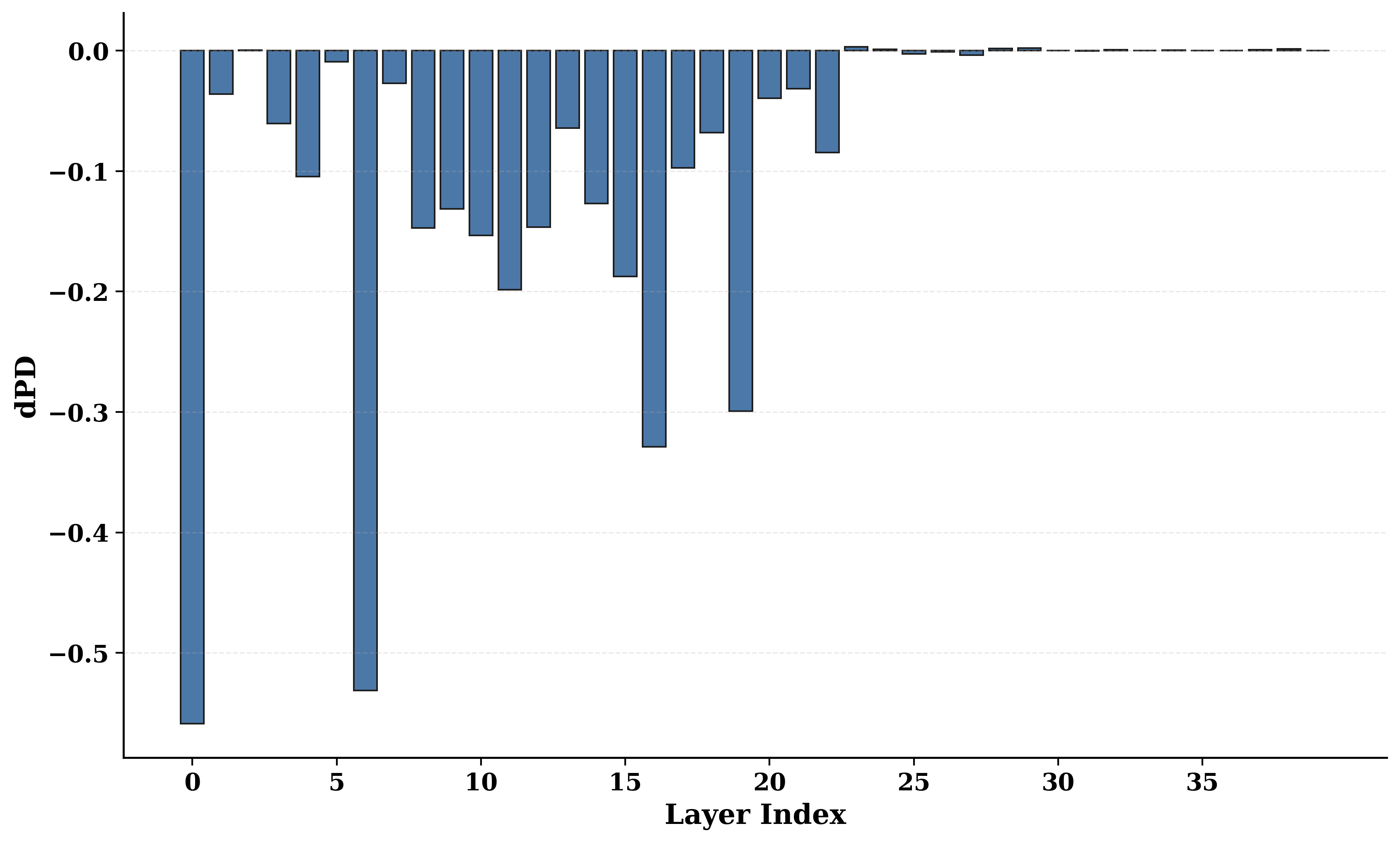}\hfill
    \includegraphics[width=0.32\textwidth]{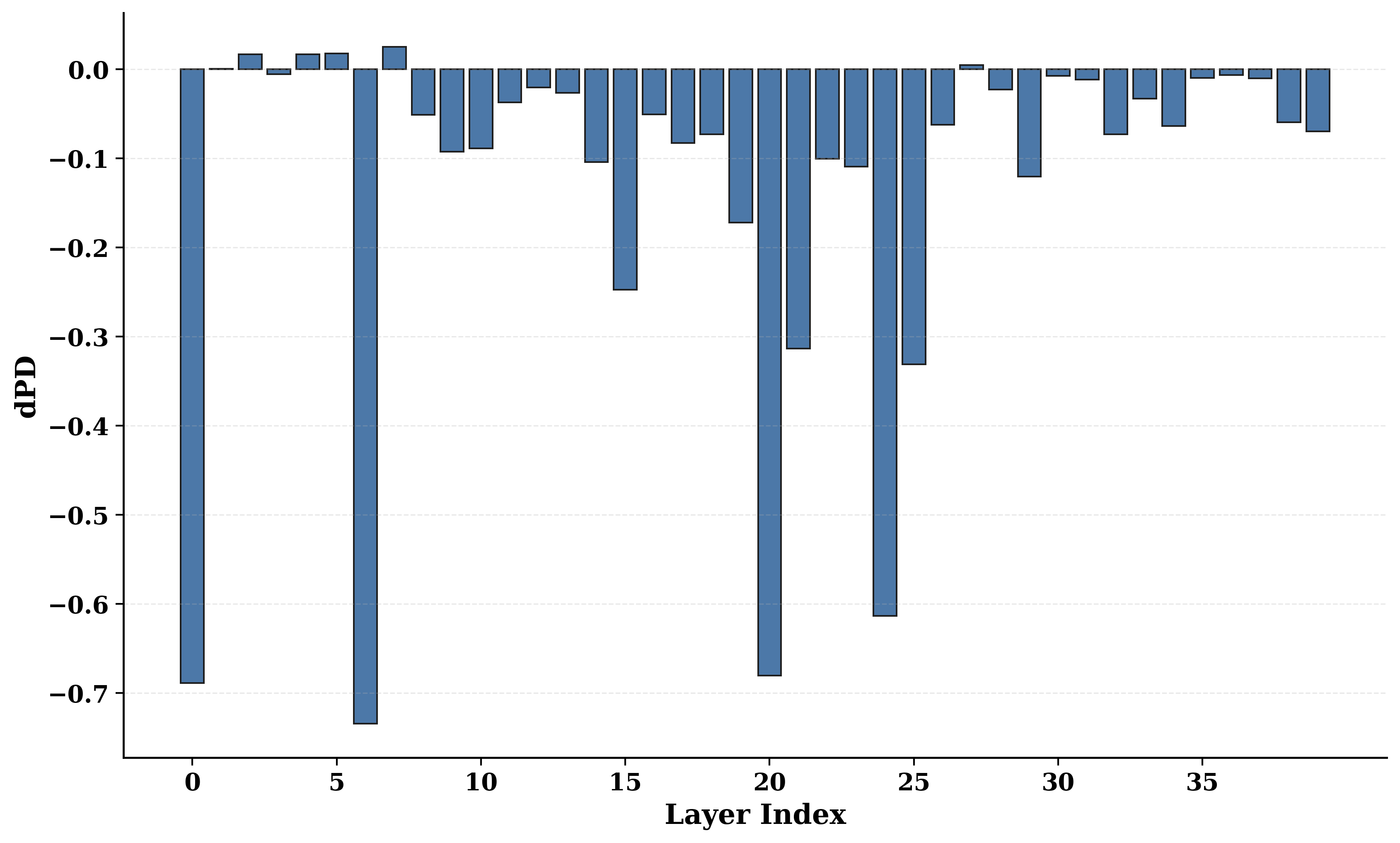}
    \caption{\textbf{Joint attention+MLP-block mean ablation on \textit{Qwen3-14B} under $\pi_{\mathrm{EF}}$.} Per-layer $\mathrm{dPD}$ when each logical block is mean-ablated. Top row: one-hop. Bottom row: two-hop. Columns: Facts / Expression / Query.}
    \label{fig:stage_joint_q14_ef}
\end{figure*}

\begin{figure*}[!htbp]
    \centering
    \includegraphics[width=0.32\textwidth]{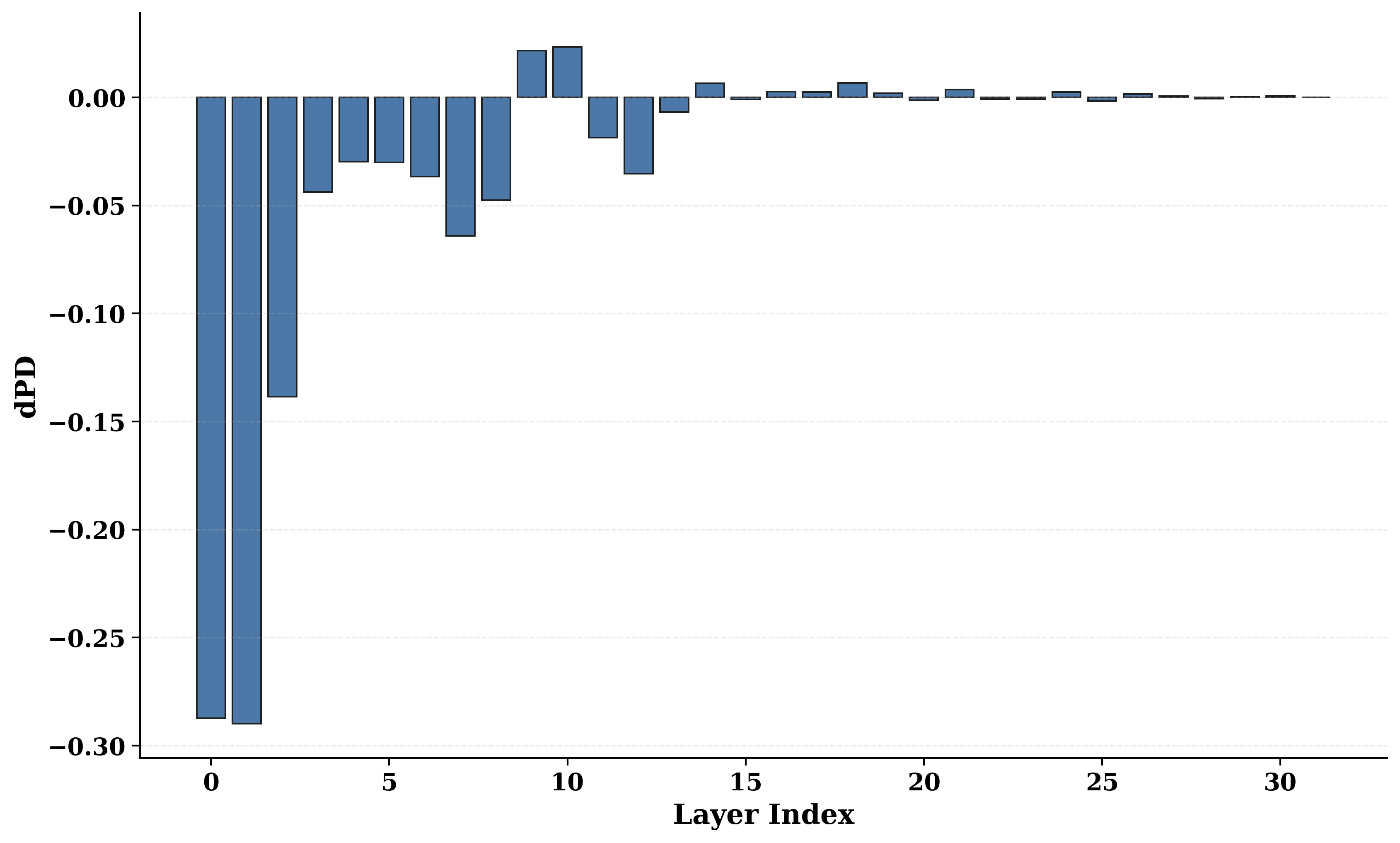}\hfill
    \includegraphics[width=0.32\textwidth]{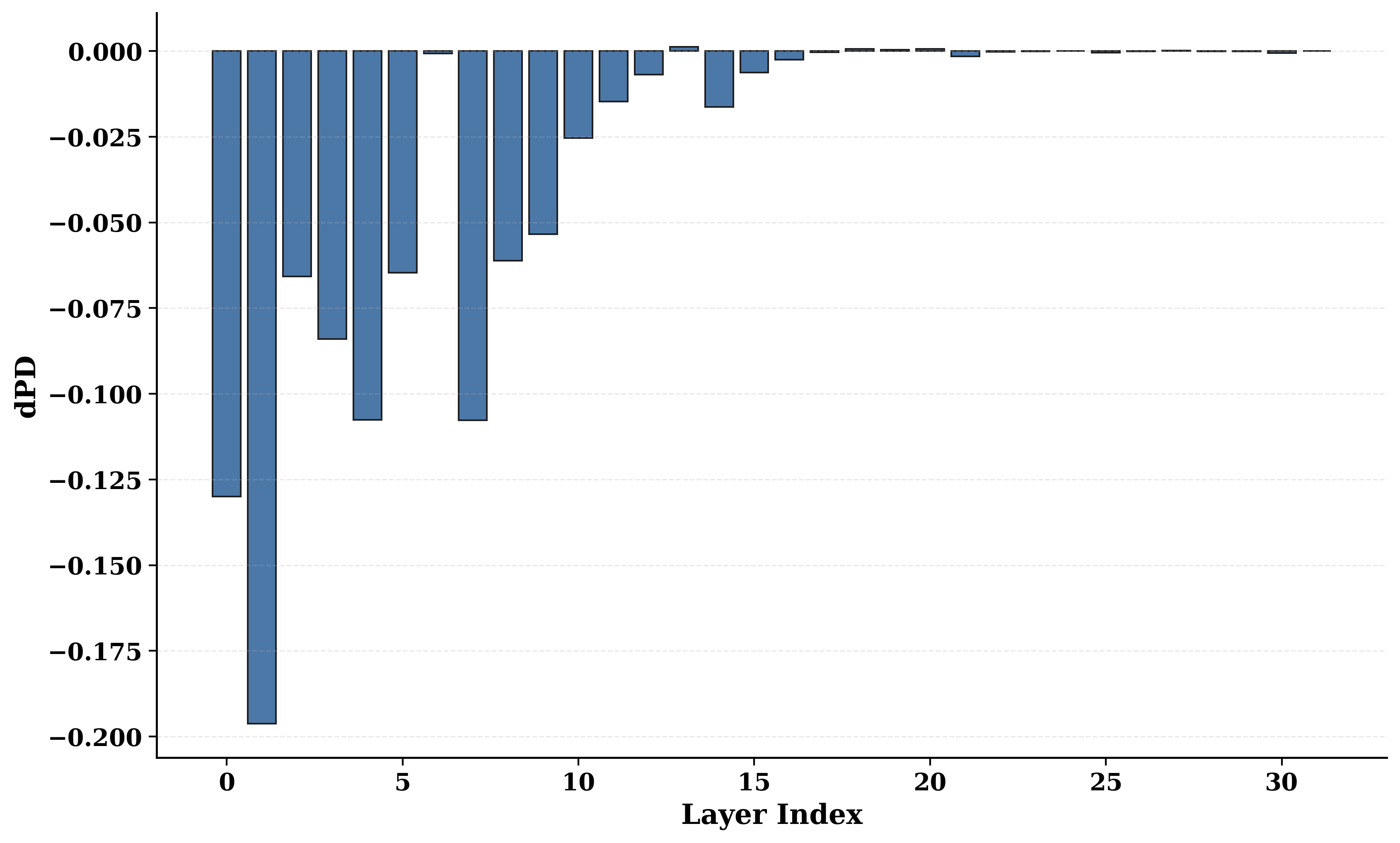}\hfill
    \includegraphics[width=0.32\textwidth]{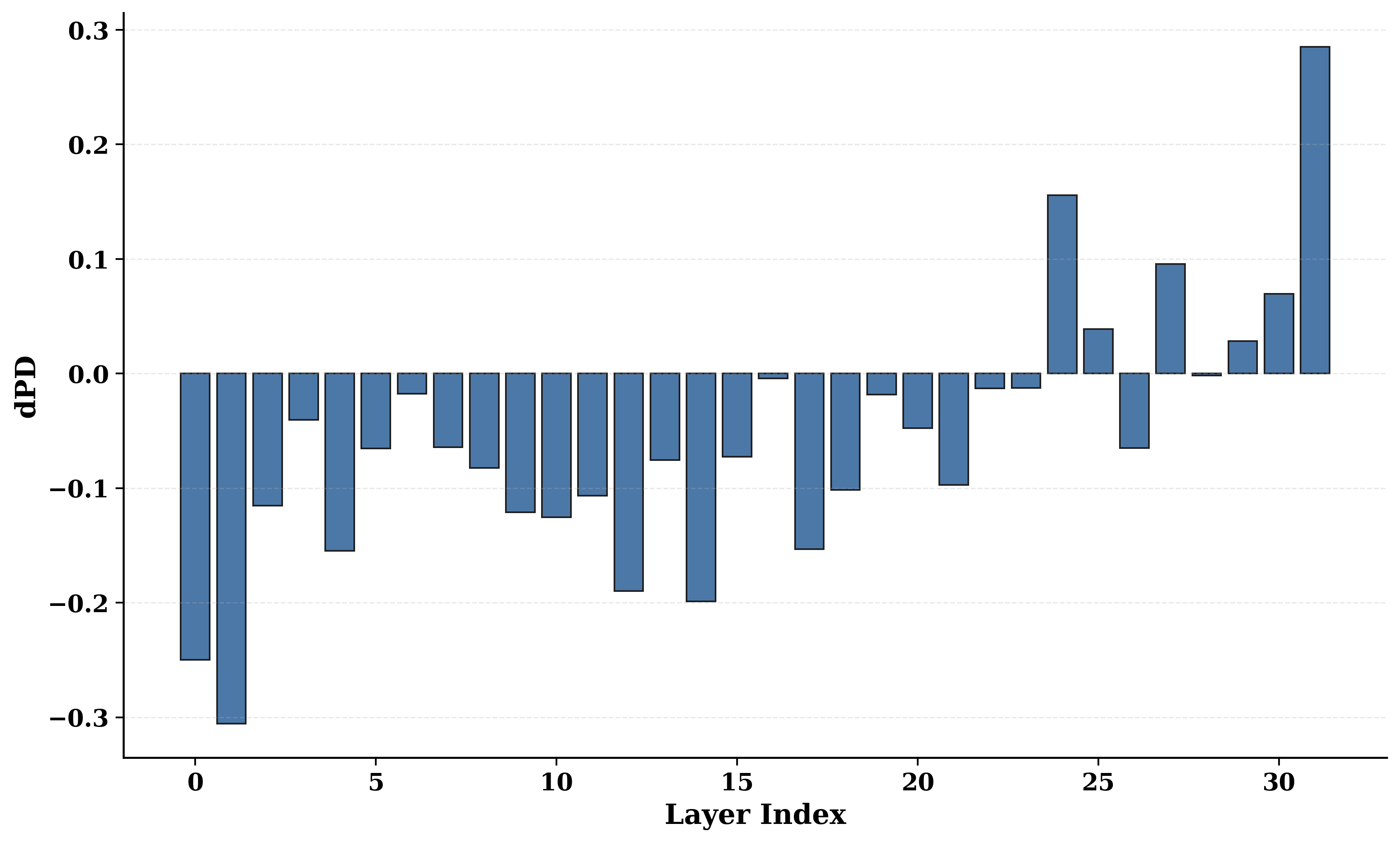}

    \includegraphics[width=0.32\textwidth]{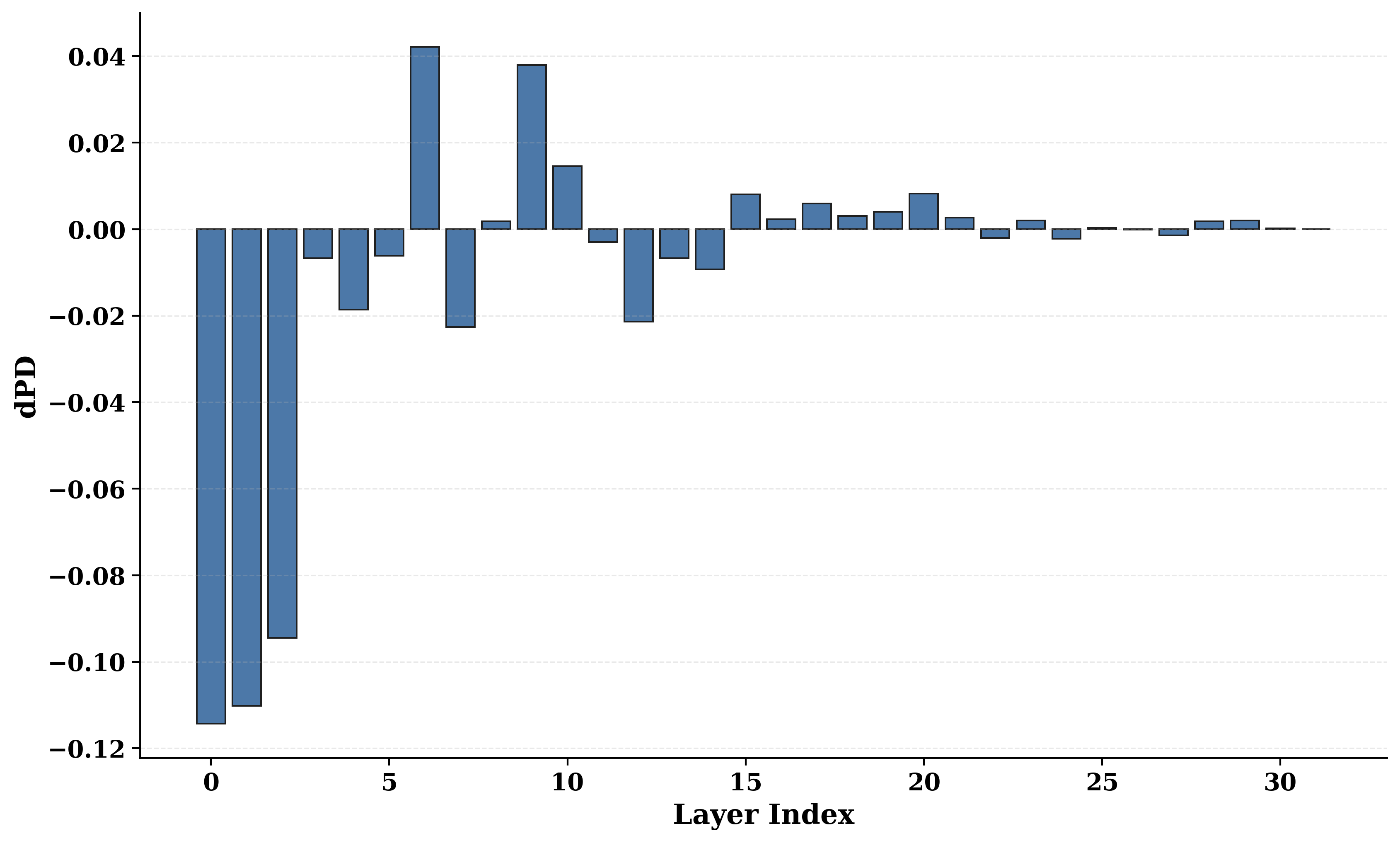}\hfill
    \includegraphics[width=0.32\textwidth]{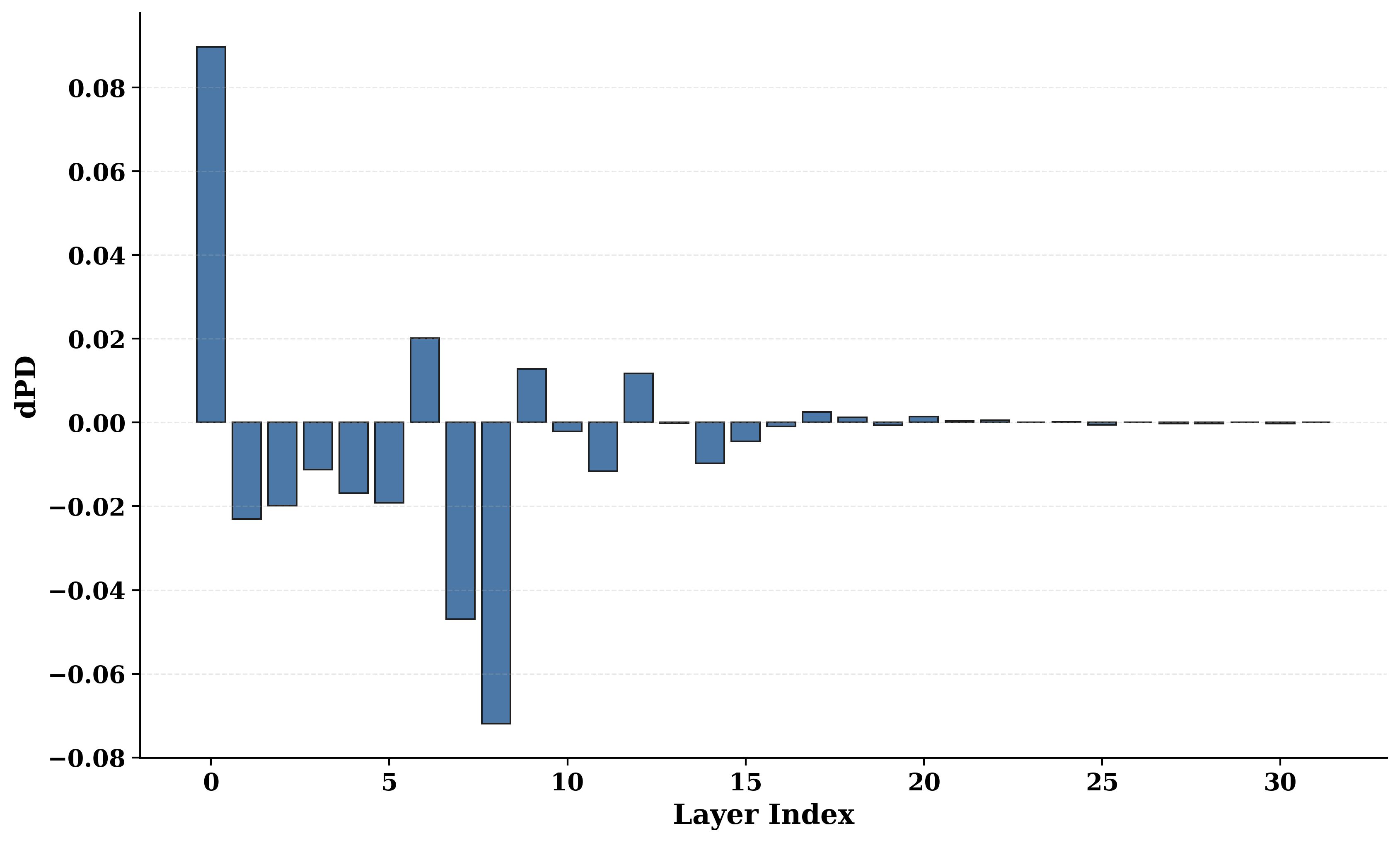}\hfill
    \includegraphics[width=0.32\textwidth]{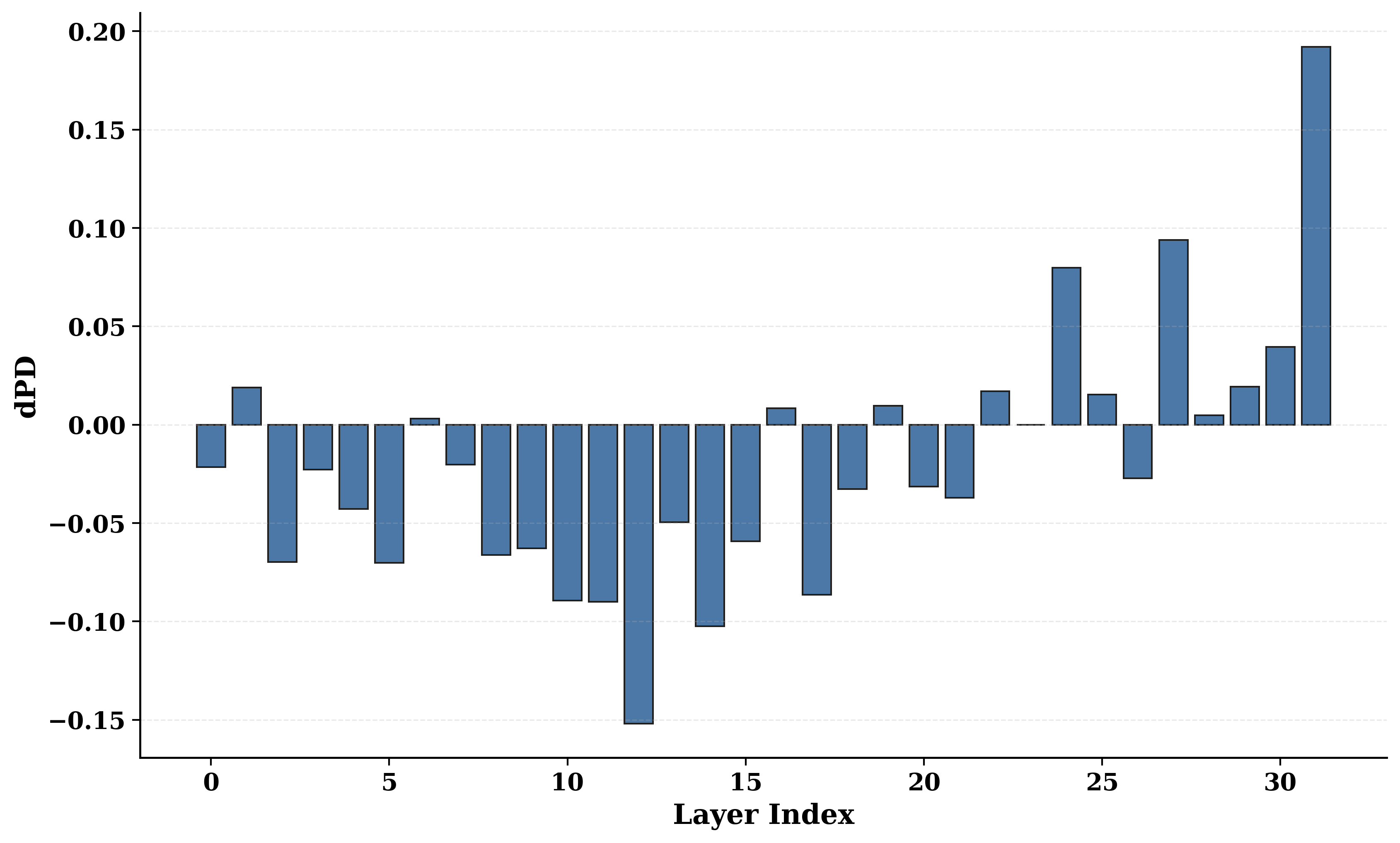}
    \caption{\textbf{Joint attention+MLP-block mean ablation on \textit{Llama-3.1-8B-Instruct} under $\pi_{\mathrm{FF}}$.} Per-layer $\mathrm{dPD}$ when each logical block is mean-ablated. Top row: one-hop. Bottom row: two-hop. Columns: Facts / Expression / Query.}
    \label{fig:stage_joint_llama_ff}
\end{figure*}

\begin{figure*}[!htbp]
    \centering
    \includegraphics[width=0.32\textwidth]{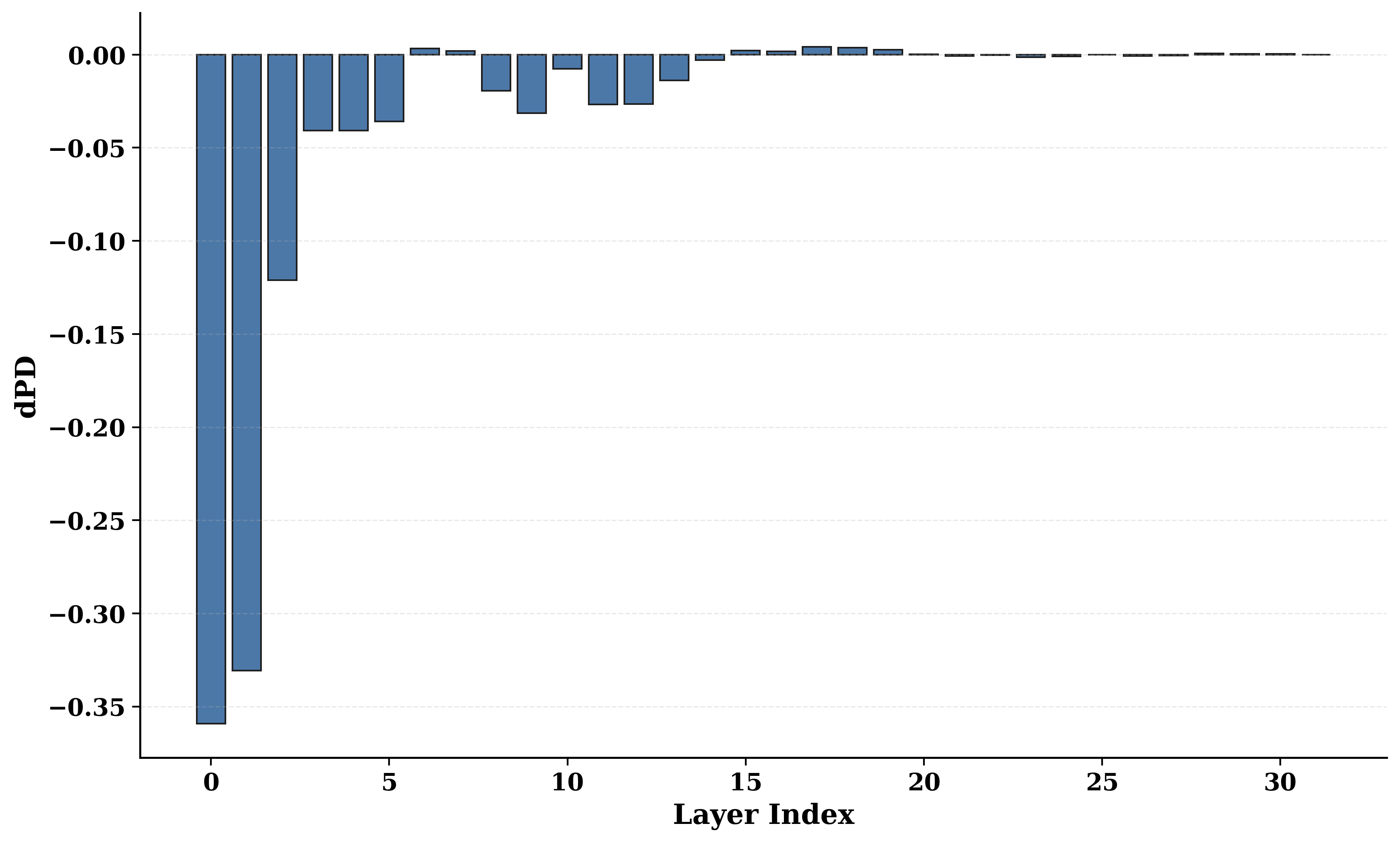}\hfill
    \includegraphics[width=0.32\textwidth]{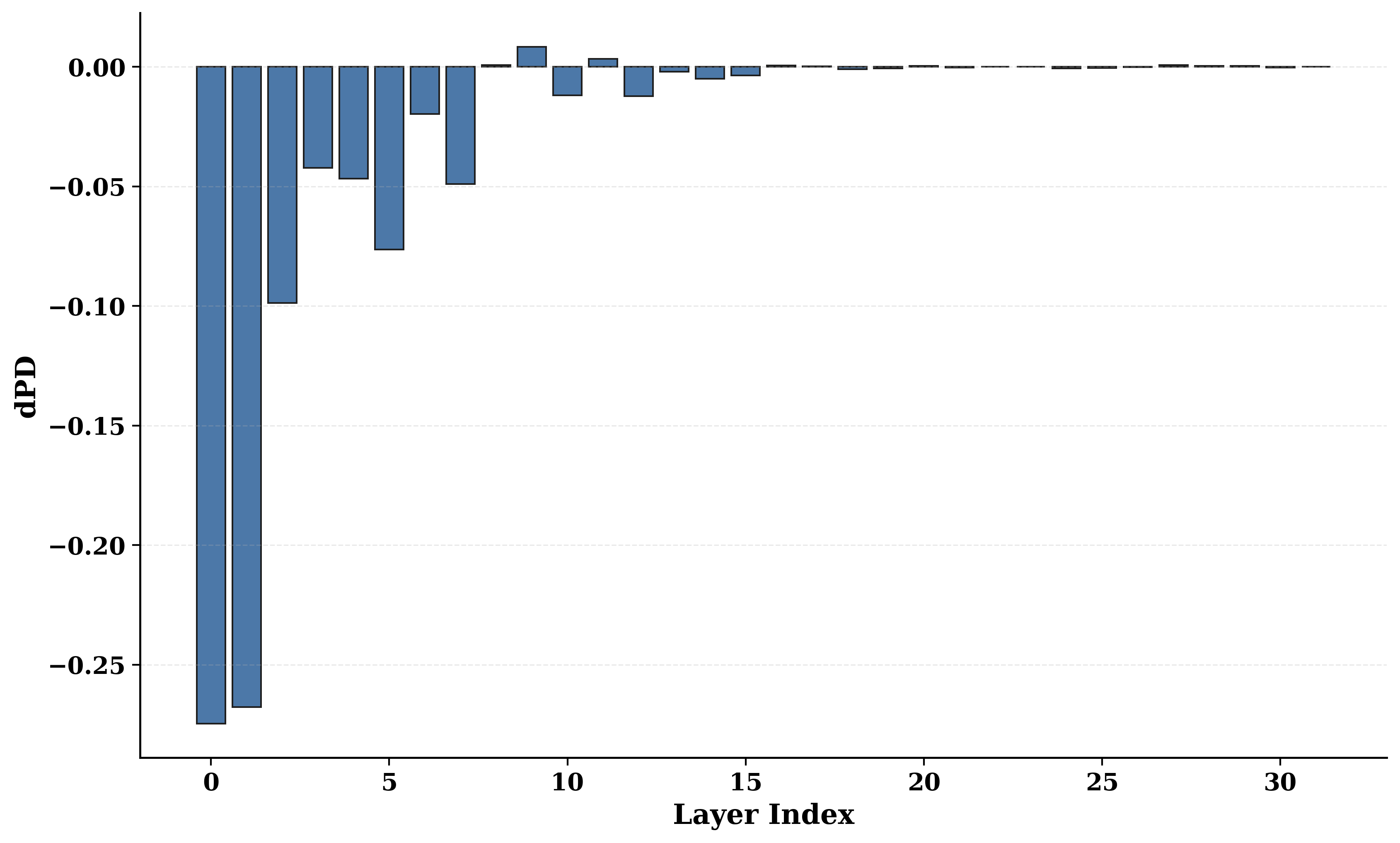}\hfill
    \includegraphics[width=0.32\textwidth]{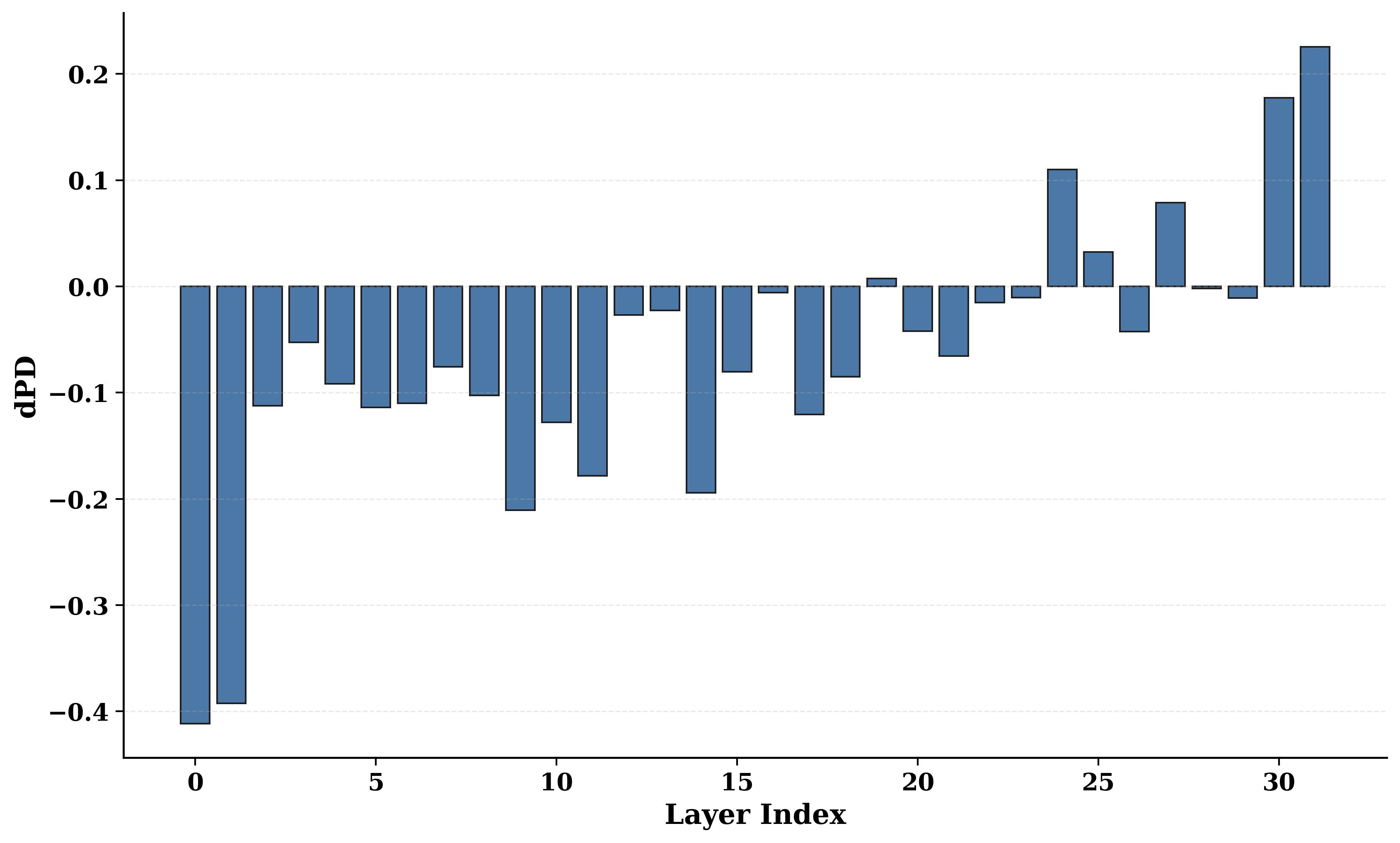}

    \includegraphics[width=0.32\textwidth]{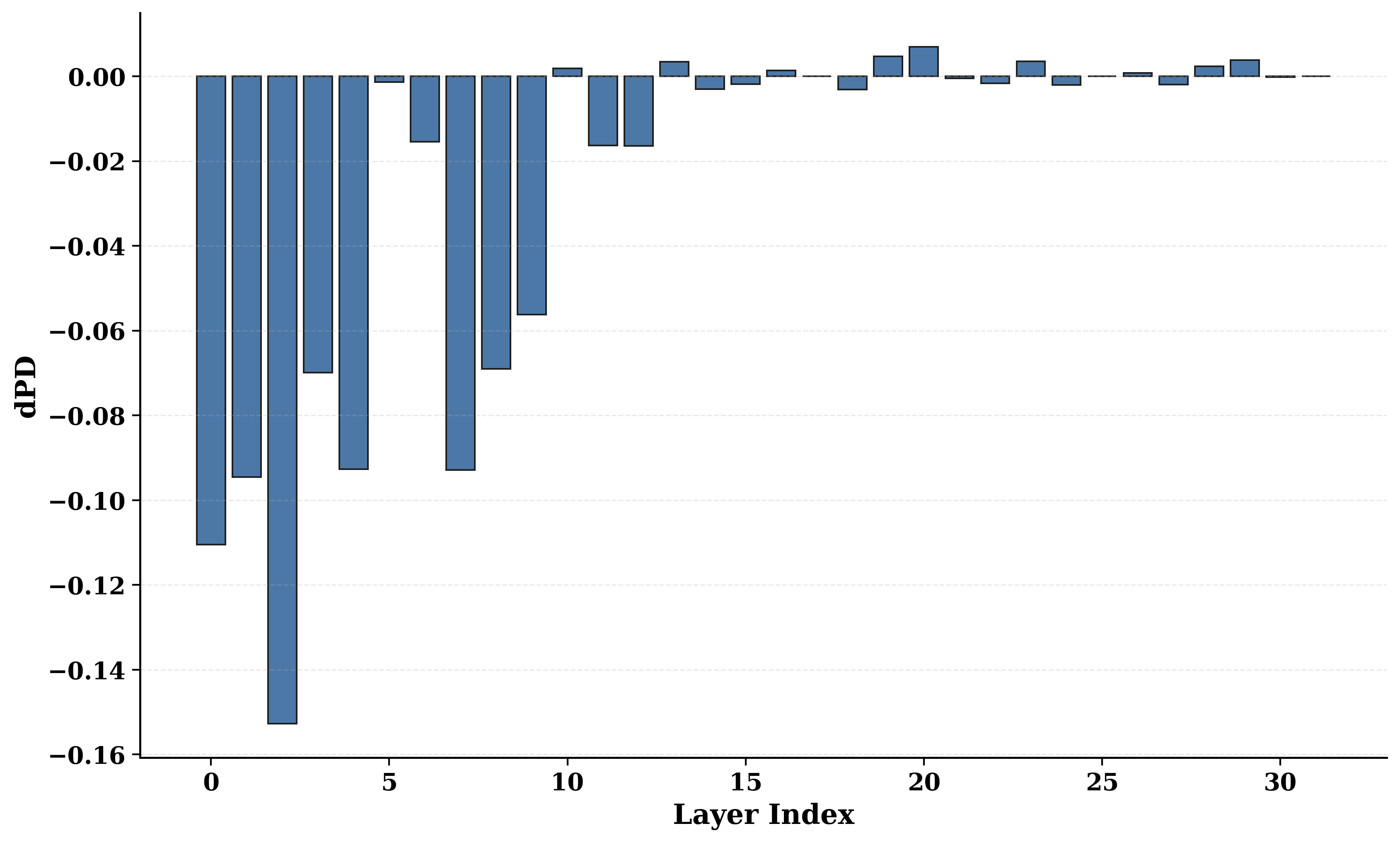}\hfill
    \includegraphics[width=0.32\textwidth]{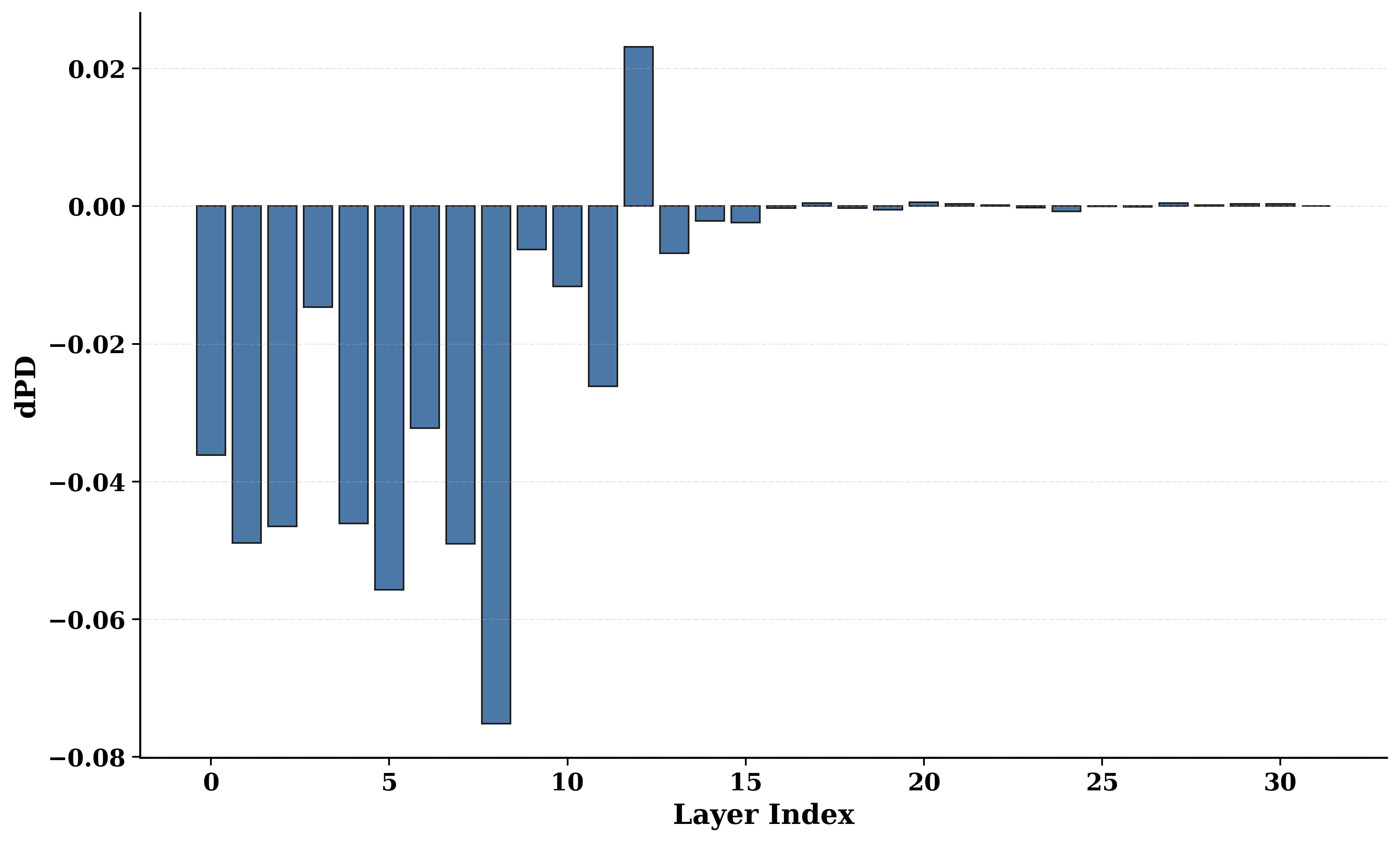}\hfill
    \includegraphics[width=0.32\textwidth]{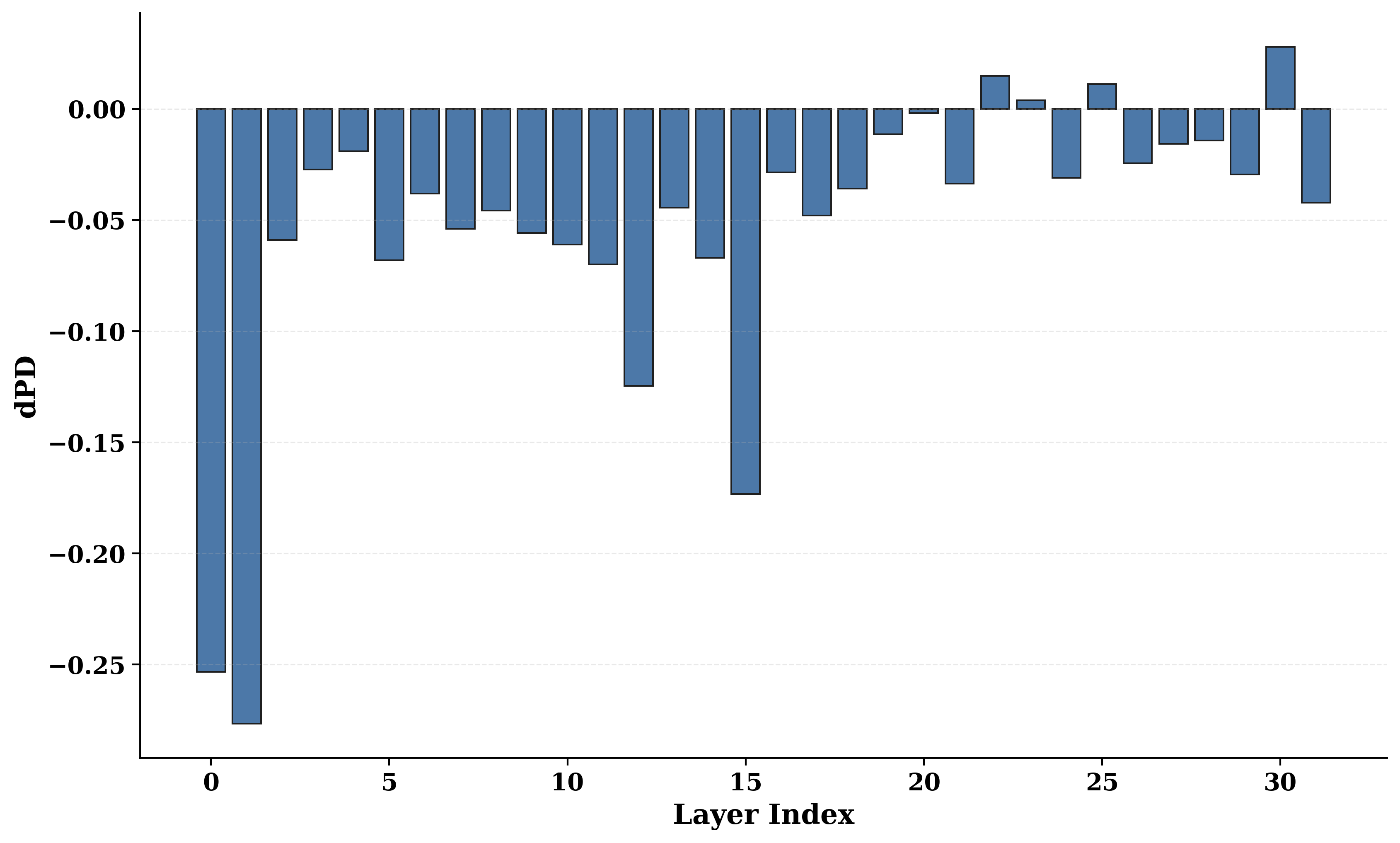}
    \caption{\textbf{Joint attention+MLP-block mean ablation on \textit{Llama-3.1-8B-Instruct} under $\pi_{\mathrm{EF}}$.} Per-layer $\mathrm{dPD}$ when each logical block is mean-ablated. Top row: one-hop. Bottom row: two-hop. Columns: Facts / Expression / Query.}
    \label{fig:stage_joint_llama_ef}
\end{figure*}

\begin{figure*}[!htbp]
    \centering
    \includegraphics[width=0.32\textwidth]{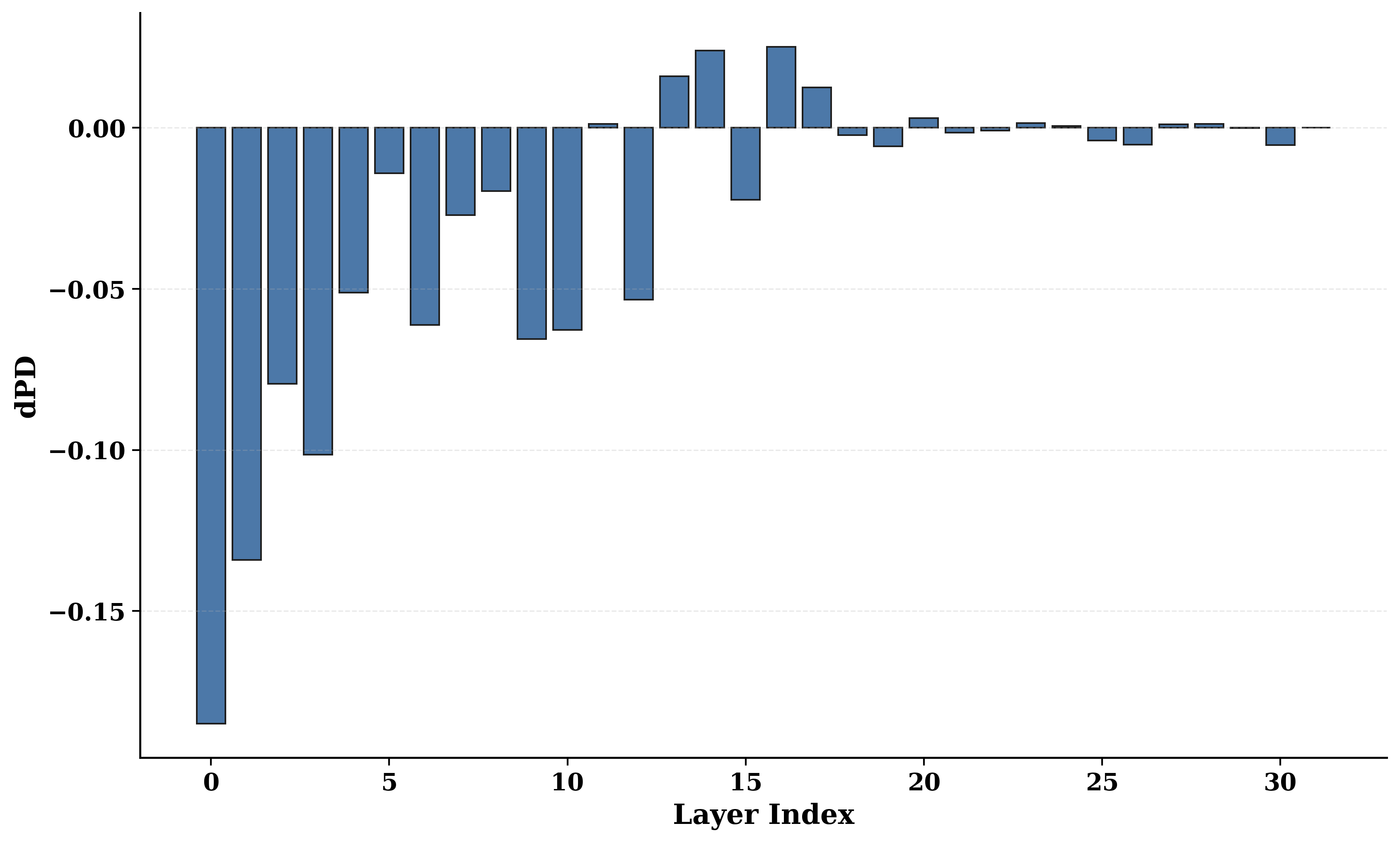}\hfill
    \includegraphics[width=0.32\textwidth]{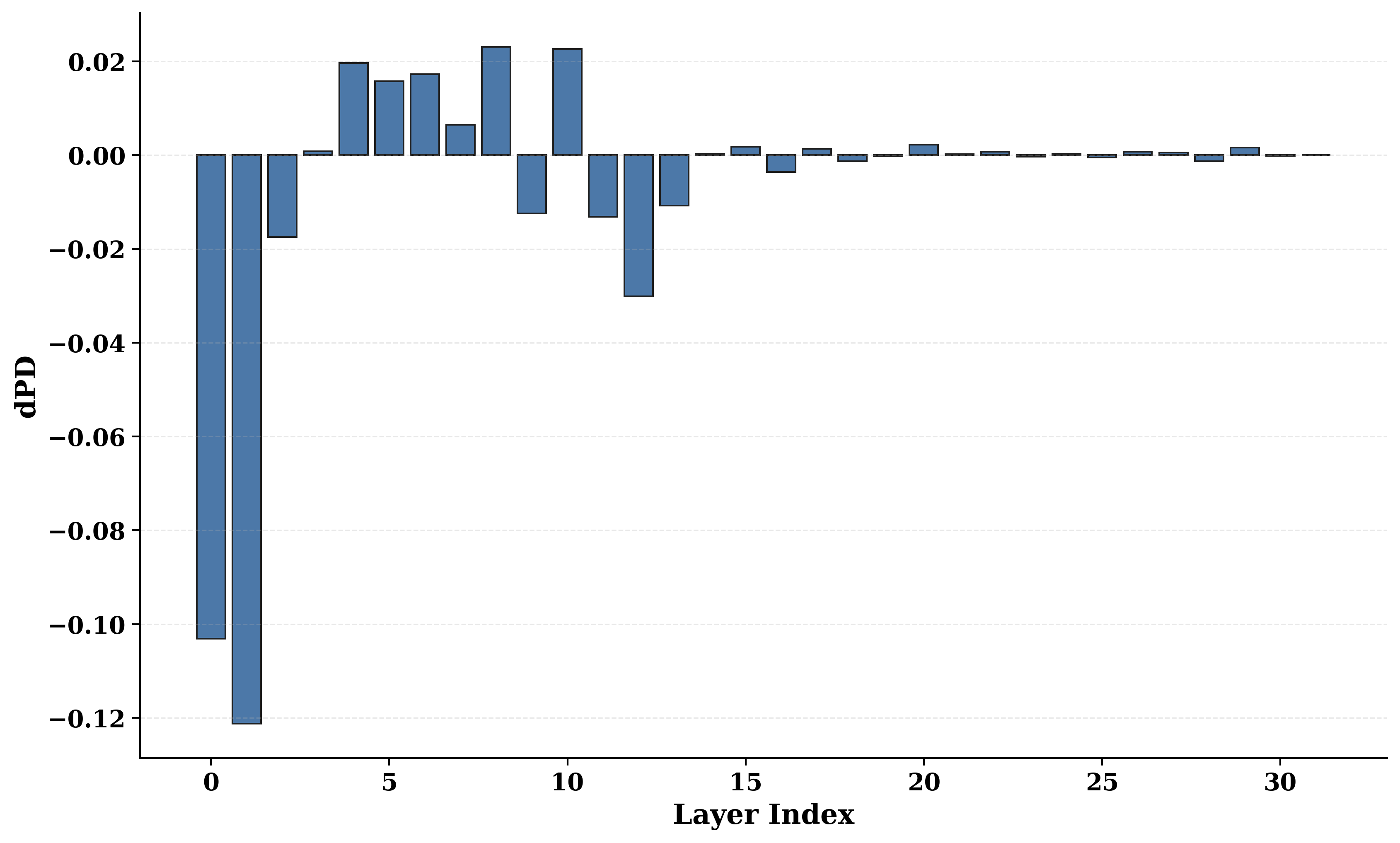}\hfill
    \includegraphics[width=0.32\textwidth]{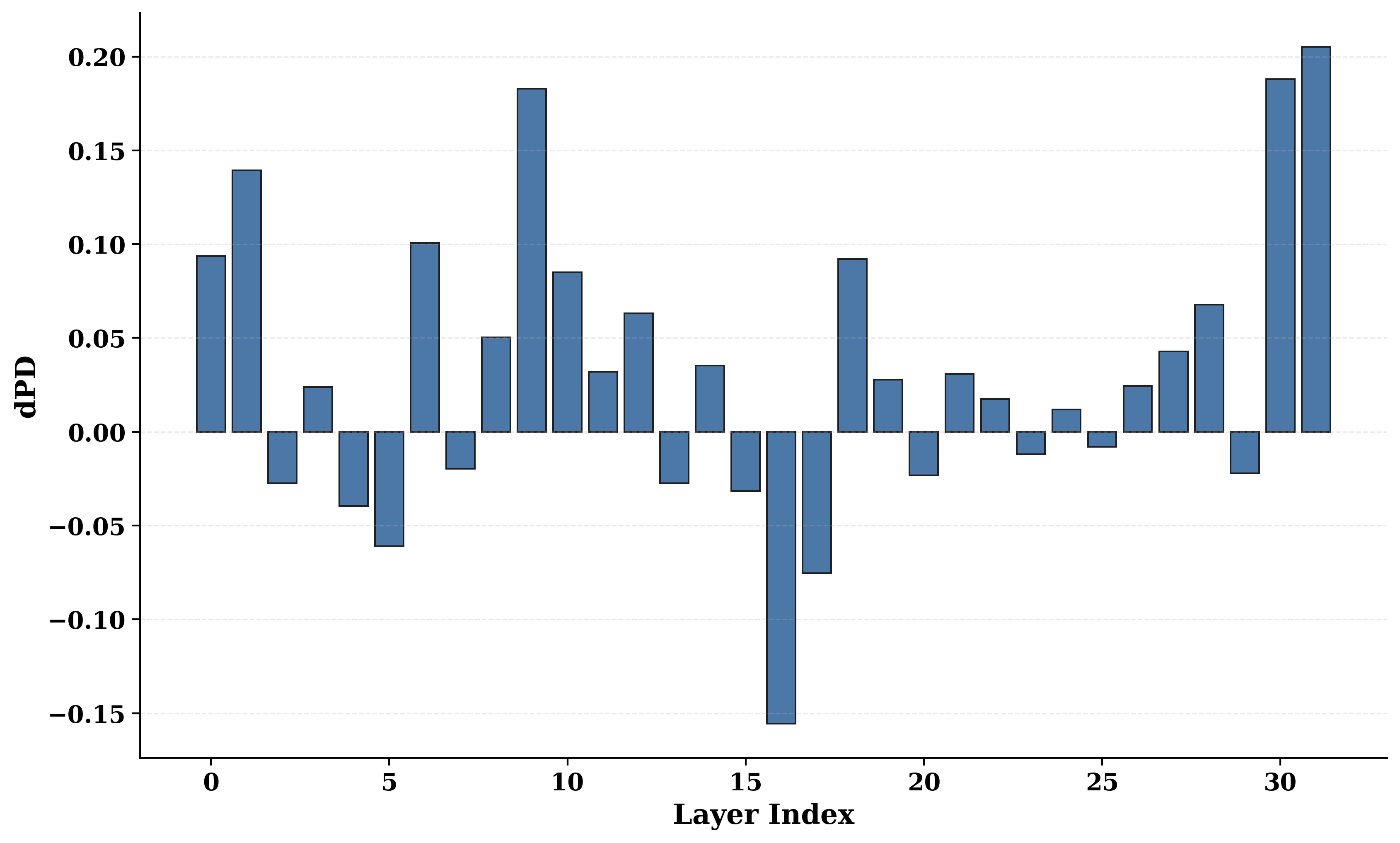}

    \includegraphics[width=0.32\textwidth]{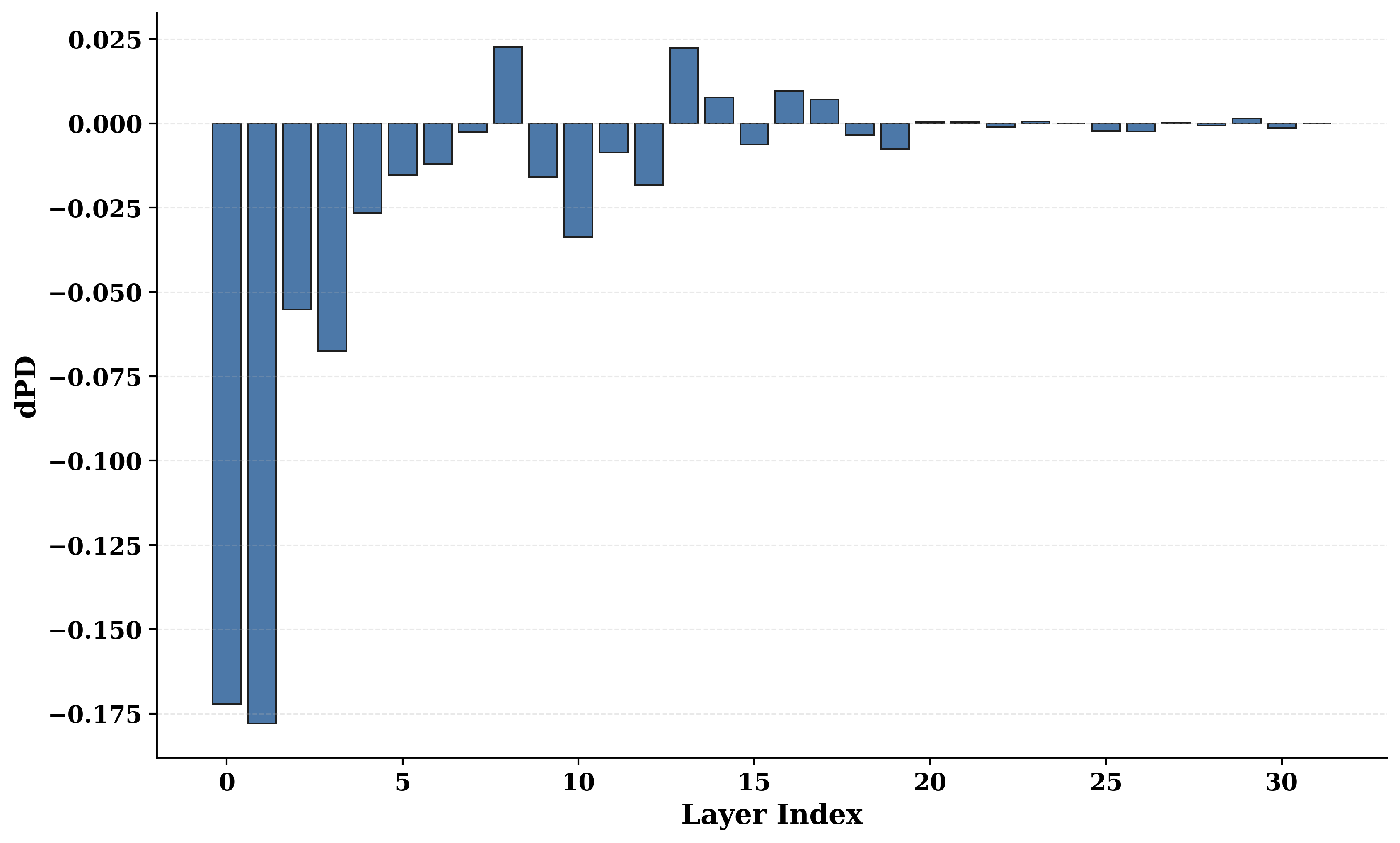}\hfill
    \includegraphics[width=0.32\textwidth]{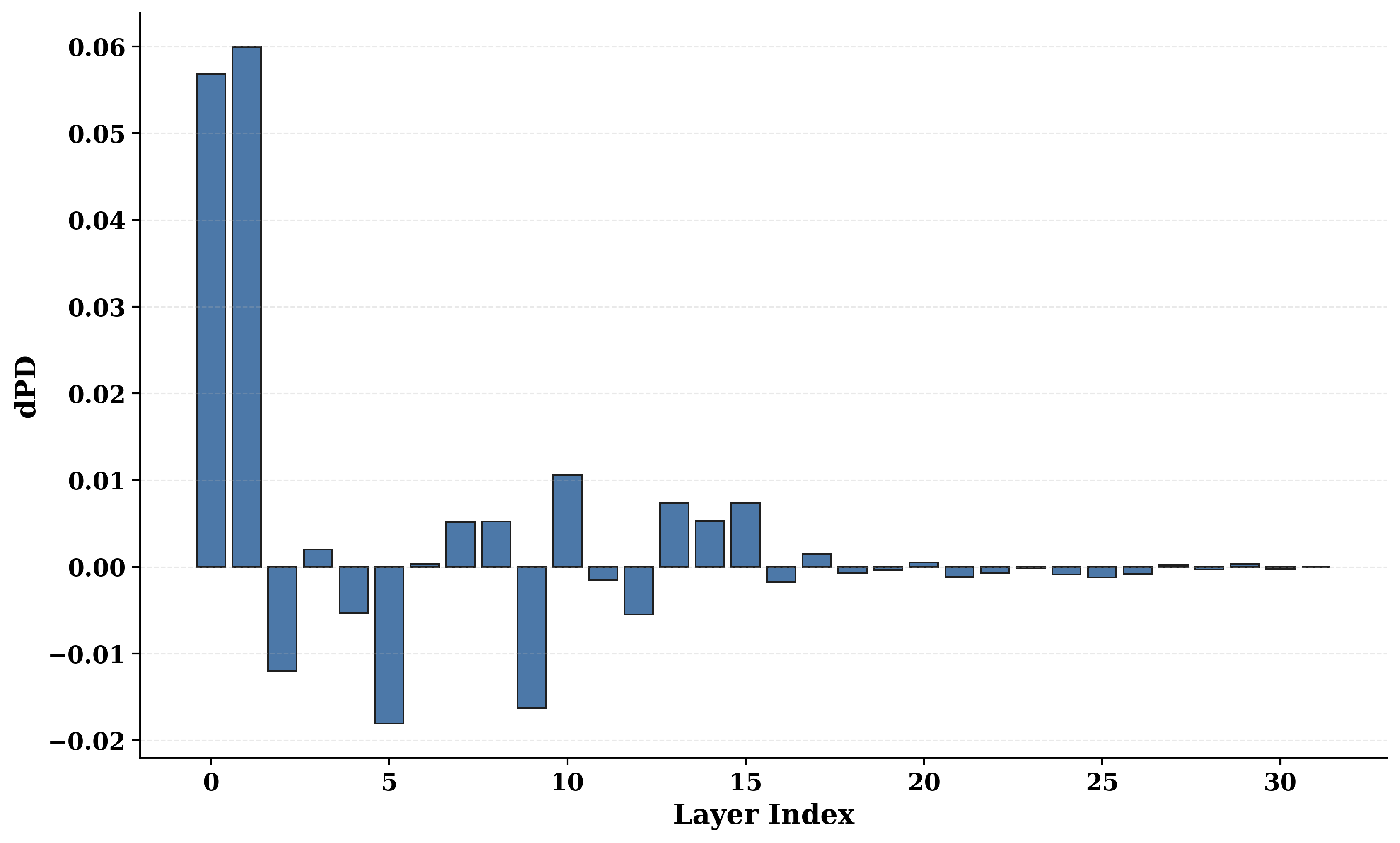}\hfill
    \includegraphics[width=0.32\textwidth]{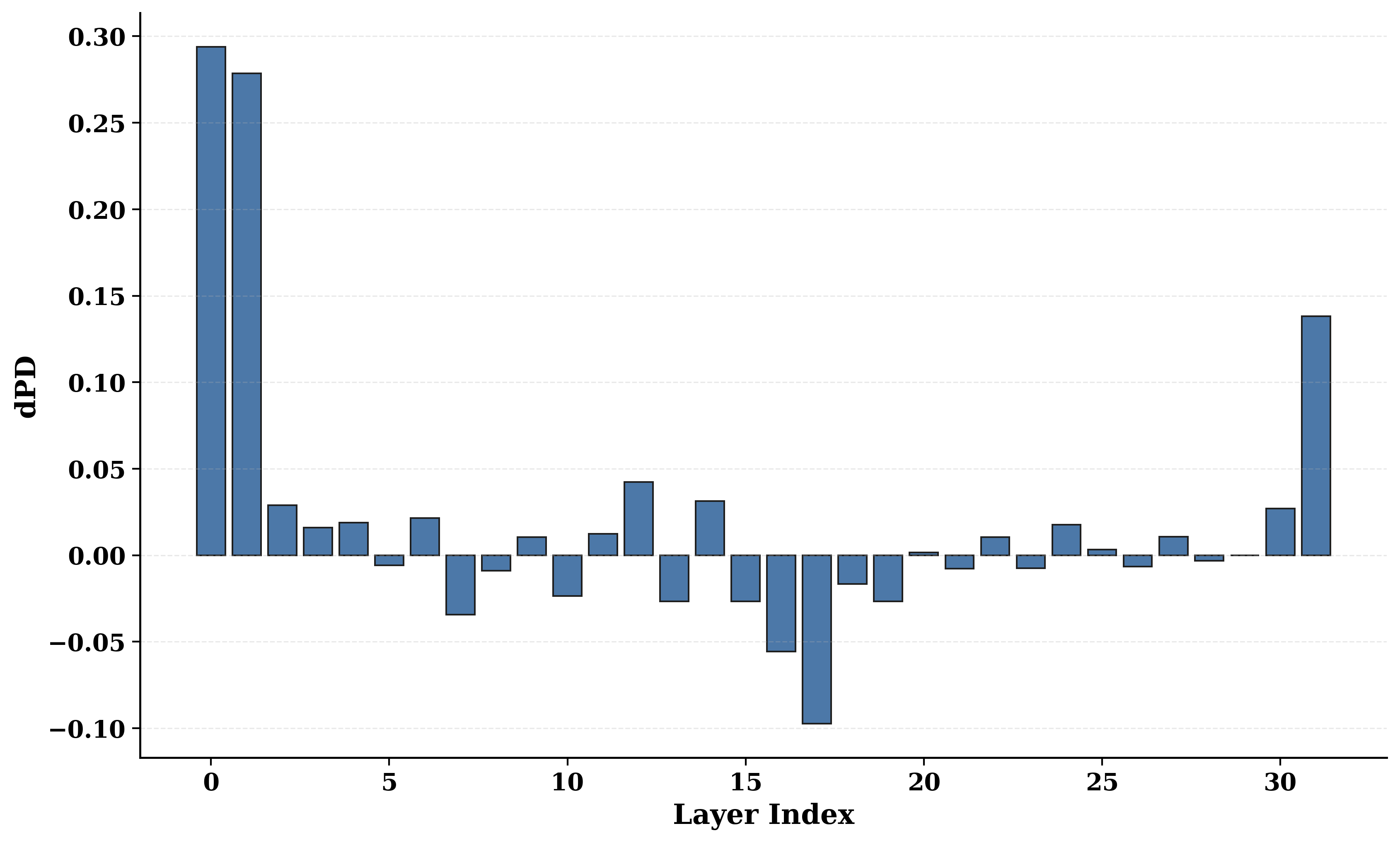}
    \caption{\textbf{Joint attention+MLP-block mean ablation on \textit{Mistral-7B-Instruct} under $\pi_{\mathrm{FF}}$.} Per-layer $\mathrm{dPD}$ when each logical block is mean-ablated. Top row: one-hop. Bottom row: two-hop. Columns: Facts / Expression / Query.}
    \label{fig:stage_joint_mistral_ff}
\end{figure*}

\begin{figure*}[!htbp]
    \centering
    \includegraphics[width=0.32\textwidth]{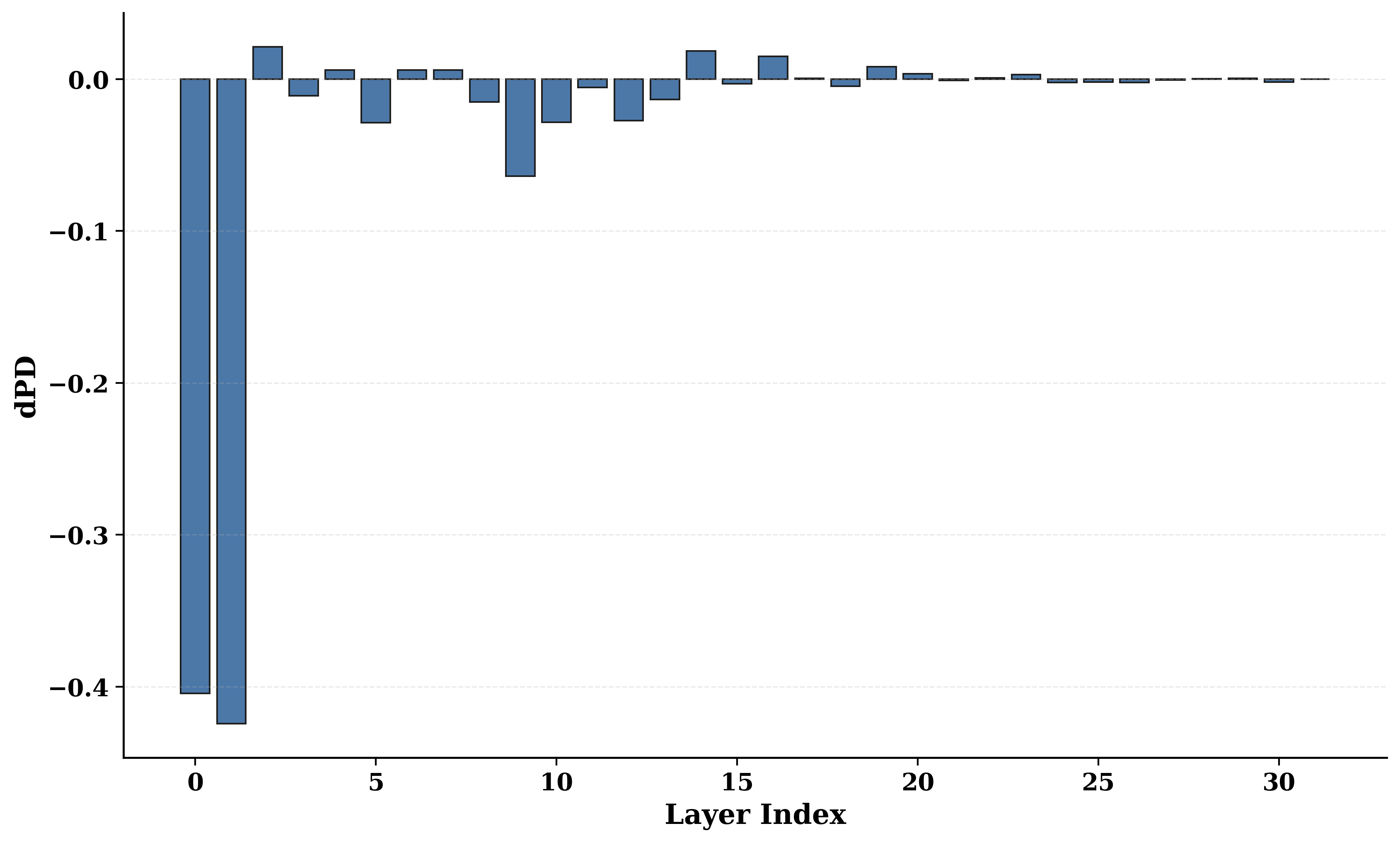}\hfill
    \includegraphics[width=0.32\textwidth]{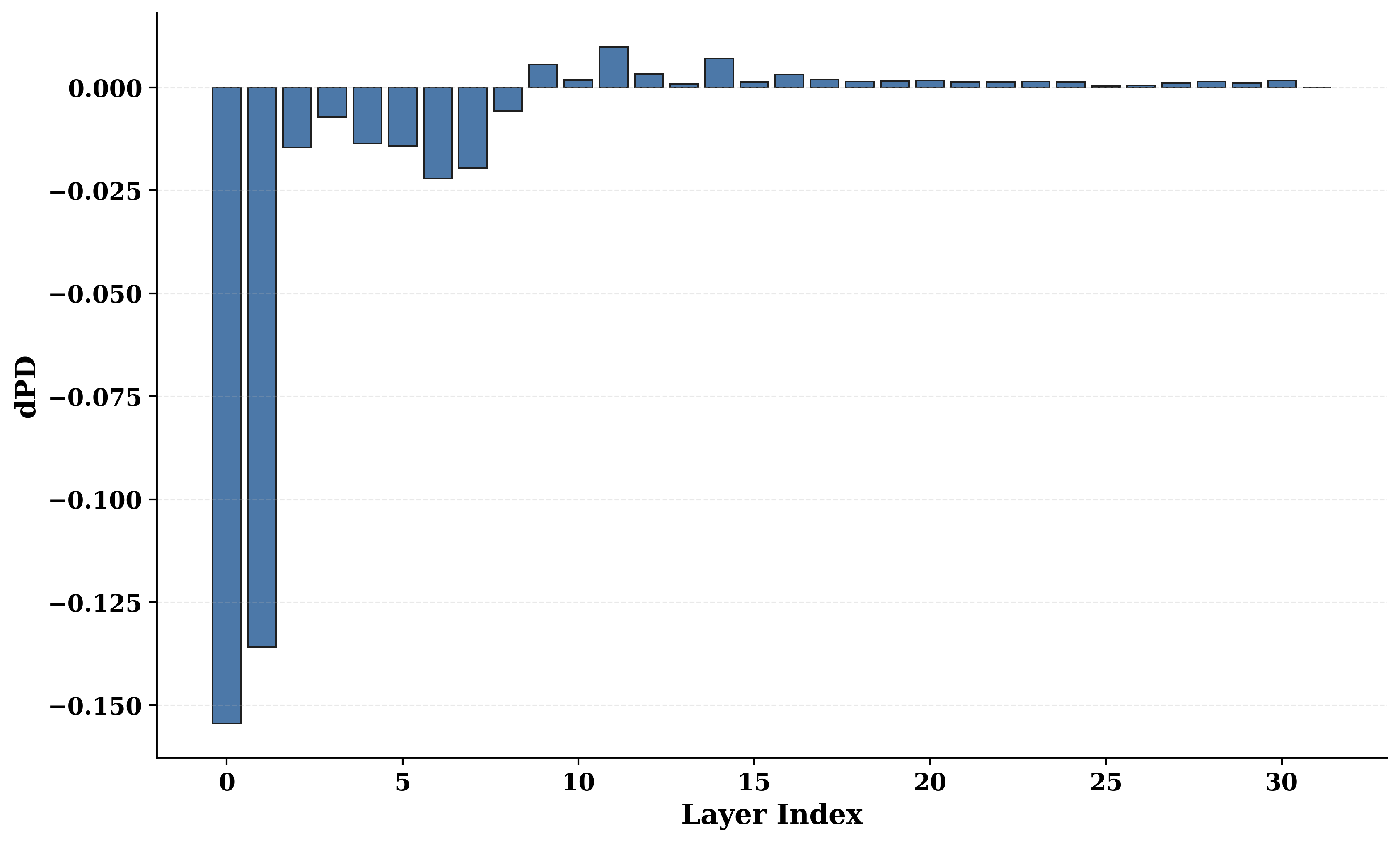}\hfill
    \includegraphics[width=0.32\textwidth]{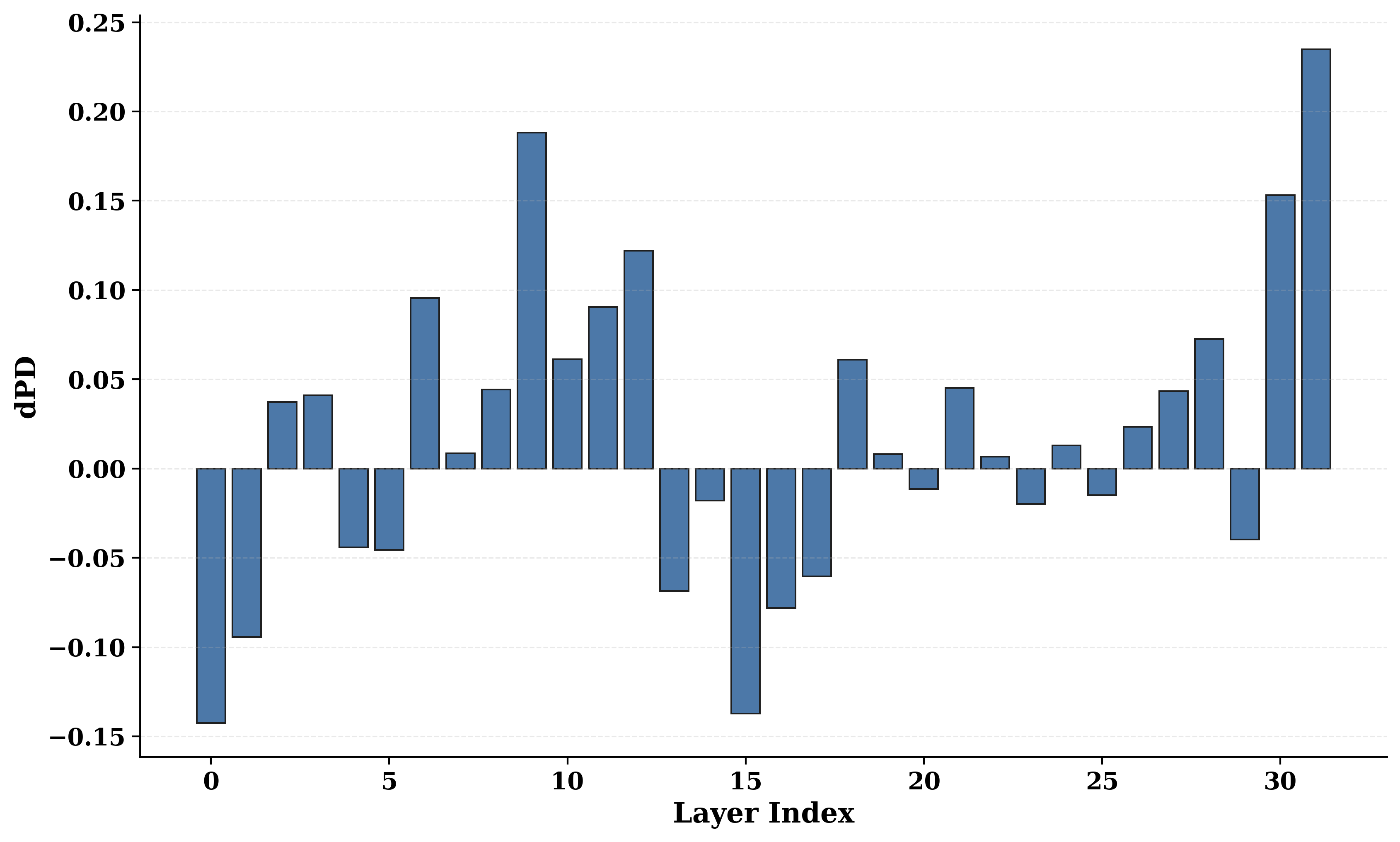}

    \includegraphics[width=0.32\textwidth]{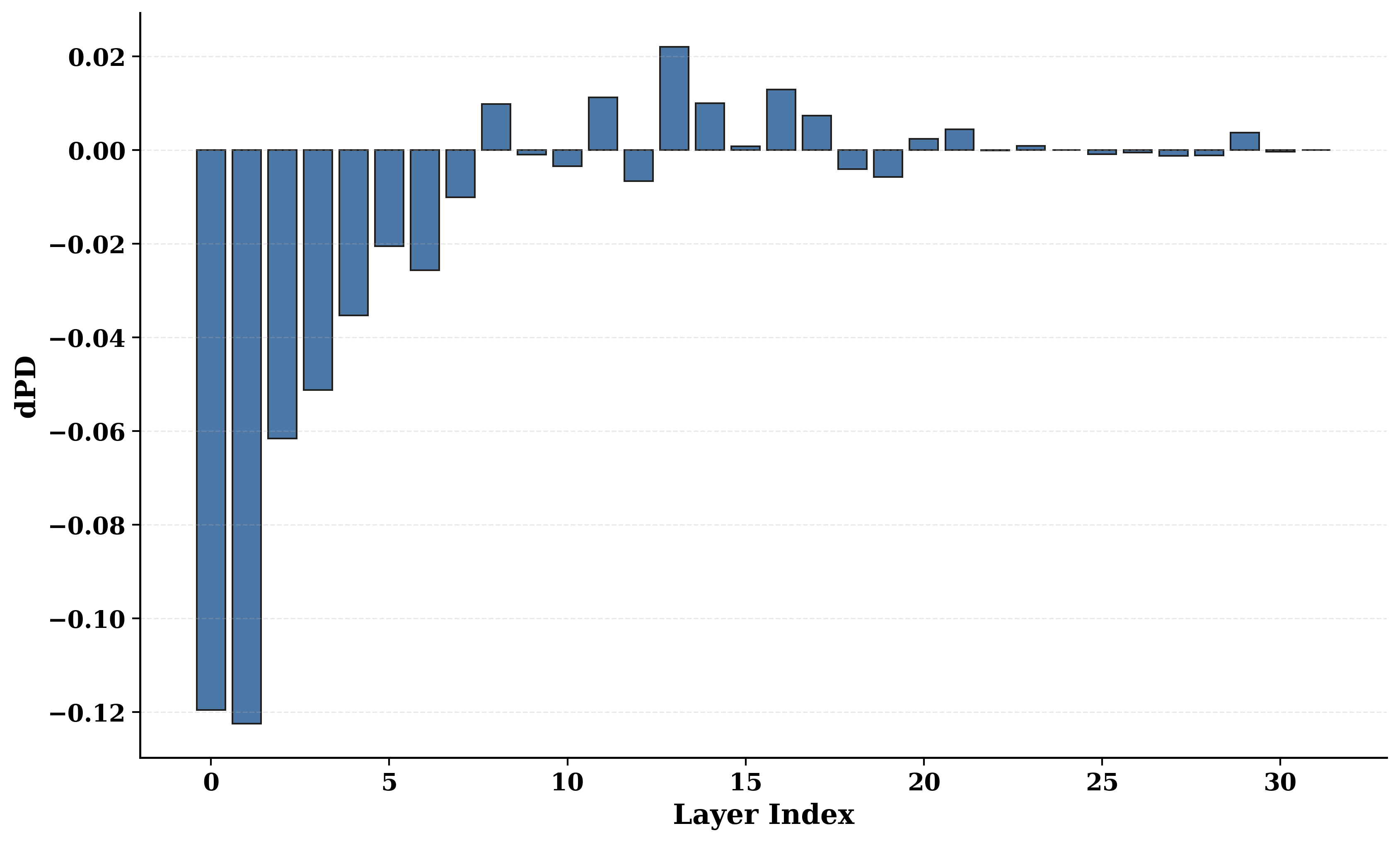}\hfill
    \includegraphics[width=0.32\textwidth]{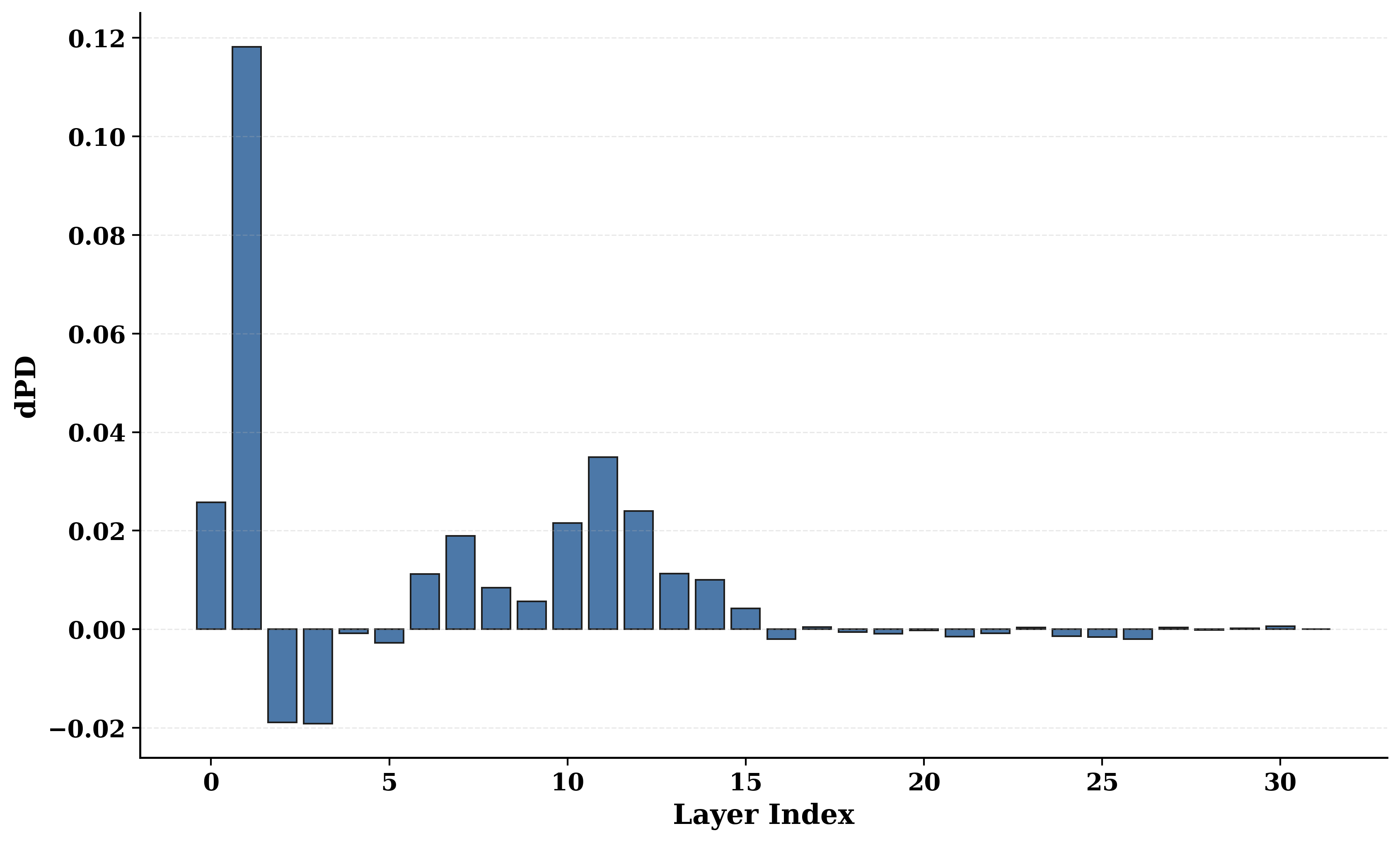}\hfill
    \includegraphics[width=0.32\textwidth]{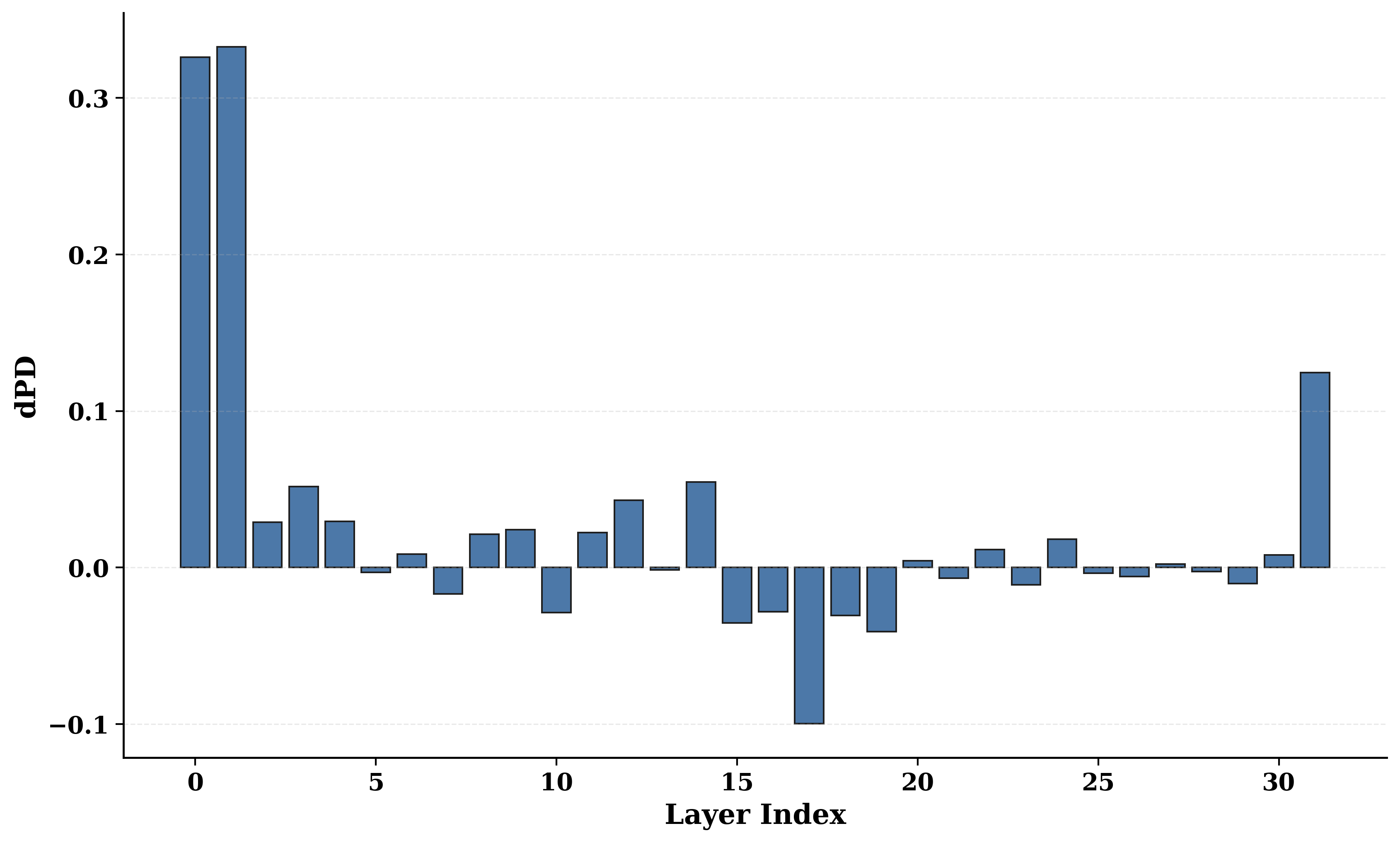}
    \caption{\textbf{Joint attention+MLP-block mean ablation on \textit{Mistral-7B-Instruct} under $\pi_{\mathrm{EF}}$.} Per-layer $\mathrm{dPD}$ when each logical block is mean-ablated. Top row: one-hop. Bottom row: two-hop. Columns: Facts / Expression / Query.}
    \label{fig:stage_joint_mistral_ef}
\end{figure*}

Figures~\ref{fig:stage_joint_q8_ef}--\ref{fig:stage_joint_mistral_ef} show the joint attention+MLP-block mean-ablation $\mathrm{dPD}$ in the same format. The body Figure~\ref{fig:qwen3_8b_attn_mlp_facts_gallery} is the \textit{Qwen3-8B} $\pi_{\mathrm{FF}}$ cell of this grid. The joint branch preserves the same coarse staged ordering as the MLP branch while sharpening early-Facts dominance.

\section{Supplementary Results: Information Routing}
\label{sec:appendix_info_converge}

This section supplements \S\ref{sec:info_and_retrieval} with the full cross-model per-token causal-density panels. All figures use the per-token aggregation $M_{c}^{(s)}$ defined in Eq.~\ref{eq:per_token_mean}; layers are partitioned into Early/Middle/Late per family as in \S\ref{sec:appendix_layer_partition}. Two robust points emerge.

\begin{figure*}[!htbp]
    \centering
    \includegraphics[width=0.49\textwidth]{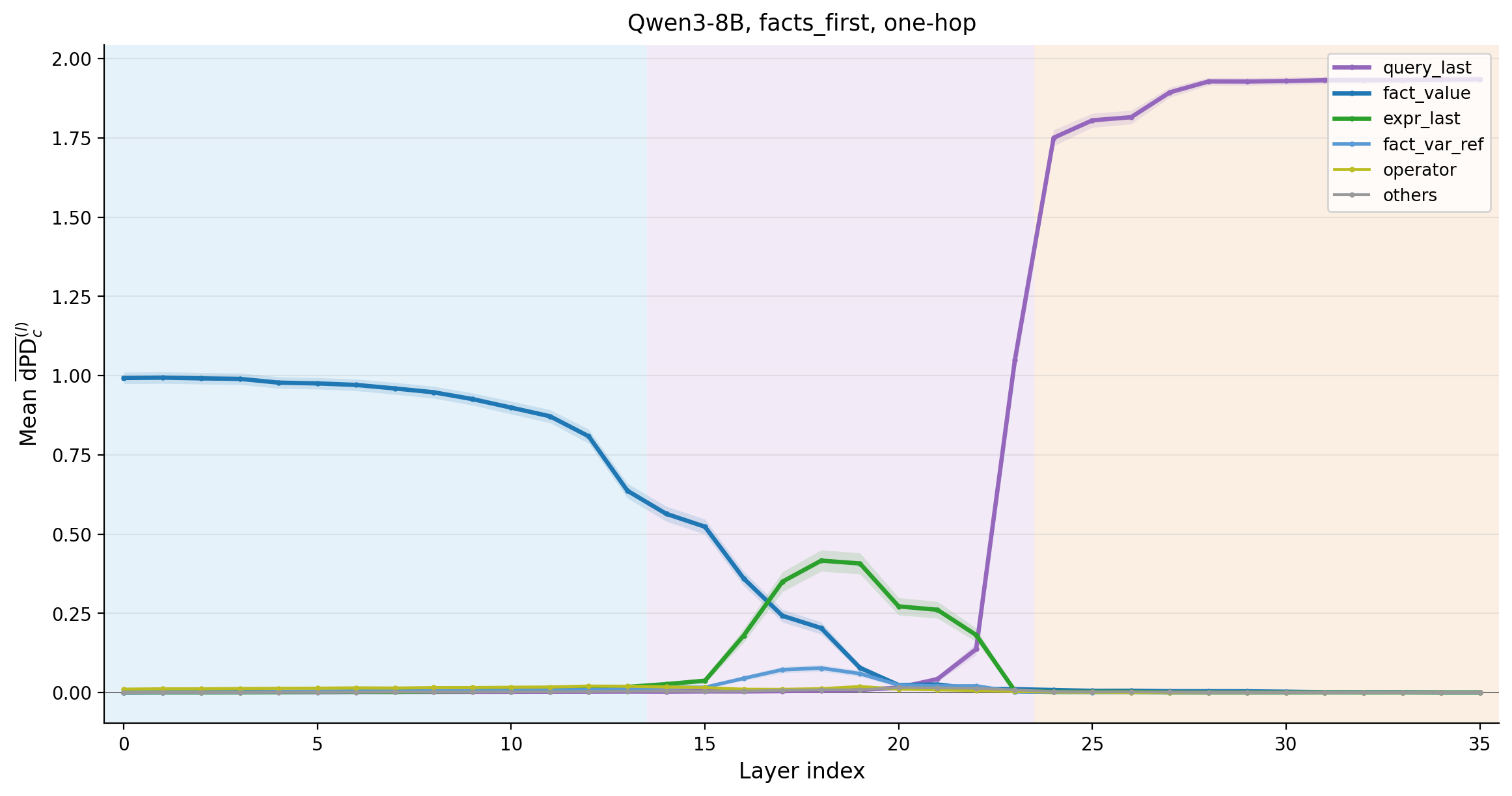}\hfill
    \includegraphics[width=0.49\textwidth]{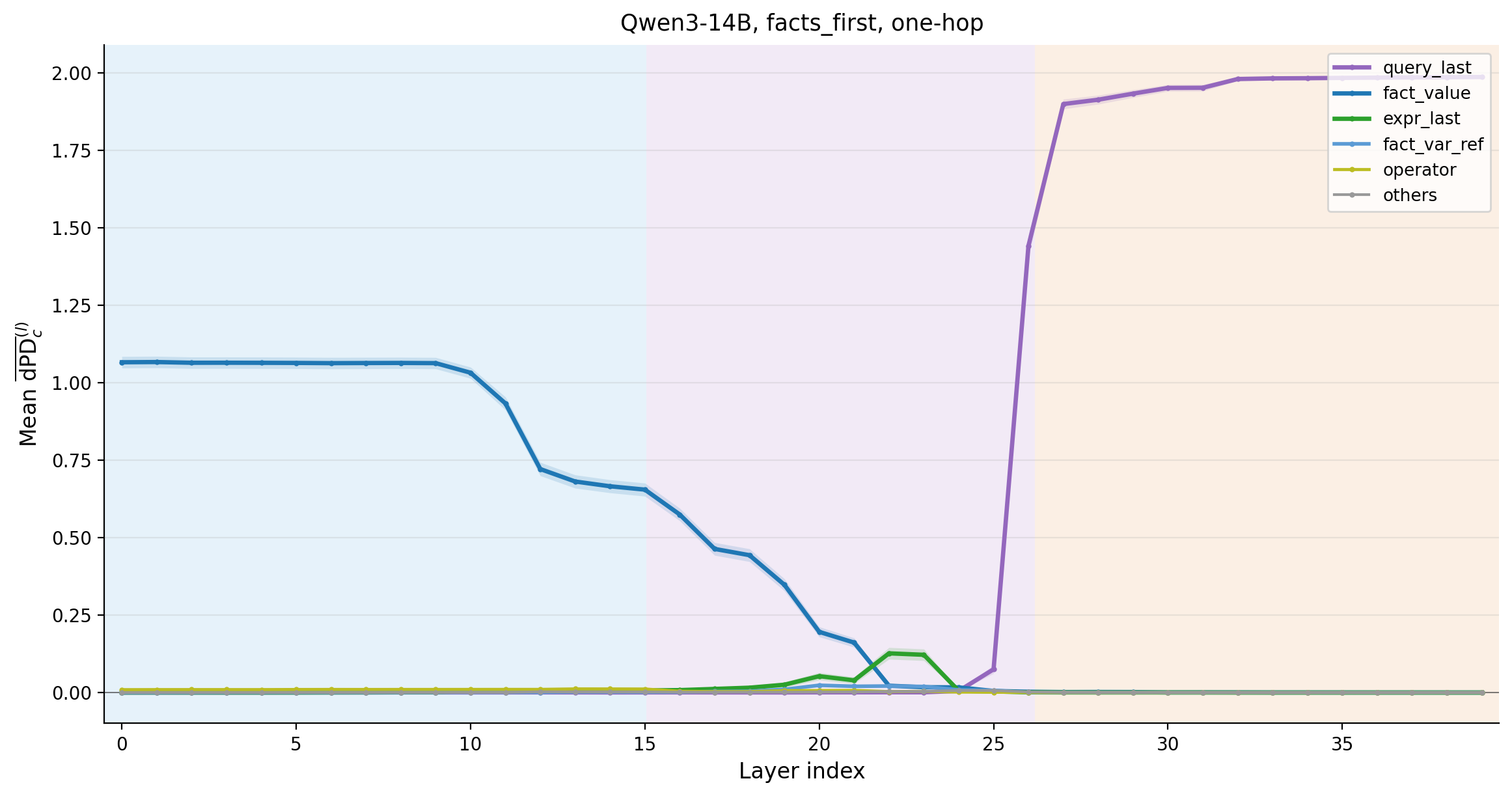}

    \includegraphics[width=0.49\textwidth]{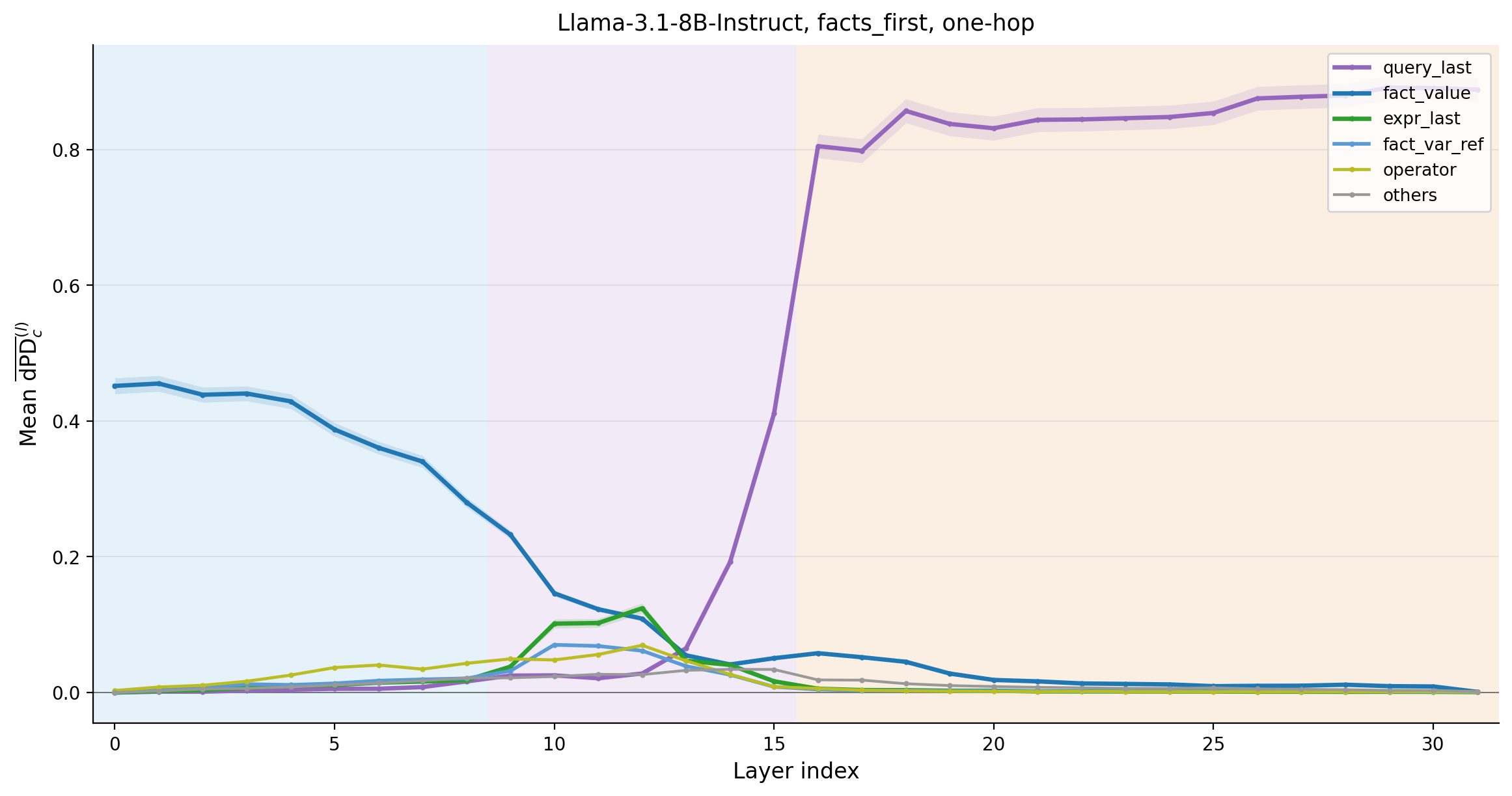}\hfill
    \includegraphics[width=0.49\textwidth]{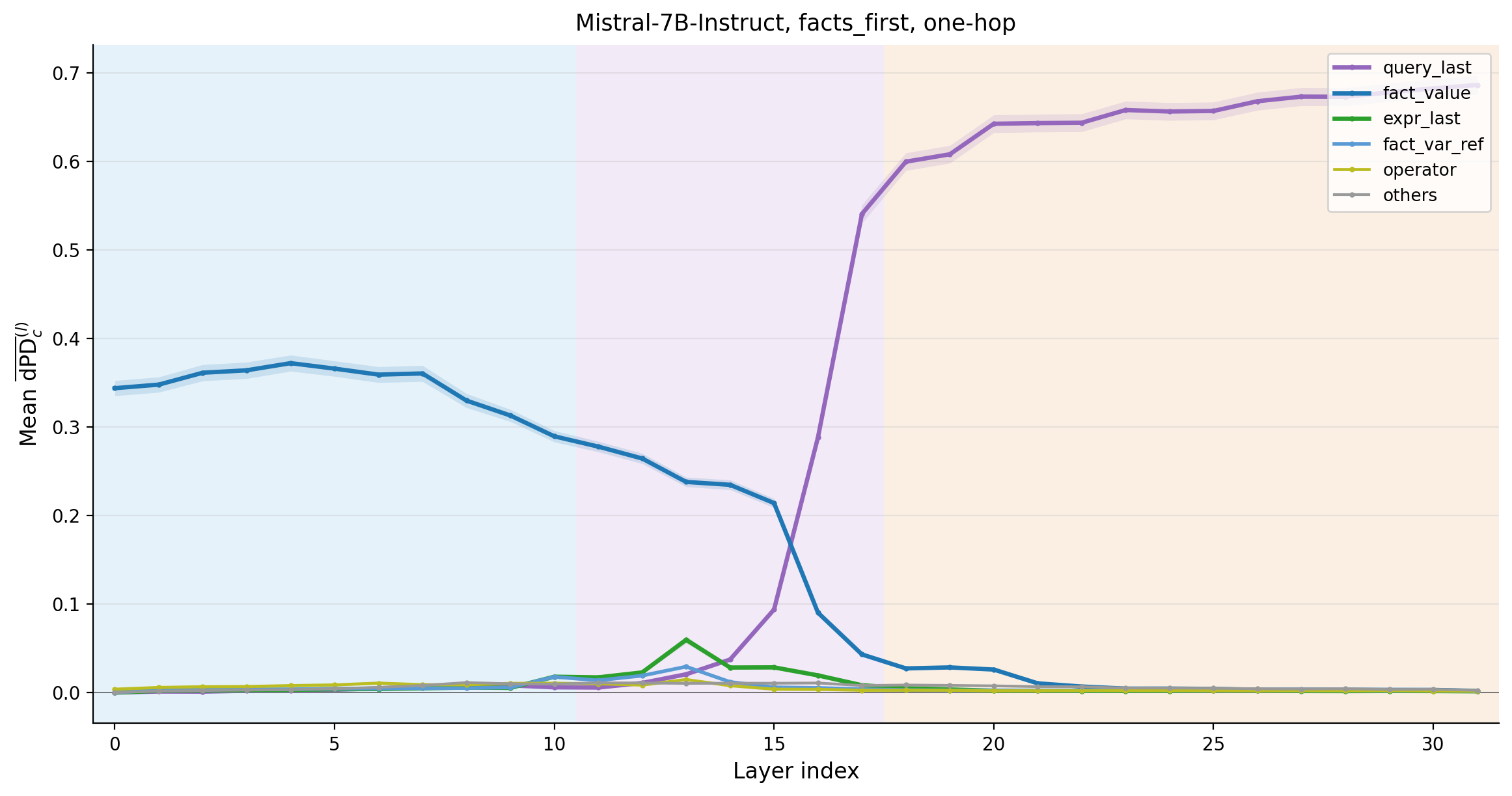}
    \caption{\textbf{Per-token mean $\mathrm{dPD}$, $\pi_{\mathrm{FF}}$, one-hop, full view.} Top-left: \textit{Qwen3-8B}; top-right: \textit{Qwen3-14B}; bottom-left: \textit{Llama-3.1-8B-Instruct}; bottom-right: \textit{Mistral-7B-Instruct}.}
    \label{fig:token_facts_first_cross_model}
\end{figure*}

\begin{figure*}[!htbp]
    \centering
    \includegraphics[width=0.49\textwidth]{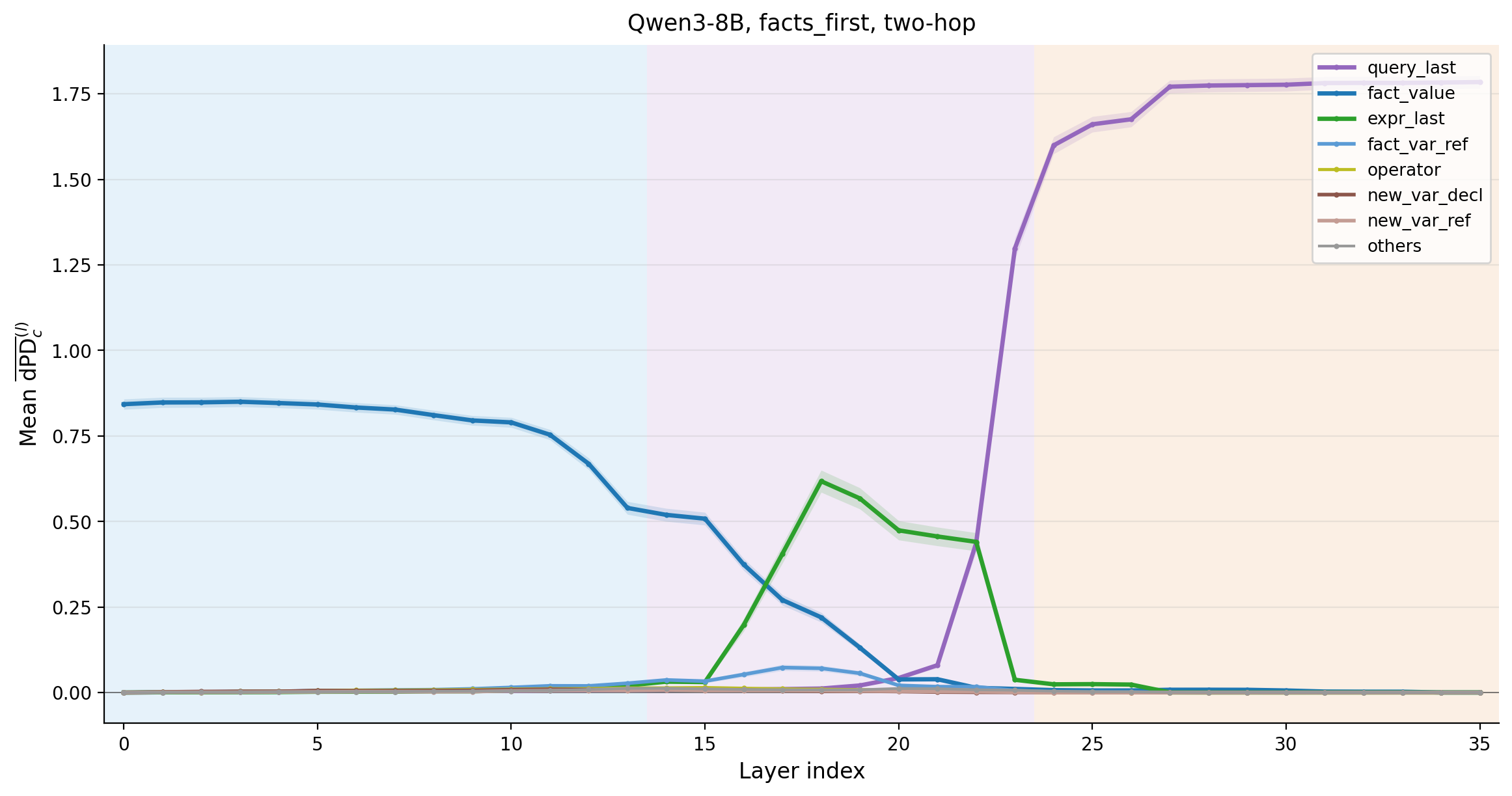}\hfill
    \includegraphics[width=0.49\textwidth]{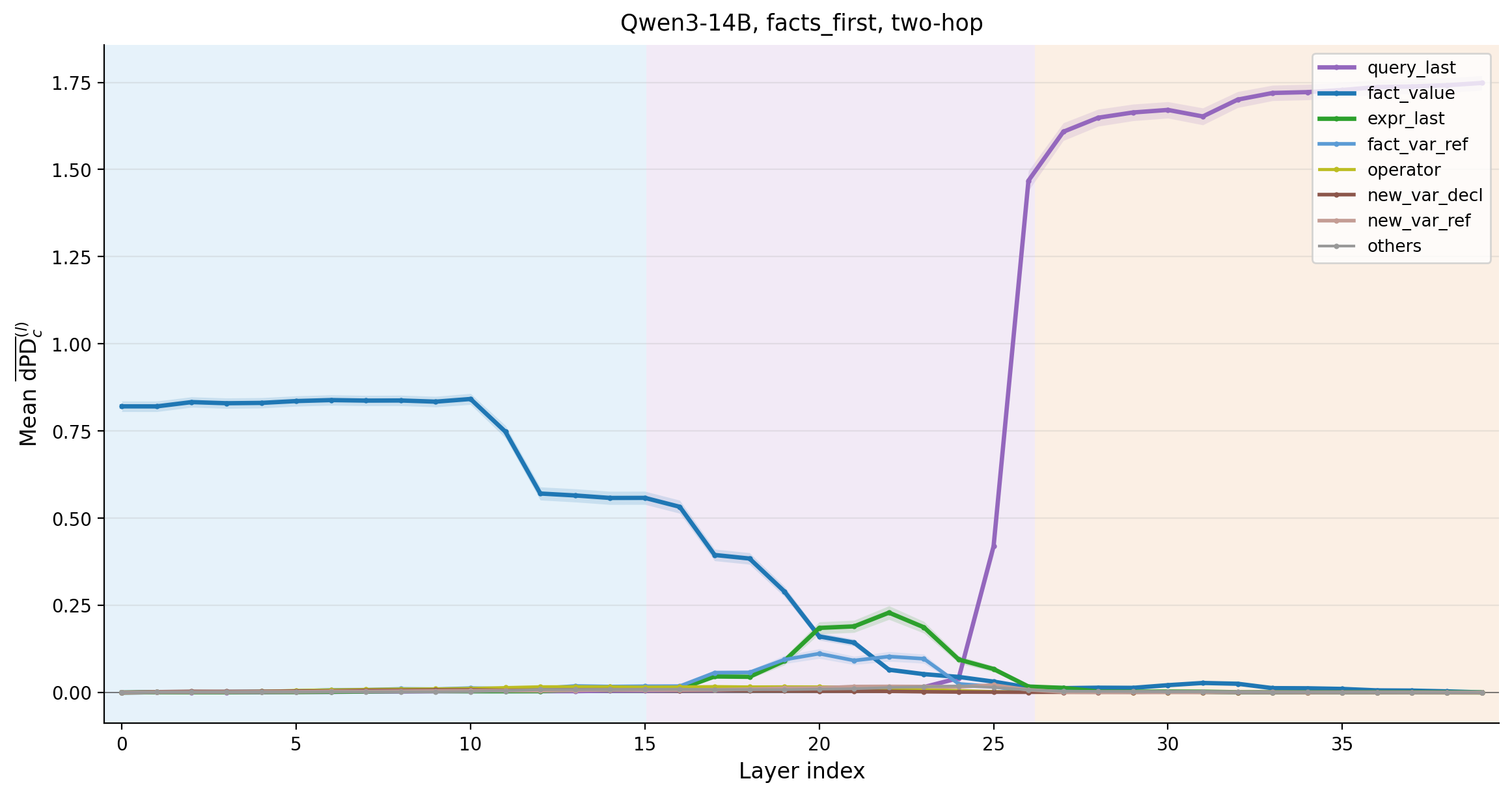}

    \includegraphics[width=0.49\textwidth]{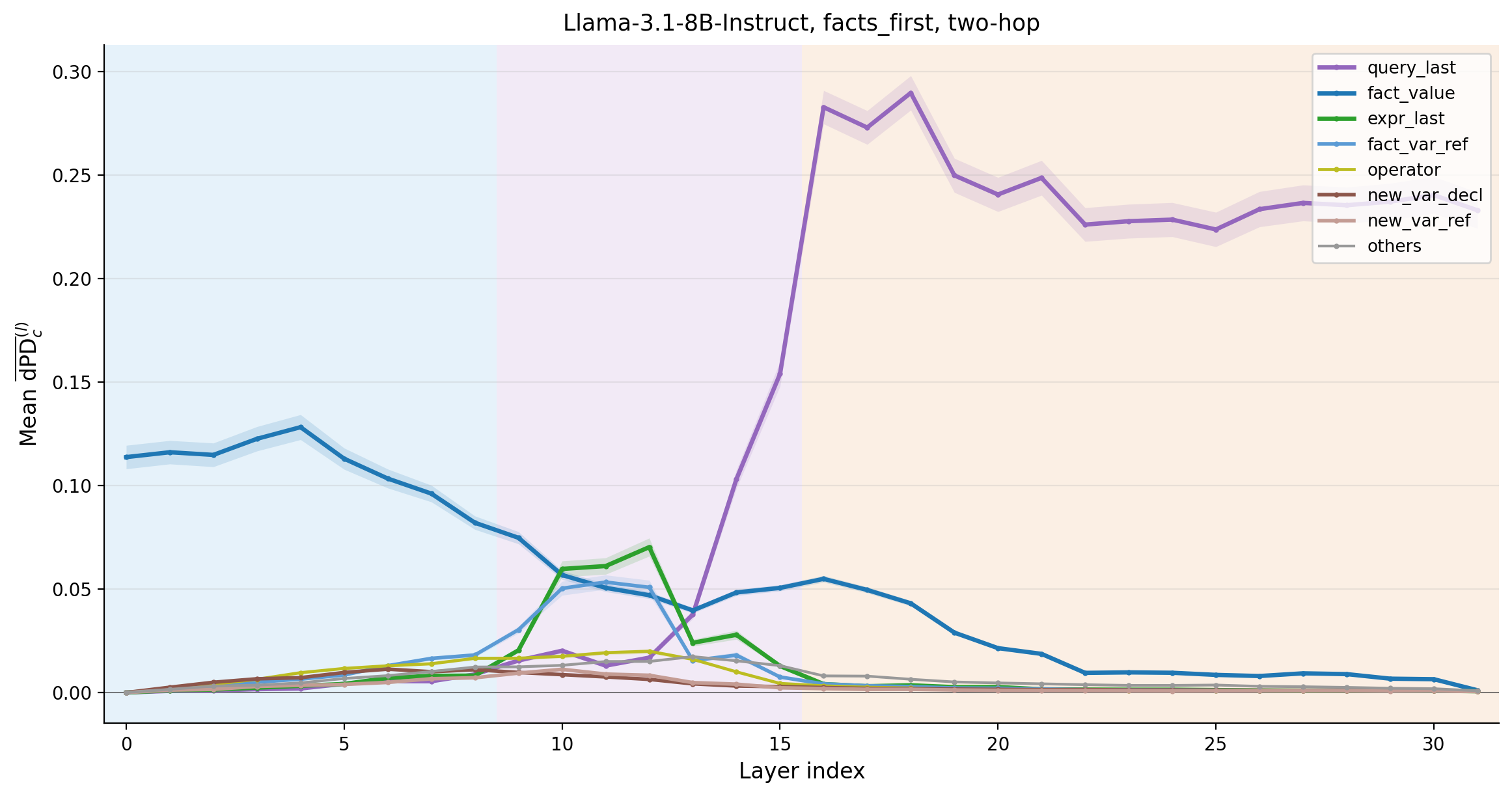}\hfill
    \includegraphics[width=0.49\textwidth]{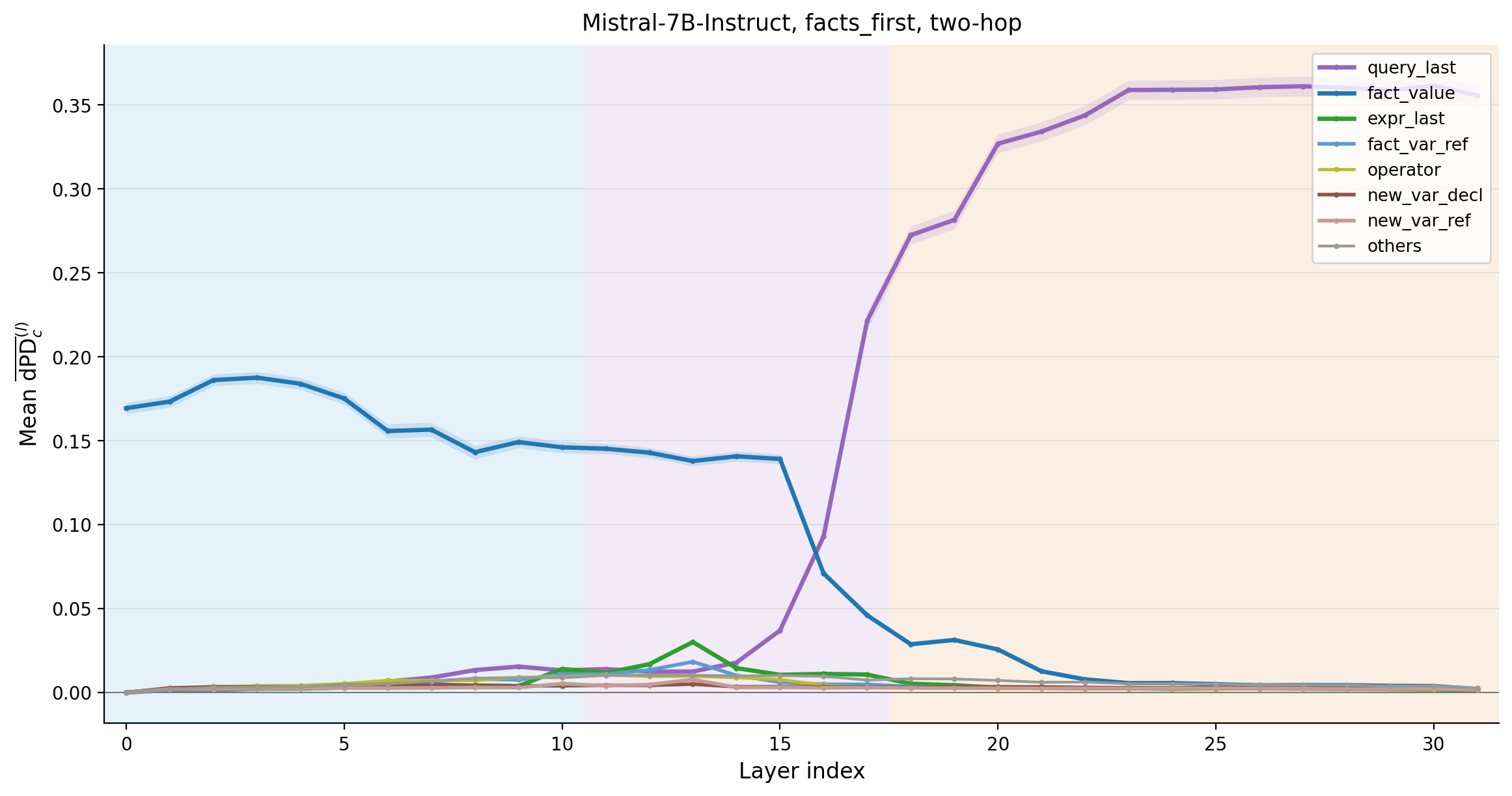}
    \caption{\textbf{Per-token mean $\mathrm{dPD}$, $\pi_{\mathrm{FF}}$, two-hop, full view.} Top-left: \textit{Qwen3-8B}; top-right: \textit{Qwen3-14B}; bottom-left: \textit{Llama-3.1-8B-Instruct}; bottom-right: \textit{Mistral-7B-Instruct}.}
    \label{fig:token_facts_first_cross_model_twohop}
\end{figure*}

\begin{figure*}[!htbp]
    \centering
    \includegraphics[width=0.49\textwidth]{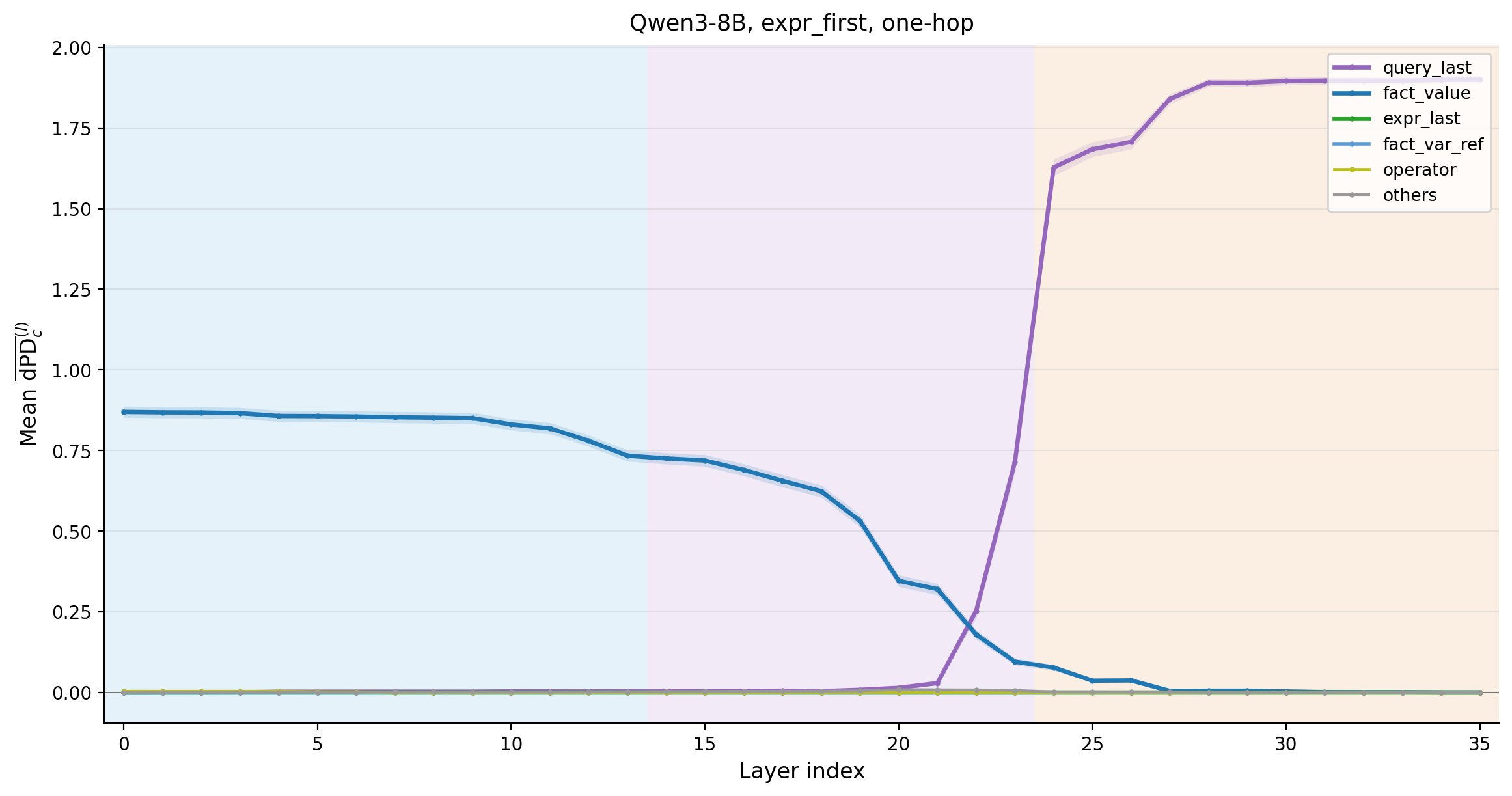}\hfill
    \includegraphics[width=0.49\textwidth]{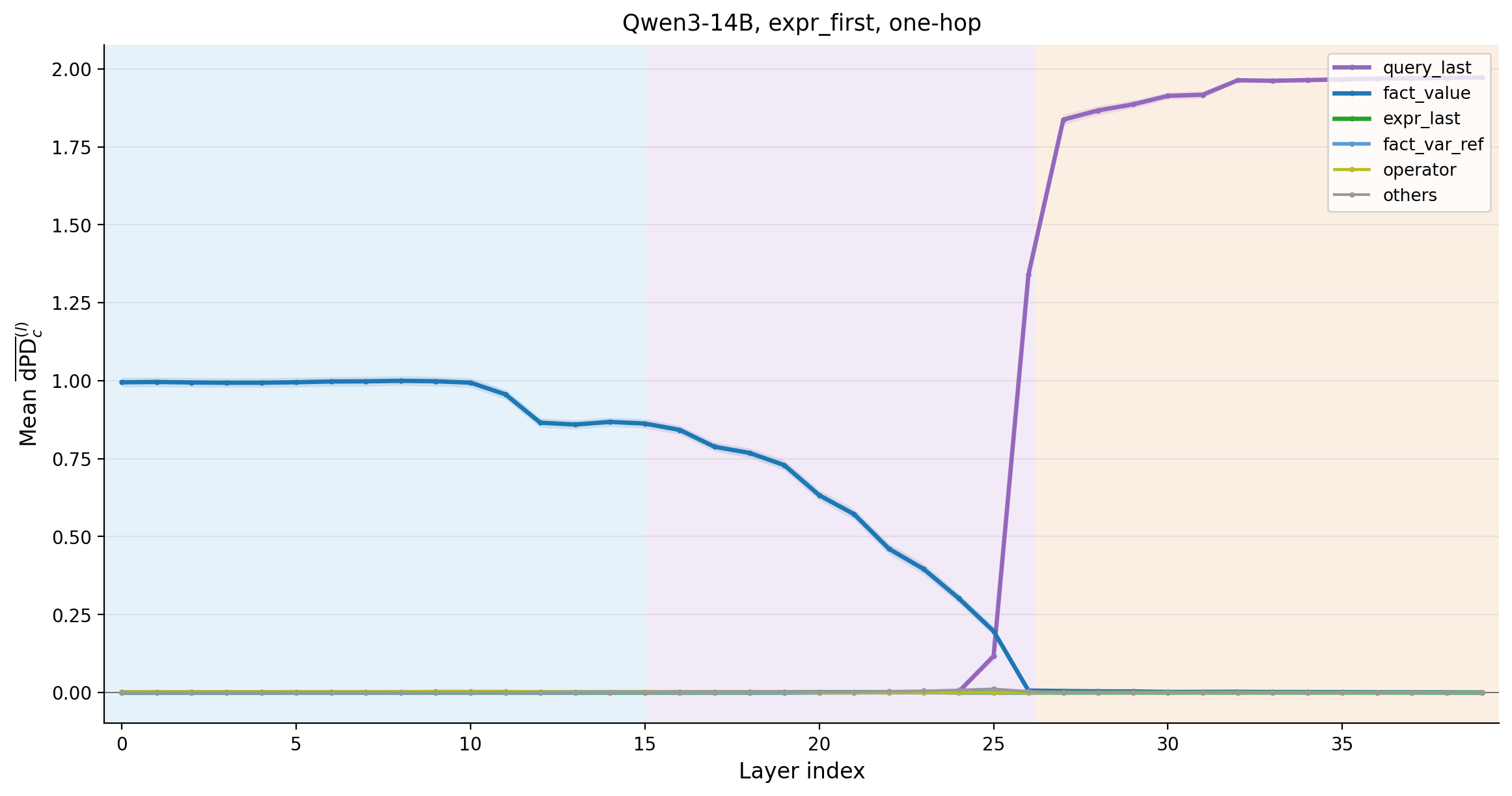}

    \includegraphics[width=0.49\textwidth]{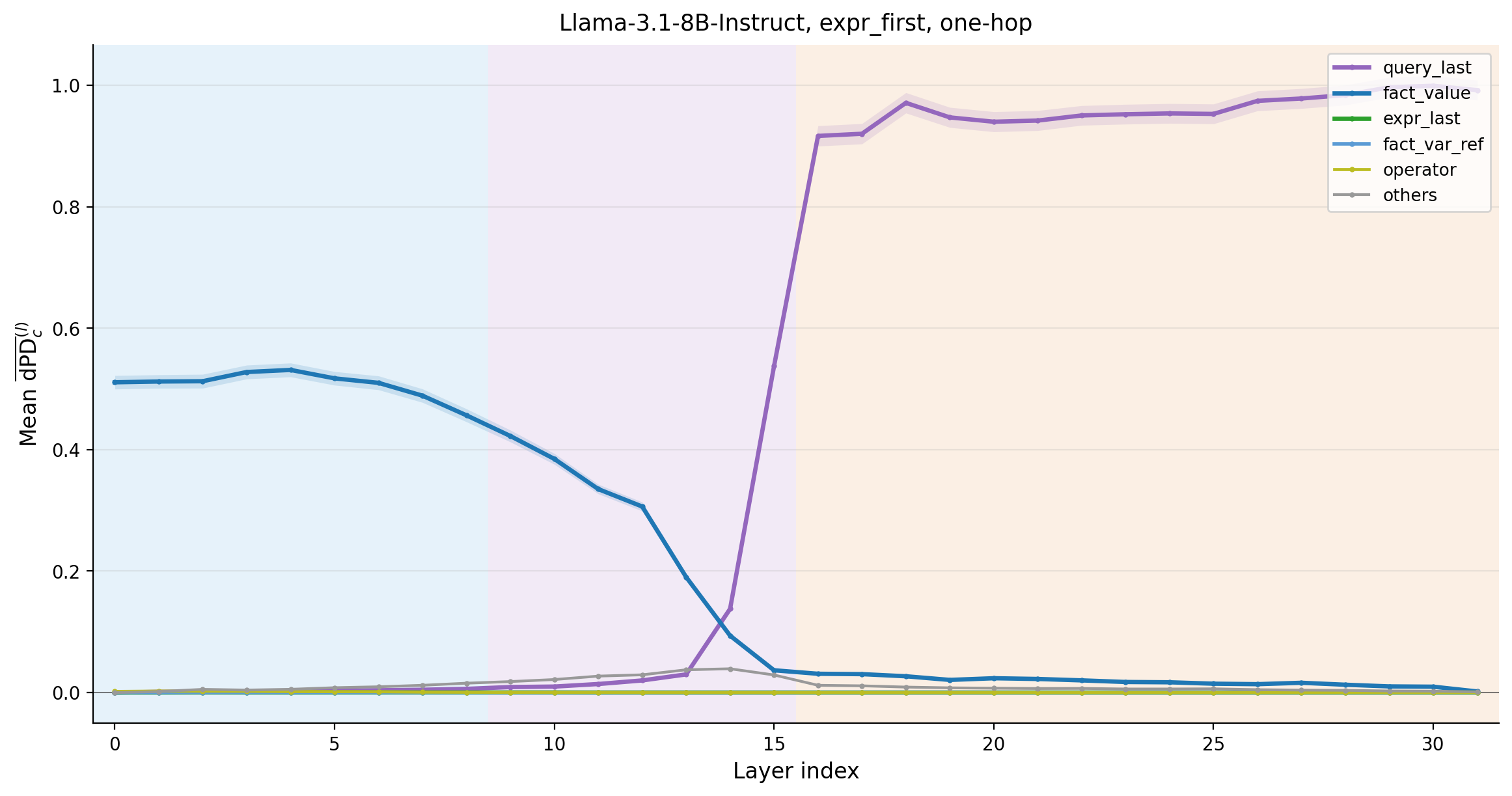}\hfill
    \includegraphics[width=0.49\textwidth]{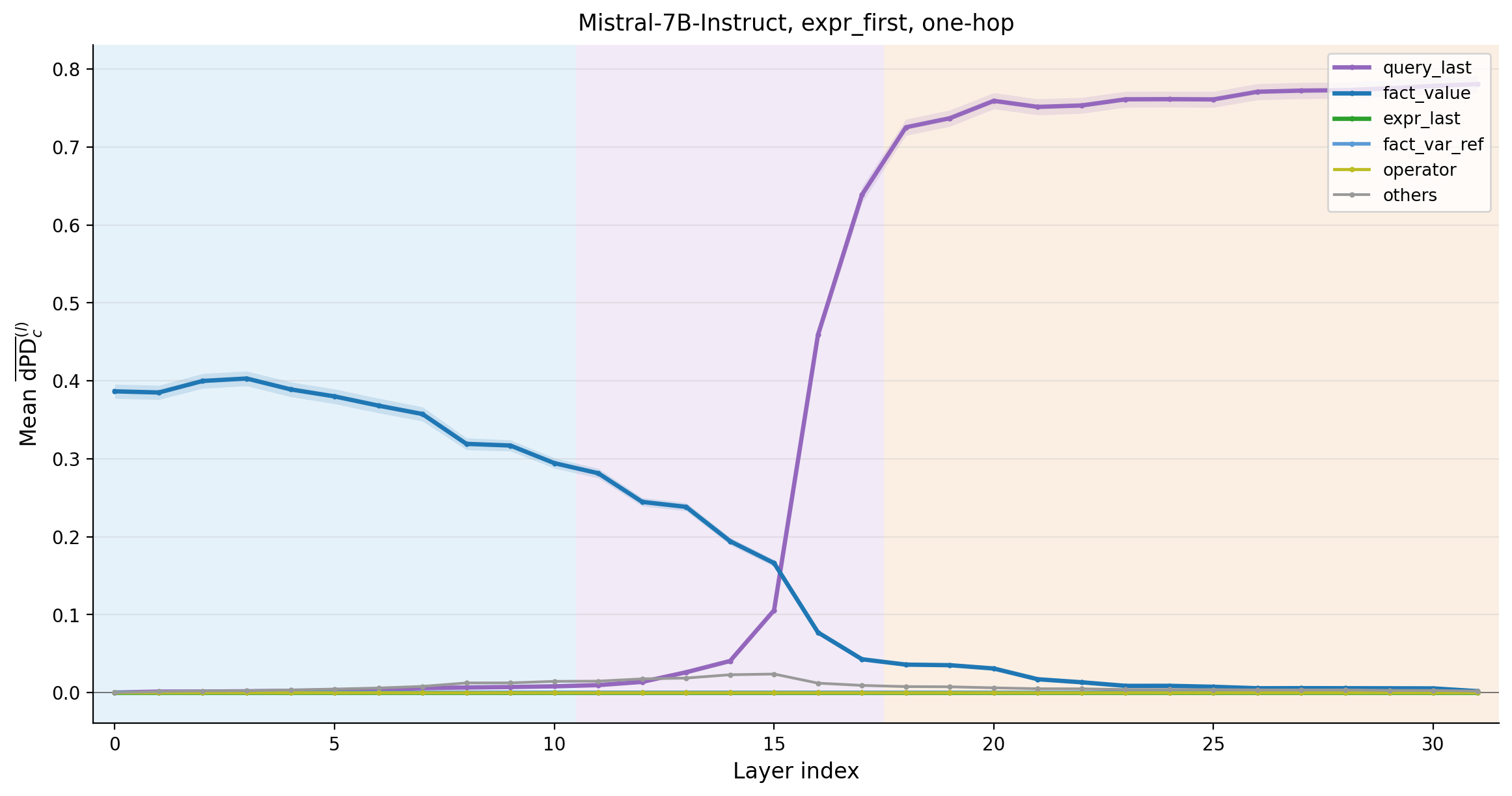}
    \caption{\textbf{Per-token mean $\mathrm{dPD}$, $\pi_{\mathrm{EF}}$, one-hop, full view.} Top-left: \textit{Qwen3-8B}; top-right: \textit{Qwen3-14B}; bottom-left: \textit{Llama-3.1-8B-Instruct}; bottom-right: \textit{Mistral-7B-Instruct}.}
    \label{fig:token_expr_first_cross_model}
\end{figure*}

\begin{figure*}[!htbp]
    \centering
    \includegraphics[width=0.49\textwidth]{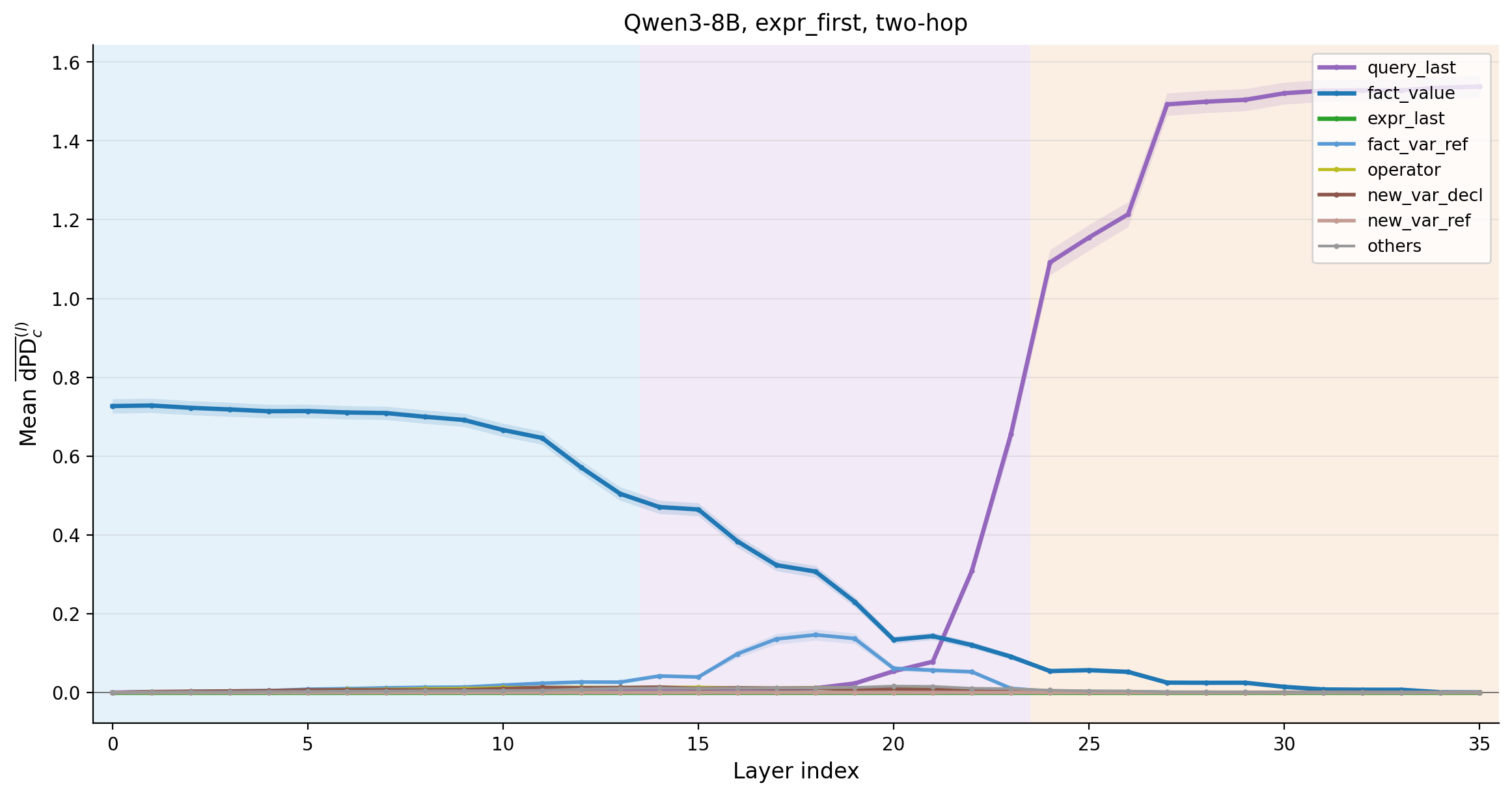}\hfill
    \includegraphics[width=0.49\textwidth]{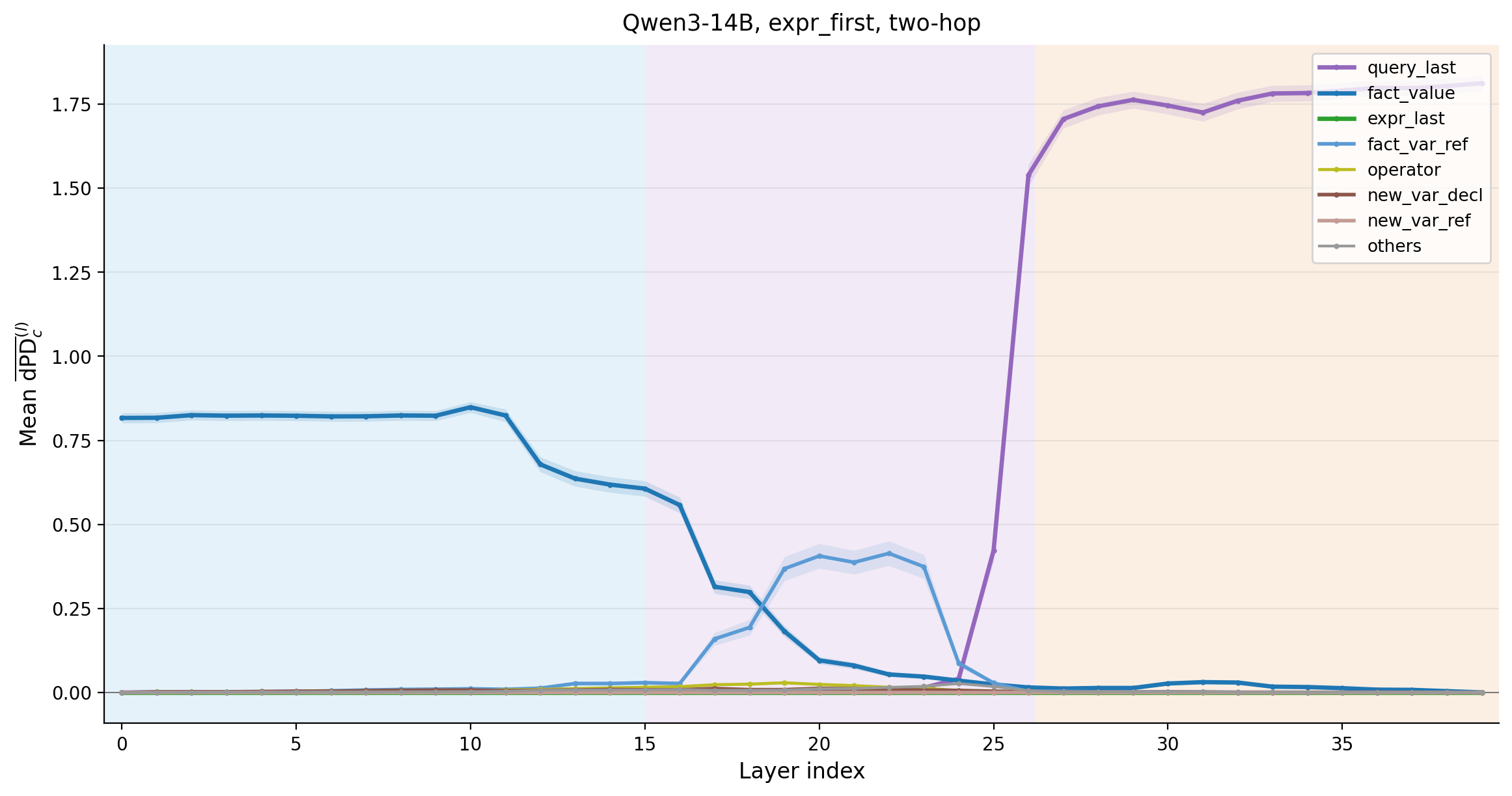}

    \includegraphics[width=0.49\textwidth]{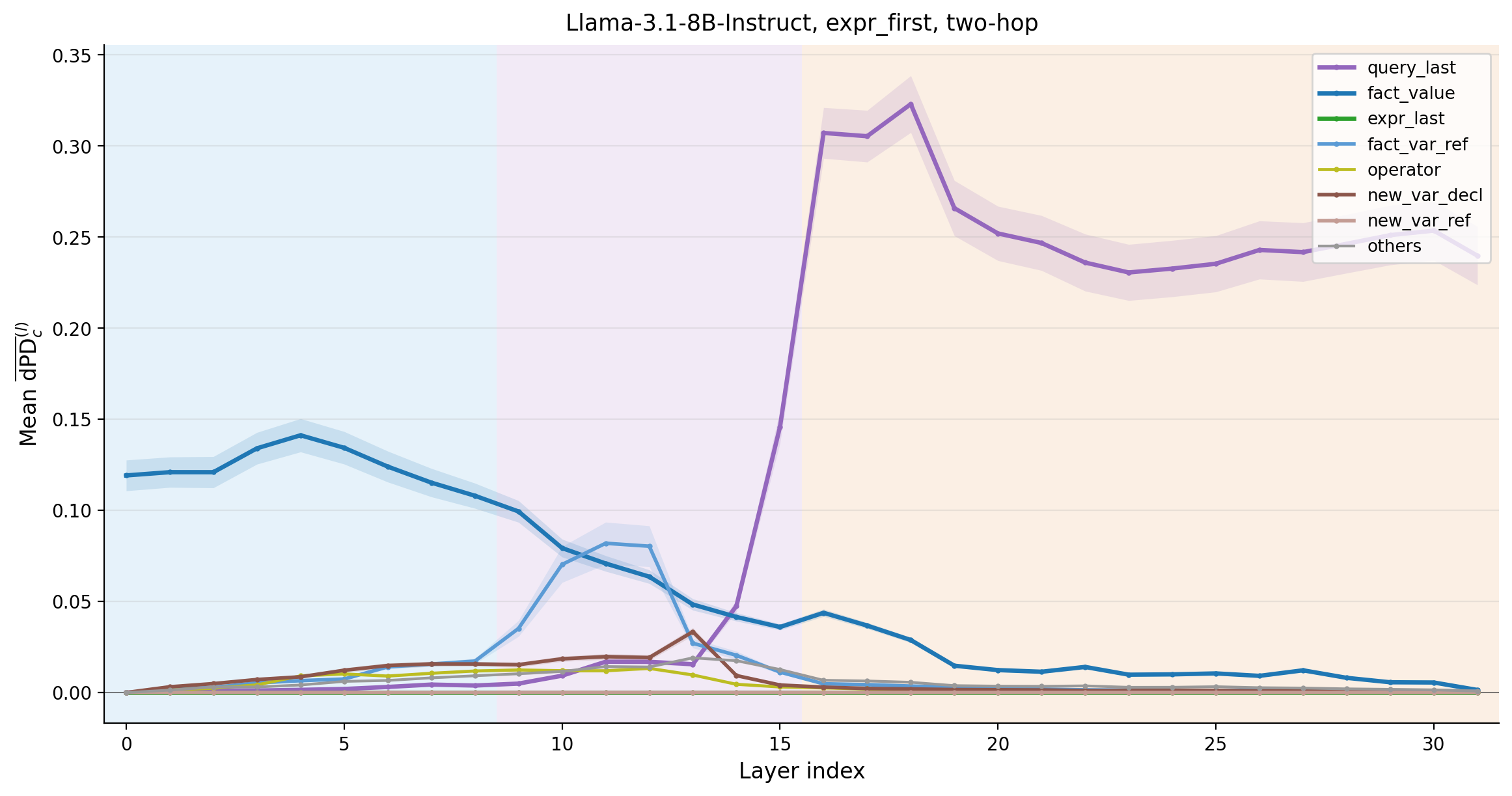}\hfill
    \includegraphics[width=0.49\textwidth]{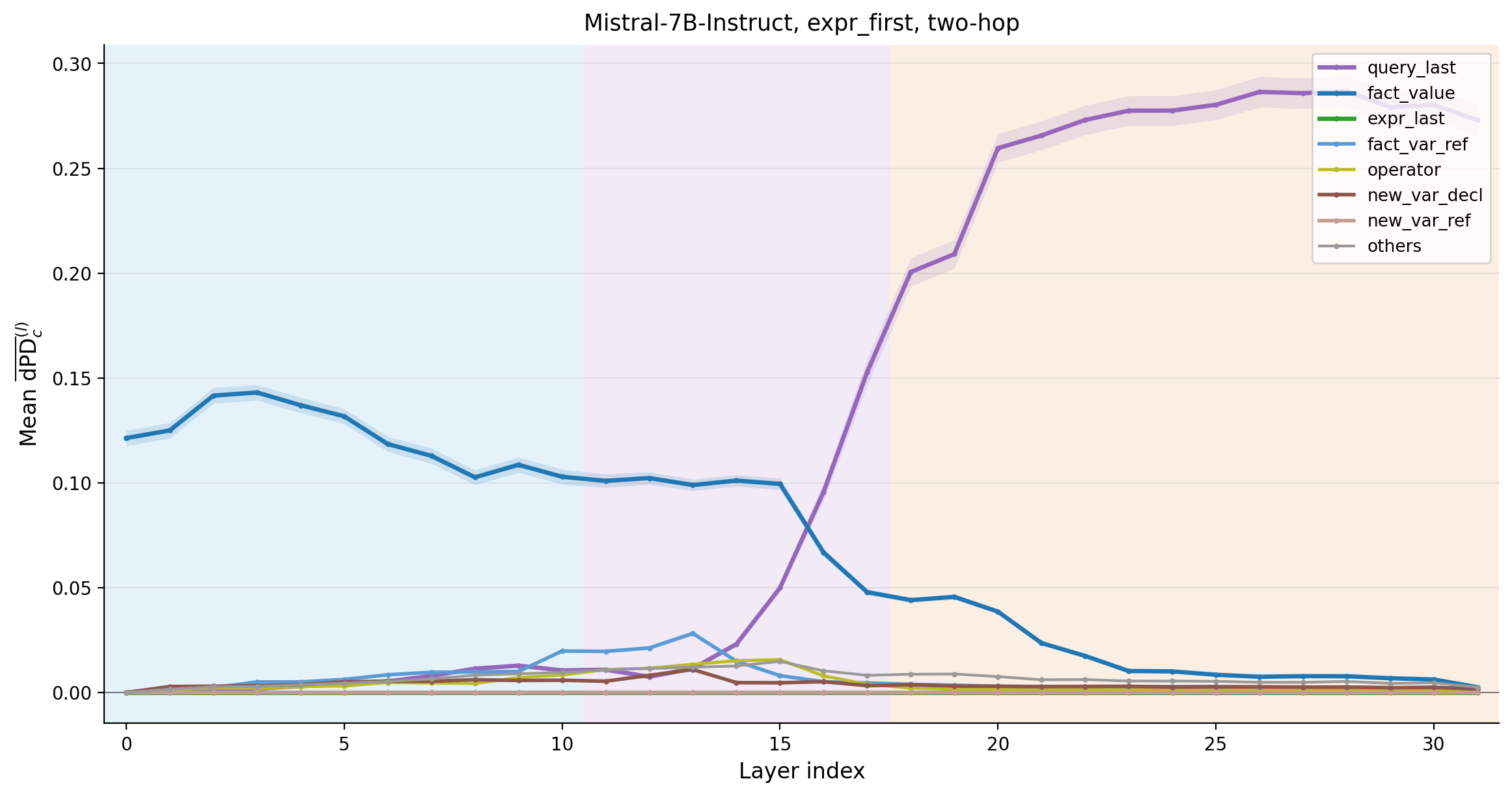}
    \caption{\textbf{Per-token mean $\mathrm{dPD}$, $\pi_{\mathrm{EF}}$, two-hop, full view.} Top-left: \textit{Qwen3-8B}; top-right: \textit{Qwen3-14B}; bottom-left: \textit{Llama-3.1-8B-Instruct}; bottom-right: \textit{Mistral-7B-Instruct}.}
    \label{fig:token_expr_first_cross_model_twohop}
\end{figure*}

\paragraph{Boundary-token convergence generalizes.}
Figures~\ref{fig:token_facts_first_cross_model}--\ref{fig:token_facts_first_cross_model_twohop} (under $\pi_{\mathrm{FF}}$) and Figures~\ref{fig:token_expr_first_cross_model}--\ref{fig:token_expr_first_cross_model_twohop} (under $\pi_{\mathrm{EF}}$) show the same boundary-token transport in all four families: \texttt{fact\_value} dominates the early layers, \texttt{expr\_last} (where present) dominates the middle layers, and \texttt{query\_last} dominates the late layers. Under $\pi_{\mathrm{EF}}$, \texttt{expr\_last} is pinned to zero by autoregressive construction, so the middle-layer carrier shifts to \texttt{fact\_value}; the early-fact and late-query dominance survive across families.

\begin{figure*}[!htbp]
    \centering
    \includegraphics[width=0.49\textwidth]{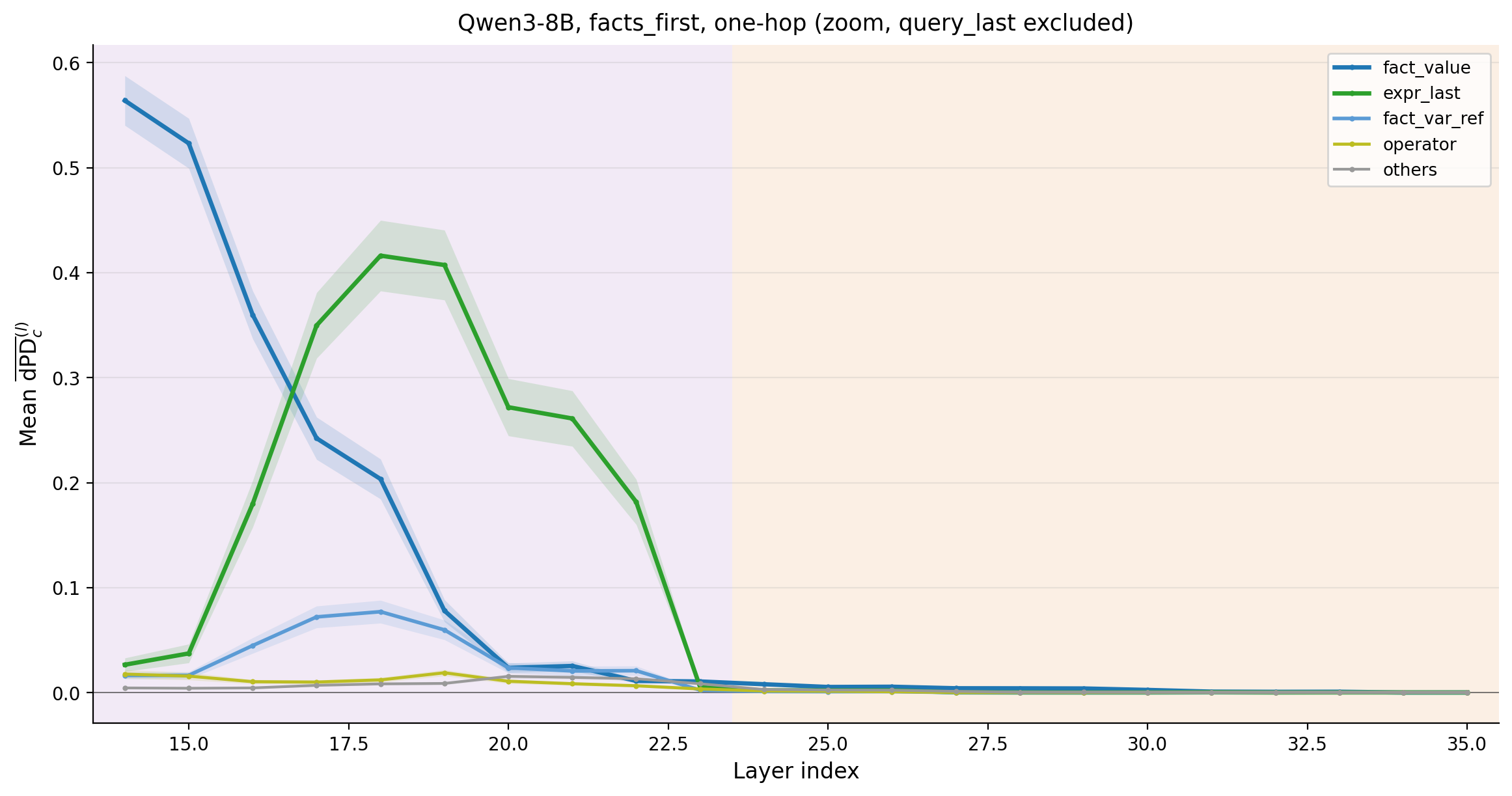}\hfill
    \includegraphics[width=0.49\textwidth]{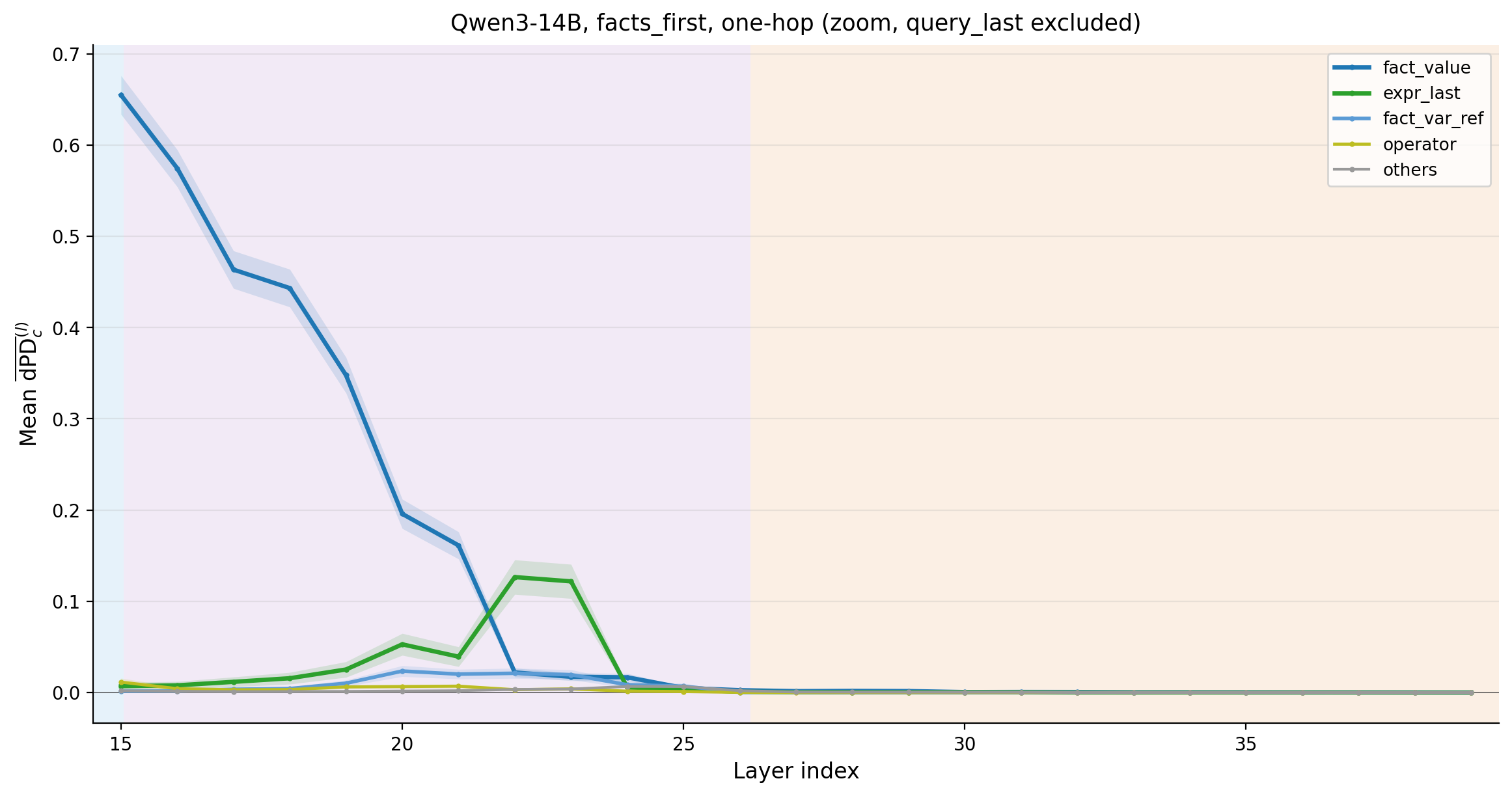}

    \includegraphics[width=0.49\textwidth]{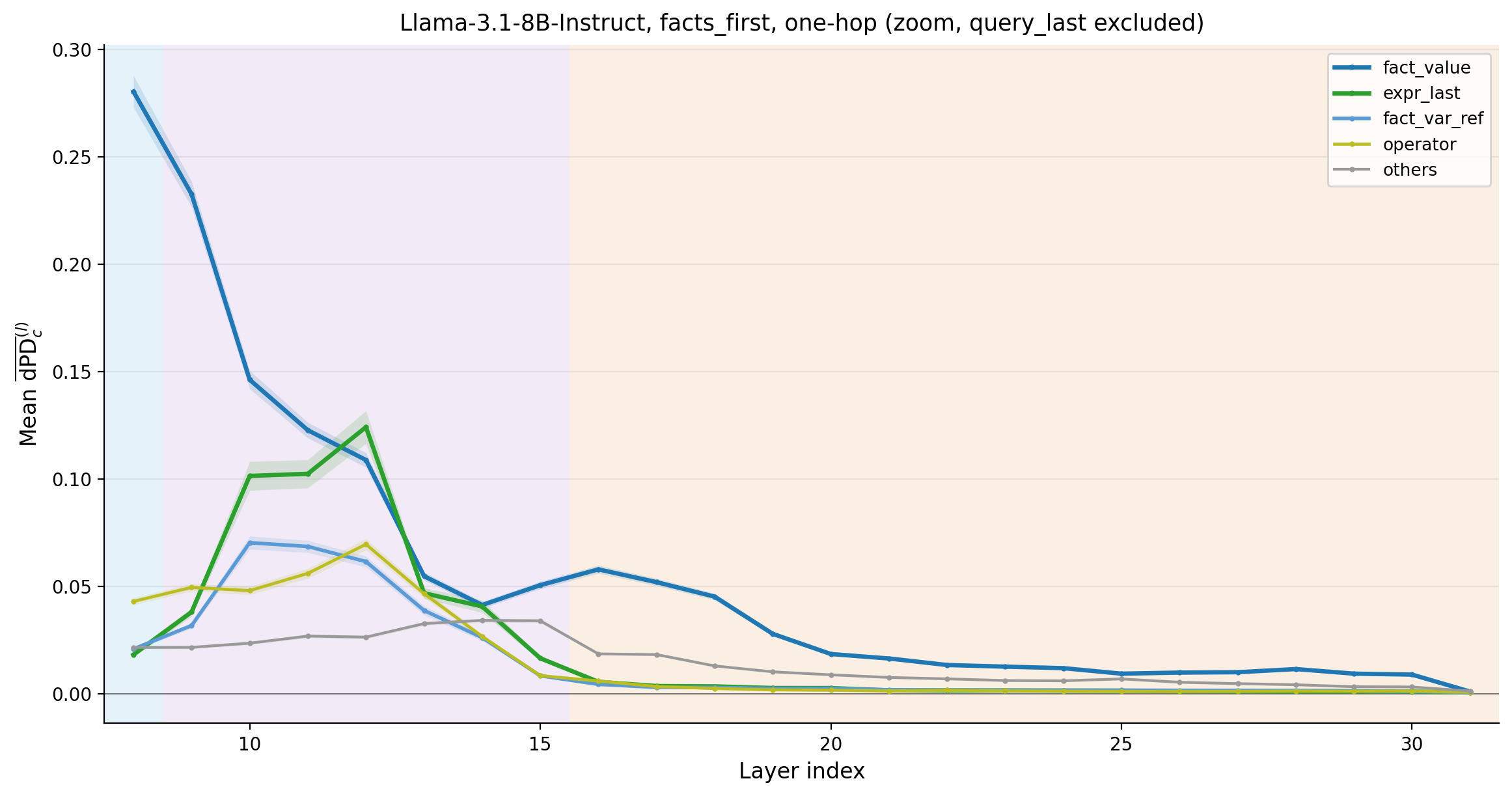}\hfill
    \includegraphics[width=0.49\textwidth]{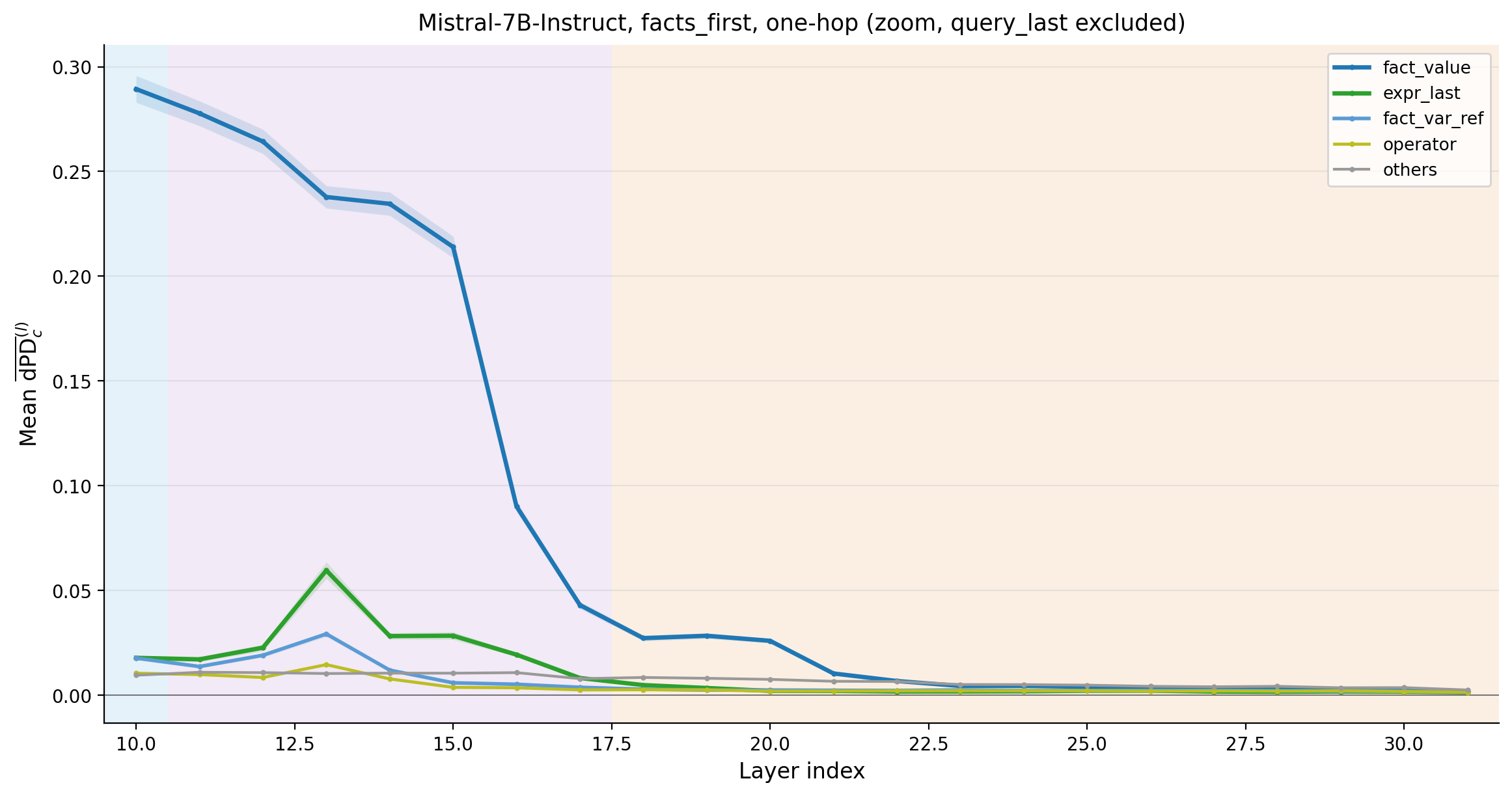}
    \caption{\textbf{Per-token mean $\mathrm{dPD}$, $\pi_{\mathrm{FF}}$, one-hop, \texttt{query\_last} excluded.} Top-left: \textit{Qwen3-8B}; top-right: \textit{Qwen3-14B}; bottom-left: \textit{Llama-3.1-8B-Instruct}; bottom-right: \textit{Mistral-7B-Instruct}.}
    \label{fig:token_facts_first_cross_model_zoom}
\end{figure*}

\begin{figure*}[!htbp]
    \centering
    \includegraphics[width=0.49\textwidth]{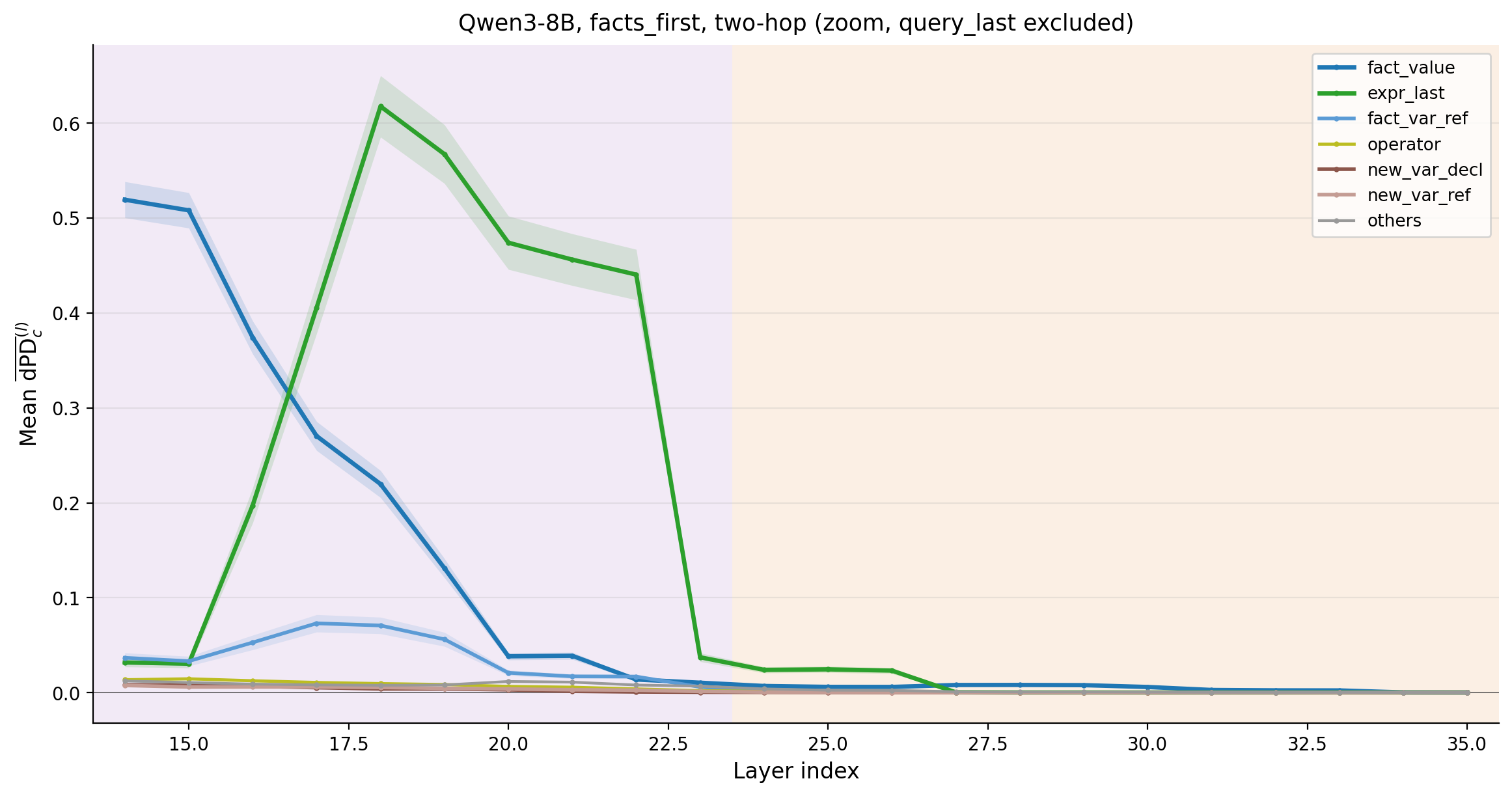}\hfill
    \includegraphics[width=0.49\textwidth]{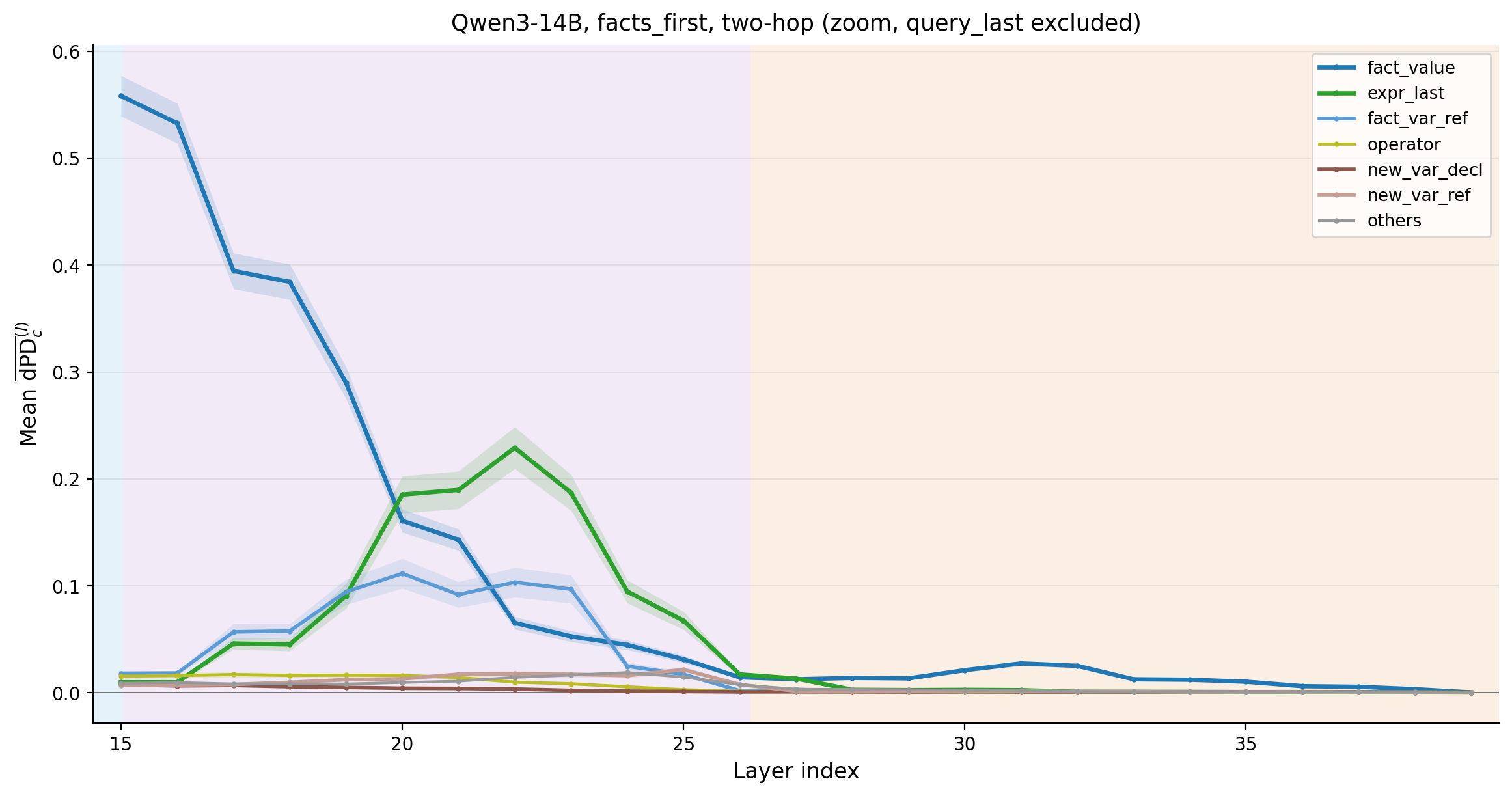}

    \includegraphics[width=0.49\textwidth]{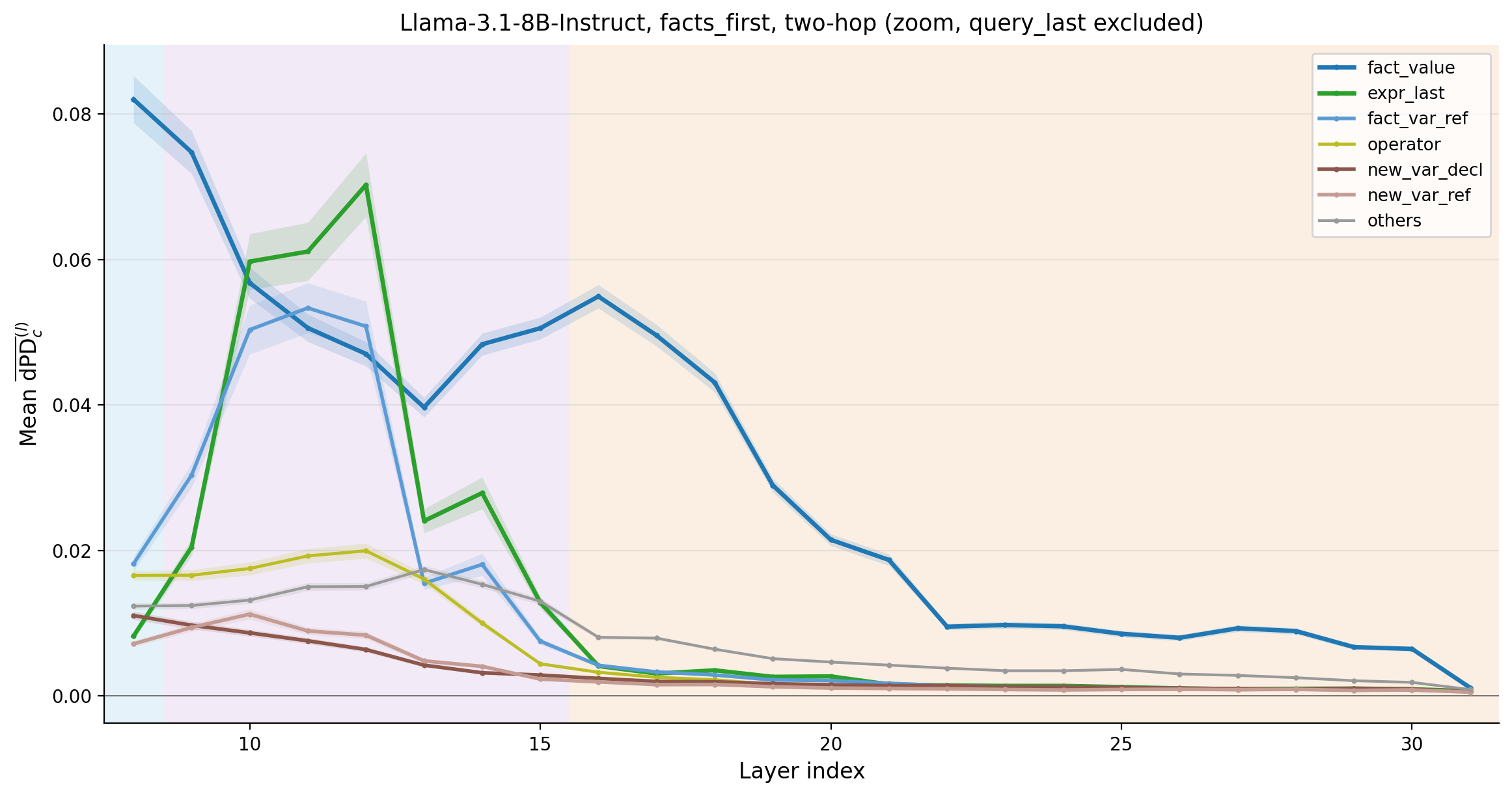}\hfill
    \includegraphics[width=0.49\textwidth]{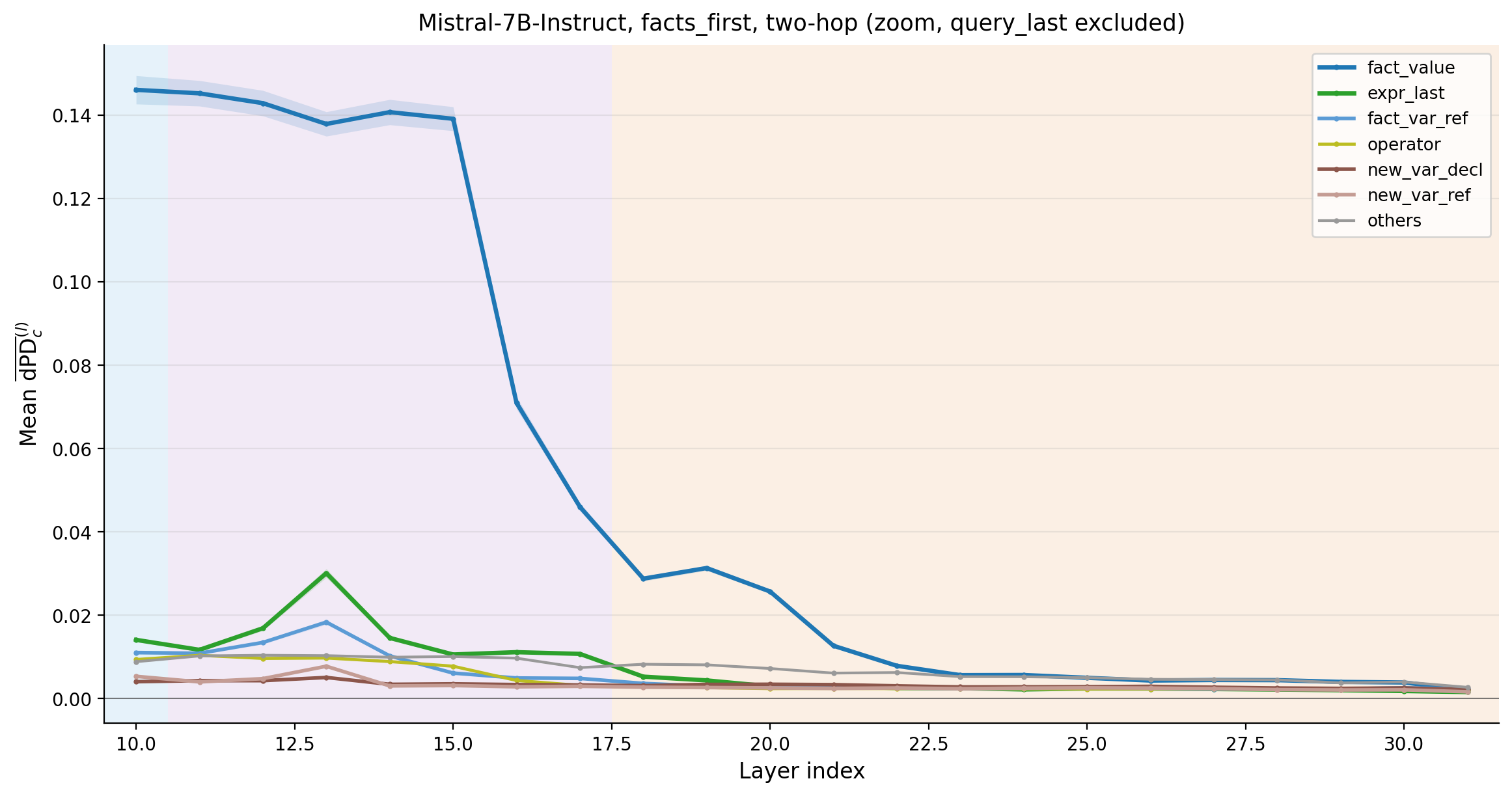}
    \caption{\textbf{Per-token mean $\mathrm{dPD}$, $\pi_{\mathrm{FF}}$, two-hop, \texttt{query\_last} excluded.} Top-left: \textit{Qwen3-8B}; top-right: \textit{Qwen3-14B}; bottom-left: \textit{Llama-3.1-8B-Instruct}; bottom-right: \textit{Mistral-7B-Instruct}.}
    \label{fig:token_facts_first_cross_model_twohop_zoom}
\end{figure*}

\begin{figure*}[!htbp]
    \centering
    \includegraphics[width=0.49\textwidth]{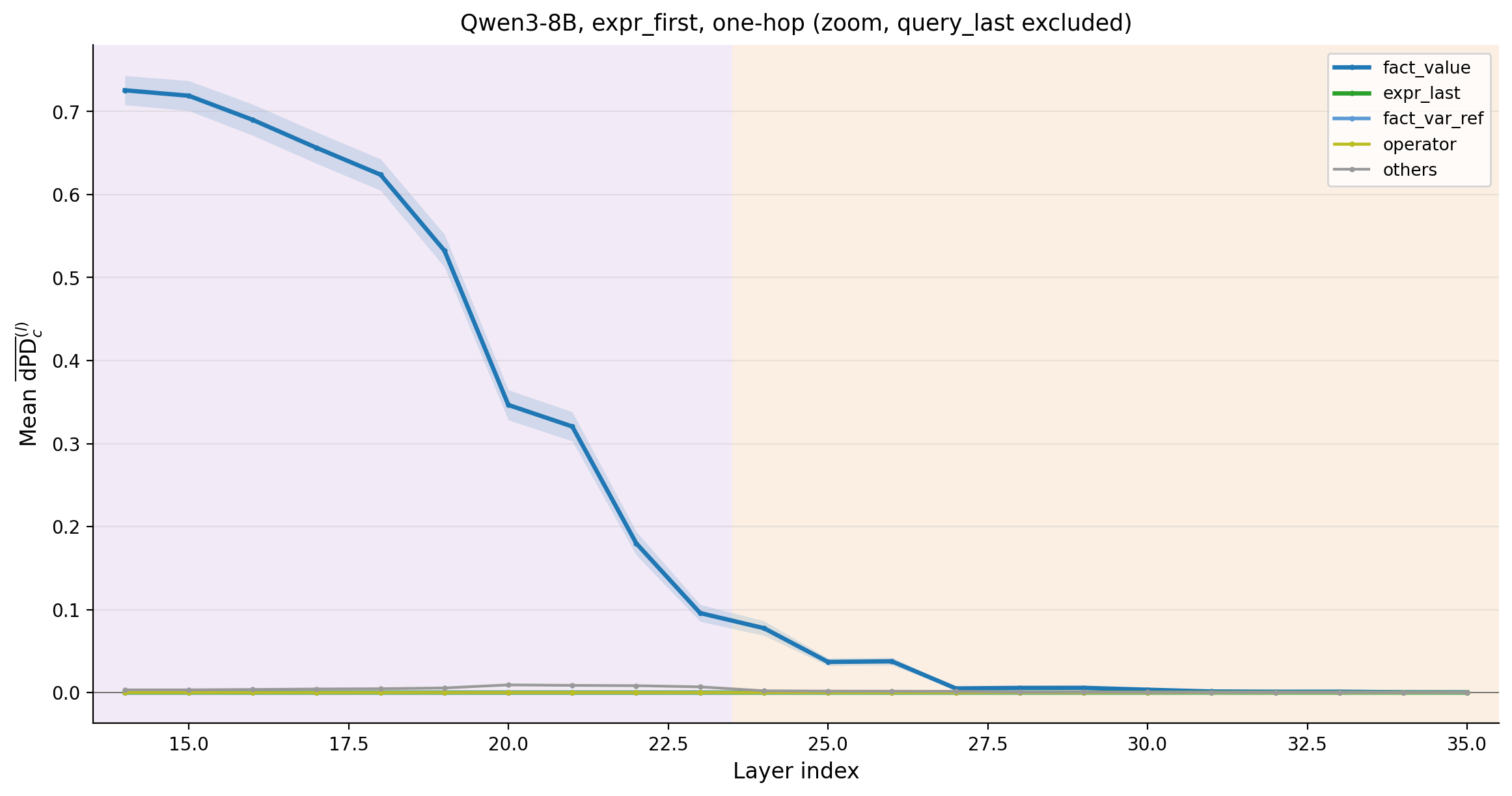}\hfill
    \includegraphics[width=0.49\textwidth]{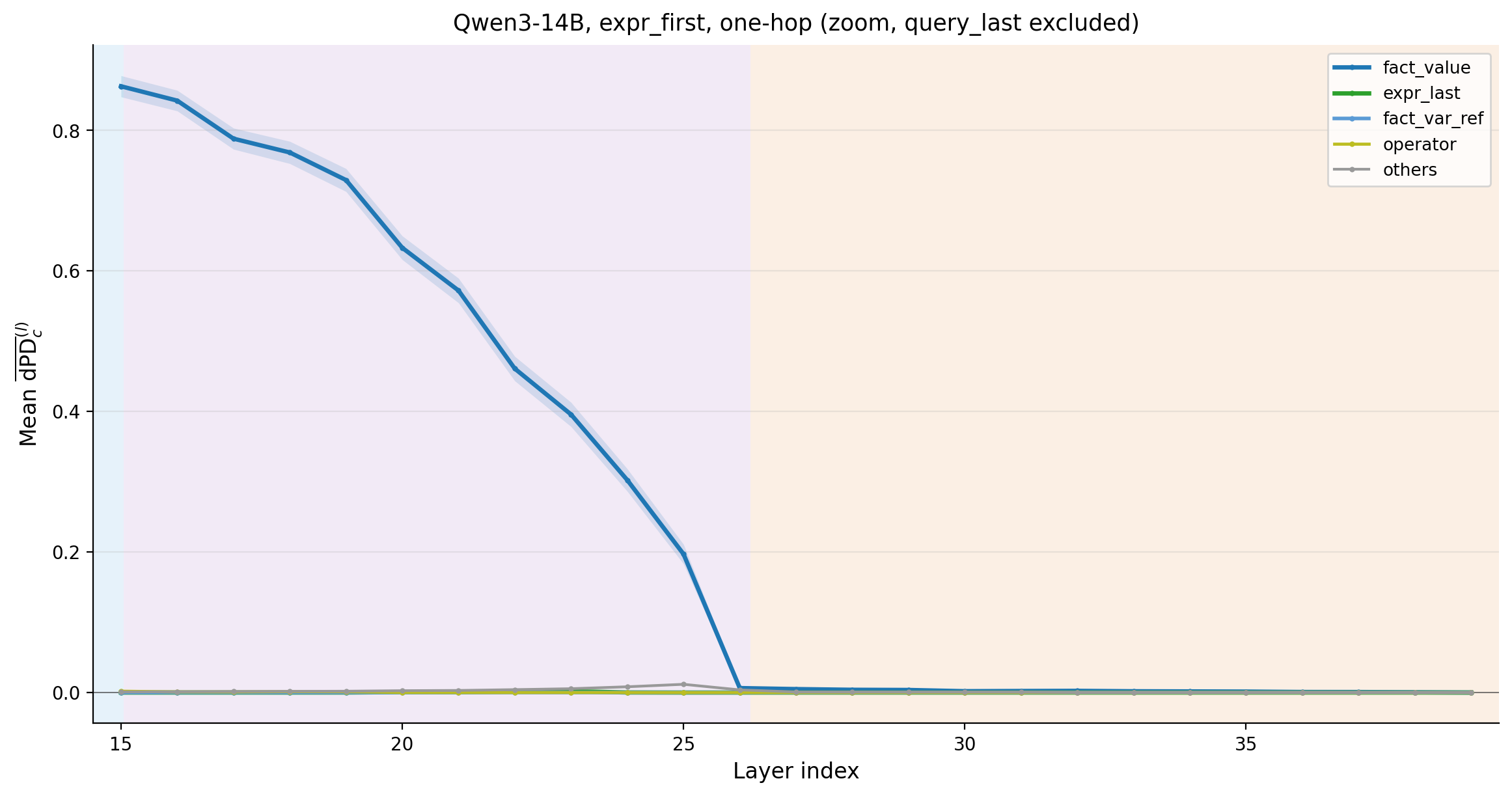}

    \includegraphics[width=0.49\textwidth]{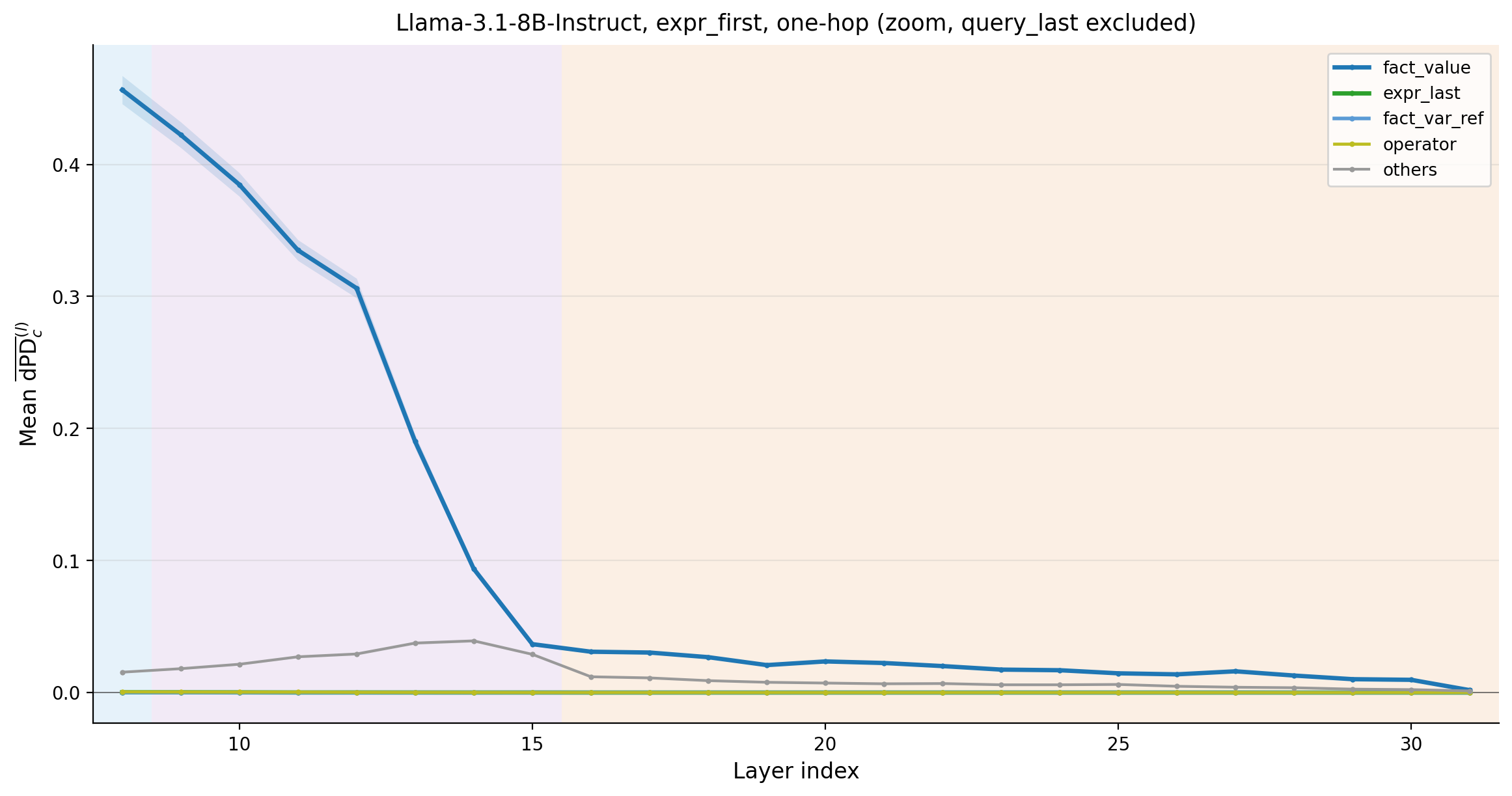}\hfill
    \includegraphics[width=0.49\textwidth]{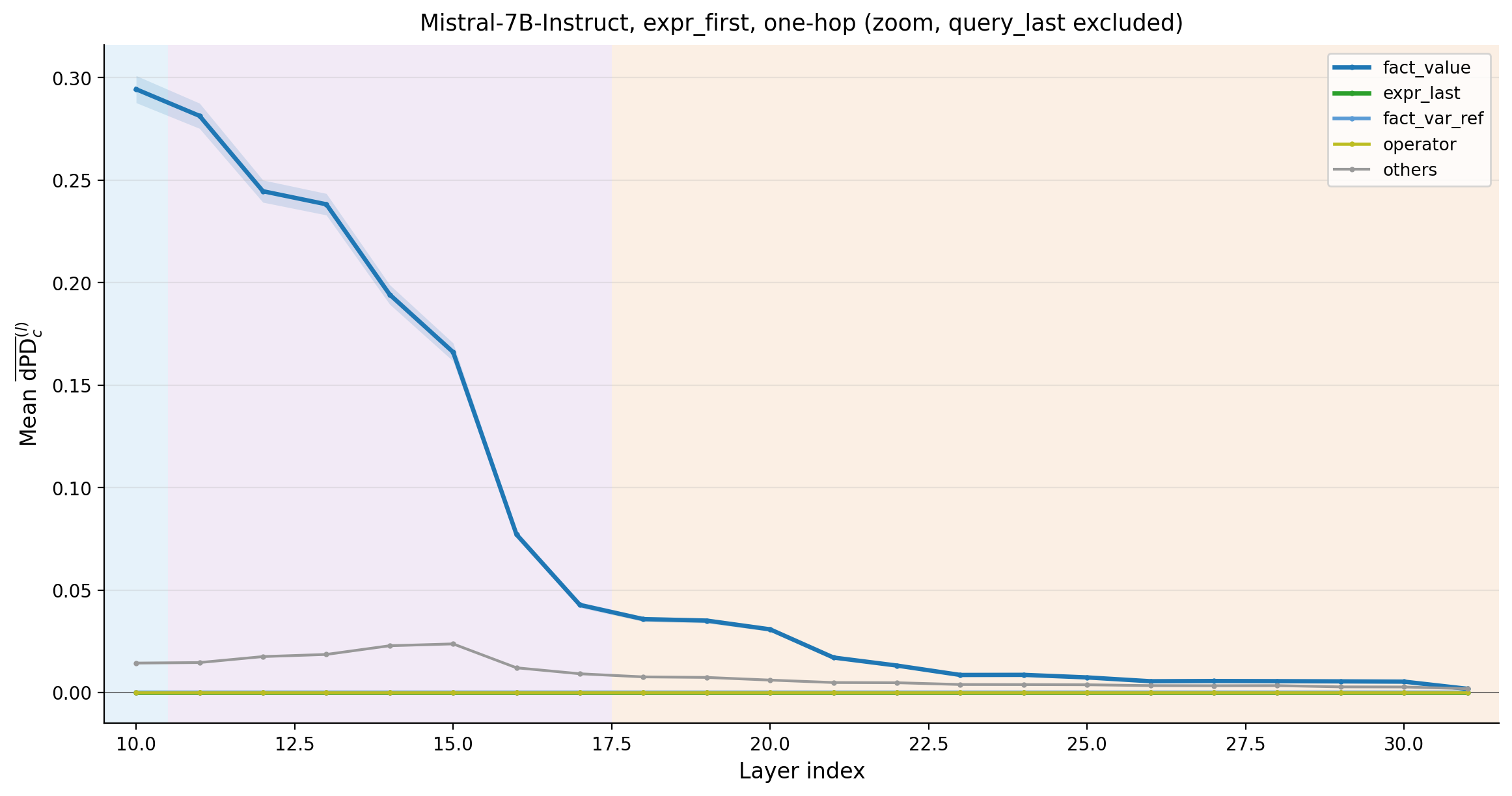}
    \caption{\textbf{Per-token mean $\mathrm{dPD}$, $\pi_{\mathrm{EF}}$, one-hop, \texttt{query\_last} excluded.} Top-left: \textit{Qwen3-8B}; top-right: \textit{Qwen3-14B}; bottom-left: \textit{Llama-3.1-8B-Instruct}; bottom-right: \textit{Mistral-7B-Instruct}.}
    \label{fig:token_expr_first_cross_model_zoom}
\end{figure*}

\begin{figure*}[!htbp]
    \centering
    \includegraphics[width=0.49\textwidth]{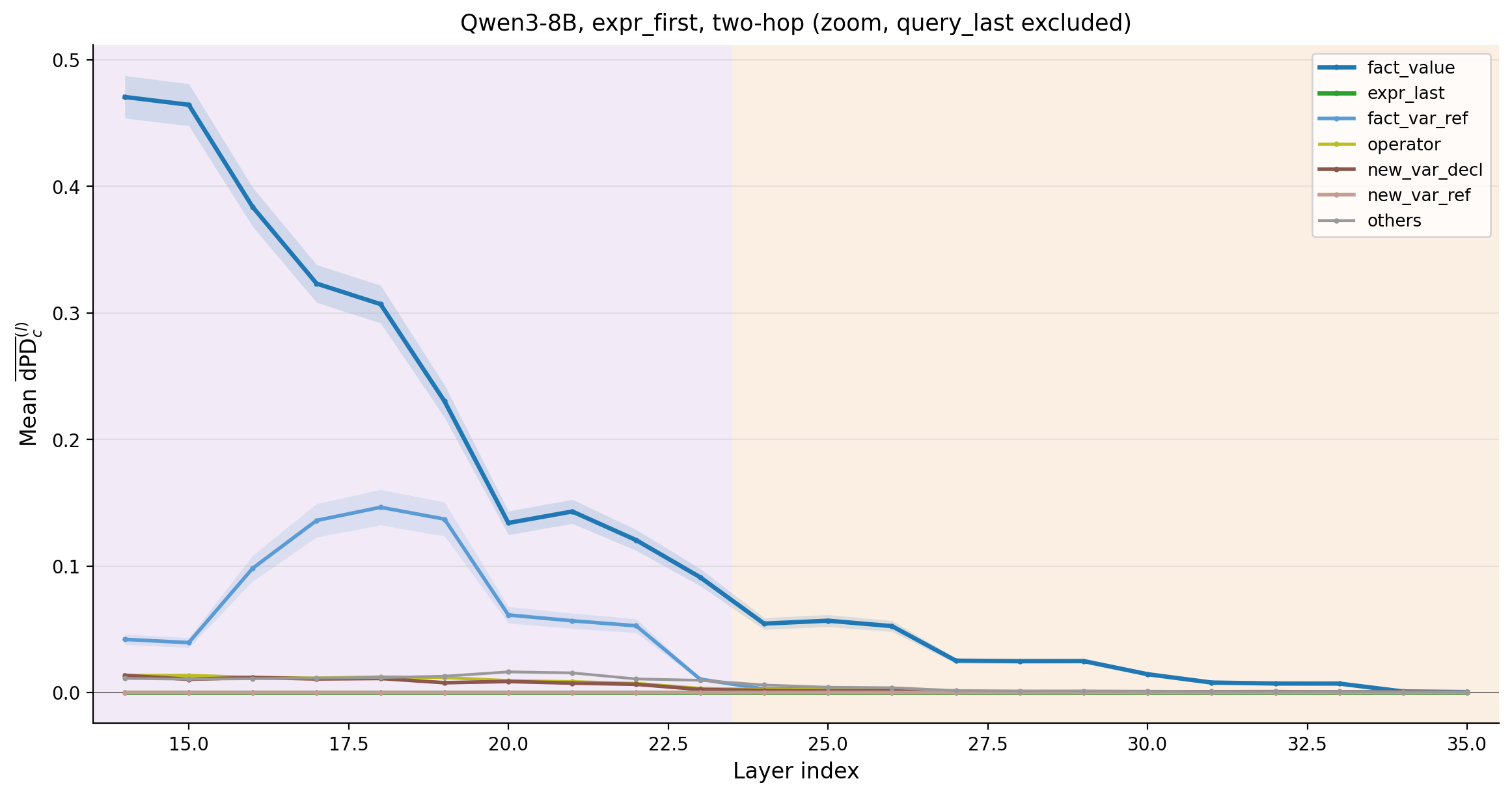}\hfill
    \includegraphics[width=0.49\textwidth]{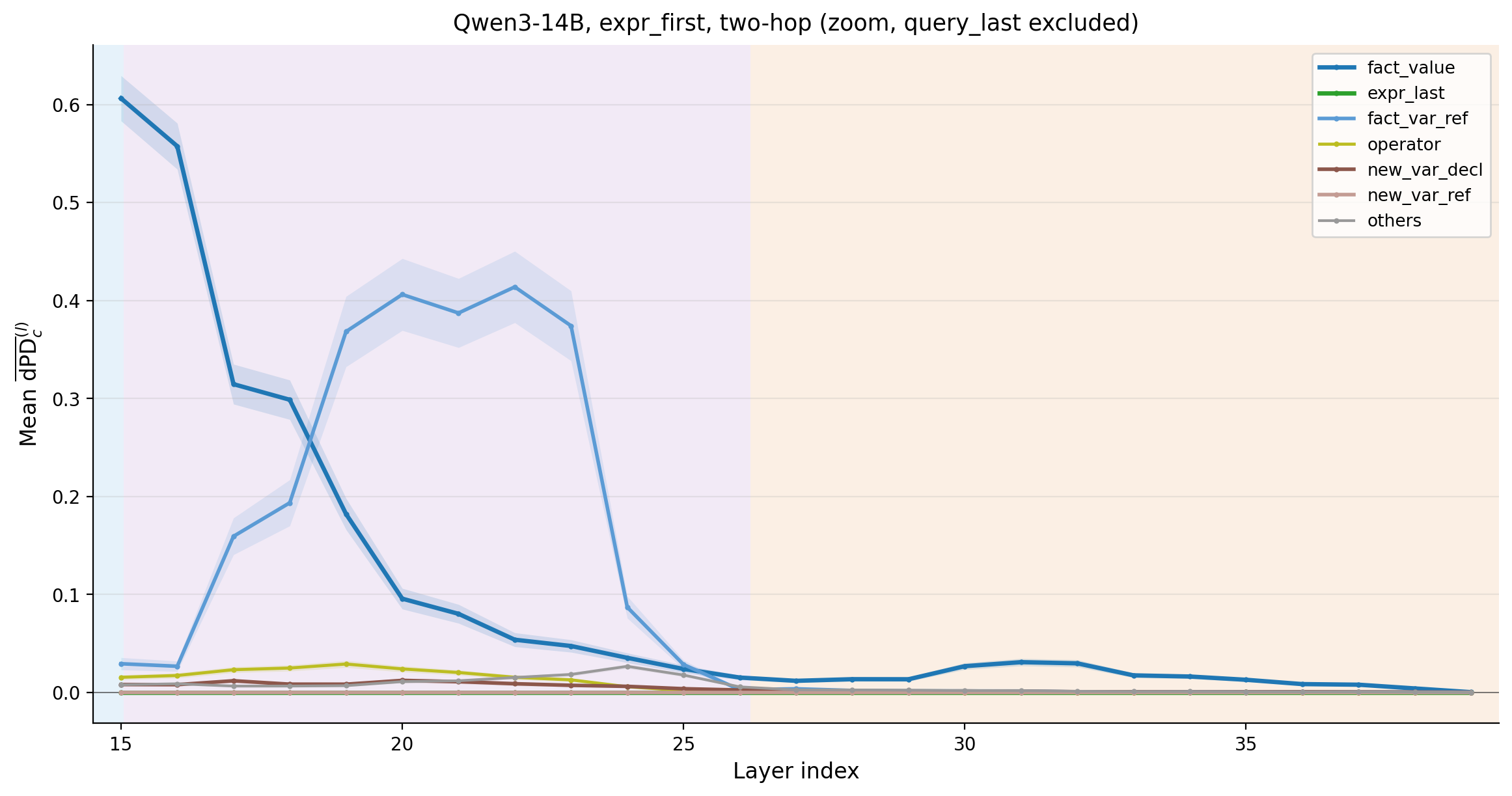}

    \includegraphics[width=0.49\textwidth]{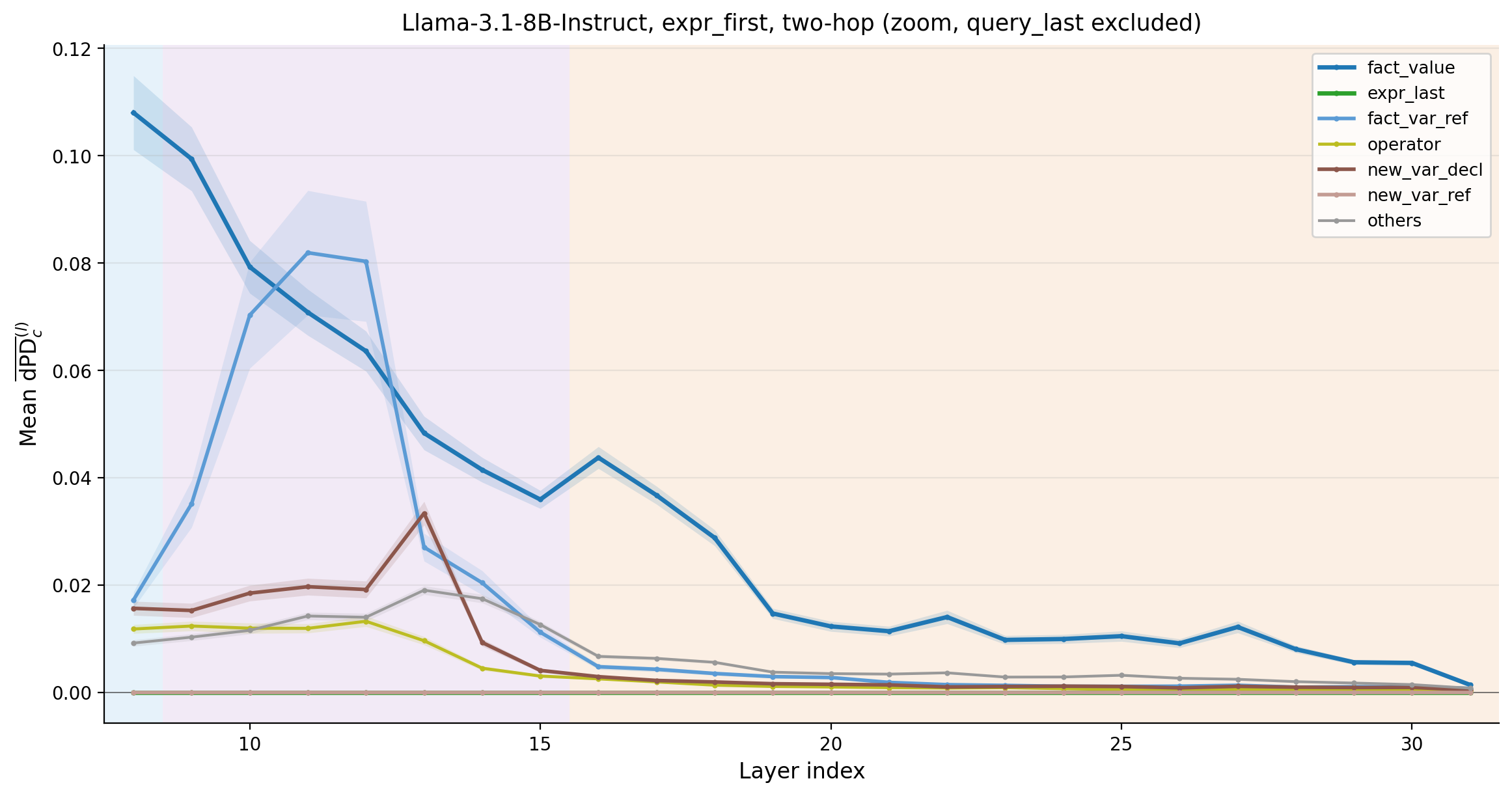}\hfill
    \includegraphics[width=0.49\textwidth]{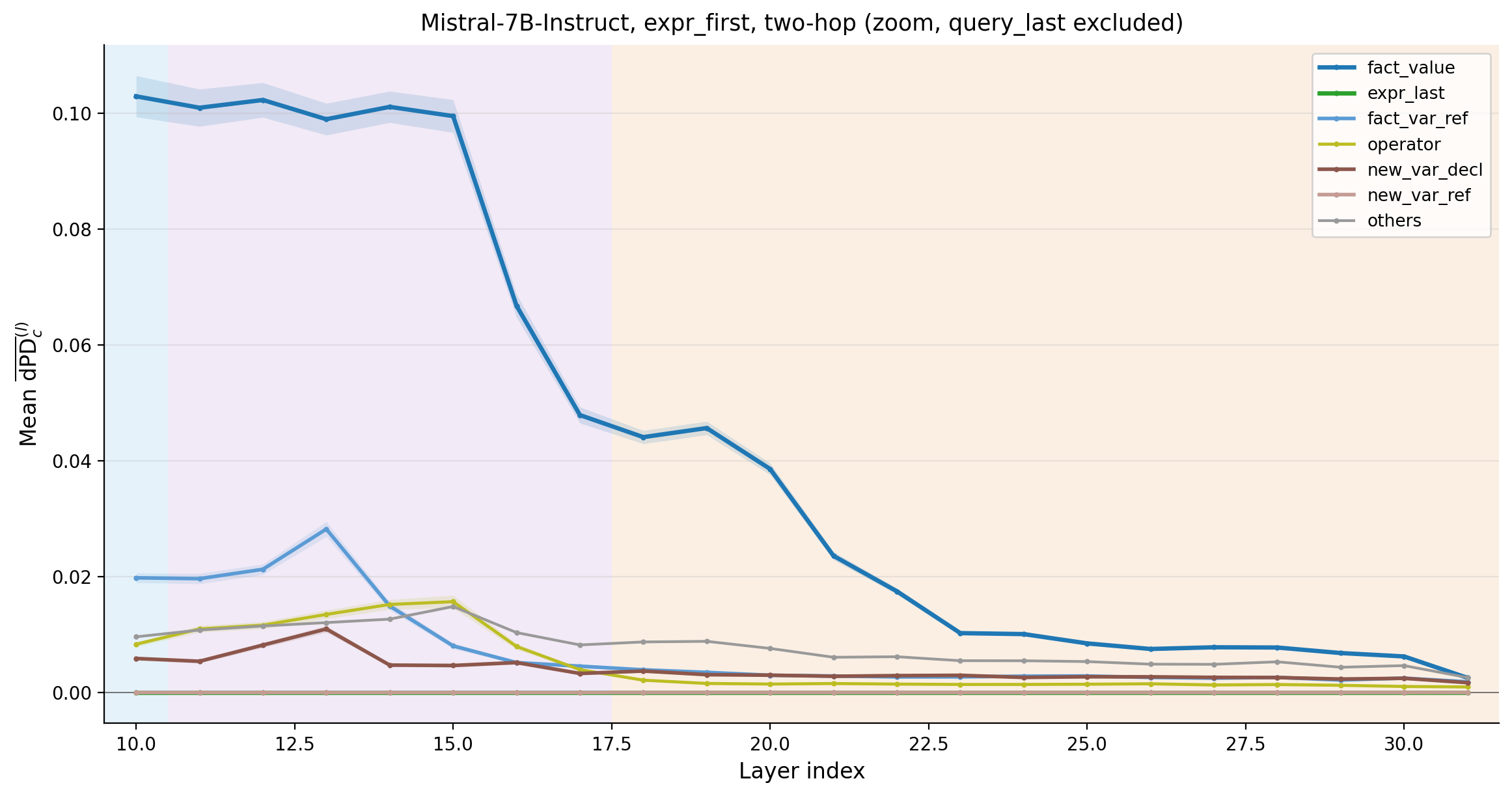}
    \caption{\textbf{Per-token mean $\mathrm{dPD}$, $\pi_{\mathrm{EF}}$, two-hop, \texttt{query\_last} excluded.} Top-left: \textit{Qwen3-8B}; top-right: \textit{Qwen3-14B}; bottom-left: \textit{Llama-3.1-8B-Instruct}; bottom-right: \textit{Mistral-7B-Instruct}.}
    \label{fig:token_expr_first_cross_model_twohop_zoom}
\end{figure*}

\paragraph{Middle-layer binding flow.}
The zoom views in Figures~\ref{fig:token_facts_first_cross_model_zoom}--\ref{fig:token_facts_first_cross_model_twohop_zoom} and Figures~\ref{fig:token_expr_first_cross_model_zoom}--\ref{fig:token_expr_first_cross_model_twohop_zoom} (\texttt{query\_last} excluded) make the middle-layer exchange between \texttt{fact\_value} and \texttt{fact\_var\_ref} visible. The signal is strongest under $\pi_{\mathrm{EF}}$ two-hop: \textit{Qwen3-14B} reaches a \texttt{fact\_var\_ref} middle-layer peak around $\mathrm{dPD}\!\approx\!0.41$ that overtakes \texttt{fact\_value}; \textit{Qwen3-8B} shows a smaller but clearly above-floor peak ($\approx 0.15$); \textit{Llama-3.1} and \textit{Mistral} retain the same shape with weaker magnitudes. The complementary late-layer \texttt{fact\_value} persistence is most visible in the harder hop $\times$ family conditions (two-hop in \textit{Llama-3.1} and \textit{Mistral}), consistent with the late-layer sub-pattern reported in \S\ref{sec:retro}.

\section{Supplementary Results: Specialized Attention Heads}
\label{appendix_heads}

\begin{figure}[!htbp]
    \centering
    \begin{subfigure}[b]{0.78\linewidth}
        \centering
        \includegraphics[width=\linewidth]{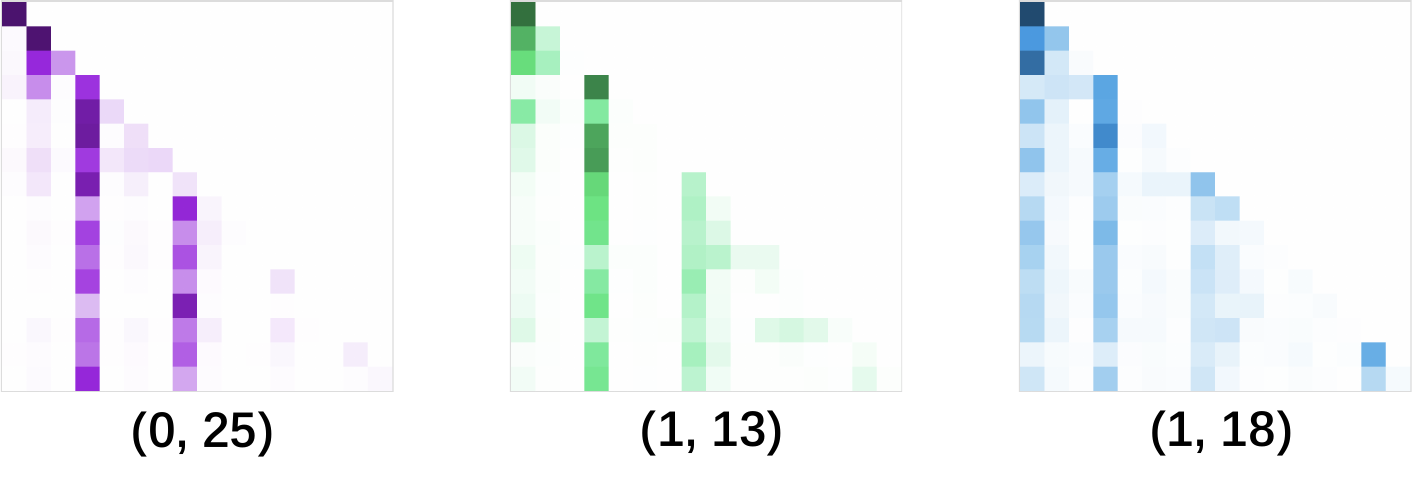}
        \caption{Splitting heads}
        \label{fig:splitting_head}
    \end{subfigure}

    \begin{subfigure}[b]{0.78\linewidth}
        \centering
        \includegraphics[width=\linewidth]{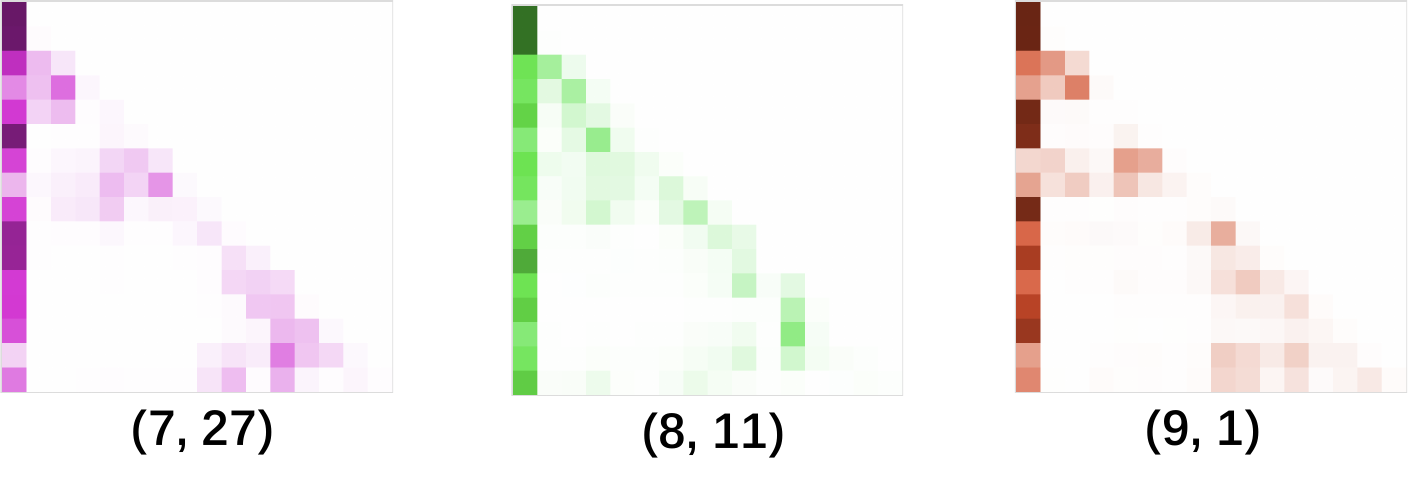}
        \caption{Transmission heads}
        \label{fig:convergence_head}
    \end{subfigure}

    \begin{subfigure}[b]{0.78\linewidth}
        \centering
        \includegraphics[width=\linewidth]{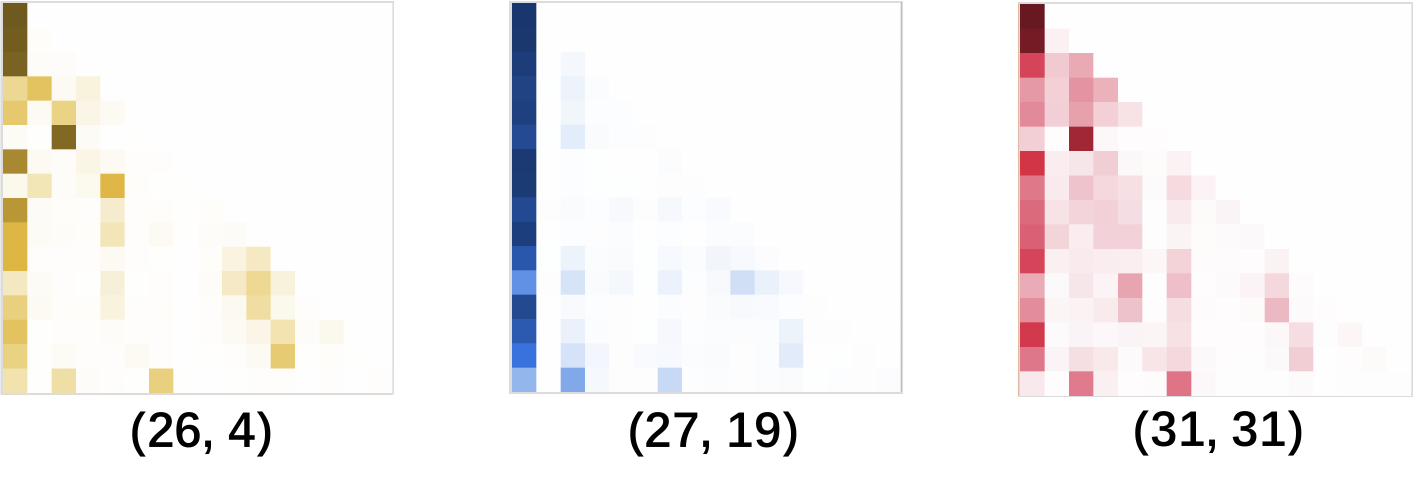}
        \caption{Fact-Retrieval heads}
        \label{fig:facts_head}
    \end{subfigure}
    \caption{\textbf{Representative attention maps for the three head roles in \textit{Qwen3-8B}.} The prompt is \texttt{A is True, B is False, ($\neg$A or $\neg$B) is}; $(x,y)$ denotes the $y$-th head in the $x$-th layer.}
    \label{fig:combined_heads}
\end{figure}

\begin{figure*}[!htbp]
    \centering
    \includegraphics[width=0.49\textwidth]{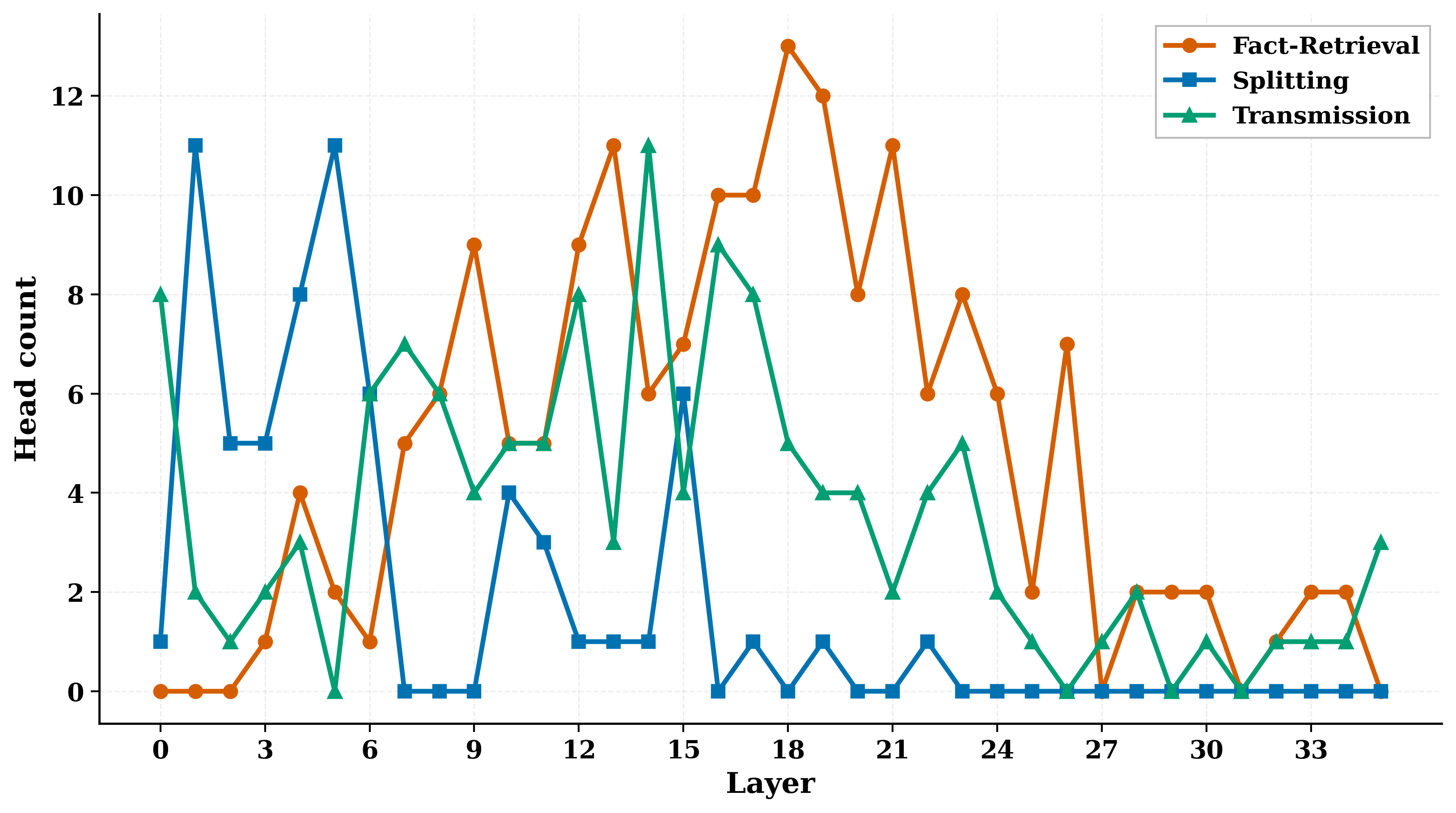}\hfill
    \includegraphics[width=0.49\textwidth]{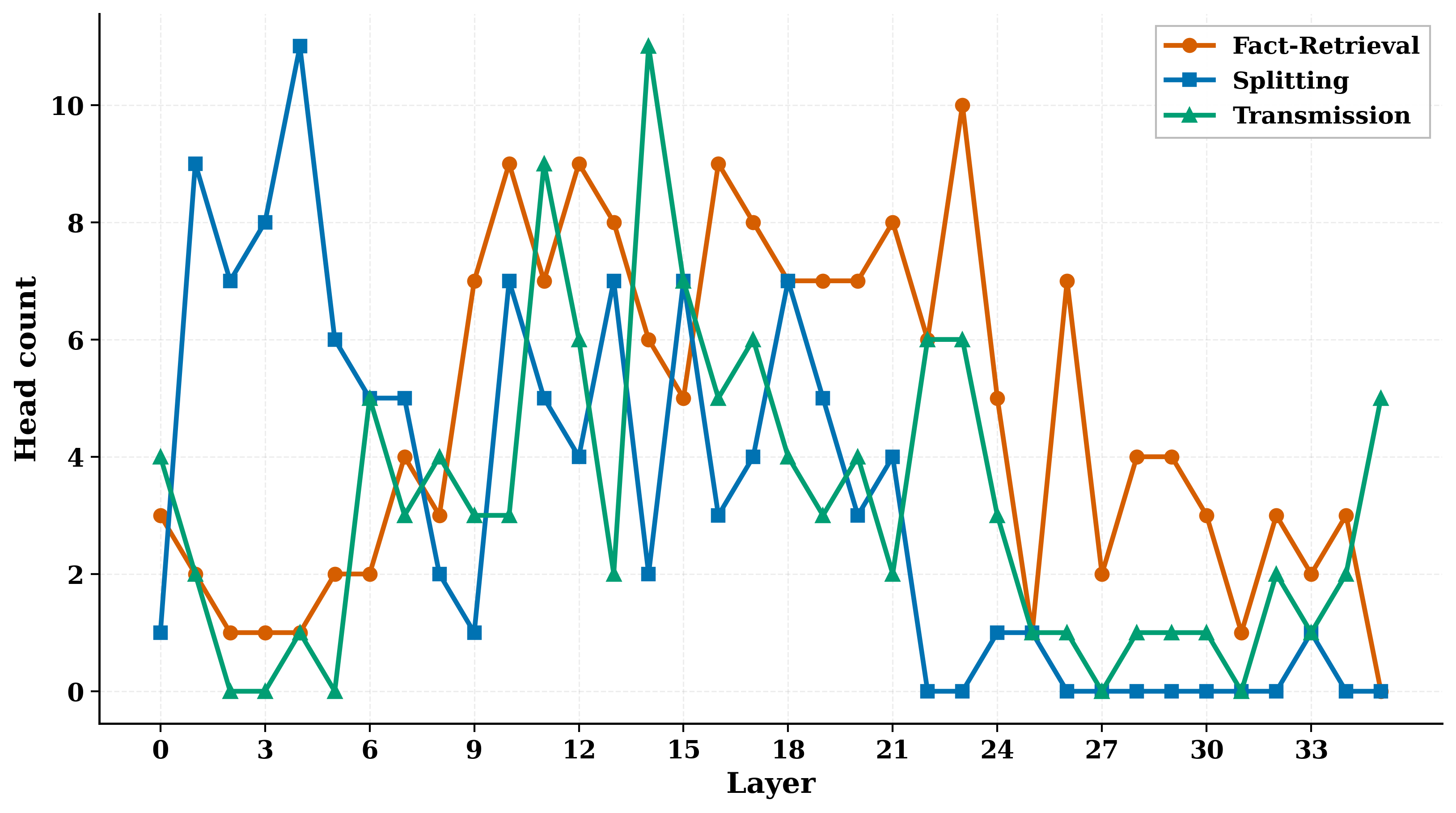}

    \includegraphics[width=0.49\textwidth]{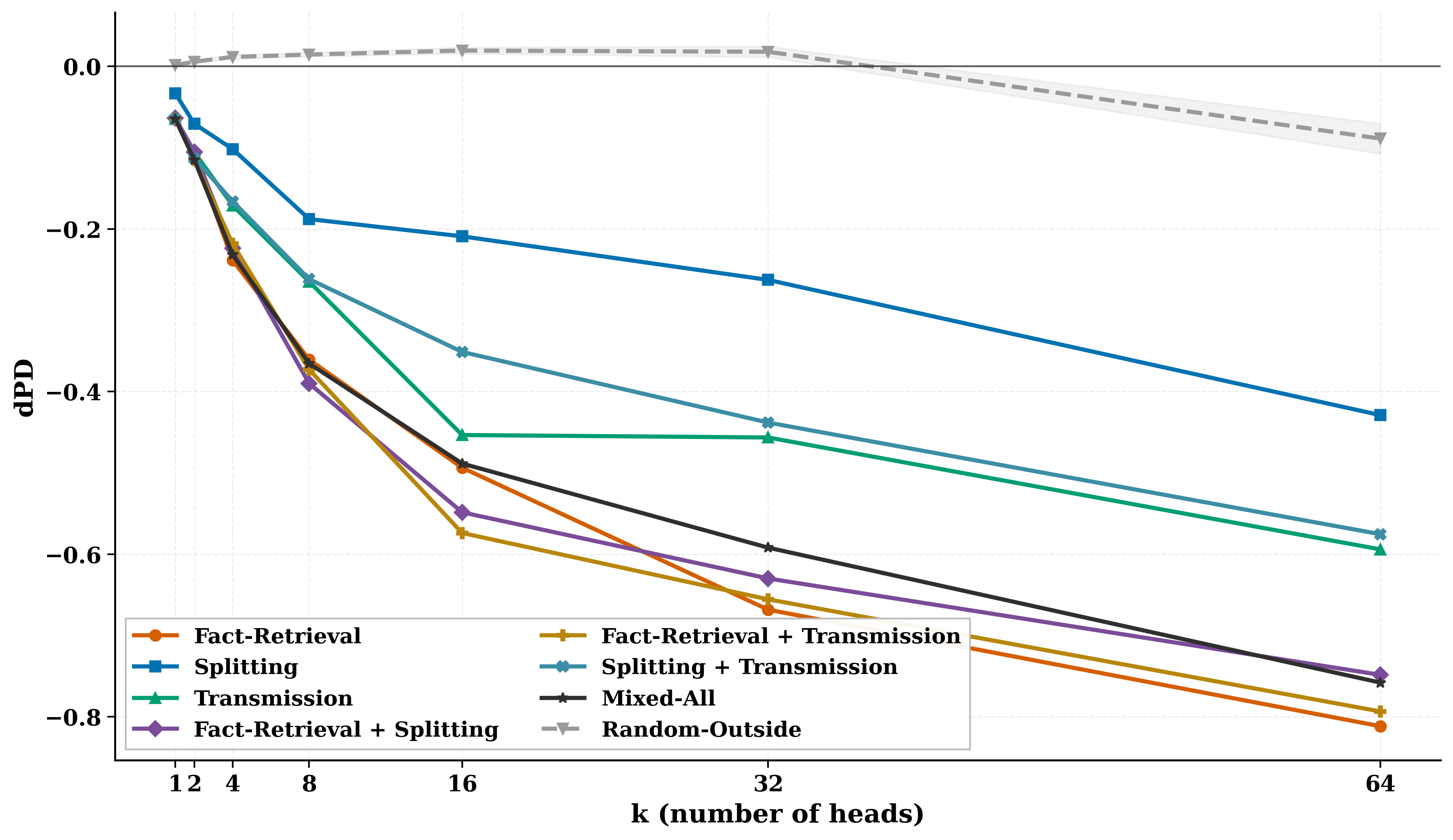}\hfill
    \includegraphics[width=0.49\textwidth]{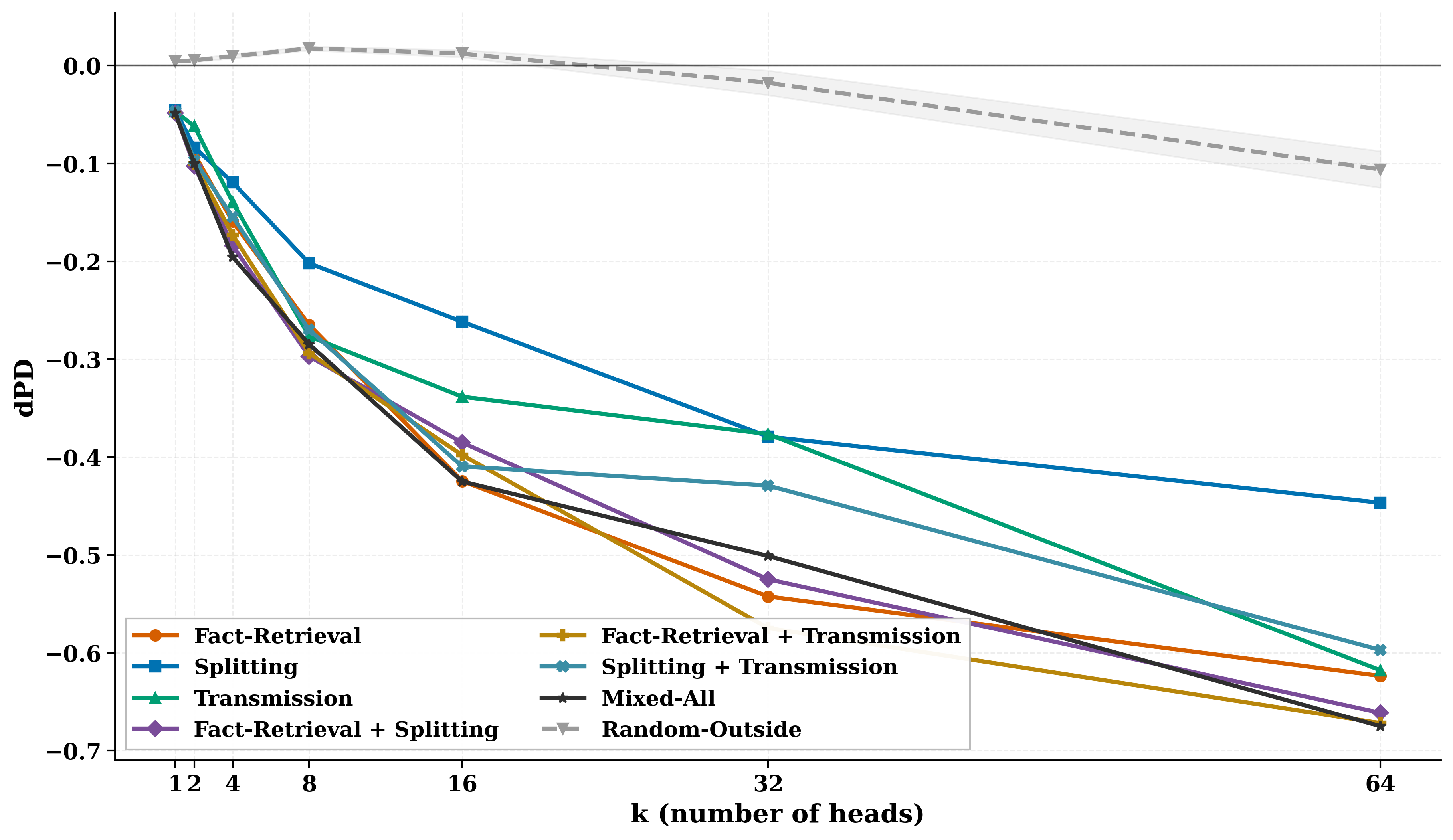}
    \caption{\textbf{\textit{Qwen3-8B}: head roles under two prompt orders.} Top row: layer-wise role distributions (left: $\pi_{\mathrm{FF}}$, right: $\pi_{\mathrm{EF}}$). Bottom row: $\mathrm{dPD}/\mathrm{PD}_{\mathrm{clean}}$ validation curves under the same prompt orders.}
    \label{fig:qwen3_8b_heads_panel}
\end{figure*}

\begin{figure*}[!htbp]
    \centering
    \includegraphics[width=0.49\textwidth]{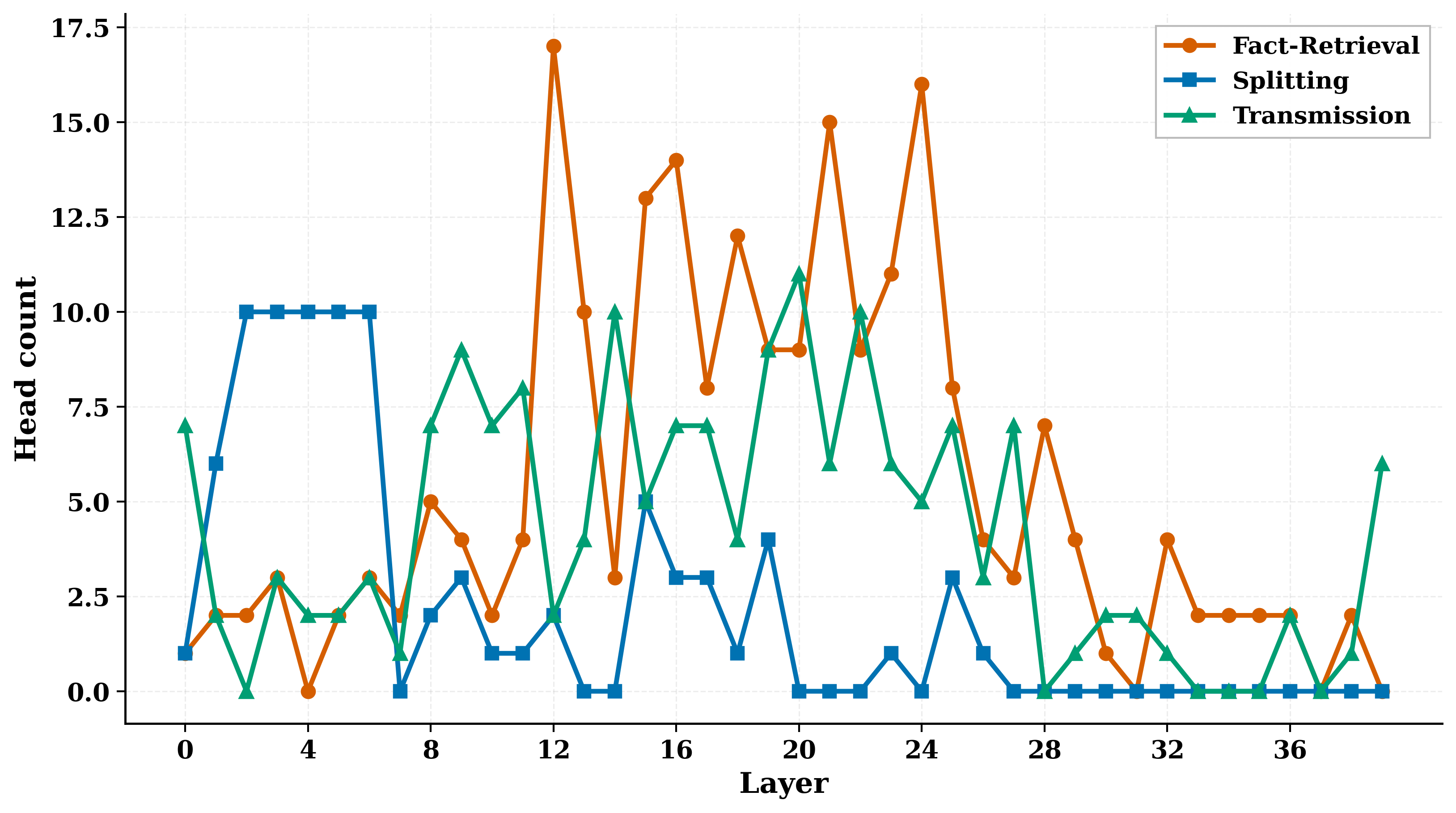}\hfill
    \includegraphics[width=0.49\textwidth]{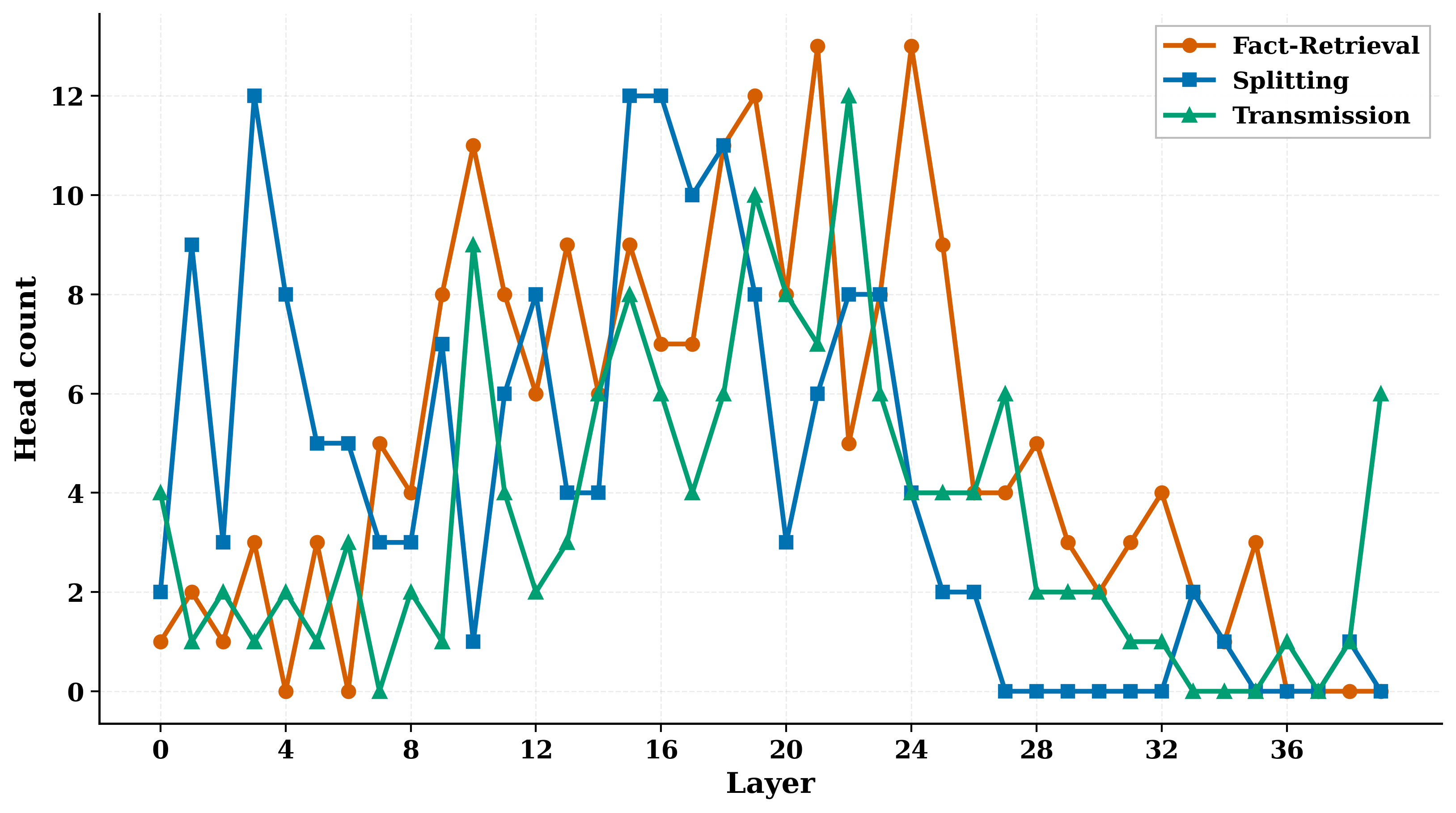}

    \includegraphics[width=0.49\textwidth]{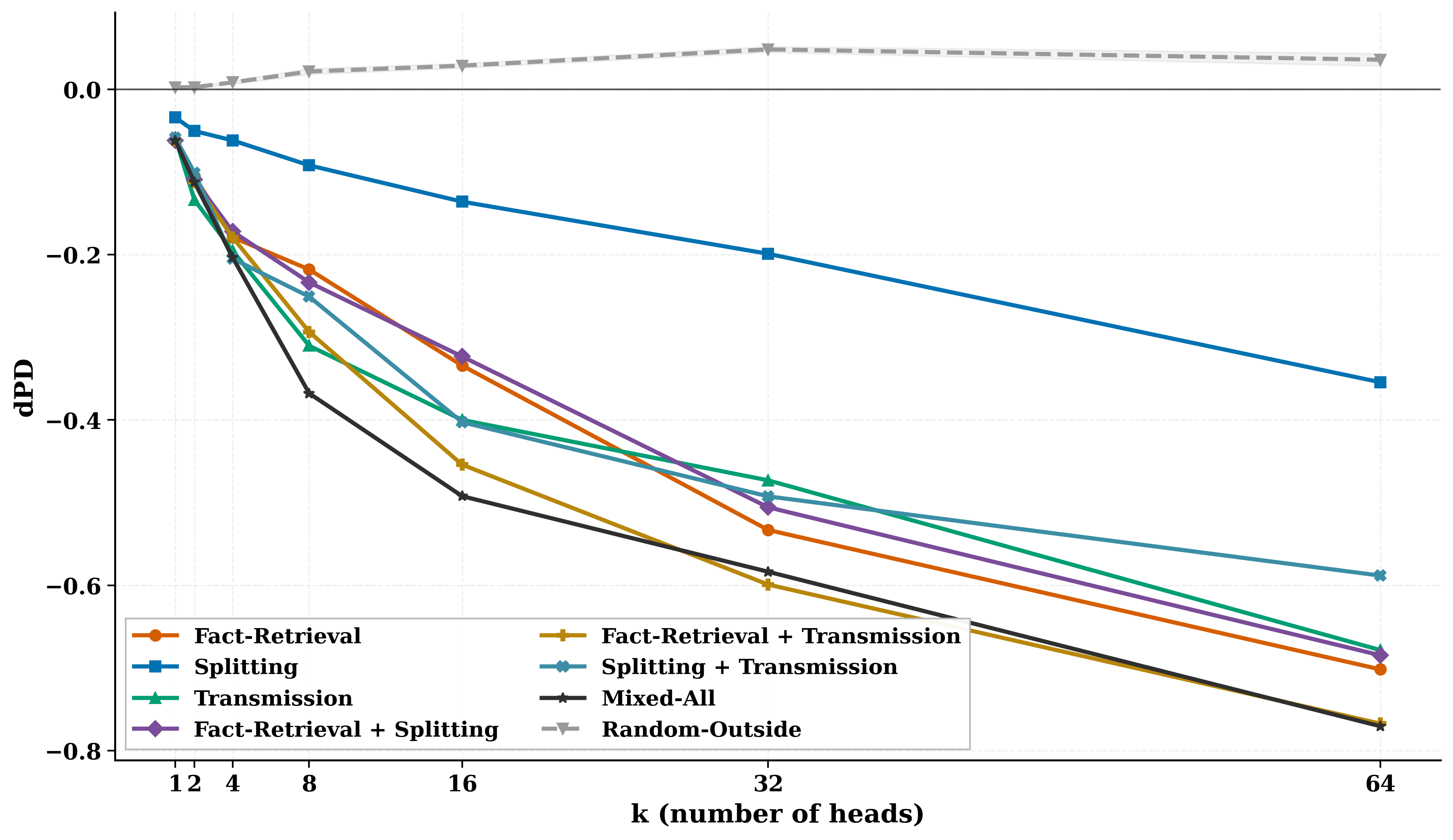}\hfill
    \includegraphics[width=0.49\textwidth]{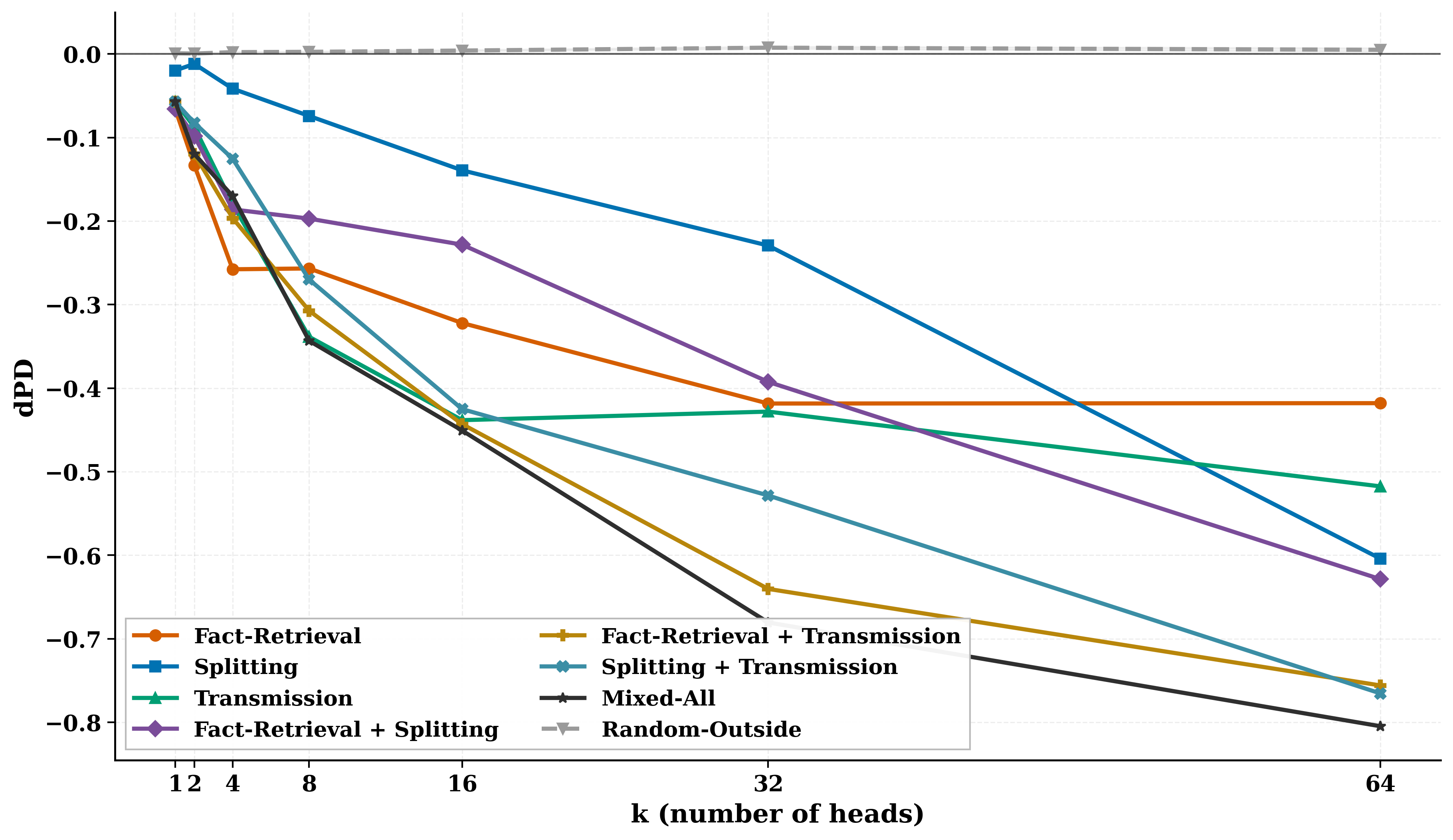}
    \caption{\textbf{\textit{Qwen3-14B}: head roles under two prompt orders.} Top row: layer-wise role distributions (left: $\pi_{\mathrm{FF}}$, right: $\pi_{\mathrm{EF}}$). Bottom row: $\mathrm{dPD}/\mathrm{PD}_{\mathrm{clean}}$ validation curves under the same prompt orders.}
    \label{fig:qwen3_14b_heads_panel}
    \label{fig:counts_14B_comparison}
    \label{fig:expr_first_heads_panel}
\end{figure*}

\begin{figure*}[!htbp]
    \centering
    \includegraphics[width=0.49\textwidth]{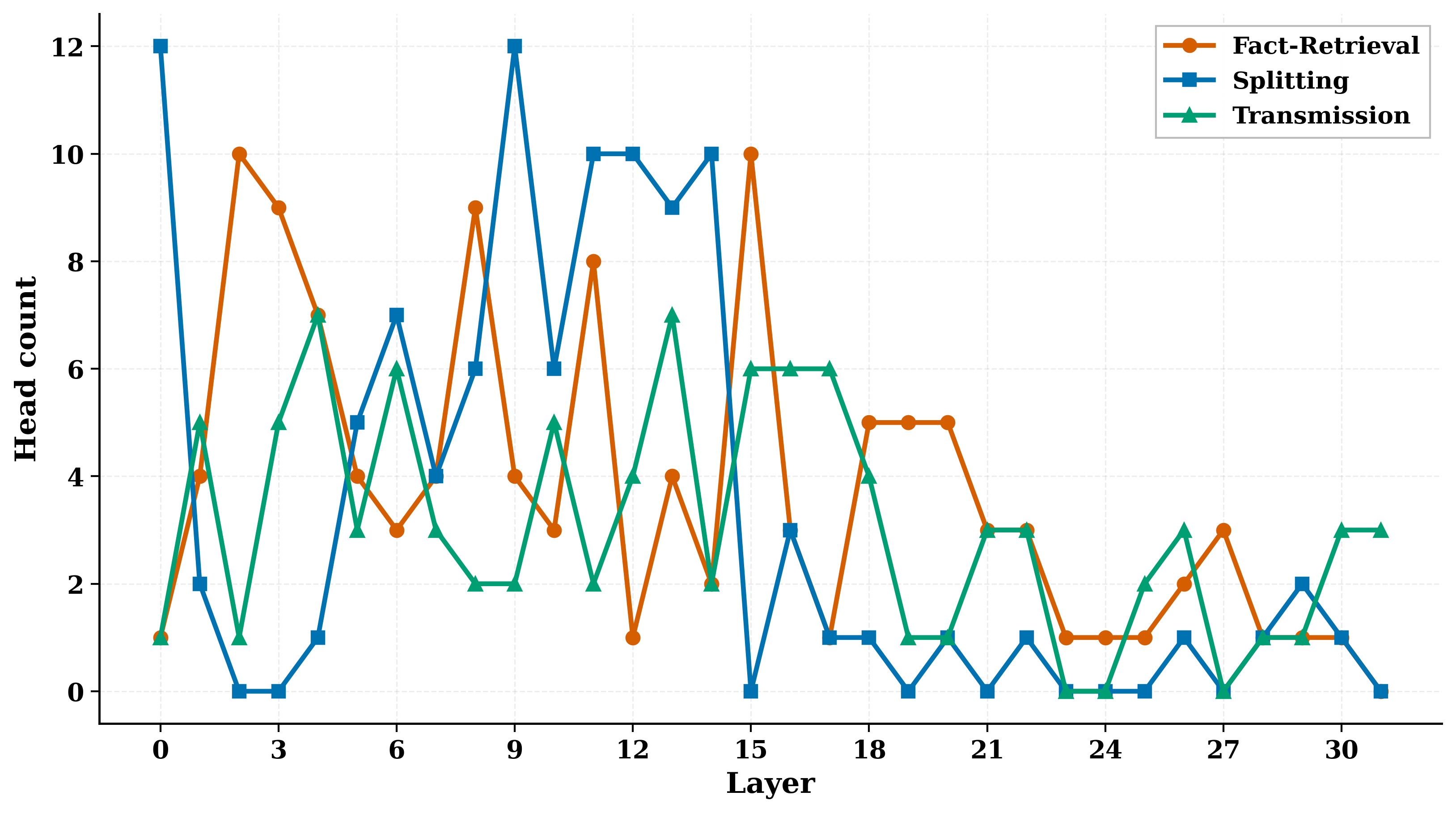}\hfill
    \includegraphics[width=0.49\textwidth]{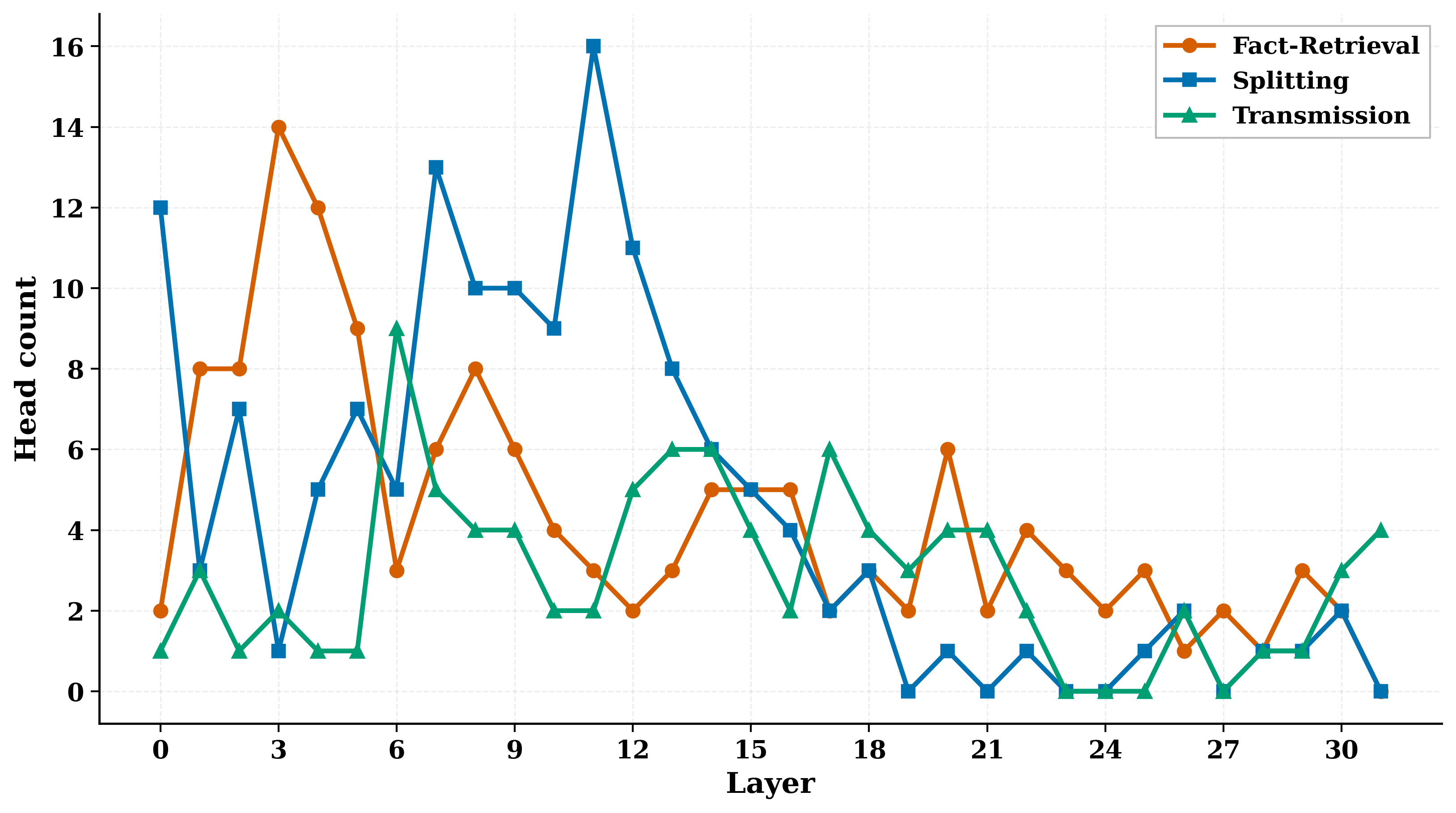}

    \includegraphics[width=0.49\textwidth]{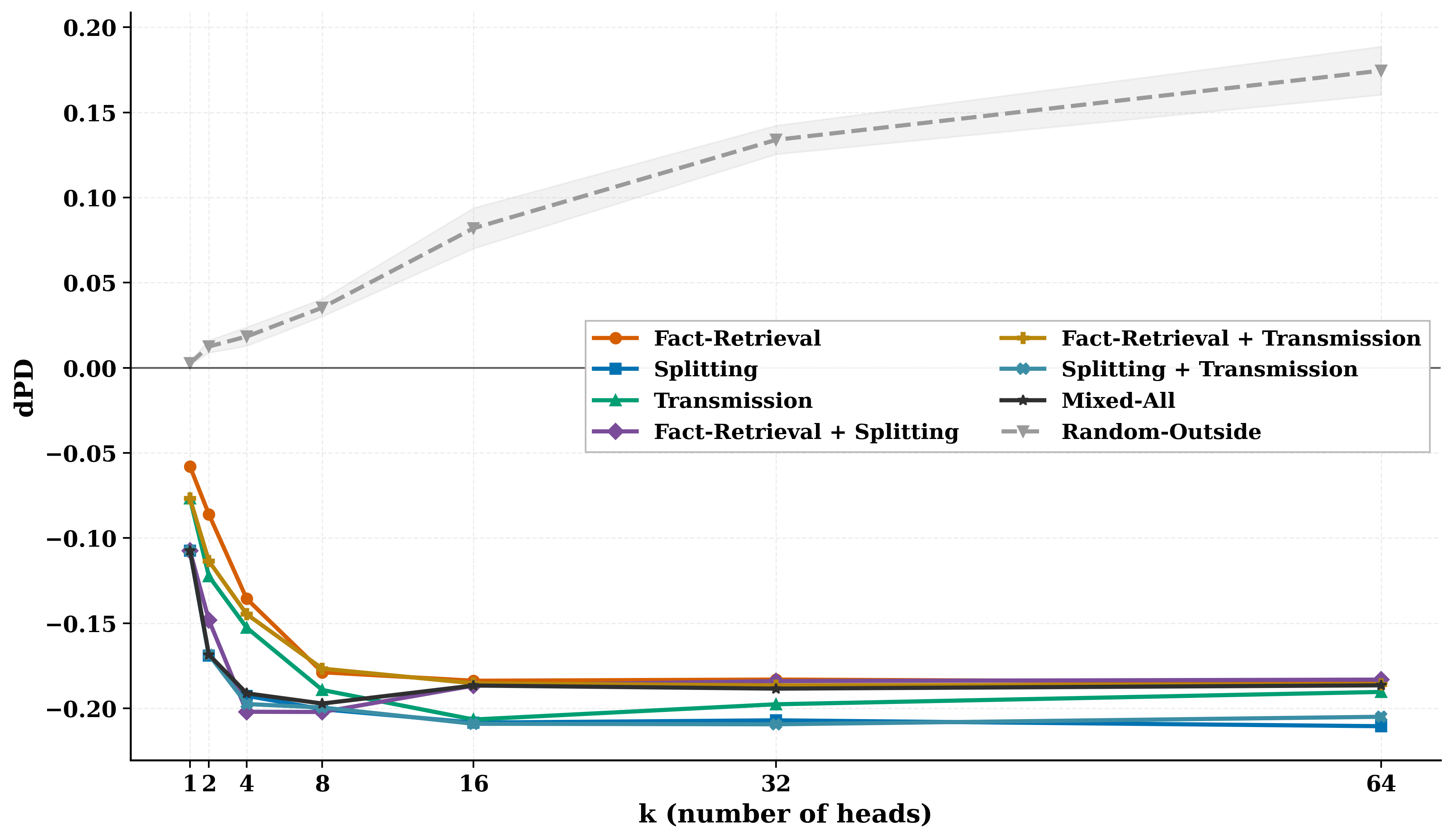}\hfill
    \includegraphics[width=0.49\textwidth]{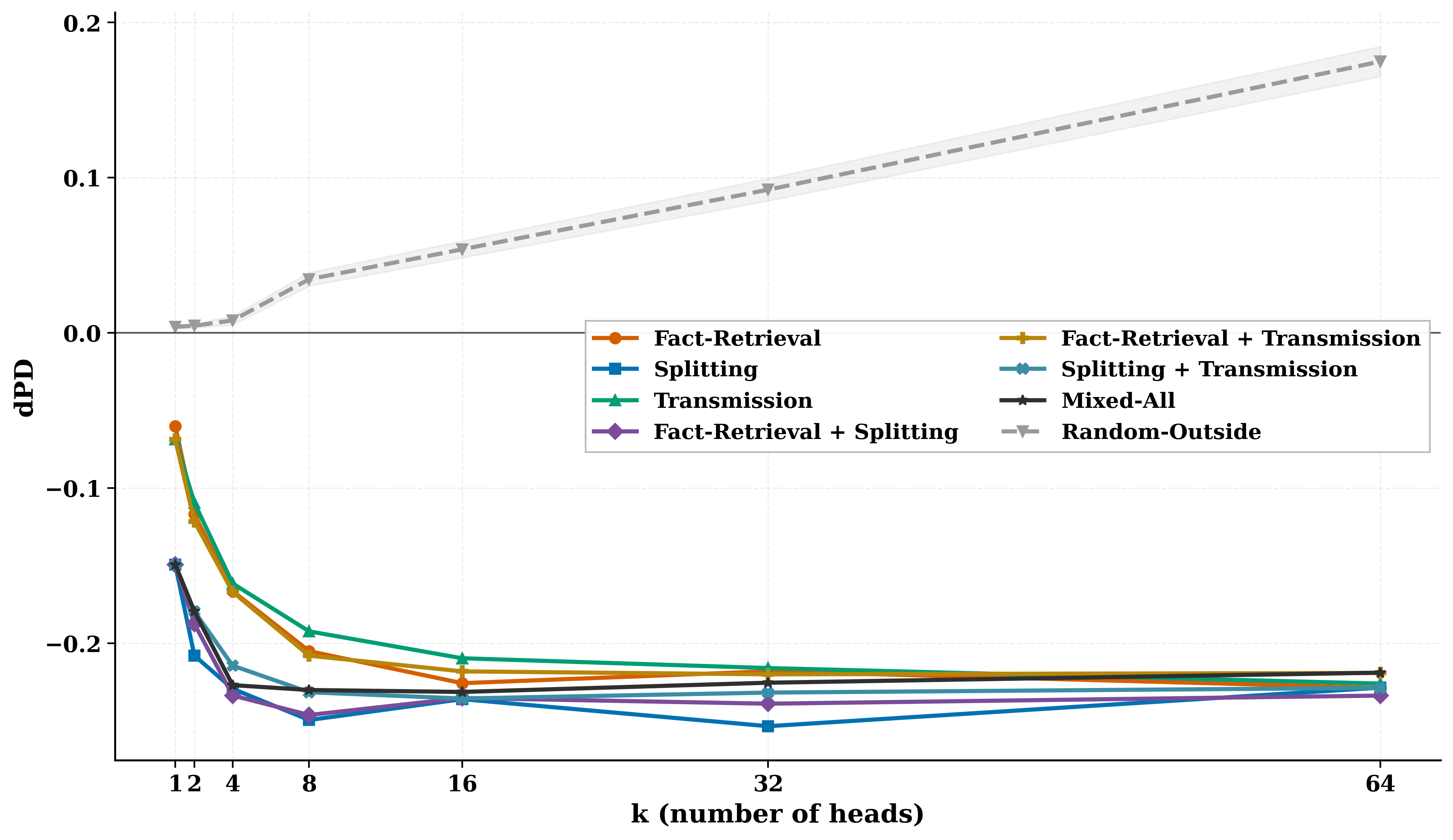}
    \caption{\textbf{\textit{Llama-3.1-8B-Instruct}: head roles under two prompt orders.} Top row: layer-wise role distributions (left: $\pi_{\mathrm{FF}}$, right: $\pi_{\mathrm{EF}}$). Bottom row: $\mathrm{dPD}/\mathrm{PD}_{\mathrm{clean}}$ validation curves under the same prompt orders.}
    \label{fig:llama_heads_panel}
\end{figure*}

\begin{figure*}[!htbp]
    \centering
    \includegraphics[width=0.49\textwidth]{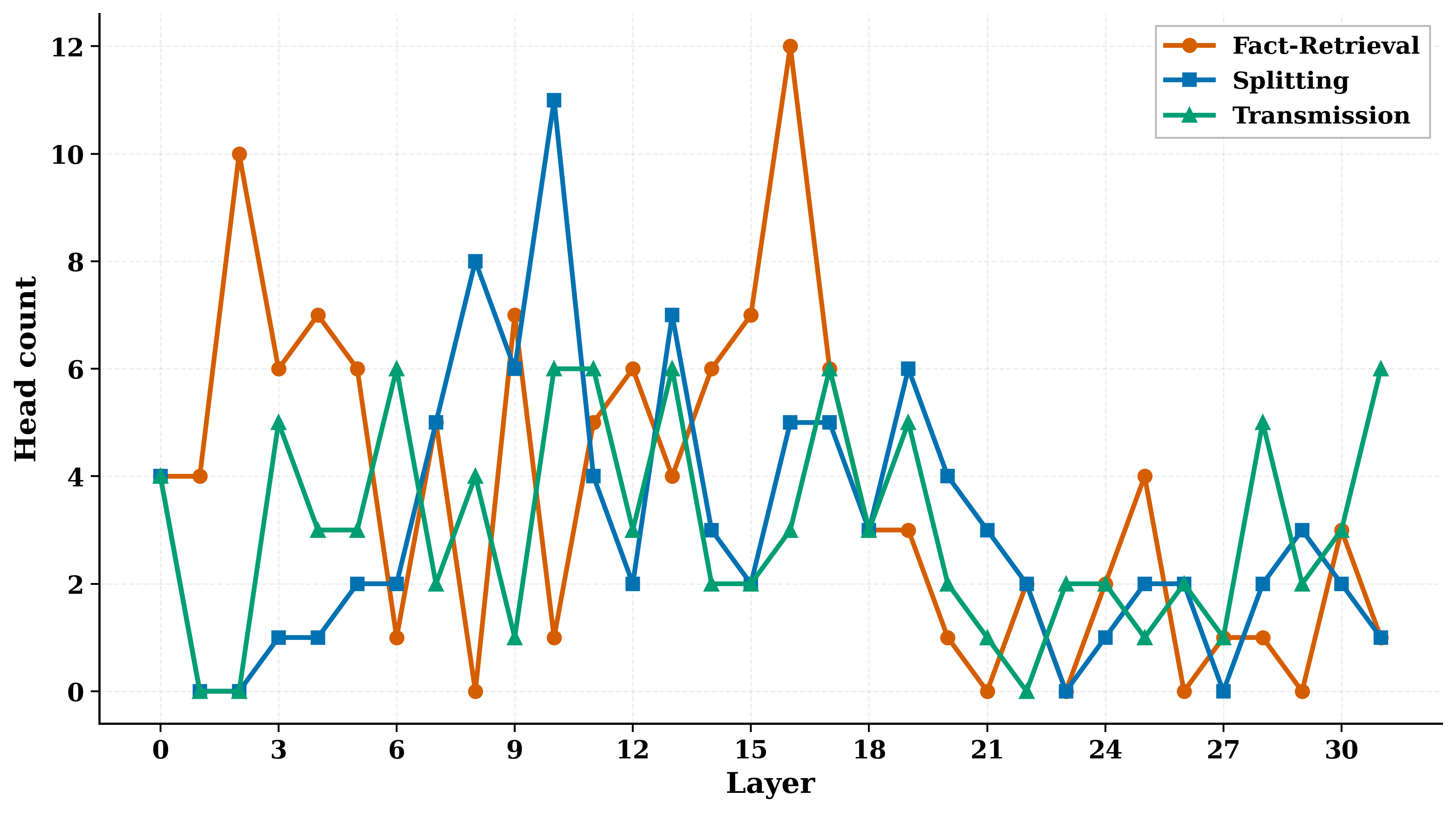}\hfill
    \includegraphics[width=0.49\textwidth]{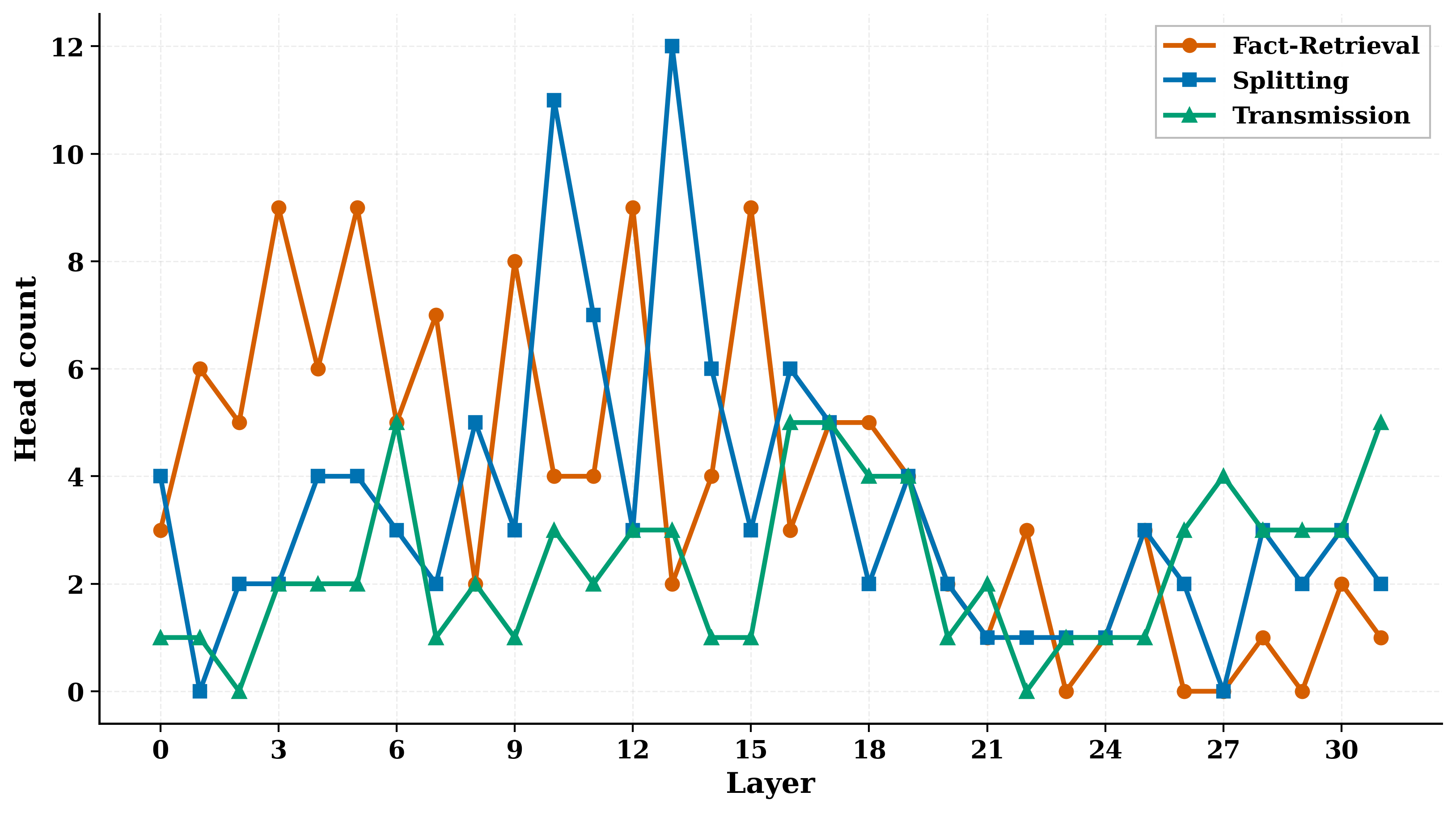}

    \includegraphics[width=0.49\textwidth]{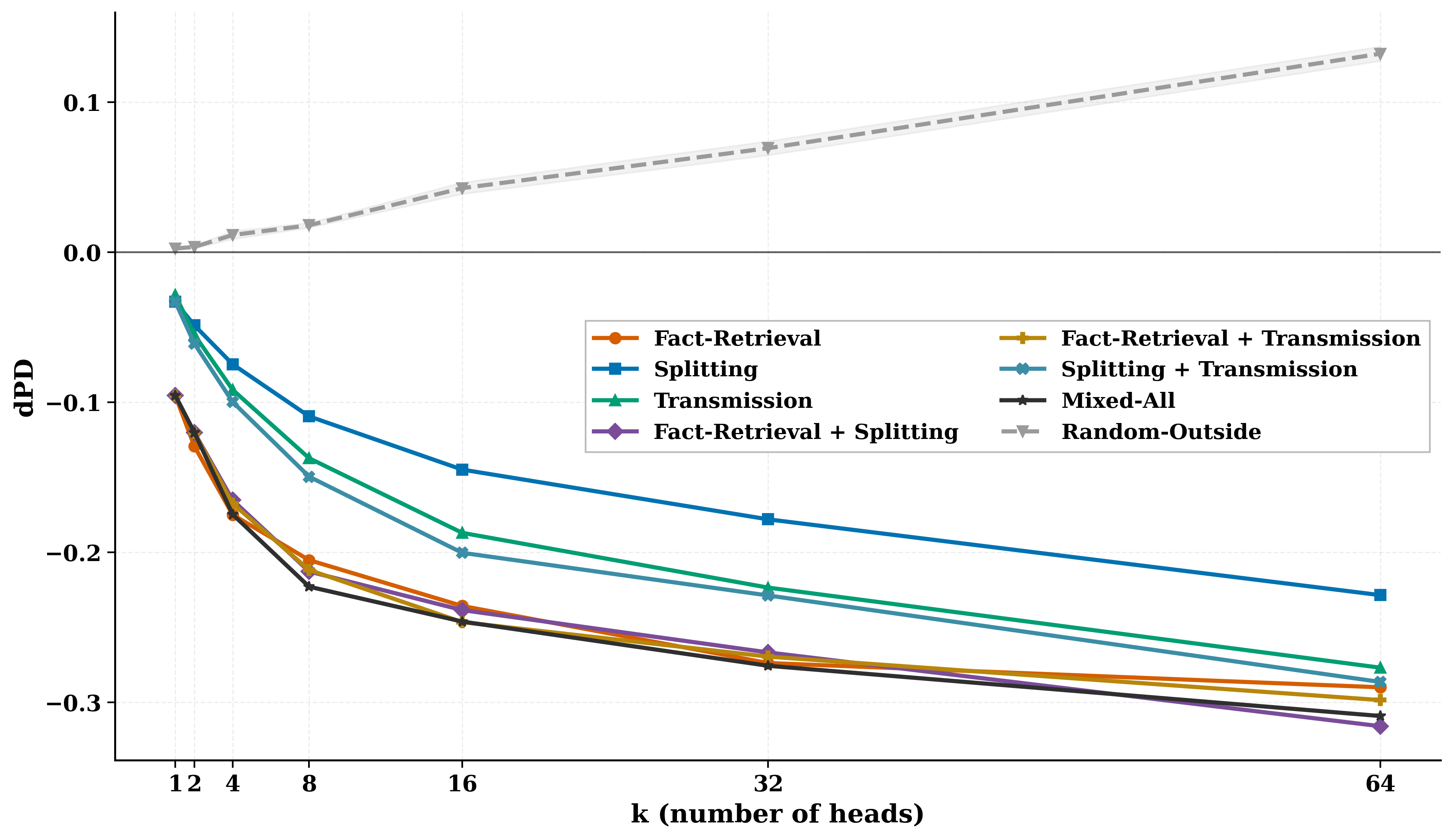}\hfill
    \includegraphics[width=0.49\textwidth]{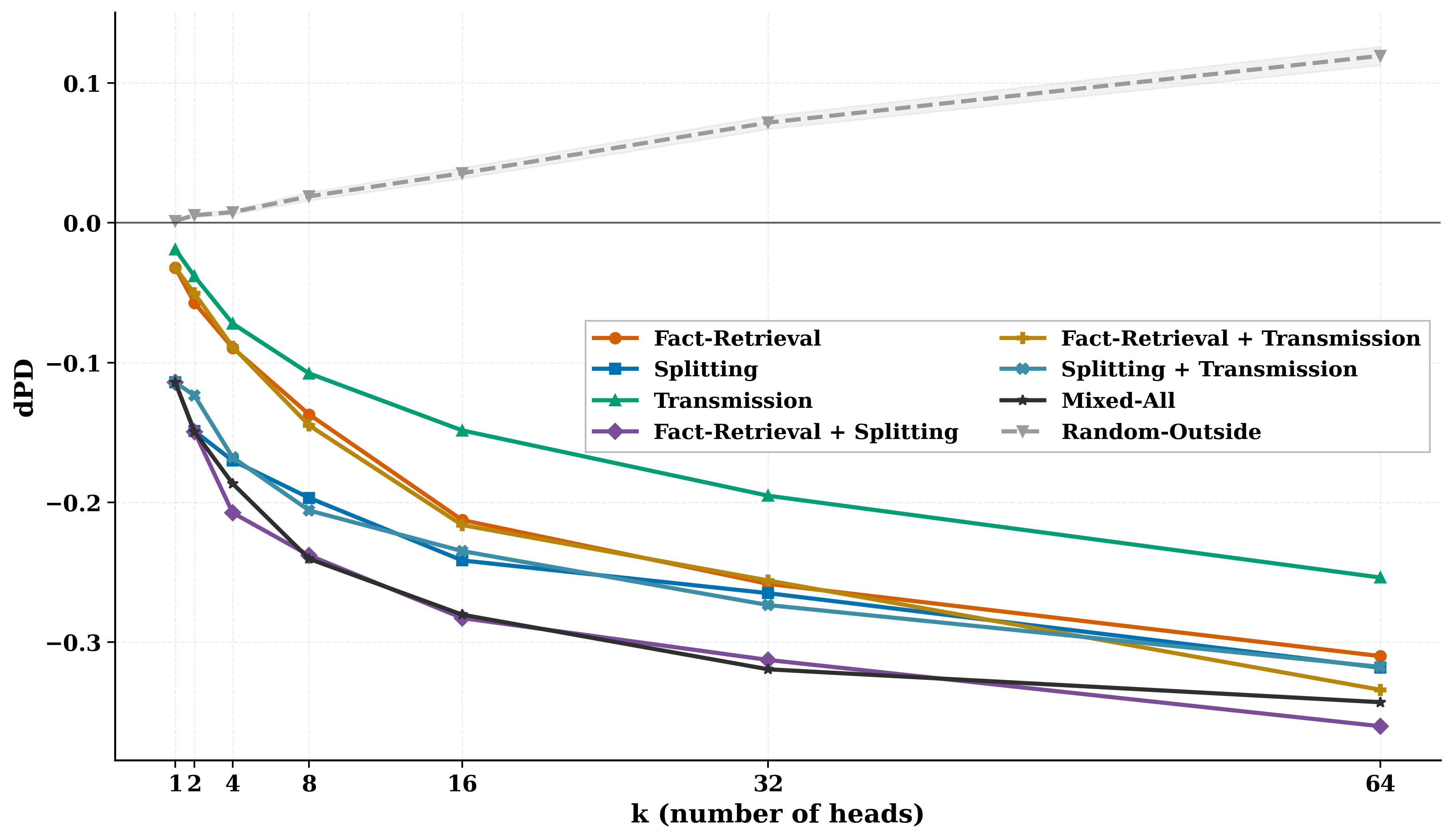}
    \caption{\textbf{\textit{Mistral-7B-Instruct}: head roles under two prompt orders.} Top row: layer-wise role distributions (left: $\pi_{\mathrm{FF}}$, right: $\pi_{\mathrm{EF}}$). Bottom row: $\mathrm{dPD}/\mathrm{PD}_{\mathrm{clean}}$ validation curves under the same prompt orders.}
    \label{fig:mistral_heads_panel}
\end{figure*}

This section supplements \S\ref{sec:heads}. We first show the representative attention patterns of the three head roles (Figure~\ref{fig:combined_heads}), then per-model 4-panel layouts of the layer-wise role distributions and validation curves under both prompt orders (Figures~\ref{fig:qwen3_8b_heads_panel}--\ref{fig:mistral_heads_panel}). Numerical role counts: under $\pi_{\mathrm{FF}}$, $66/129/175$ (Splitting/Transmission/Fact-Retrieval) in \textit{Qwen3-8B} and $87/169/217$ in \textit{Qwen3-14B}; under $\pi_{\mathrm{EF}}$, $116/114/167$ and $170/142/200$. \textit{Llama} yields $323$ classified heads under $\pi_{\mathrm{FF}}$ ($119/106/98$) and $377$ under $\pi_{\mathrm{EF}}$ ($139/146/92$); \textit{Mistral} yields $314$ under $\pi_{\mathrm{FF}}$ ($118/99/97$) and $307$ under $\pi_{\mathrm{EF}}$ ($123/109/75$). The Splitting / Transmission / Fact-Retrieval taxonomy is recoverable in every cell; the family-level differences lie in the sharpness of the validation curves, not in the existence of the roles.

\end{document}

%% file: experiments.tex
\section{Experiments}
\label{sec:results}
We analyze propositional reasoning using coarse-to-fine intervention granularity. We first use mean ablation on MLP and Attention (§\ref{sec:staged}) to when each \textit{logical block} (\textit{i.e.}, Facts, Expression, Query) is processed temporally. Next we zoom in to the \textit{structural categories} and ask \emph{how} information routes
(§\ref{sec:info_and_retrieval}). We finally identify \emph{which} attention heads implement the above patterns as discrete \textit{computational operators} \citep{wang2022interpretabilitywildcircuitindirect} (§\ref{sec:heads}). 
We run the full pipeline on four LLMs: \textit{Qwen3-8B/14B}, \textit{Llama-3.1-8B-Instruct}, and \textit{Mistral-7B-Instruct}. Main-text figures mainly use \textit{Qwen3} as representative cases.

\subsection{Coarse-grained Analysis: Staged Computation}
\label{sec:staged}
We begin with a fundamental question: \textit{whether LLMs organize propositional reasoning into computational stages given different logical  blocks?}
Therefore, we mean-ablate each logical block with the token-wise mean computed over the input prompt at every layer in three complementary pathways (MLP, Attention, and the joint Attention+MLP) to remove that block's information without injecting any outlier signal and isolate the layers at which the block carries necessity load.

\begin{figure*}[!htbp]
    \centering
    \begin{subfigure}[b]{0.285\textwidth}
        \centering
        \includegraphics[width=\linewidth]{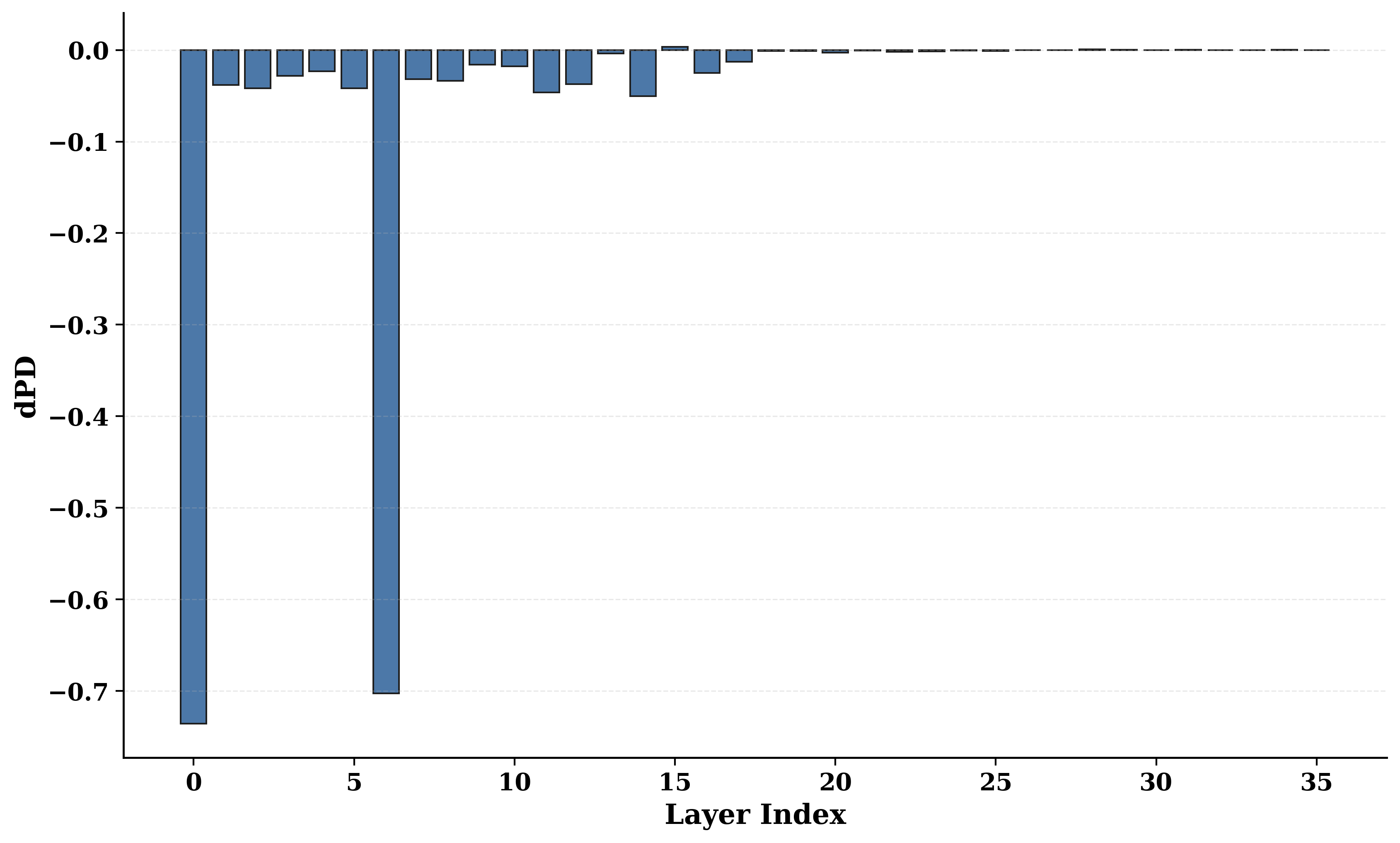}
        \caption{Facts (One-hop)}
    \end{subfigure}
    \hfill
    \begin{subfigure}[b]{0.285\textwidth}
        \centering
        \includegraphics[width=\linewidth]{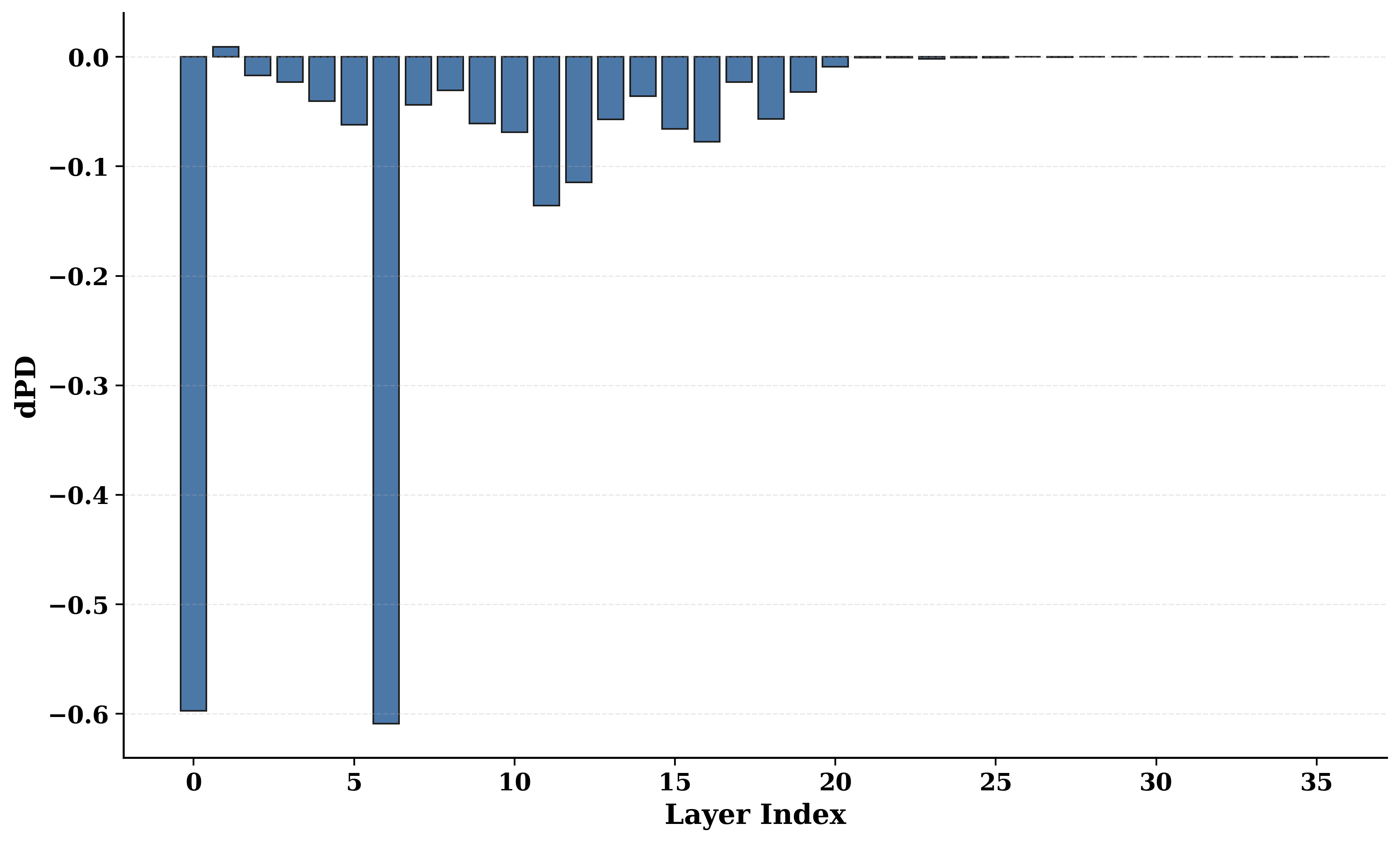}
        \caption{Expression (One-hop)}
    \end{subfigure}
    \hfill
    \begin{subfigure}[b]{0.285\textwidth}
        \centering
        \includegraphics[width=\linewidth]{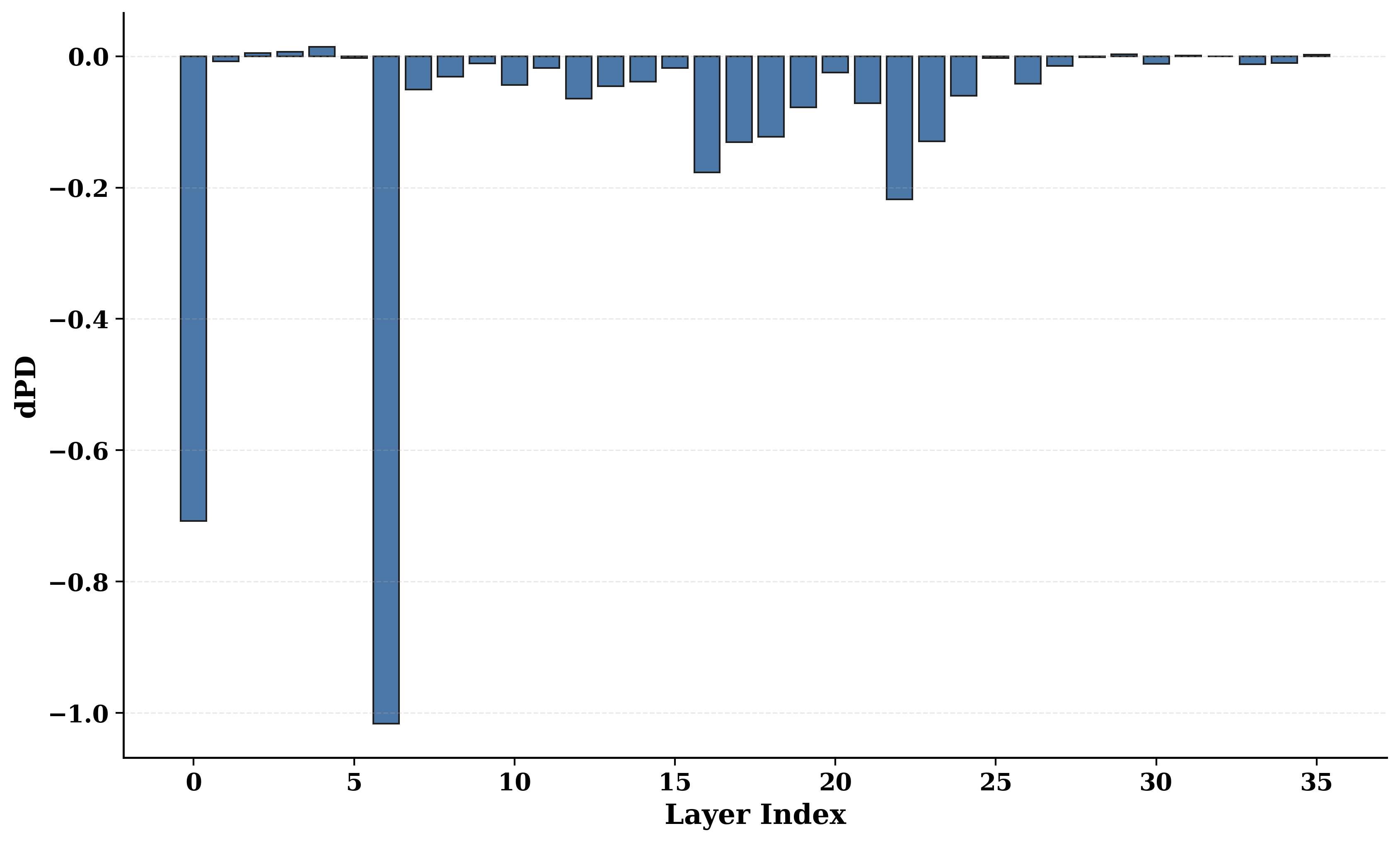}
        \caption{Query (One-hop)}
    \end{subfigure}

    \vspace{0.15em}

    \begin{subfigure}[b]{0.285\textwidth}
        \centering
        \includegraphics[width=\linewidth]{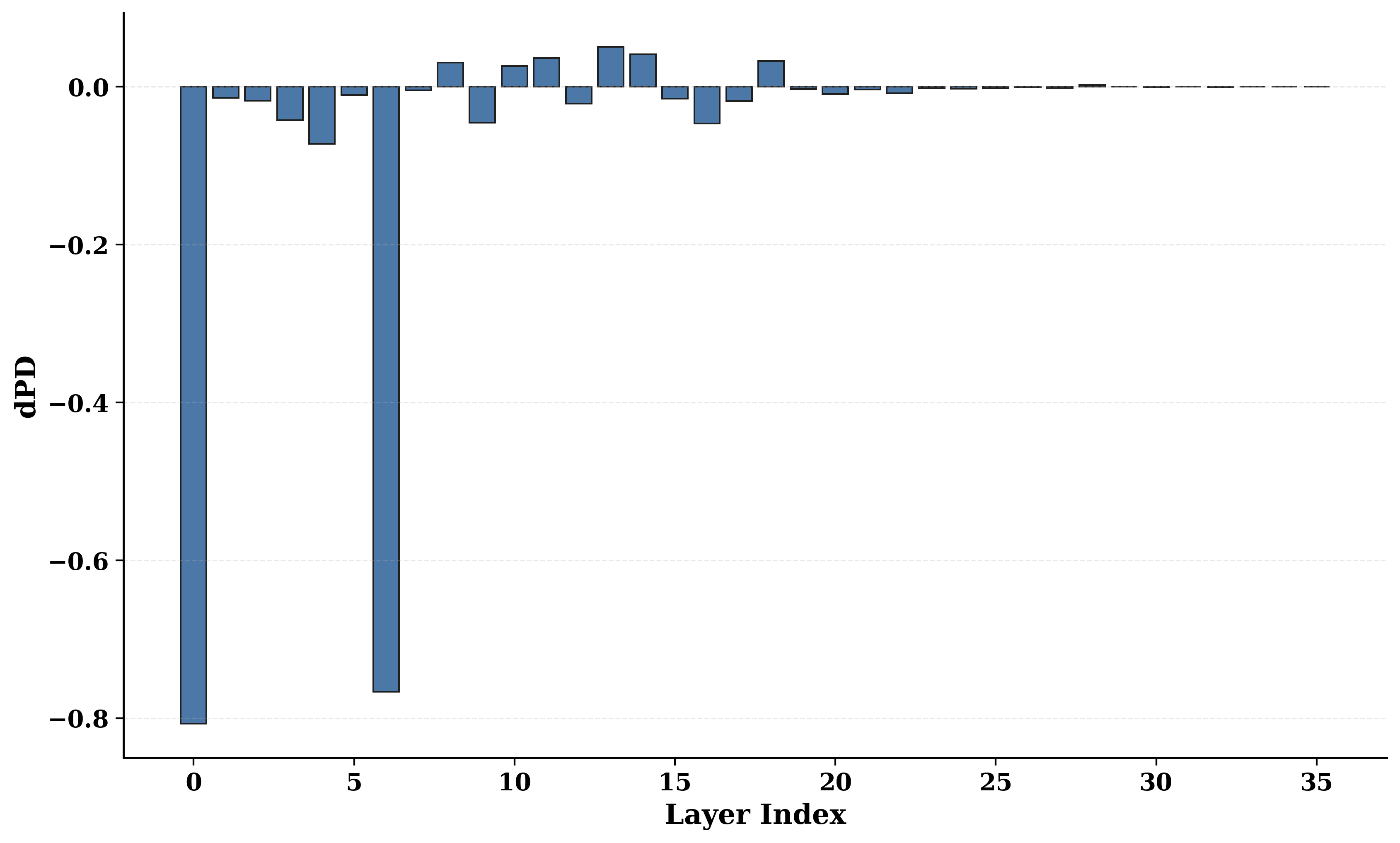}
        \caption{Facts (Two-hop)}
    \end{subfigure}
    \hfill
    \begin{subfigure}[b]{0.285\textwidth}
        \centering
        \includegraphics[width=\linewidth]{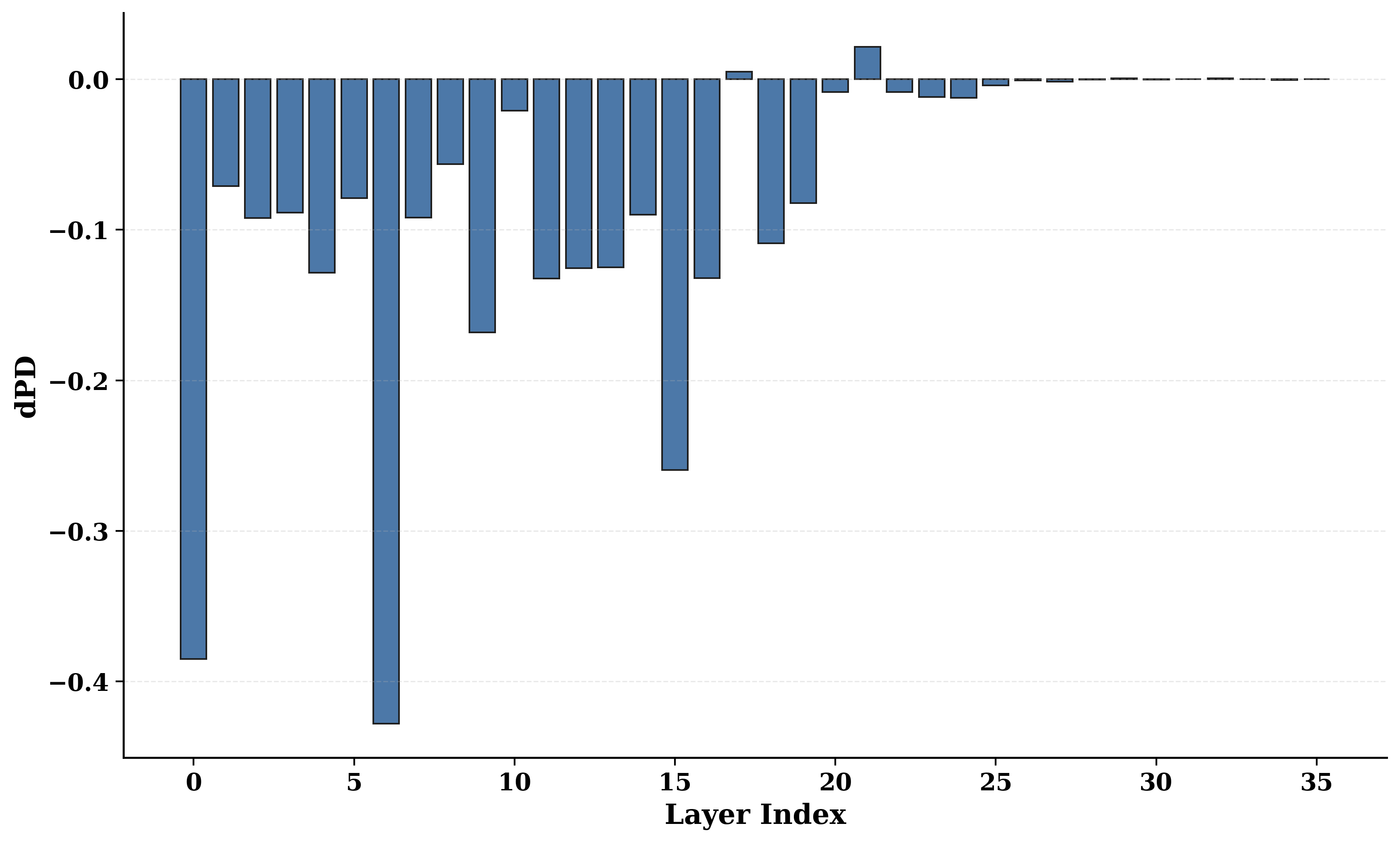}
        \caption{Expression (Two-hop)}
    \end{subfigure}
    \hfill
    \begin{subfigure}[b]{0.285\textwidth}
        \centering
        \includegraphics[width=\linewidth]{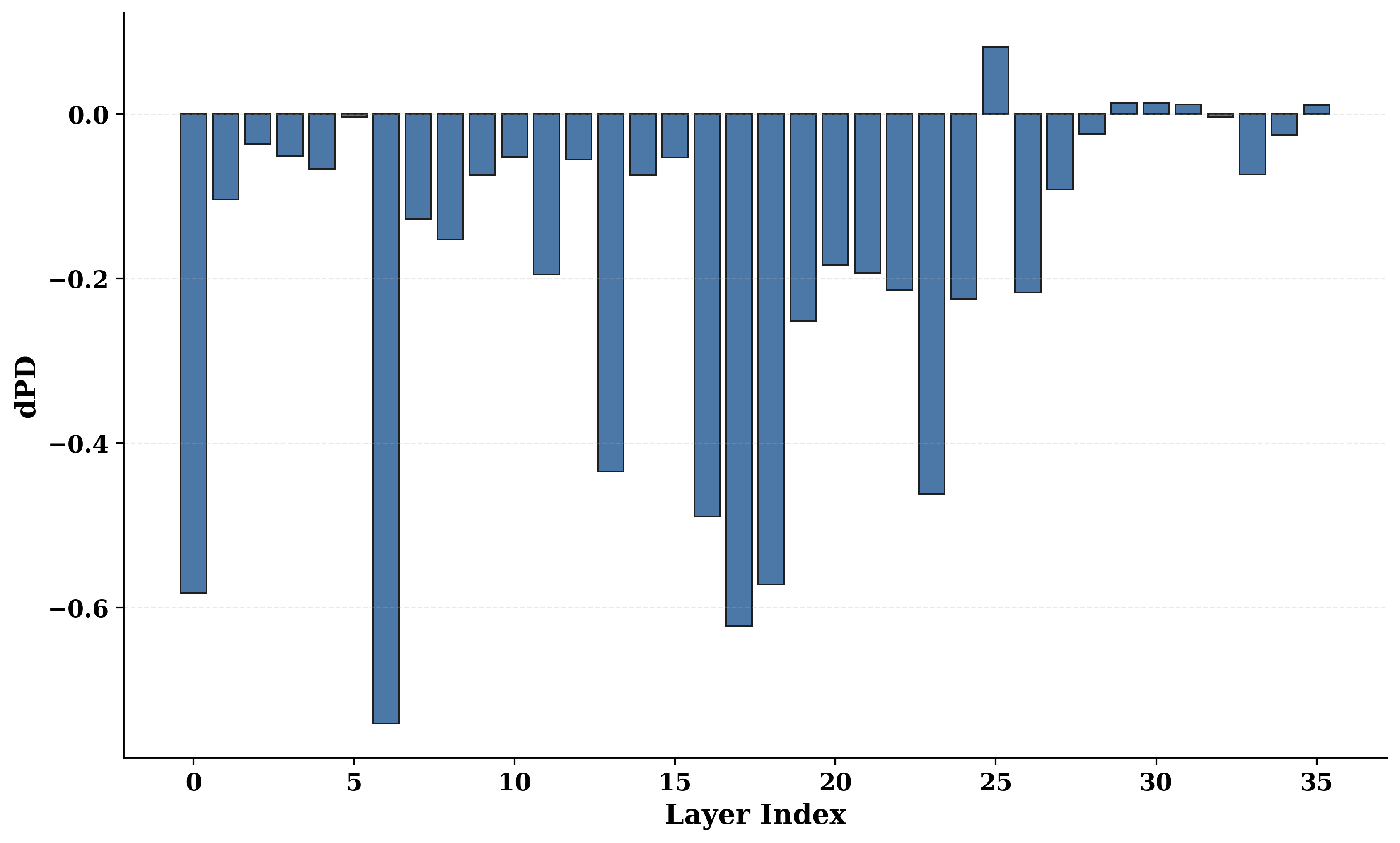}
        \caption{Query (Two-hop)}
    \end{subfigure}
    \caption{\textbf{Joint Attention+MLP mean ablation on \textit{Qwen3-8B} under $\pi_{\mathrm{FF}}$.} Per-layer $\mathrm{dPD}$ when the activations of each logical block (Facts/Expression/Query) are mean-replaced. Facts effects concentrate in early layers, Expression in middle layers, and Query in late layers; two-hop preserves the ordering while broadening the Query(§\ref{sec:staged}).}
    \label{fig:qwen3_8b_attn_mlp_facts_gallery}
    \vspace{-0.5em}
\end{figure*}

The same staged pattern appears across all three ablation pathways. Figure~\ref{fig:qwen3_8b_attn_mlp_facts_gallery}
shows the representative \textit{Qwen3-8B} joint ablation. \texttt{Fact\_Block} concentrates in the early layers, while \texttt{Expression\_Block} dominates in the middle layers and \texttt{Query\_Block} in the late layers. Notably, few early layers carry significant effects for all logical blocks simultaneously, so we concentrated distributions rather than peaks, and the meaningful signal is the \emph{relative order} of the three distributions instead of absolute layer cutoff. Concretely, the \texttt{Expression} distribution lies after the \texttt{Fact} distribution, and the \texttt{Query} lies after the \texttt{Expression}; two-hop reasoning broadens the later computation but preserves this computational ordering.
Results of MLP ablation and Attention ablation are in
(Figure~\ref{fig:attn_region_triples}, \ref{fig:attn_mlp_region_triples}; Appendix~\ref{sec:appendix_region_variants}).

The same ordering survives prompt-order reversal and model change. Under $\pi_{\mathrm{EF}}$ the three distributions keep the same order as under $\pi_{\mathrm{FF}}$, but the staging is more compact: the \texttt{Fact} concentration arrives roughly one to two layers later, while the \texttt{Expression} and \texttt{Query} concentrations move few layers earlier, so the three stages occupy a narrower depth window (Appendix~\ref{sec:appendix_stage_mlp}; compare Figures~\ref{fig:stage_mlp_q8_ff} and~\ref{fig:stage_mlp_q8_ef}). Scaling to \textit{Qwen3-14B} reproduces the same template (Figure~\ref{fig:staged_computation_14B}; Appendix~\ref{sec:appendix_stage}). The ordering also holds across model families, with two coexisting family differences: \textit{Llama-3.1-8B} and \textit{Mistral-7B} place the \texttt{Fact} and \texttt{Expression} stages at earlier layers than \textit{Qwen3}, while devoting more overall mass to the \texttt{Query} block and keeping a broader late-integration tail (Figures~\ref{fig:llama_mlp_facts_gallery}, \ref{fig:mistral_mlp_facts_gallery}; Appendix~\ref{sec:appendix_stage_mlp}). 

\subsection{Fine-grained Analysis: Information Routing}
\label{sec:info_and_retrieval}
This section investigates \textit{how information routes between computational stages} through fine-grained activation patching on residual stream, as it serves as the primary information pathway and accumulates representations across layers, By patching residual stream activations at specific token positions, we can trace where stage-relevant information resides as it propagates through the network. 

For each item we build a clean/corrupt pair that flips the label. The clean prompt is, \textit{e.g.}, \texttt{A is True, B is False, C = ($\neg$A or $\neg$B). C is} (answer \texttt{True}); the corrupt is \texttt{A is True, B is True, C = ($\neg$A or $\neg$B). C is}'' (answer \texttt{False}). We use metric~\ref{eq:aggregated mean dPD} to aggregate token-wise causal influence with each structural categories (Figure~\ref{fig:info_converge_token_main}).

\subsubsection{Information Converges at Boundary Tokens}
\label{sec:info}
\begin{figure*}[!htbp]
    \centering
    \begin{subfigure}{0.32\textwidth}
        \centering
        \includegraphics[width=\linewidth]{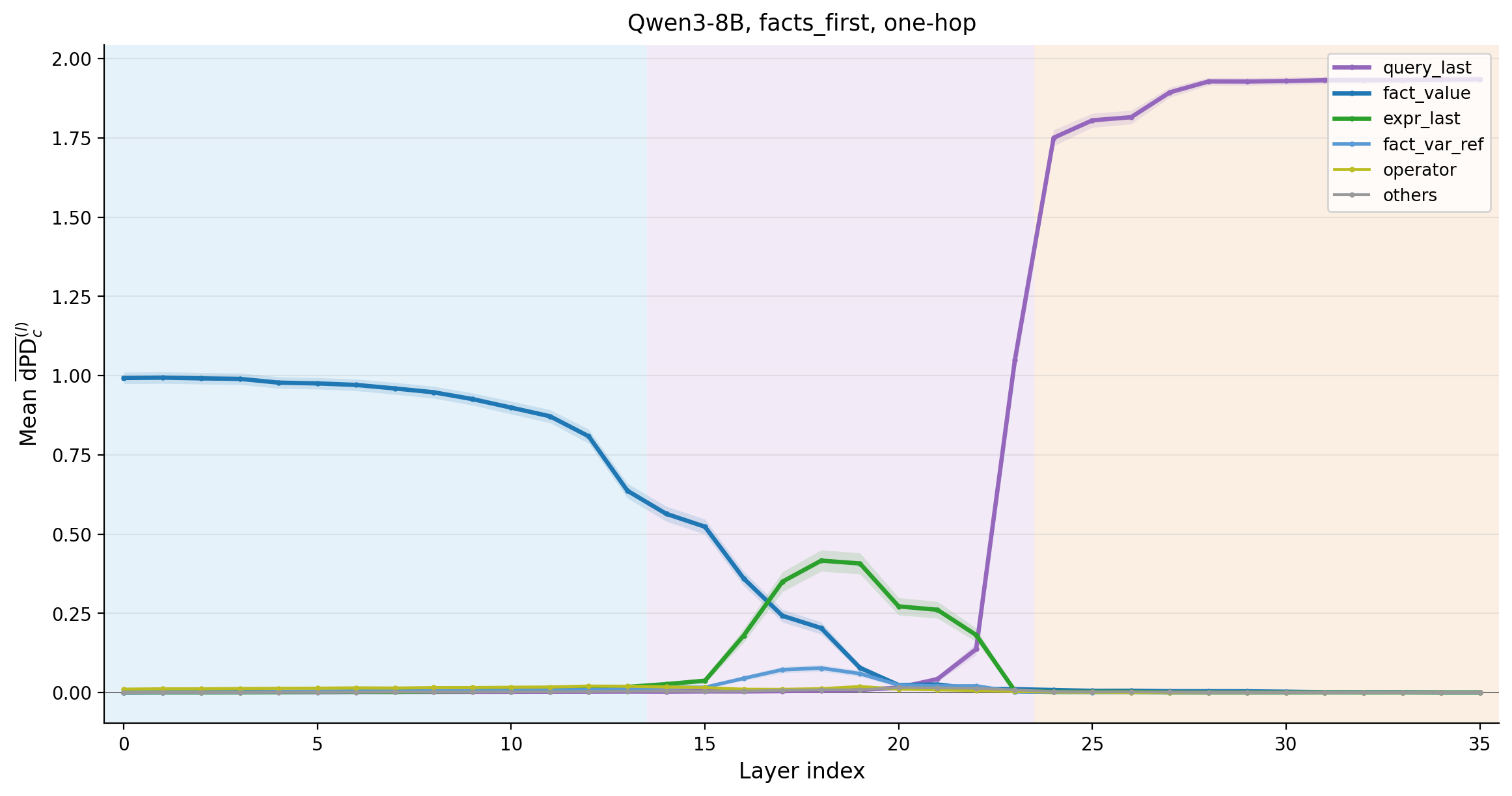}
        \caption{\textit{Qwen3-8B}, $\pi_{\mathrm{FF}}$, one-hop}
        \label{fig:info_converge_token_main}
    \end{subfigure}
    \hfill
    \begin{subfigure}{0.32\textwidth}
        \centering
        \includegraphics[width=\linewidth]{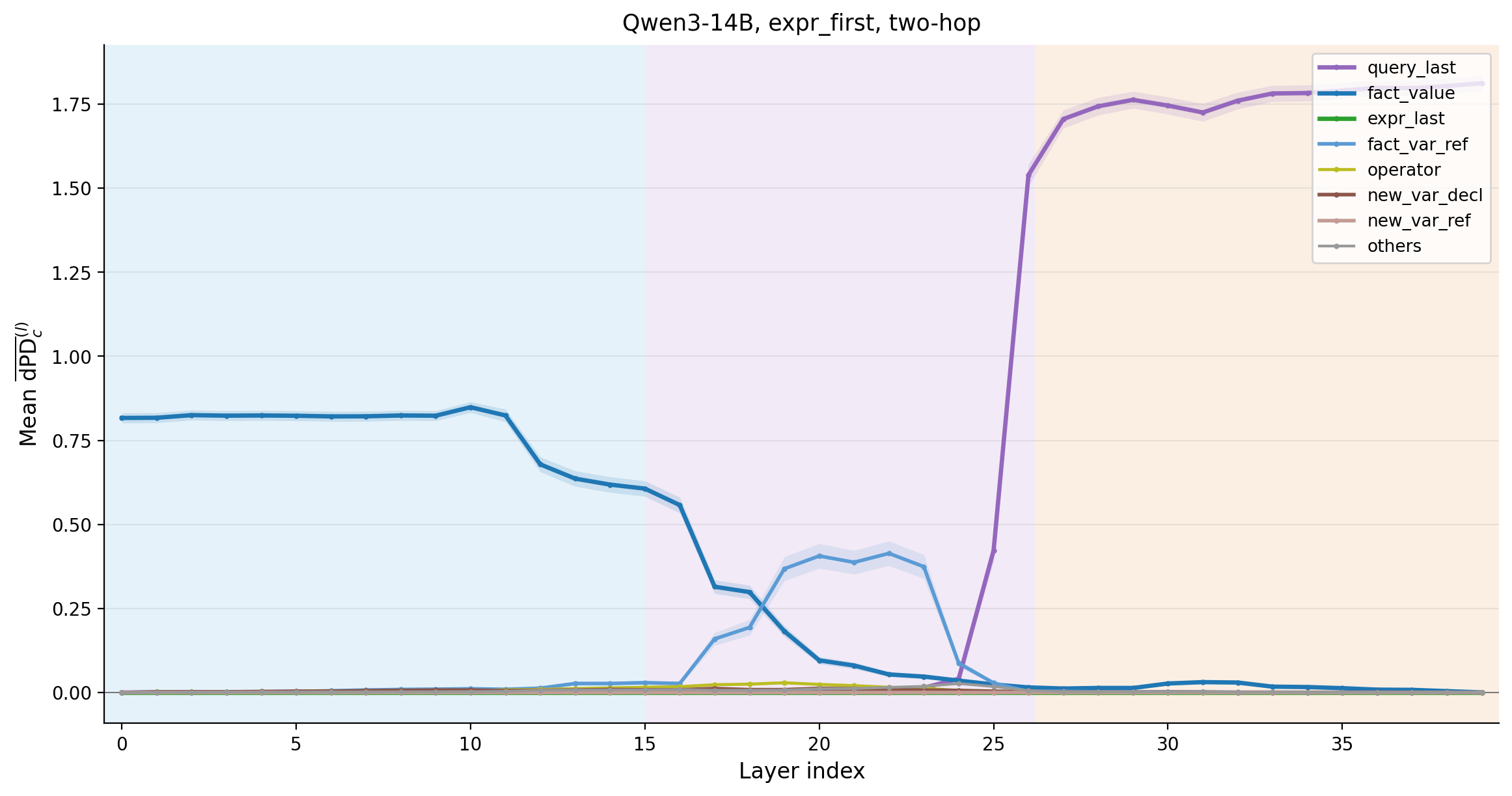}
        \caption{\textit{Qwen3-14B}, $\pi_{\mathrm{EF}}$, two-hop}
        \label{fig:info_converge_token_ef_main}
    \end{subfigure}
    \hfill
    \begin{subfigure}{0.32\textwidth}
        \centering
        \includegraphics[width=\linewidth]{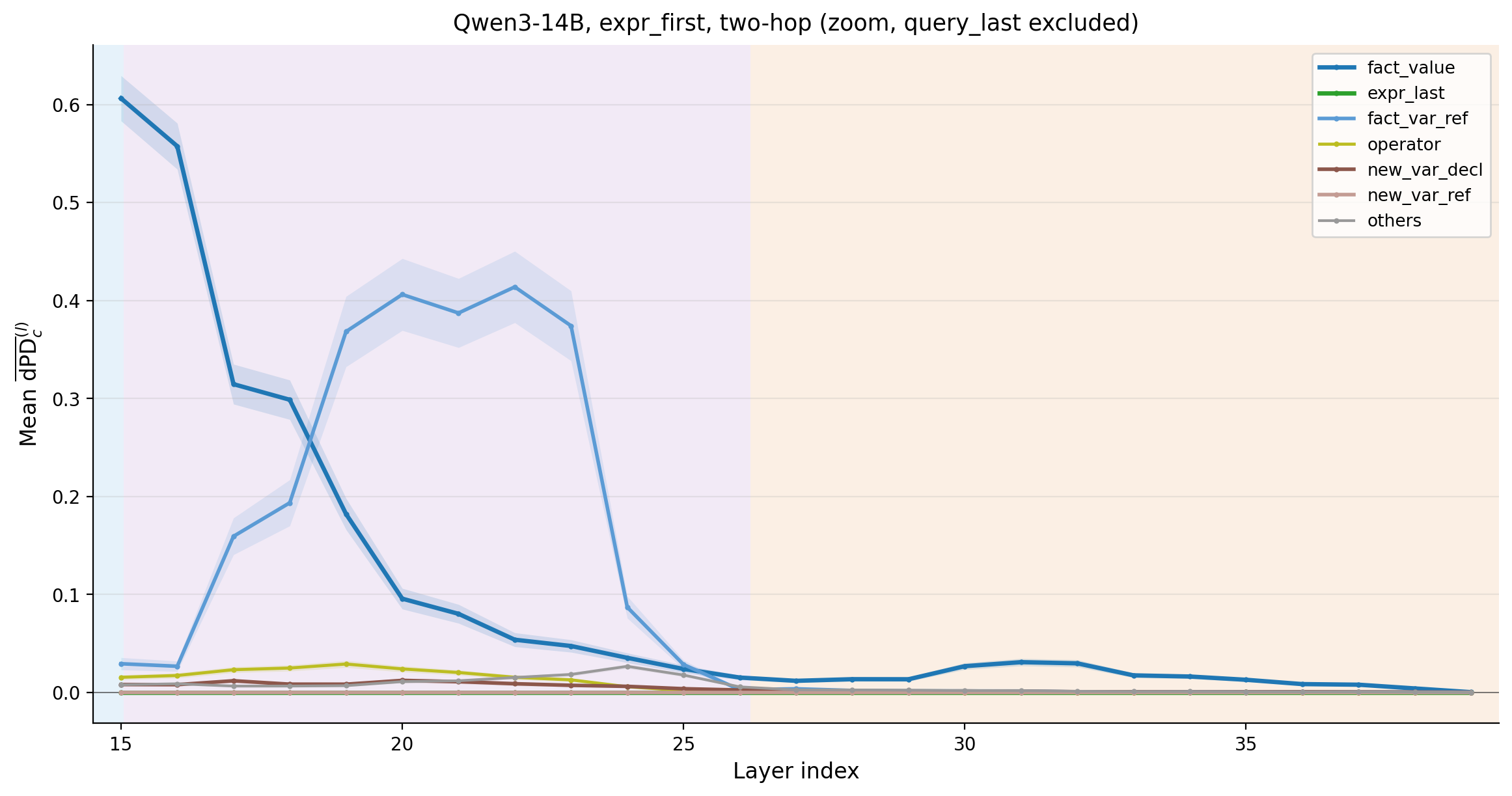}
        \caption{\textit{Qwen3-14B}, $\pi_{\mathrm{EF}}$, two-hop (zoom)}
        \label{fig:fact_retrieval_zoom_main}
    \end{subfigure}
    \caption{\textbf{Per-token mean $\mathrm{dPD}$ by structural category.} Residual-stream activation-patching effects averaged per token within each category; shading marks Early/Middle/Late layers. (a) Under $\pi_{\mathrm{FF}}$, boundary tokens dominate their respective layers---\texttt{fact\_value} early, \texttt{expr\_last} in middle, \texttt{query\_last} late---while the remaining categories stay near zero. (b) Under $\pi_{\mathrm{EF}}$, \texttt{expr\_last} is pinned to zero by autoregressive construction, but \texttt{fact\_value} and \texttt{query\_last} still dominate the early and late layers, confirming that boundary-token convergence persists under prompt reordering. (c) Zoom of (b) into middle and late layers with \texttt{query\_last} excluded: the binding-flow signature during expression process between \texttt{fact\_value} and \texttt{fact\_var\_ref}.}
    \vspace{-0.5em}
    \label{fig:info_converge_combined}
\end{figure*}

Figure~\ref{fig:info_converge_token_main} shows the representative \textit{Qwen3-8B} run under $\pi_{\mathrm{FF}}$. The three boundary-token structural categories dominate their respective computational stages: \texttt{fact\_value} in early layers, \texttt{expr\_last} in middle layers, and \texttt{query\_last} in late layers, while the remaining categories (\texttt{fact\_var\_ref}, \texttt{operator}, \texttt{others}) stay negligible. Causal influence concentrates on a small set of \emph{boundary tokens} that close or anchor each logical block, rather than spreading uniformly across the block. The token-level transport runs Fact $\rightarrow$ Expression $\rightarrow$ Query through these positions, mirroring the layer-level staging in Figure~\ref{fig:qwen3_8b_attn_mlp_facts_gallery}.

Under $\pi_{\mathrm{EF}}$, the clean and corrupt prompts differ only at the \texttt{fact\_value} position (the flipped truth assignment) and the model is autoregressive, so patching the clean residual of any expression token (\textit{e.g.}, \texttt{expr\_last}) into the corrupt forward pass transfers no information and produces no $\mathrm{dPD}$ by construction. The other two boundary tokens continue to dominate their respective layers under $\pi_{\mathrm{EF}}$ (Figure~\ref{fig:info_converge_token_ef_main}, Appendix~\ref{sec:appendix_info_converge}), confirming that boundary-token convergence generalizes across different prompt orderings.



The same boundary-token rule survives two-hop reasoning and replication across all four models. The relative weight of each boundary differs by family: \textit{Qwen3} keeps the strongest fact-side signal, \textit{Llama} hands off to query tokens earliest, and \textit{Mistral} lies in between. Details are in Appendix~\ref{sec:appendix_info_converge}.

\subsubsection{Fact Retrospection}
\label{sec:retro}

\begin{figure}[!htbp]
    \centering
    \includegraphics[width=\columnwidth]{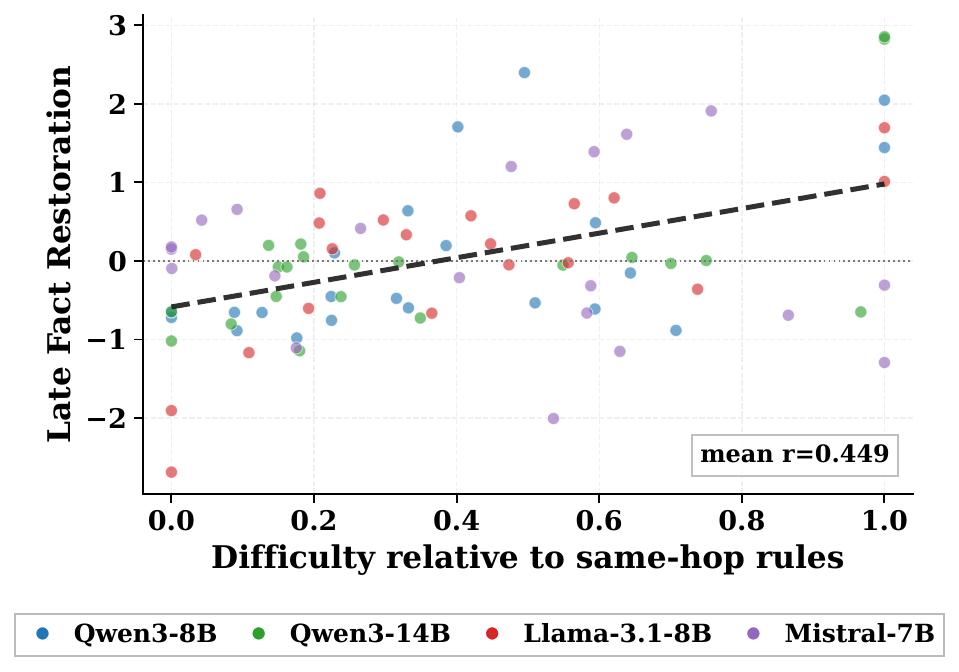}
    \caption{\textbf{Harder tasks show stronger Fact Retrospection.} Each point is one rule $\times$ hop group, colored by model and plotted against its difficulty relative to other rules at the same hop. Both axes are rescaled separately for each model and prompt order. The dashed line is the mean trend.}
    \vspace{-0.5em}
    \label{fig:lfr_difficulty}
\end{figure}

The \texttt{query\_last} in late layers peak visually compresses what happens in the later layers. To reveal the information flow during expression processing, where expression-variable tokens bind to fact values, we exploit the autoregressive suppression of \texttt{expr\_last} under $\pi_{\mathrm{EF}}$ and zoom in middle and late layers with \texttt{query\_last} excluded.


As shown in figure~\ref{fig:fact_retrieval_zoom_main},  \texttt{fact\_value} and \texttt{fact\_var\_ref} exchange dynamically across the middle layers: as the layer index advances, \texttt{fact\_value}'s $\mathrm{dPD}$ declines from its early-layer peak while \texttt{fact\_var\_ref} rises to a middle-layer peak within the same layer range. This indicates information routing during fact binding: at these depths, attention heads use \texttt{fact\_var\_ref} as the query position and copy the truth value from \texttt{fact\_value}, shifting causal effects from \texttt{fact\_value} to \texttt{fact\_var\_ref} across middle layers.
The exchange between \texttt{fact\_value} and \texttt{fact\_var\_ref} generalizes across reasoning hop, prompt order, and models, although magnitude differs. \textit{Qwen3} shows the clearest middle-layer exchange, \textit{Llama-3.1} is weaker, and \textit{Mistral} is negligible (Appendix~\ref{sec:appendix_info_converge}). 
A complementary yet counterintuitive pattern emerges as tasks get harder: factual influence outlasts the binding step. Specifically, in the \emph{late} layers, \texttt{fact\_value} stays well above the other structural categories excepts \texttt{query\_last}, and is markedly significant under two-hop compared to one-hop for the same model and prompt order (Appendix ~\ref{sec:appendix_info_converge}). 
We call the pattern as \textit{Fact Retrospection} that LLMs keep attending back to \texttt{fact\_value} positions even after most information has already routed forward to \texttt{query\_last}.

To quantify the relationship between task difficulty and fact retrospection, we group tasks by hop and logical rule within each model and prompt order. A group's difficulty is one minus the average of its clean and corrupted accuracy, centered within hop so that each group is scored against other rules at the same depth. \textit{Late Fact Restoration} (LFR) is the mean $\mathrm{dPD}$ from patching clean \texttt{fact\_value} activations into the paired corrupted run, averaged over the fact-value positions and over the late layers of each model. Using $1{,}000$ patching examples per model and prompt order, difficulty is positively associated with LFR (Figure~\ref{fig:lfr_difficulty}; $\bar r = 0.449$ over the $8$ model $\times$ prompt-order conditions, stratified permutation $p = 0.0004$). Therefore, \textit{Fact Retrospection} strengthens as tasks get harder.


\subsection{Computational Operaters Analysis: Specialized Attention Heads} 
\subsubsection{Head Taxonomy}
\label{sec:heads}

\begin{figure}[!htbp]
    \centering
    \includegraphics[width=\linewidth]{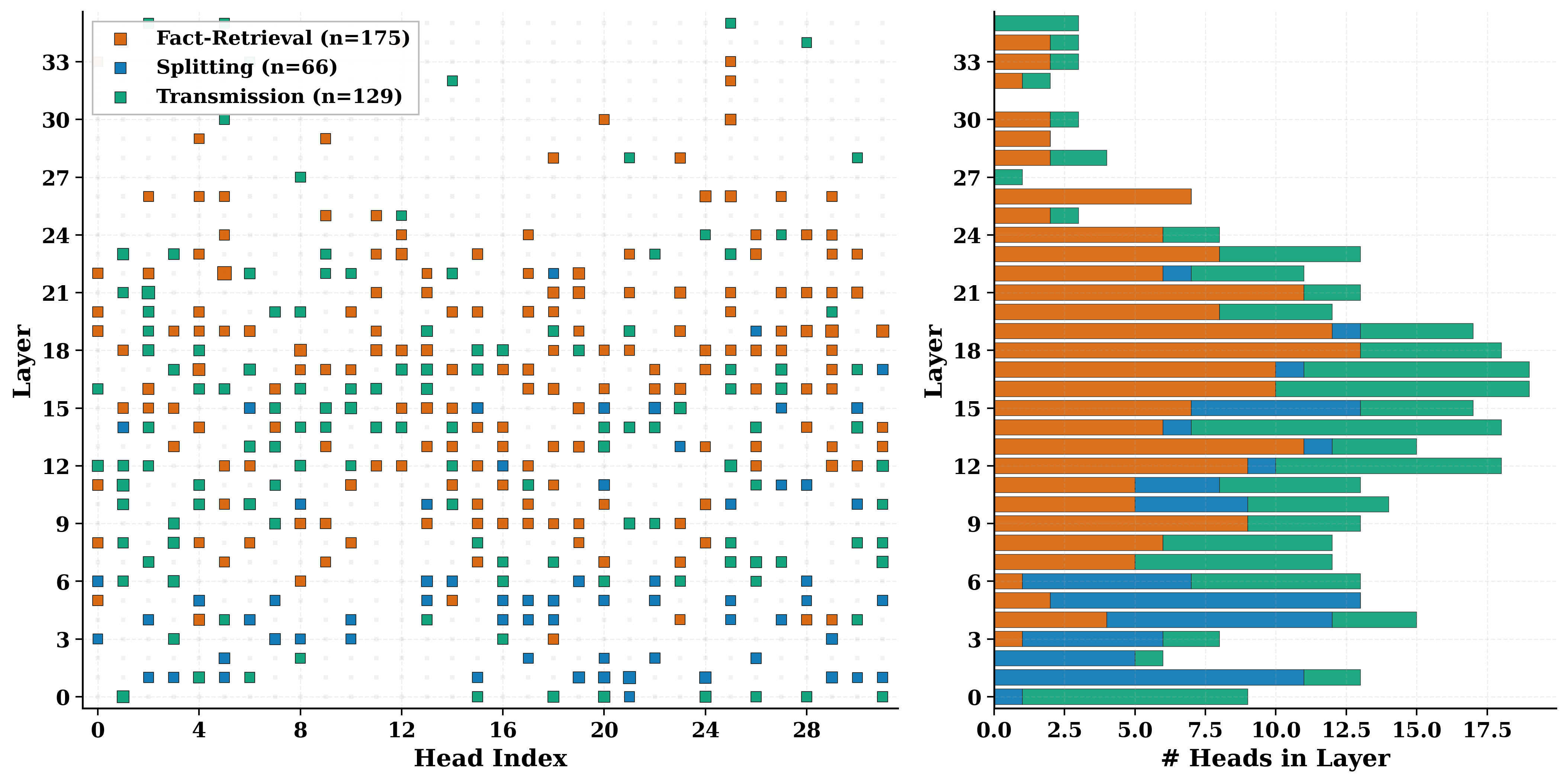}
    \caption{\textbf{Layer-wise distribution of the three head roles in \textit{Qwen3-8B} under $\pi_{\mathrm{FF}}$.} Counts of heads classified as Splitting, Transmission, and Fact-Retrieval at each layer. Splitting Heads concentrate in early layers, Transmission Heads in middle layers, and Fact-Retrieval Heads peak from the middle into late layers. Scale, prompt-order, and cross-family extensions are in Appendix~\ref{appendix_heads}.}
    \label{fig:counts}
    \vspace{-0.5em}
\end{figure}



With the macroscopic computational patterns characterized in the preceding sections, the last core question remains: \textit{which attention heads implement these patterns as \emph{computational operators}?}
Analyzing the attention patterns, We find Specialized Attention Heads, \textit{i.e.}, a consistent, layer-wise division of labor in which heads at different layers assume specialized, stable functional roles. We group these heads into three main categories, each corresponding to a macroscopic mechanism identified earlier. Detailed attention patterns of these heads could refer to figure \ref{fig:combined_heads}. We further compute the layer-wise distribution of these heads (Figure~\ref{fig:counts}). The depth ranges where each role concentrates mirror the macroscopic staging of §\ref{sec:staged}--§\ref{sec:retro}.


\begin{table*}[!htbp]
\centering
\scriptsize
\renewcommand{\arraystretch}{0.95}
\setlength{\tabcolsep}{2.5pt}
\resizebox{\textwidth}{!}{%
\begin{tabular}{@{} c l | cc cc cc cc | cc cc | cc cc @{}}
\toprule
& & \multicolumn{8}{c|}{\textbf{Qwen3}} & \multicolumn{4}{c|}{\textbf{Llama-3.1-8B}} & \multicolumn{4}{c@{}}{\textbf{Mistral-7B}} \\
\cmidrule(lr){3-10} \cmidrule(lr){11-14} \cmidrule(lr){15-18}
& & \multicolumn{2}{c}{8B-FF} & \multicolumn{2}{c}{8B-EF} & \multicolumn{2}{c}{14B-FF} & \multicolumn{2}{c|}{14B-EF}
  & \multicolumn{2}{c}{FF} & \multicolumn{2}{c|}{EF}
  & \multicolumn{2}{c}{FF} & \multicolumn{2}{c@{}}{EF} \\
\cmidrule(lr){3-4}\cmidrule(lr){5-6}\cmidrule(lr){7-8}\cmidrule(lr){9-10}\cmidrule(lr){11-12}\cmidrule(lr){13-14}\cmidrule(lr){15-16}\cmidrule(lr){17-18}
\textbf{$k$} & \textbf{Role}
  & Acc & $\mathrm{dPD}$ & Acc & $\mathrm{dPD}$ & Acc & $\mathrm{dPD}$ & Acc & $\mathrm{dPD}$
  & Acc & $\mathrm{dPD}$ & Acc & $\mathrm{dPD}$
  & Acc & $\mathrm{dPD}$ & Acc & $\mathrm{dPD}$ \\
\midrule
-- & Baselines & 96.6 & -- & 92.8 & -- & 94.8 & -- & 97.6 & -- & 60.2 & -- & 69.2 & -- & 99.4 & -- & 99.8 & -- \\
\midrule
  \multirow{5}{*}{$8$} & Fact-Retrieval & 74.6 & $-0.40$ & 77.4 & $-0.26$ & 83.2 & $-0.24$ & 86.0 & $-0.23$ & 55.2 & $-0.16$ & 58.4 & $-0.20$ & 41.6 & $-0.21$ & 65.4 & $-0.13$ \\
   & Transmission   & 80.4 & $-0.29$ & 78.0 & $-0.26$ & 76.8 & $-0.34$ & 79.8 & $-0.33$ & 55.2 & $-0.17$ & 58.4 & $-0.19$ & 45.2 & $-0.14$ & 77.6 & $-0.10$ \\
   & Splitting      & 84.6 & $-0.21$ & 81.2 & $-0.19$ & 90.4 & $-0.09$ & 93.4 & $-0.09$ & 55.2 & $-0.18$ & 58.4 & $-0.24$ & 47.0 & $-0.11$ & 50.4 & $-0.19$ \\
   & Mixed-All      & 74.6 & $-0.40$ & 77.6 & $-0.27$ & 74.2 & $-0.39$ & 81.6 & $-0.34$ & 55.2 & $-0.18$ & 58.4 & $-0.22$ & 41.6 & $-0.23$ & 47.4 & $-0.24$ \\
   & Random-Outside & 97.1 & $+0.01$ & 94.7 & $+0.02$ & 96.5 & $+0.02$ & 98.0 & $+0.00$ & 62.4 & $+0.03$ & 72.6 & $+0.03$ & 95.9 & $+0.02$ & 94.5 & $+0.02$ \\
\midrule
  \multirow{5}{*}{$16$} & Fact-Retrieval & 67.4 & $-0.55$ & 70.0 & $-0.41$ & 76.8 & $-0.37$ & 81.4 & $-0.31$ & 55.2 & $-0.16$ & 58.4 & $-0.22$ & 41.2 & $-0.24$ & 44.6 & $-0.21$ \\
   & Transmission   & 69.4 & $-0.51$ & 75.0 & $-0.32$ & 72.2 & $-0.43$ & 76.0 & $-0.44$ & 55.2 & $-0.18$ & 58.4 & $-0.20$ & 41.6 & $-0.19$ & 56.4 & $-0.14$ \\
   & Splitting      & 83.4 & $-0.23$ & 78.2 & $-0.25$ & 88.2 & $-0.14$ & 90.6 & $-0.14$ & 55.0 & $-0.19$ & 58.4 & $-0.23$ & 42.2 & $-0.15$ & 43.6 & $-0.24$ \\
   & Mixed-All      & 68.4 & $-0.54$ & 71.2 & $-0.40$ & 65.8 & $-0.53$ & 76.2 & $-0.43$ & 55.2 & $-0.17$ & 58.4 & $-0.22$ & 41.4 & $-0.25$ & 43.4 & $-0.27$ \\
   & Random-Outside & 97.4 & $+0.02$ & 95.0 & $+0.01$ & 96.9 & $+0.03$ & 97.9 & $+0.00$ & 65.5 & $+0.07$ & 74.1 & $+0.05$ & 92.8 & $+0.04$ & 90.7 & $+0.04$ \\
\midrule
  \multirow{5}{*}{$32$} & Fact-Retrieval & 57.0 & $-0.74$ & 63.2 & $-0.54$ & 63.8 & $-0.58$ & 77.2 & $-0.42$ & 55.2 & $-0.16$ & 58.4 & $-0.21$ & 41.2 & $-0.28$ & 43.8 & $-0.25$ \\
   & Transmission   & 69.2 & $-0.52$ & 73.2 & $-0.36$ & 69.2 & $-0.51$ & 76.6 & $-0.44$ & 55.2 & $-0.18$ & 58.4 & $-0.21$ & 41.2 & $-0.23$ & 44.8 & $-0.19$ \\
   & Splitting      & 80.4 & $-0.30$ & 73.2 & $-0.36$ & 84.8 & $-0.21$ & 86.8 & $-0.23$ & 55.2 & $-0.19$ & 58.4 & $-0.25$ & 41.4 & $-0.18$ & 43.2 & $-0.26$ \\
   & Mixed-All      & 62.0 & $-0.66$ & 67.4 & $-0.48$ & 61.4 & $-0.62$ & 62.0 & $-0.68$ & 55.2 & $-0.17$ & 58.4 & $-0.22$ & 41.2 & $-0.28$ & 43.2 & $-0.31$ \\
   & Random-Outside & 97.2 & $+0.02$ & 92.5 & $-0.02$ & 98.3 & $+0.05$ & 98.1 & $+0.01$ & 72.0 & $+0.12$ & 79.2 & $+0.09$ & 88.4 & $+0.07$ & 85.2 & $+0.07$ \\
\midrule
  \multirow{5}{*}{$64$} & Fact-Retrieval & 50.6 & $-0.88$ & 58.4 & $-0.61$ & 57.2 & $-0.74$ & 78.0 & $-0.41$ & 55.2 & $-0.16$ & 58.4 & $-0.22$ & 41.2 & $-0.29$ & 43.2 & $-0.30$ \\
   & Transmission   & 62.6 & $-0.65$ & 60.0 & $-0.62$ & 57.4 & $-0.72$ & 69.4 & $-0.56$ & 55.2 & $-0.17$ & 58.4 & $-0.22$ & 41.2 & $-0.28$ & 43.2 & $-0.24$ \\
   & Splitting      & 70.4 & $-0.48$ & 69.2 & $-0.43$ & 75.2 & $-0.39$ & 64.0 & $-0.66$ & 55.2 & $-0.20$ & 58.4 & $-0.23$ & 41.2 & $-0.23$ & 43.2 & $-0.31$ \\
   & Mixed-All      & 52.2 & $-0.83$ & 57.4 & $-0.67$ & 53.0 & $-0.81$ & 56.8 & $-0.85$ & 55.2 & $-0.16$ & 58.4 & $-0.21$ & 41.2 & $-0.31$ & 43.2 & $-0.33$ \\
   & Random-Outside & 91.1 & $-0.09$ & 86.7 & $-0.11$ & 97.1 & $+0.04$ & 98.3 & $+0.01$ & 78.9 & $+0.16$ & 84.4 & $+0.17$ & 79.5 & $+0.14$ & 77.6 & $+0.12$ \\
\bottomrule
\end{tabular}%
}
\caption{\textbf{Post-ablation accuracy and mean $\mathrm{dPD}$ shift after top-$k$ head ablation, across four models and both prompt orders.} For each setting we report \textbf{Acc}, the post-ablation top-1 accuracy on held-out clean prompts (\%), and \textbf{dPD}, the mean shift in the two-answer probability difference $\mathrm{PD}_{\mathtt{True}}-\mathrm{PD}_{\mathtt{False}}$ relative to the unintervened baseline (more negative $=$ stronger push away from the correct answer). FF $=\pi_{\mathrm{FF}}$, EF $=\pi_{\mathrm{EF}}$. \textit{Mixed-All} ablates the union of the top heads of all three roles; \textit{Random-Outside} control is sampled from heads not assigned to any role.
}
\label{tab:qwen3_head_accuracy_main}
\end{table*}

\paragraph{Splitting Heads.} In early layers, these heads concentrate on comma tokens (Figure~\ref{fig:splitting_head}), which act as boundary markers between logical blocks. They materialize Staged Computation: by marking clause boundaries, they separate the factual premises from the expression to be solved.
\paragraph{Transmission Heads.} These heads aggregate information within a logical block and route it toward the block's terminal token. Their attention maps form lower-triangular blocks inside the Facts and Expression blocks (Figure~\ref{fig:convergence_head}), matching the boundary-token convergence pattern of §\ref{sec:info}. 
\paragraph{Fact-Retrieval Heads.} These heads route attention from the query token or expression tokens back toward truth-value tokens (\texttt{True}/\texttt{False}) (Figure~\ref{fig:facts_head}), and they concentrate in middle-to-late layers (Figure~\ref{fig:counts}). This is the head-level counterpart of the information routing in §\ref{sec:retro}: the same depth range where fact-value tokens remain above zero is occupied by heads which retrieve those fact tokens.

\subsubsection{Ablation Study}
To verify whether the head taxonomy genuinely implements the above mechanisms, we perform targeted mean-ablation of the top-$k$ heads of each role and compare them with a \textit{Random-Outside} control sampled from heads not assigned to any role (Appendix~\ref{sec:appendix_interventions}). We report the model's post-ablation accuracy on held-out clean prompts together with the mean $\mathrm{dPD}$ shift, the change in the two-answer probability difference $\mathrm{PD}_{\mathtt{True}}-\mathrm{PD}_{\mathtt{False}}$ relative to the unintervened baseline.

Table~\ref{tab:qwen3_head_accuracy_main} reports both metrics across $k \in \{8, 16, 32, 64\}$ for all models and both prompt orders. In \textit{Qwen3}, role-targeted ablations consistently drop accuracy well below the \textit{Random-Outside} control, with the ordering Fact-Retrieval $\geq$ Transmission $>$ Splitting robust across sizes and prompt orders; the cross-family rows show two distinct failure modes, with \textit{Llama3.1} saturating by $k=8$ (roles not cleanly separated) and \textit{Mistral} degrading sharply under both role-targeted and \textit{Random-Outside} ablations (globally sensitive).